\documentclass[
12pt,
english,
infoheadfooters 
version=v1.0.0 
]{phDissertation}

\usepackage{algorithm}
\usepackage{algorithmic}

\usepackage{amssymb}
\usepackage{amsfonts,amstext,amsmath,amssymb}
\usepackage{booktabs}
\usepackage{multirow, subfig}
\usepackage{comment}
\usepackage{adjustbox}
\usepackage{amsmath,amssymb,epsfig,graphicx,url}
\usepackage{xspace,stmaryrd}
\usepackage{nicefrac}
\usepackage{caption}
\usepackage{textcomp}
\usepackage{lscape}
\usepackage{rotating}
\usepackage{diagbox}
\usepackage{pgfplots}
\pgfplotsset{compat=newest}
\usetikzlibrary{decorations.pathreplacing,angles,quotes}
\usetikzlibrary{shapes.misc, shapes.geometric,positioning}
\tikzset{
	cross/.style={cross out, draw=black, minimum size=2*(#1-\pgflinewidth), inner sep=0pt, outer sep=0pt}, 
	cross/.default={1pt}
}

\let\Oldendproof\endproof%
\def\endproof{\hfill$\blacksquare$\Oldendproof}%
\usepackage{stmaryrd}

\usepackage{epigraph} 
\newcommand{\ttpm}{\textnormal{\texttt{\textpm}}}
\newcommand{\new}[1]{{\color{black}{#1}}} 
\newcommand{\bx}{\boldsymbol{x}}
\newcommand{\bbeta}{\boldsymbol{\beta}}
\newcommand{\bp}{\boldsymbol{p}}
\newcommand{\btheta}{\boldsymbol{\theta}}

\def\somega#1{\{\omega_#1\}}
\newcommand{\acc}{\textsf{acc}}
\newcommand{\conf}{\textsf{conf}}

\usepackage{bibentry}
\usepackage{lipsum}
\usepackage{proofread}

\usetikzlibrary{cd}
\usepackage{nicefrac}
\usepackage{tikz}
\usepackage{pgfplots}
\usetikzlibrary{chains, arrows, automata, shapes.misc}
\pgfplotsset{compat=newest}
\tikzset{
	treenode/.style = {shape=rectangle, rounded corners,
	 draw, align=center,
	 top color=white, bottom color=blue!20},
	root/.style     = {treenode, font=\Large, bottom color=red!30},
	env/.style      = {treenode, font=\ttfamily\normalsize},
	dummy/.style    = {circle,draw}
}

\thesistitle{ Medical Image Segmentation with Belief Function Theory and Deep Learning}
\examiner{}
\degree{Doctor of Philosophy} 
\author{Ling \textsc{HUANG}}
\addresses{}
\subject{Biological Sciences} 
\university{\href{http://www.utc.fr}{University of Technology of Compiègne}}
\department{~}
\group{\href{https://www.hds.utc.fr/en/research/research-teams/cid-team-knowledge-uncertainty-data.html}{CID Team (Knowledge, Uncertainty, Data) \\ Heudiasyc Laboratory}} 
\faculty{\href{http://faculty.university.com}{Faculty Name}}
\AtBeginDocument{
	\hypersetup{pdftitle=\ttitle} 
	\hypersetup{pdfauthor=\authorname} 
	\hypersetup{pdfkeywords=\keywordnames} 
}

\begin{document}
\newgeometry{
  margin=1in, 
  footskip=1pt
}
\frontmatter 
\begin{titlepage}
	\begin{center}
		\includegraphics[scale=0.23]{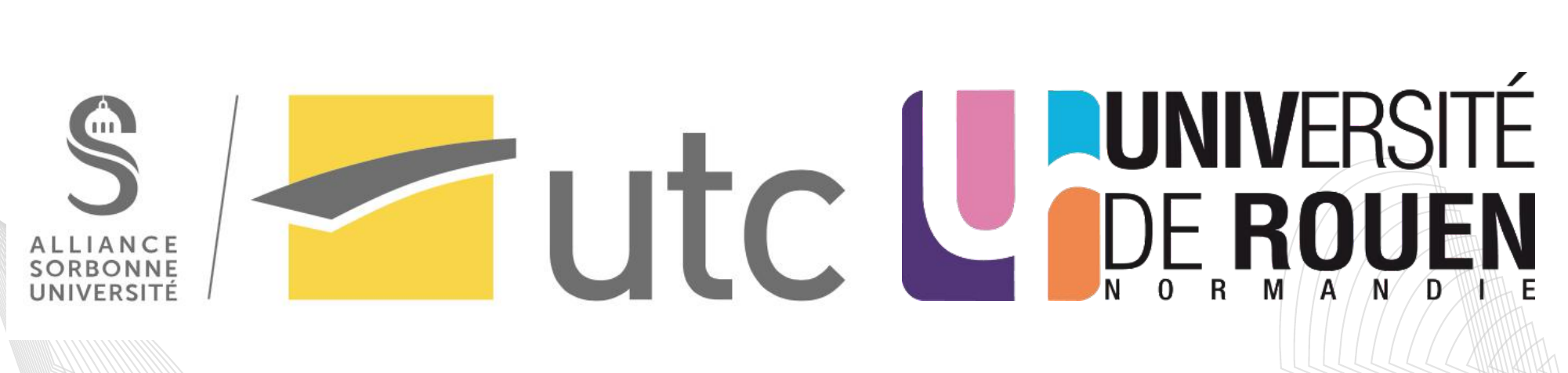}
		\hfill
		\includegraphics[scale=0.35]{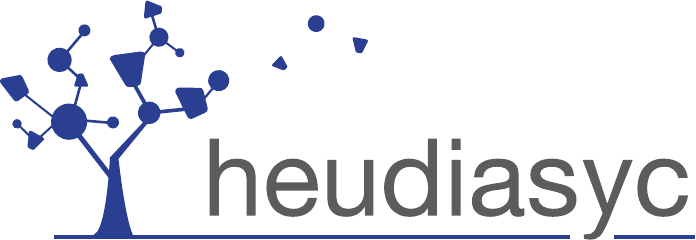}
		\hfill
		\includegraphics[scale=0.045]{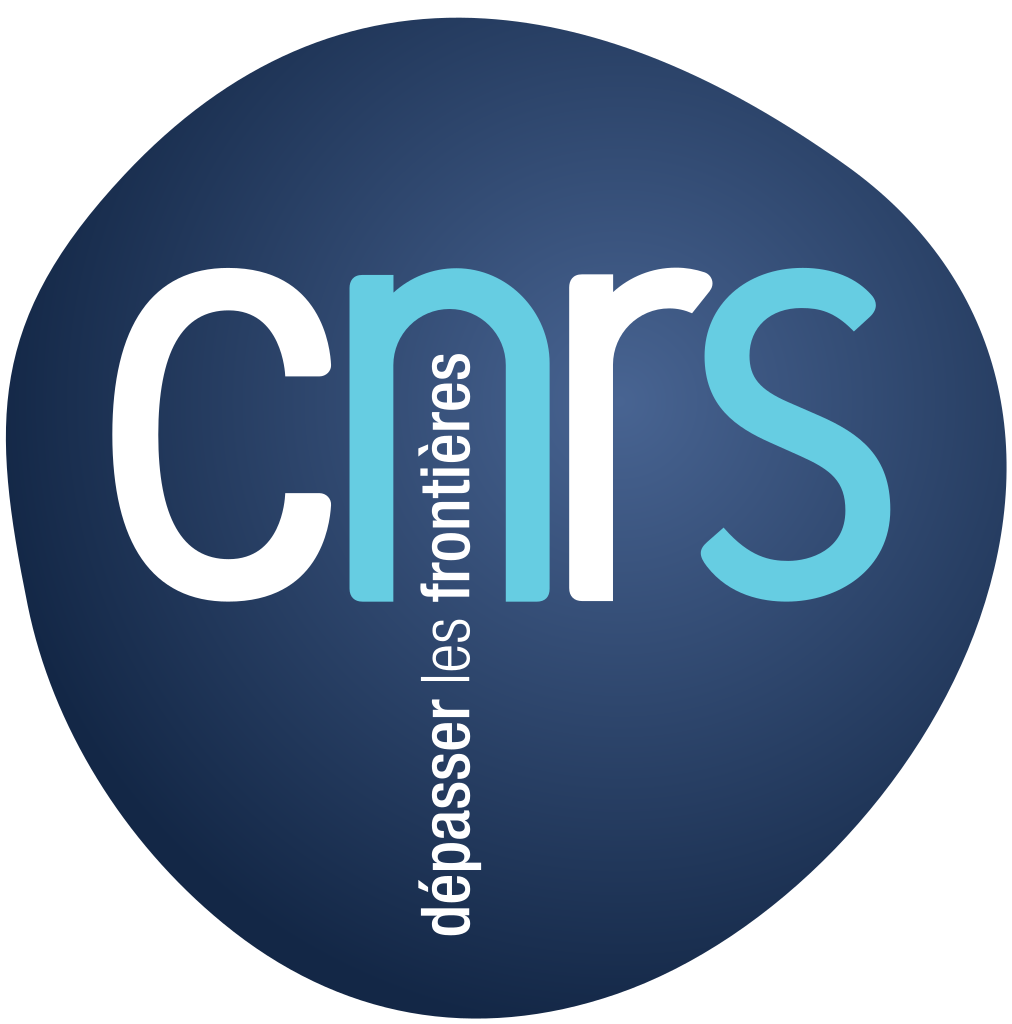}
         \hfill
         \includegraphics[scale=0.12]{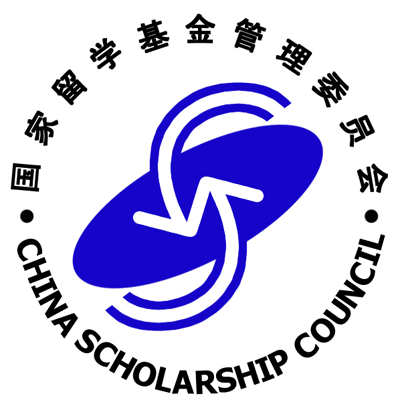}\\
		\vspace*{.02\textheight}
		{\scshape\LARGE \univname\par}\vspace{0.7cm} 
		\textsc{\Large Doctoral Thesis}\\[0.5cm] 
	
		\HRule \\[0.2cm] 
		{\huge \bfseries \ttitle\par}\vspace{0.2cm} 
		\HRule \\[0.5cm] 
	
		\begin{minipage}[t]{0.45\textwidth}
			\begin{flushleft} \large
				\emph{Author:}\\
				\href{https://www.hds.utc.fr/~hualing/dokuwiki/}{Ling \textsc{Huang}} 
			\end{flushleft}
		\end{minipage}
		\begin{minipage}[t]{0.45\textwidth}
			\begin{flushright} \large
				\emph{Supervisors:} \\
				Prof. Thierry \textsc{Den{oe}ux} \\
				Prof. Su \textsc{Ruan} 
			\end{flushright}
		\end{minipage}\\[1cm]
		
		\begin{minipage}[t]{0.9\textwidth}
			\begin{flushleft} 
				\large\emph{Jury:}\\[0.1cm]\normalsize
				\begin{tabular}{llr} 	
              		\hspace{-2mm}\href{hds.utc.fr/~fdavoine/dokuwiki/doku.php}{Prof. Franck DAVOINE} & 
			{UTC}  &  
					President of jury\\    
					
					\hspace{-2mm}\href{https://www.researchgate.net/profile/Sylvie-Le-Hegarat}{Prof. Sylvie LE HÉGARAT} & 
					{University of Paris-Saclay} &
					Reviewer\\
					
					\hspace{-2mm}\href{https://davidmercier.fr/}{Prof. David MERCIER} &
					{University of Artois} &
					Reviewer\\
					
					\hspace{-2mm}\href{https://www.femto-st.fr/en/femto-people/emmanuelramasso}{Asso Prof. Emmanuel RAMASSO} &
					{ENSMM School} &
					Examiner\\
					
					\hspace{-2mm}\href{https://pagesperso.litislab.fr/suruan/}{Prof. Su RUAN} & 
					{University of Rouen} &
					Co-supervisor\\
					
				\hspace{-2mm}\href{https://www.hds.utc.fr/~tdenoeux/dokuwiki/}{Prof. Thierry DEN{OE}UX} & {UTC}  &  
					Supervisor\\

				\end{tabular}
			\end{flushleft}
		\end{minipage}\\[0.6cm]

		\vspace{-0.1cm}\vfill\large 
		\textit{A thesis submitted in fulfillment of the requirements\\ 
		for the degree of \degreename}\\[0.2cm] 
		\textit{in the}\\[0.2cm]
		\groupname\\\deptname\\
		{\large \today}
	\end{center}
	\cleardoublepage
\end{titlepage}


\begin{abstract}
\addchaptertocentry{\abstractname}
Deep learning has shown promising contributions in medical image segmentation with powerful learning and feature representation abilities. However, it has limitations for reasoning with and combining imperfect (imprecise, uncertain, and partial) information. In this thesis, we study medical image segmentation approaches with belief function theory and deep learning, specifically focusing on information modeling and fusion based on uncertain evidence. 

First, we review existing belief function theory-based medical image segmentation methods and discuss their advantages and challenges. Second, we present a semi-supervised medical image segmentation framework to decrease the uncertainty caused by the lack of annotations with evidential segmentation and evidence fusion. Third, we compare two evidential classifiers, evidential neural network and radial basis function network, and show the effectiveness of belief function theory in uncertainty quantification; we use the two evidential classifiers with deep neural networks to construct deep evidential  models for lymphoma segmentation. Fourth, we present a multimodal medical image fusion framework taking into account the reliability of each MR image source when performing different segmentation tasks using mass functions and contextual discounting.

\mbox{}
\end{abstract}
	\begin{acknowledgements}\addchaptertocentry{\acknowledgementname} 



First and foremost, I am extremely grateful to my two supervisors, Prof. Thierry Den{\oe}ux from UTC and Prof. Su Ruan from University of Rouen Nornamdie for their invaluable advice and continuous support during my Ph.D. study. It is my great luck to be supervised by Prof. Thierry Den{\oe}ux and Prof.Su Ruan. Without their assistance and dedicated involvement, this thesis would have never been accomplished. 

I would like to express my great gratitude to my supervisor, Prof. Thierry Den{\oe}ux, for his constructive guidance during my PhD study. As a scrupulous and creative scholar working in the field of uncertainty modeling, belief function theory, information fusion, Prof. Thierry Den{\oe}ux has taught me tremendous expertise. He is the leader of my scientific research career and is always encourage me to do a better work by his insightful comments. It is him that helped me to build my inner confidence about scientific research and helped me to find a solid, long-term and sustainable way to move forward. He has set a very good example to me about being a conscientious and professional researcher. I am also very grateful for his  understanding and supports for every important moment during my Ph.D.

I would like to thank my co-supervisor, Prof. Su Ruan, for her nuanced and constructive guidance. Thanks to her strong expertise and experience in medical image analysis and machine learning, my research background has been enhanced to face future challenges. I am impressed by her passion for work and the quality of hardworking. I appreciate her cheerful encouragement when I face difficulties in research. I also appreciate her help and support in my life during my stay in Rouen as well as in Compiègne.

I would also like to thanks all the jury members, Prof. Sylvie Le Hegarat-MASCL from Université Paris-Sud, and Prof. David MERCIER from Université of Artois, Prof. Emmanuel Ramasso from ENSMM School for spending their valuable time to review my thesis manuscript and give me valuable comments. I would like to give a special thanks to Dr. Franck DAVOINE from UTC for his help during my Ph.D. and also for accepting to be the president of jury in my thesis defense. 

I would to thank my colleagues from Heudiasyc in Compiègne with whom I spent a huge amount of time during office hours and beyond. Thanks to all of you for the moments we shared. I would like to say thanks to my colleagues from LITIS Quantif in Rouen for their help during my stay in Rouen. I would also like to thank my friends from China and France, for their warm help when I need. 
 
I would like to gratefully acknowledge the support provided for this thesis under the framework of the Labex MS2T. I would also like to thank the Chinese Scholarship Council for supporting my study in France.

I would like to express great thanks to my family, in particular to my beloved parents, my brother and my sister-in-law, for their continuous support and encouragement throughout the years of my study. 

Last but not least, I would also like to thank myself for dedicating my gold age to finishing this thesis. All efforts are worth it.

\end{acknowledgements}
\cleardoublepage

\begin{publication}
	\addchaptertocentry{\publicationname} 

\textbf{Journal papers}
	\begin{itemize} 
		\item \textbf{Ling Huang}, Su Ruan and Thierry Den{\oe}ux. "Application of belief functions to medical image segmentation: A review." \emph{Information Fusion}, 91 (2023), pp. 737-756,  \url{https://doi.org/10.1016/j.inffus.2022.11.008}.
		
		\item \textbf{Ling Huang}, Su Ruan, Pierre Decazes, Thierry Den{\oe}ux. "Lymphoma segmentation from 3D PET-CT images using a deep evidential network." \emph{International Journal of Approximate Reasoning}, 149 (2022), pp. 39-60, \url{https://doi.org/10.1016/j.ijar.2022.06.007}.
		
		\item \textbf{Ling. Huang}, Su Ruan and Thierry Den{\oe}ux. "Semi-Supervised Multiple Evidence Fusion for Brain Tumor Segmentation." Neurocomputing. (Accepted).
		
	\end{itemize}
	
\textbf{Conferences papers}
	\begin{itemize} 
		\item \textbf{Ling Huang}, Thierry Den{\oe}ux, Pierre Vera, Su Ruan. "Evidence fusion with contextual discounting for multi-modality medical image segmentation." \emph{International Conference on Medical Image Computing and Computer-Assisted Intervention} (MICCAI). Springer, Cham, 2022.
		
		\item \textbf{Ling Huang}, Su Ruan, Pierre Decazes, Thierry Den{\oe}ux. "Evidential segmentation of 3D PET/CT images."
		\emph{International Conference on Belief Functions} (BELIEF). Springer, Cham, 2021.
		
		\item \textbf{Ling Huang}, Su Ruan, and Thierry Den{\oe}ux. "Belief function-based semi-supervised learning for brain tumor segmentation." \emph{18th International Symposium on Biomedical Imaging} (ISBI). IEEE, 2021.
		
		\item \textbf{Ling Huang}, Su Ruan, and Thierry Den{\oe}ux. "Covid-19 classification with deep neural network and belief functions." \emph{The Fifth International Conference on Biological Information and Biomedical Engineering} (BIBE). 2021.
		
	\end{itemize}
	
	\textbf{Workshop paper}
	\begin{itemize} 
		\item \textbf{Ling Huang}, Thierry Den{\oe}ux, David Tonnelet, Pierre Decazes and Su Ruan. "Deep PET/CT fusion with Dempster-Shafer theory for lymphoma segmentation." \emph{18th International Workshop on Machine Learning in Medical Imaging} (MLMI). Springer, Cham, 2021.
	\end{itemize}
	
\textbf{Available codes}
\begin{itemize}
\item The python code using open-source software
library Pytorch and MONAI can be downloaded at \url{https://github.com/iWeisskohl}.
\end{itemize}
 
 
\end{publication}
\cleardoublepage	


\tableofcontents 

\cleardoublepage

\addchaptertocentry{\symbolsname}
\begin{symbols}{llll} 

	$\Omega$ && frame of discernment  \\
	$m$ && mass function  \\
	$Bel$ && belief function  \\
	$Pl$ &&  plausibility function \\
	$pl$ && contour function \\
	$BetP$ &&  Pignistic probability \\
	$\oplus$ && orthogonal sum\\
	$\kappa$ &&degree of conflict&   \\
	$\beta$ && degree of belief that the source mass function is reliable\\
	$^{\beta} m$ &&  discounted mass function \\
	$\bbeta$ && a vector of reliability values   \\
	$m_{?}$&& vacuous mass function \\

	$p_i$ &&  a prototype in an ENN or RBF classifier \\
	$d_i$ && distance between $x$ and prototype $p_i$\\
	$s_i$ &&  distance-based support between $x$ and prototype $p_i$ \\
	$\underline{E}_m(u), \overline{E}_m(u)  $ &&lower, upper expected utilities associated to $m$\\
	$pCNN, mENN$&& segmentation outputs of probability module and mass function module\\
	$S, G$ && output results and ground truth\\
	$x_t$&& the transformed data based on input $x$\\
	$v_i$ && weight of the connection between hidden unit $i$ and output unit \\
	$w_i$ && as weights of evidence for class 1 or 2, depending on the sign of $v_i$\\
	$\lambda$ &&regularization coefficient\\
	$R$&&regularizer\\

\end{symbols}

\cleardoublepage
\mainmatter 
\restoregeometry



\begin{introduction}
\addchaptertocentry{\introductionname}

\section*{Research background and challenges}

In clinical routine, the segmentation or delineation of the target region is performed manually by physicians. However, the manual segmentation operation has limited efficiency and accuracy. First, image segmentation is time-consuming, especially with 3D imagesv. It leads to a waste of medical resources. Second, the segmentation performance is limited by the quality of medical images, the difficulty of the disease, and the domain knowledge of the experts. Thus, the study on automatic medical image segmentation methods with powerful artificial intelligence technologies is necessary to address the problem of limited medical resources and improve the segmentation performance.

Classical medical image segmentation
approaches~\cite{batenburg2009adaptive,kimmel2003fast,salvador2004determining,StroblBias,kleinbaum1996logistic} focus on low-level feature analysis, e.g.,~gray and textual features or
hand-crafted features. These approaches have limitations in terms of segmentation accuracy, efficiency, and reliability, which creates a big gap between experimental performance and clinical application. More recently, the success of deep learning in the medical domain brought a lot of contributions to medical image segmentation
tasks~\cite{ronnebergerconvolutional,myronenko20183d,isensee2018nnu,bahdanau2014neural}. Deep learning solves the segmentation efficiency problem, allowing for large-scale medical image segmentation, and contributes greatly to its accuracy. However, large-scale labeled training data are needed to reach a satisfying segmentation performance. Obtaining precisely annotated data is particularly challenging in the medical image segmentation domain, which has become one of the bottlenecks of learning-based segmentation approaches. 

Despite the excellent performance of deep learning-based medical image segmentation methods, doubts about the reliability of the segmentation results still remain~\cite{hullermeier2021aleatoric}, which explains why their application to therapeutic decision-making for complex oncological cases is still limited. A reliable segmentation model should be well calibrated, i.e., its confidence should match its accuracy. Therefore, a trustworthy representation of uncertainty is desirable and should be considered a key feature of any deep learning method, especially in safety-critical application domains, e.g., medical image segmentation. In general, deep models have two sources of uncertainty: aleatory uncertainty and epistemic uncertainty~\cite{hora1996aleatory,der2009aleatory}. Aleatory uncertainty refers to the notion of randomness, i.e., the variability in an experimental outcome due to inherently random effects. In contrast, epistemic uncertainty refers to uncertainty caused by a lack of knowledge (ignorance) about the best model, i.e., the ignorance of the learning algorithm or decision-maker. As opposed to uncertainty caused by randomness, uncertainty caused by ignorance can be reduced based on additional information or the design of a suitable learning algorithm. In medical image segmentation, the uncertainty can be decomposed into three levels~\cite{lakshminarayanan2017simple}. \emph{Pixel/voxel-level} uncertainty is useful for interaction with physicians by providing additional guidance for correcting segmentation results. \emph{Instance-level} is the uncertainty aggregate by pixel/voxel-level uncertainty, which can be used to reduce the false discovery rate. \emph{Subject-level} uncertainty offers information on whether the segmentation model is reliable. Early approaches to quantify the segmentation uncertainty were based on Bayesian theory~\cite{hinton1993keeping,mackay1992practical}. The popularity of deep segmentation models has revived  research on model uncertainty estimation and has given rise to specific methods such as variational dropout~\cite{gal2016dropout,tran2018bayesian}, and model ensembles~\cite{lakshminarayanan2017simple,rupprecht2017learning}. However, probabilistic segmentation models capture knowledge in terms of a single probability distribution and cannot distinguish between aleatory and epistemic uncertainty, which limits the exploitation of the results. 

Furthermore, single-modality biomedical data does not provide a complete representation of disease information, and information sources are usually imperfect. The advances in medical imaging machines and technologies now allow us to obtain medical images with several modalities, such as Magnetic Resonance Imaging (MRI)/Positron Emission Tomography (PET), multi-sequence MRI, or PET/Computed Tomography (CT). The different modalities can provide complementary information about cancer and other abnormalities in the human body. Thus, the fusion of multiple information is vital to improve the accuracy of diagnosis and help radiotherapy. The success of information fusion depends on how well the fused knowledge represents reality \cite{white1991data}. More specifically, it depends on how adequate the input information is, how accurate and appropriate prior knowledge is, and on the quality of the uncertainty model used \cite{rogova2004reliability}. Though deep learning has shown promising performance in information representation, modeling imperfect data and combining unreliable sources need further study.

In this thesis, we address medical image segmentation using belief function theory (BFT) \cite{dempster1967upper,shafer1976mathematical,denoeux20b} and deep neural networks, with a focus on uncertainty quantification and information fusion. BFT is a theoretical framework for modeling, reasoning with and combining imperfect (imprecise, uncertain, and partial) information. With BFT, we can quantify epistemic uncertainty directly and further explore the possibility of improving the model reliability based on the quantified uncertainty. Besides uncertainty quantification, Dempster's rule of BFT offers a way to fuse multiple uncertain information, e.g., multimodal medical images, to achieve more promising segmentation performance \cite{lian2018joint, ghesu2021quantifying}. Motivated by the high expressivity of BFT, the goals of this thesis are 1) to develop evidential medical image segmentation models in the framework of BFT and deep neural networks, with the cooperation of powerful learning algorithms such as semi-supervised learning, 2) to study the performance of deep evidential segmentation with uncertainty quantification and 3) to study the effectiveness of BFT for constructing deep multimodal segmentation model.

\section*{Contributions of the thesis}
The contributions of this thesis are organized in three parts:
\begin{itemize}
\item \textbf{Semi-supervised medical image segmentation}: The main idea of this part is to address the annotation limitation with the design of semi-supervised learning and decrease the uncertainty caused by the lack of annotations with evidential segmentation and evidence fusion. We first propose a semi-supervised learning algorithm based on an image transformation strategy by producing pseudo labels for unannotated data. Then a probabilistic-based deep neural network and a BFT-based evidential neural network are used to compute segmentation results with uncertainty quantified by probabilities and mass functions. Finally, these probabilities and mass functions are combined by Dempster's rule. 

\item \textbf{Uncertainty quantification in medical image segmentation}:
A reasonable and reliable quantification of segmentation uncertainty is important to optimize the segmentation framework and further improve performance. In this part, an automatic evidential segmentation model based on BFT and deep learning is proposed to segment lymphomas from 3D PET-CT images, which not only focuses on lymphoma segmentation accuracy but also on uncertainty quantification using belief functions. The model is composed of a deep feature-extraction module and an evidential layer. The feature extraction module uses an encoder-decoder framework to extract semantic feature vectors from 3D inputs. The evidential layer then uses prototypes in the feature space to compute a belief function at each voxel, quantifying the uncertainty about the presence or absence of lymphoma at this location. Two evidential layers are compared based on different ways of using distances to prototypes for computing mass functions.
    
\item \textbf{Multimodal medical image segmentation with contextual discounting}: Different modality biomedical data have different reliabilities to segment specific diseases. Thus, quantifying context-based modality-level uncertainty and combining multimodal information is essential to reach a reliable decision and explain the decision. In this part, we construct a multi-MR image brain tumor segmentation framework by modeling the reliability of each source of MR images when doing different segmentation tasks using mass functions and contextual discounting. Different modality MR images are taken as an independent source of input. One feature extraction and one evidential segmentation module are used for each modality input to assign each voxel a mass function. Then a contextual discounting layer is designed to take into account the reliability of the single modality  MR images when classifying different diseases. Finally, the discounted evidence from different modalities is combined by Dempster's rule to obtain a final segmentation. 

\end{itemize}

\section*{Layout of the thesis}
This thesis is structured in two parts and seven chapters:
\begin{itemize}
\item Part I introduces the theoretical background and general context that supports the thesis. Chapter \ref{Chapter1} first introduces medical image segmentation and its significance in clinical treatment and then introduces classical deep segmentation methods and their limitations. Chapter \ref{Chapter2} recalls the fundamentals of belief function theory. It first describes how imperfect (uncertain and imprecise) information can be modeled and combined in the framework of BFT, and then introduces the basic belief assignment methods that generate mass function for medical image segmentation. Chapter \ref{Chapter3} comprehensively reviews existing medical image segmentation methods using BFT.

\item Part II is devoted to our three main contributions. Chapter \ref{Chapter4} introduces the semi-supervised evidence fusion model with the application of MRI brain tumor segmentation. Chapter \ref{Chapter5} describes the deep evidential segmentation framework to quantify segmentation uncertainty with 3D PET-CT lymphoma segmentation application. Chapter \ref{Chapter6} presents the contextual discounting-based multimodal medical image fusion model with the application of multi-MRI brain tumor segmentation. Finally, we conclude our work and give some perspectives for future work in Chapter \ref{Chapter7}.
\end{itemize}

\end{introduction}


\part{Theoretical Background and General Context}

\chapter{Medical image segmentation}
\label{Chapter1}


\tocpartial


\section{Introduction to medical image segmentation}
Medical image segmentation is a critical step in radiotherapy planning, where organs at risk and tumors must be precisely located in images. The segmentation of medical images involves the extraction of regions of interest (ROI) from 2D/3D image data (e.g., pathological or optical imaging with color images, MRI, PET, and CT scans). The main goal of segmenting medical images is to identify areas of organs, cancer, and other abnormalities in the human body, for example, brain tumor, lymphomas, the interior of the human body (such as lungs, spinal canal, and vertebrae), and skin and cell lesions. 

\subsection{Common medical imaging techniques}
\begin{figure}
    \centering
    \includegraphics[width=\textwidth]{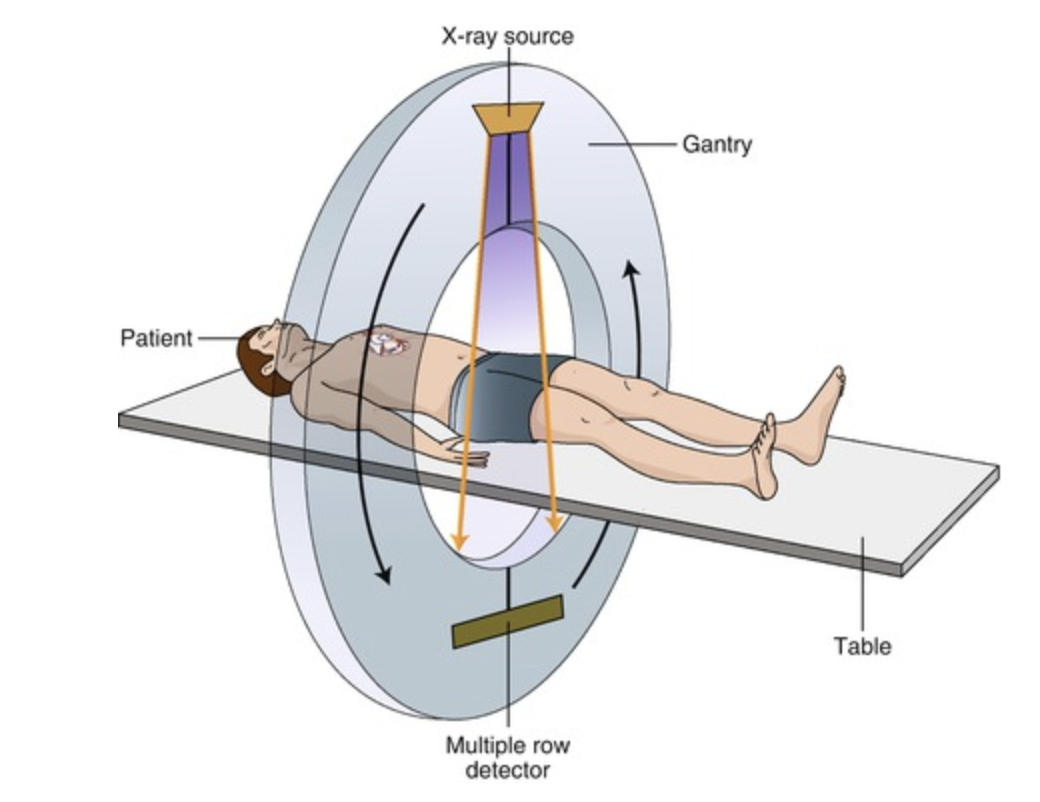}
    \caption{A depiction of a typical CT scanner \cite{pelberg2015basic}. The X-ray source and row detectors are housed within the gantry, which rotates around the patient (helical scanning technique). The cone-shaped X-ray beams emanate from the tube source on one side of the gantry, pass through the patient and terminate at the detectors on the other side of the gantry.}
    \label{fig:ct_scaner}
\end{figure}

Medical imaging is playing an increasingly central role in radiotherapy practice, thanks to the advances in imaging techniques. It influences the effectiveness of radiotherapy protocols, including the delineation of radiation target volumes and adjacent normal tissues, the design of the radiation dose distribution, the monitoring of treatment response, etc. A treatment that adopts advanced imaging technology to reduce uncertainties and assist decision-making during a course of treatment is referred to as image-guided radiotherapy. In this section, we introduce three commonly used medical imaging technologies: Computed Tomography (CT), Positron Emission Tomography (PET), and Magnetic resonance imaging (MRI).
\paragraph{Computed Tomography (CT)} CT is a gold standard image modality in radiation oncology, and its imaging devices are widely available at almost all cancer centers. CT images can provide anatomical information to show the geometric positions of the target tumor and adjacent organs at risk. CT imaging is based on the measurement of the X-rays attenuation between the source and the detector. The principle of CT imaging is briefly described in Figure \ref{fig:ct_scaner}. The use of CT in radiotherapy allows three-dimensional dose calculation and optimization.

\begin{figure}
    \centering
    \includegraphics[width=\textwidth]{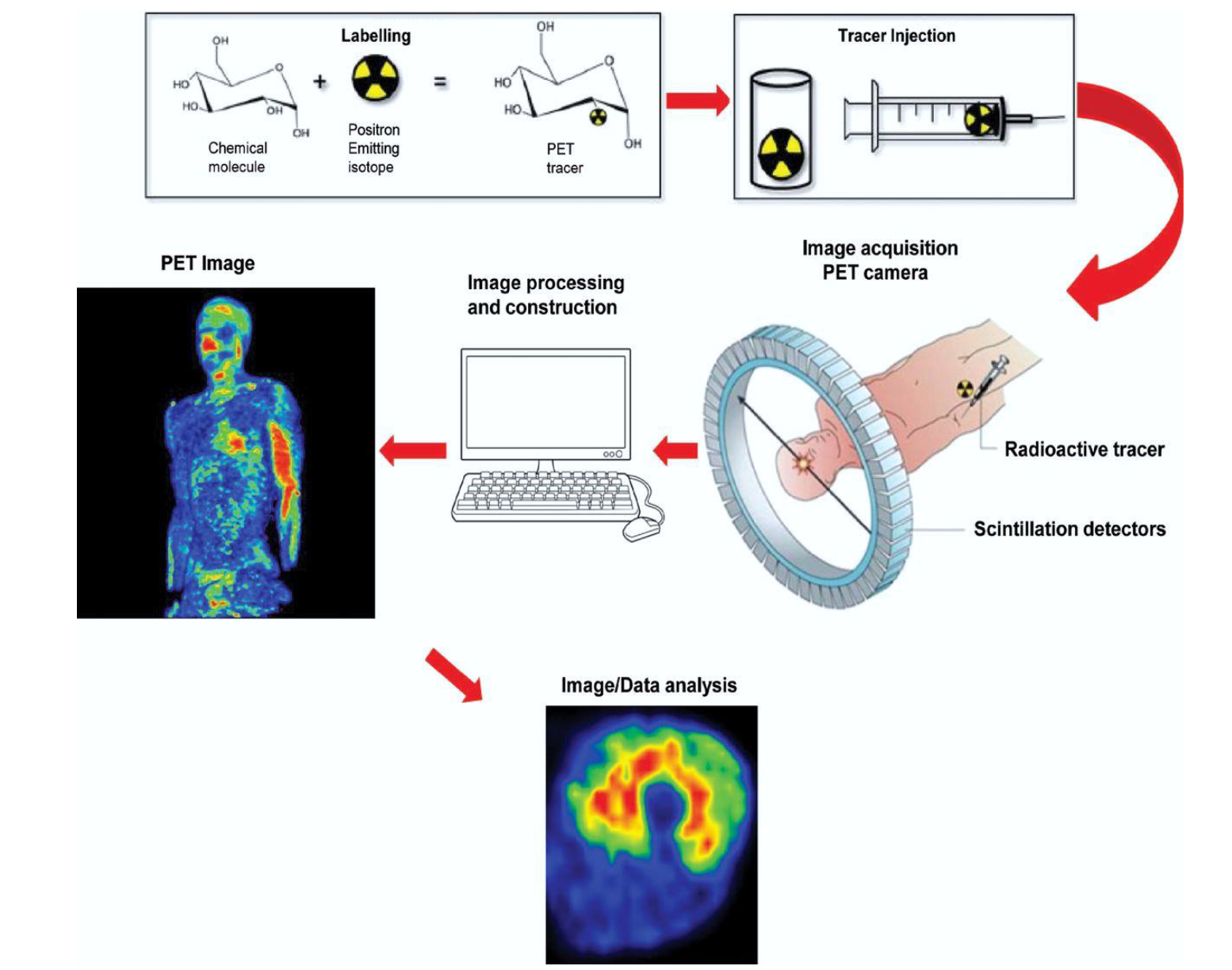}
    \caption{The fundamental principles of PET imaging \cite{rudroff201518f}. PET tracer is first selected, and the information, either the tracer's kinetics within the tissues with dynamic imaging or the spatial distribution of the tracer with static imaging, is then detected by scintillation detectors of the PET scanner and is used to reconstruct images.}
    \label{fig:pet_scaner}
\end{figure}

\begin{figure}
    \centering
    \includegraphics[width=\textwidth]{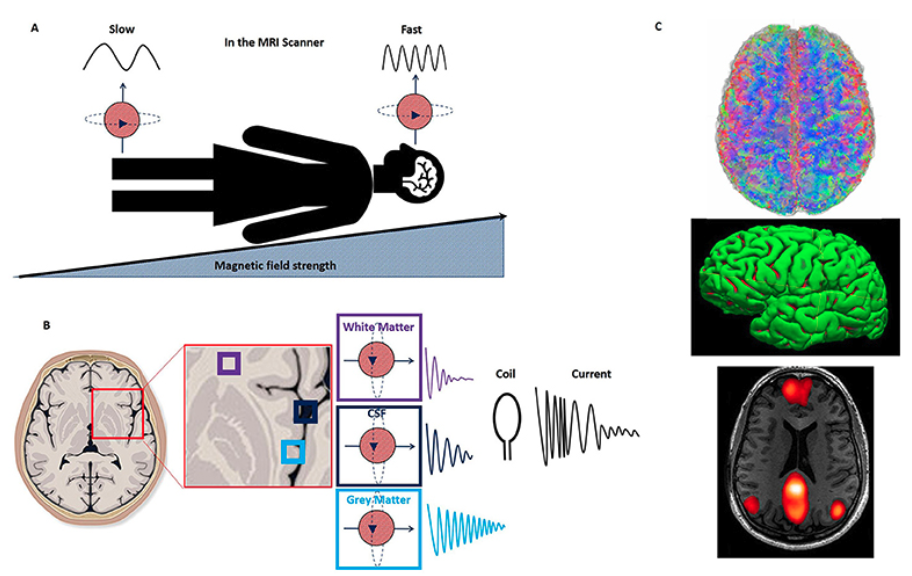}
    \caption{The physics of MRI and its revelation of the brain \cite{broadhouse2020physics}. (A) An example of an MRI scanner with a B1 field (magnetic field strength varies across the body), which increases across the body, from foot to head. Hydrogen protons in the head will then be spinning faster than those in the feet, allowing us to define the spatial position according to the frequencies. (B) Different tissues of a brain scan, such as white matter and gray matter (C) Examples of different brain images: the structural connection of the brain via white matter—the information (top right), the volume of gray matter regions of the brain (middle right), and the functional connection and communication of the brain (bottom right).}
    \label{fig:mri_scaner}
\end{figure}

\paragraph{Positron Emission Tomography (PET)} PET is a functional imaging technique used in nuclear medicine that can measure tissue metabolic activity in vivo through an injected radioactive tracer. Complementary to CT, PET can provide critical functional information on target tissues or tumors for more precisely guarding the procedure of radiotherapy. The principle of PET imaging is briefly described in Figure \ref{fig:pet_scaner}. With different radioactive tracers, PET imaging makes it possible to monitor the different functional activities of a target tumor (e.g., metabolism, proliferation, and oxygen delivery) on a molecular scale. The most commonly used radioactive tracer in clinical oncology practice is fluorine-18 (F-18) fluorodeoxyglucose (FDG). PET scanning with FDG, i.e., FDG-PET, can highlight tumor tissues with high metabolic rate. It has been widely used for diagnosis, staging, and re-staging of most cancers, such as non-small cell lung cancer, esophageal carcinoma, or lymphomas, etc.
\paragraph{Magnetic resonance imaging (MRI)} MRI is a widely used technique to generate high-resolution and high-contrast medical images of soft tissues. With MRI, we can see inside the human body in detail by using magnets and radio to map the location of the water and then use this information to generate a detailed image. Figure \ref{fig:mri_scaner} gives an introduction to the physics of MRI and its revelation of the brain. 
Different protocols or different MRI sequences defined by changing the repetition time (TR) and the echo time (TE) can be used to generate different modalities of MR images. For example, tissue can be characterized by two different relaxation times: T1 (longitudinal relaxation time) and T2 (transverse relaxation time). T1 determines the rate at which excited protons return to equilibrium in the direction of the main magnetic field, and T2 determines the rate at which excited protons reach equilibrium in the direction perpendicular to the main magnetic field or go out of phase with each other. The most common MRI sequences are T1-weighted and T2-weighted scans. T1-weighted images are produced by using the short TE and RT. This image weighting is useful for obtaining morphological information such as the cerebral cortex, fatty tissue, and focal liver lesions. In contrast, T2-weighted images are produced by using longer TE and TR. This image weighting is useful for detecting edema and inflammation, revealing white matter lesions, and assessing zonal anatomy in the prostate and uterus. Fluid-attenuated inversion recovery (FLAIR) is an MRI sequence with an inversion recovery set to null fluids. The FLAIR sequence is similar to a T2-weighted image except that the repetition time and TR times are very long. The FLAIR sequence is very sensitive to pathology and makes the differentiation between Cerebrospinal fluid (CSF) and an abnormality much easier.
Radiotherapy is one of the five principal methods used in the clinical treatment of malignant tumors, the other methods being surgery, chemotherapy, immunotherapy, and hormonal therapy. Accurate tumor segmentation in medical images is a critical step for diverse objectives in clinical oncology, including reliable diagnosis and tumor staging, as well as solid radiotherapy planning. Thanks to the sustained advancement of medical imaging techniques, as well as progress made in medical image analysis, the effectiveness of radiotherapy for cancer treatment is being continuously improved.

\subsection{Significance of automatic medical image segmentation}
In the practical process of radiotherapy planning, the delineation of target tumor volumes is usually carried out manually by experienced clinicians with a computer interface. However, manual segmenting of medical images is time-consuming, labor-intensive, and operator-dependent. Moreover, the segmentation performance is sensitive to the operator's variability and may lead to imprecise and unreliable tumor contours. Recent advances in machine learning techniques, especially deep learning, make it possible to perform automatic segmentation with promising performance. In the rest of this chapter, we first briefly recall deep learning-based medical image segmentation methods in Section \ref{sec: deep learning}. Then we introduce three deep segmentation models that be used as the baseline models in this thesis in Section \ref{sec: baseline model}. Finally, in Section \ref{sec: conclusion}, we give a conclusion of existing deep learning-based segmentation methods and explain the advantages of using BFT with deep neural network in medical image segmentation.

\section{Deep learning approaches to medical image segmentation}
\label{sec: deep learning}
Early image segmentation methods used the information provided by the image itself, e.g.,~gray, textual, contrast, and histogram features, as well as segmenting ROI based on threshold~\cite{batenburg2009adaptive}, edge detection~\cite{kimmel2003fast}, graph partitioning~\cite{onoma2014segmentation}, clustering~\cite{salvador2004determining}, etc. More recently, researchers have been interested in hand-crafted features, e.g.,~Scale Invariant Feature Transform (SIFT)~\cite{lowe1999object}, Features from Accelerated Segment Test (FAST)~\cite{rosten2006machine}, and Geometric hashing~\cite{mian2006three}, as well as segmenting ROI using machine learning-based methods such as support vector machine (SVM)~\cite{suykens1999least}, random forest (RF)~\cite{StroblBias}, logistic regression (LR)~\cite{kleinbaum1996logistic}, etc. Those methods have attracted great interest for a while, but their accuracy cannot meet clinical application requirements because only low-level or middle-level features are considered.

\begin{figure}
    \centering
    \includegraphics[width=\textwidth]{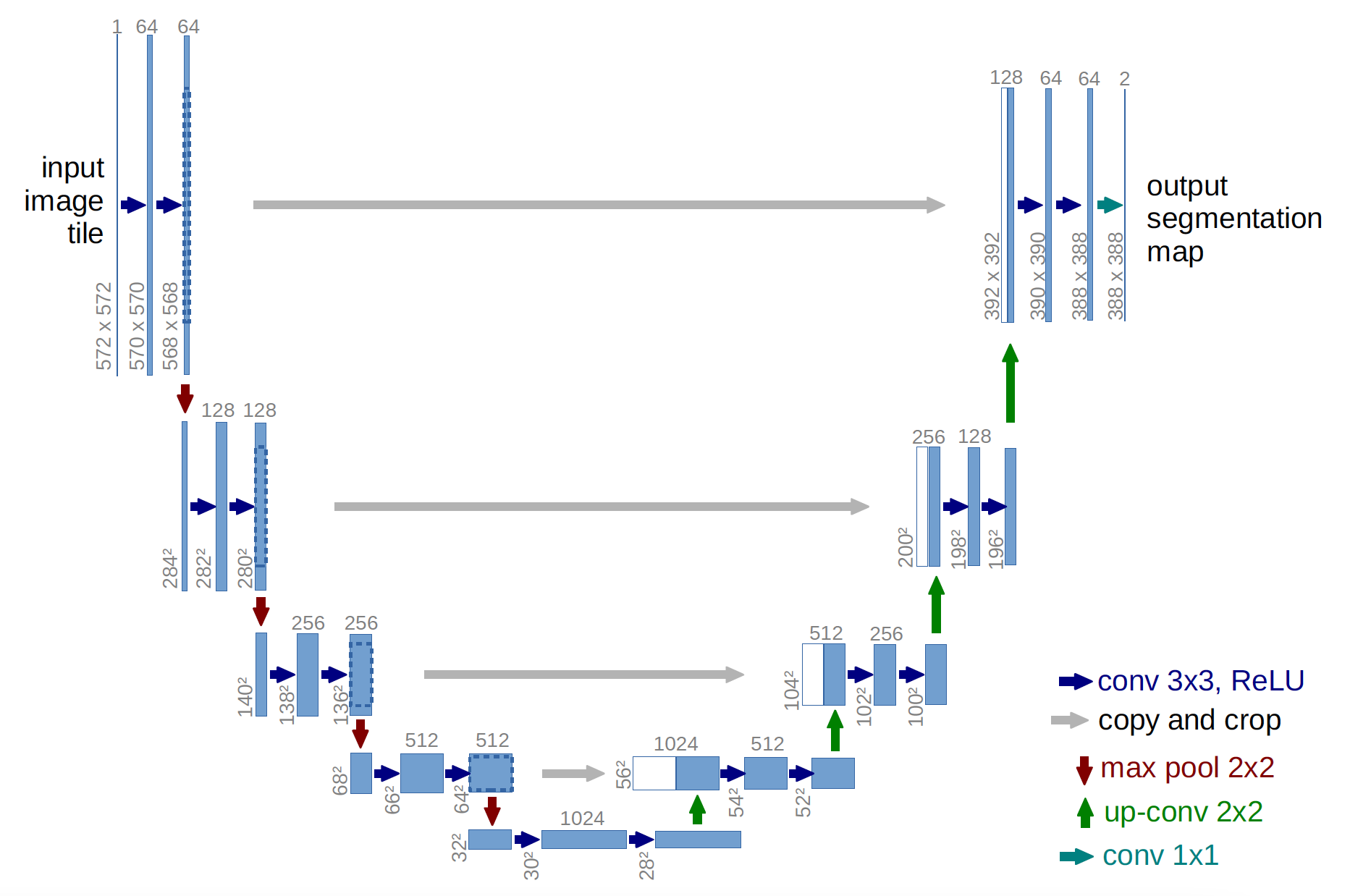}
    \caption{Example of an UNet architecture \cite{ronnebergerconvolutional}. Each blue box corresponds to feature maps ($32\times 32$ in the lowest resolution). The number of channels in the feature maps is denoted at the top of the box. The width and height of the convoluted feature maps are provided at the lower-left edge of the box. White boxes represent copied feature maps from the dotted box.}
    \label{fig:unet}
\end{figure}

Deep learning, a sub-field of machine learning concerned with neural networks, has achieved significant achievements in computer vision tasks by its ability to represent high-level semantic features \cite{Krizhevsky2012ImageNet, bai2018optimization}, as well as in medical image segmentation domain \cite{ciresan2012deep, seyedhosseini2013image, hariharan2015hypercolumns, long2015fully, ronnebergerconvolutional}. In \cite{ciresan2012deep}, Ciresan et al. proposed a pixel-level classification network for electron microscopy (EM) image segmentation. This network won the competition of EM segmentation challenge at ISBI 2012 by a large margin. In \cite{seyedhosseini2013image}, Seyedhosseini et al. proposed a multiresolution contextual model called the cascaded hierarchical model (CHM). In \cite{hariharan2015hypercolumns}, Hariharan et al. used hypercolumns as pixel descriptors to make the best use of semantic context and allow precise location. Long et al. \cite{long2015fully} were the first authors to show that a fully convolutional network (FCN) \cite{long2015fully} could be trained end-to-end for semantic segmentation, exceeding the state-of-the-art when the paper was published in 2015. UNet \cite{ronnebergerconvolutional}, a successful modification and extension of FCN, has become the most popular model for medical image segmentation in recent years. Figure \ref{fig:unet} shows an example framework of UNet with a down-sampling path and an up-sampling path. Based on UNet, research for deep learning-based medical image segmentation can be summarized in two major directions: the design of the segmentation model and the optimization (loss function).

\subsection{Deep segmentation model}
Based on UNet, there has been many studies on deep segmentation networks with an encoder-decoder architecture (e.g., 3D UNet \cite{cciccek20163d}, Deep3D-UNet \cite{zhu2018anatomynet}, V-Net \cite{milletari2016v}, SegResNet \cite{myronenko20183d}). Recently, some popular approaches, such as attention mechanism \cite{bahdanau2014neural} and transformer mechanism \cite{carion2020end, han2021transformer} have achieved promising performance with deep neural networks (e.g., Attention UNet \cite{oktay2018attention, trebing2021smaat} and Transformer-UNet \cite{cao2021swin, hatamizadeh2022unetr, hatamizadeh2022swin}), and have also been applied to medical image segmentation. Although the design of deep segmentation models with a research focus on the model structure has achieved great success, there are still some doubts about whether the so-called innovation of the network structure over the years is really just overfitting \cite{isensee2021nnu}, and some researchers suggest that more attention should be paid not only to the structure, but also to other aspects such as training and inference strategies. Thus, deep analysis of the segmentation performance, e.g., segmentation accuracy and uncertainty, is necessary to achieve a better segmentation performance.

\subsection{Model optimization}
\label{subsec: optimization}
\begin{figure}
    \centering
    \includegraphics[width=\textwidth]{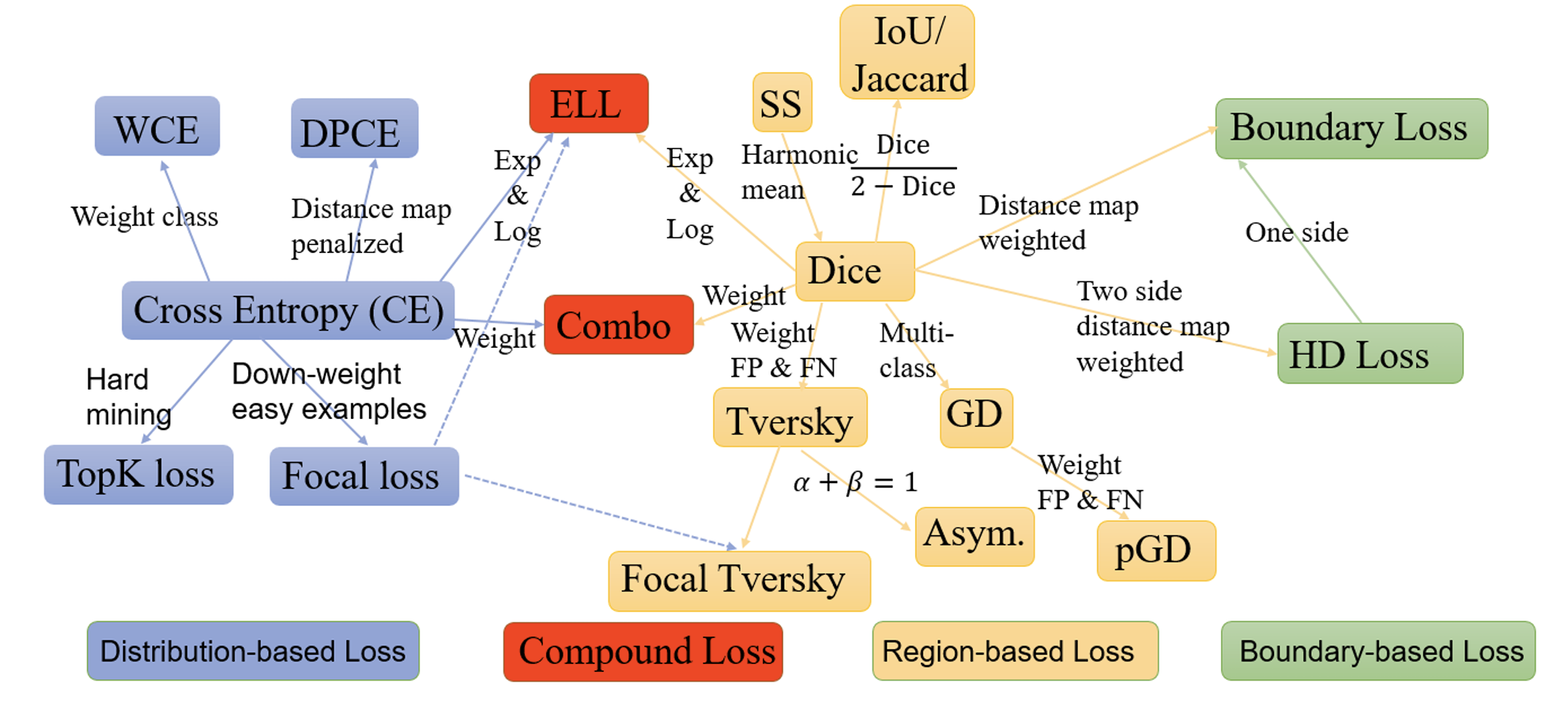}
    \caption{Overview of loss functions for medical image segmentation \cite{ma2020segmentation}.}
    \label{fig:loss}
\end{figure}

Model optimization consists of designing loss functions relevant to the problems to be solved. Different from nature image segmentation, the segmented objects in a medical image are usually tiny compared with the background, which causes the unbalanced label distribution problem. The specific design of loss functions is one of the crucial ingredients in deep learning-based medical image segmentation. The existing loss functions for the deep segmentation model can be classified into four categories: distribution-based loss, region-based loss, boundary-based loss, and compound loss. Figure \ref{fig:loss} shows the overview of loss functions for medical image segmentation. The distribution-based loss aims to minimize the dissimilarity between two distributions. Cross entropy (CE) loss is the fundamental function. Based on this, a lot of variants, such as Weighted CE loss \cite{ronnebergerconvolutional}, Focal loss \cite{lin2017focal}, distance penalized CE loss \cite{caliva2019distance}, and CE with KL divergence \cite{zhou2022tri} are used for medical image segmentation. The region-based loss functions aim to minimize the mismatch or maximize the overlap regions between ground truth and predicted segmentation. The Dice loss \cite{milletari2016v} is the most commonly used region-based loss function because it directly optimizes the Dice coefficient. Based on Dice loss, there are a lot of variants, e.g., IoU loss (also called Jaccard loss) \cite{rahman2016optimizing}, Tversky loss \cite{salehi2017tversky}, Generalized Dice loss \cite{sudre2017generalised}, ContrastiveLoss \cite{chen2020simple}, etc. The recently-proposed boundary-based loss functions aim to minimize a distance between ground truth and predicted segmentation. Representative examples are Boundary (BD) loss \cite{ma2020segmentation} and Hausdorff Distance (HD) loss \cite{karimi2019reducing}. 

In addition to the above three categories of loss functions, compound loss functions are popular and efficient in medical image segmentation; they consist in combining different types of loss functions, i.e., the weighted sum between weighted CE and Dice loss. A general summary can be obtained from the existing literature that mildly imbalanced problems are well handled by Dice loss or generalized Dice loss \cite{sudre2017generalised}. Highly imbalanced segmentation tasks are much more complicated and require more robust loss functions. More details about loss functions for deep segmentation can be found in \cite{ma2020segmentation}.

\subsection{Evaluation criteria}
\label{subsec: evaluation cri}
\begin{figure}
\centering
\includegraphics[width=0.5\textwidth]{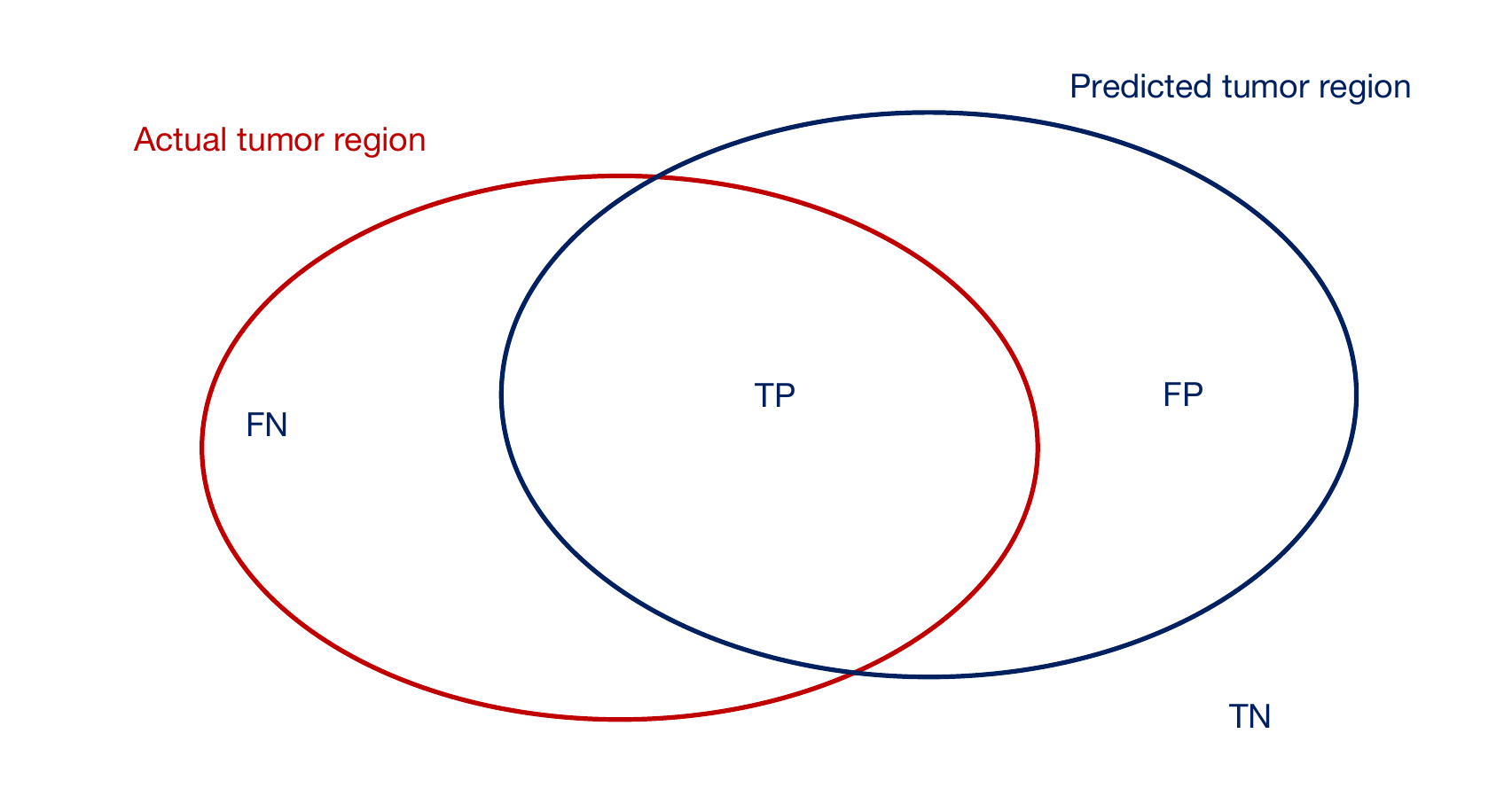}
\caption{Interpretation of the true positive (TP), false positive (FP), true negative (TN), and false negative (FN) used for the definition of evaluation criteria. \label{fig:criteria}}
\end{figure}

Dice score, Sensitivity, Precision, and Hausdorff distance (HD) are the most commonly used evaluation criteria to assess the quality of deep medical image segmentation methods; they are defined as follows:
\begin{equation}
    \textsf{Dice}(P,T)=\frac{2\times TP}{FP+2\times TP+FN}, 
    \label{eq: dice_score}
\end{equation}
\begin{equation}
    \textsf{Sensitivity}(P,T)=\frac{TP}{TP+FN},
    \label{eq: sensi}
\end{equation}
\begin{equation}
    \textsf{Precision}(P,T)=\frac{TP}{TP+FP},
    \label{eq: precision}
\end{equation}
where $P$ is the number of real positive voxels in the data, $N$ is the number of real negative voxels in the data; $TP$, $FP$, and $FN$ denote the numbers of true positive, false positive, and false negative voxels, respectively (See Figure \ref{fig:criteria}), and 

\begin{equation}
\textsf{HD} = 
\max\left (\max_{i\in S}\min_{j\in G} d(i,j),\max_{j \in G}\min_{i \in S} d(i,j)\right ),
\label{eq:hds}
\end{equation}
where $G$ denotes the actual tumor region, $S$ denotes the segmented tumor region, and $d$ represents the distance of voxels between $S$ and $G$.

\begin{figure}
    \centering
    \includegraphics[width=0.5\textwidth]{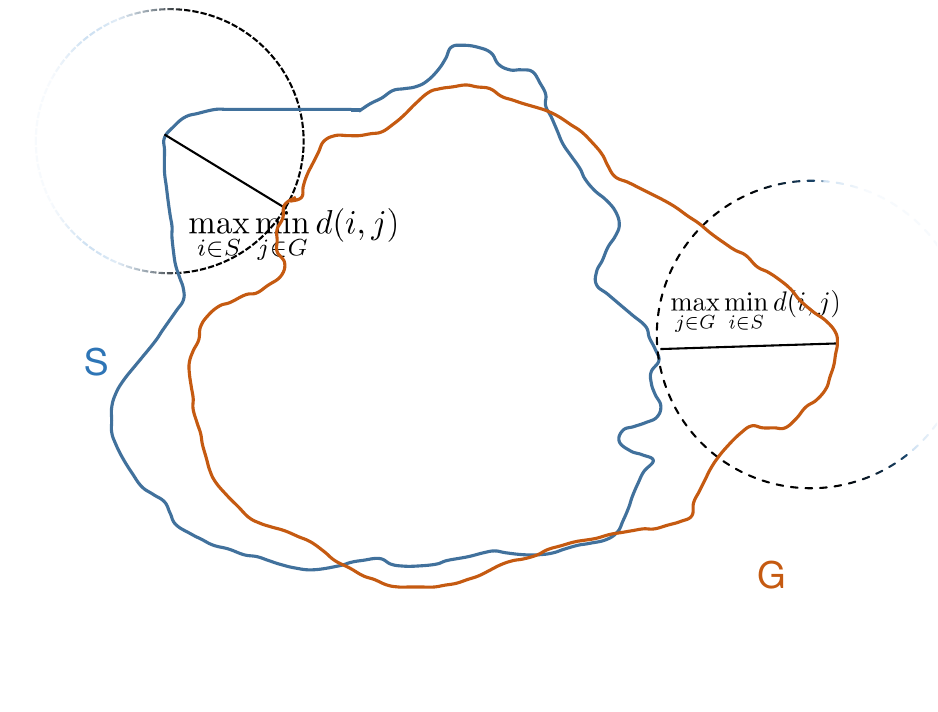}
    \caption{The calculation of the Hausdorff distance between the blue region $S$ and the orange region $G$.}
    \label{fig: hd_dis}
\end{figure}

The Dice score is a global measure of segmentation performance. Sensitivity is the proportion, among actual tumor voxels, of voxels correctly predicted as tumors. Precision is the proportion, among predicted tumor voxels, of voxels that actually belong to the tumor region; it is, thus, an estimate of the probability that the model is correct when it predicts that a voxel is in a tumor region. We note that neither sensitivity nor precision is global performance criteria. We can increase sensitivity by predicting the tumor class more often (at the expense of misclassifying a lot of background pixels), and we can increase precision by being very cautious and predicting the tumor class only when it has a high probability (at the expense of missing a lot of tumor voxels). These two criteria, thus, have to be considered jointly. Finally, we can also remark that a fourth criterion can also be defined: specificity, which is the proportion, among background voxels, of voxels, correctly predicted as background (i.e., $TN/(TN+FP)$). However, as there are much more background voxels than tumor ones, this criterion is not informative in tumor segmentation applications (it is always very close to 1). The Hausdorff distance measures how far two subsets of a metric space are from each other with the definition in \eqref{eq:hds}. More precisely, the Hausdorff distance is the greatest of all the distances from a point in one set to the closest point in the other set. Figure \ref{fig: hd_dis} shows an example of the calculation of the Hausdorff distance. A more comprehensive introduction to the evaluation criteria for medical image segmentation can be found in \cite{taha2015metrics}.


\section{Three baseline deep segmentation models}
\label{sec: baseline model}
In this section, we describe three deep segmentation networks: Residual-UNet \cite{kerfoot2018left}, MFNet \cite{chen2018multi}, and No-New-UNet \cite{isensee2021nnu}, which will be used in the rest of the thesis as the baseline models.

\subsection{Residual-UNet} 
\label{subsec:Unet}
\begin{figure}
\includegraphics[width=\textwidth]{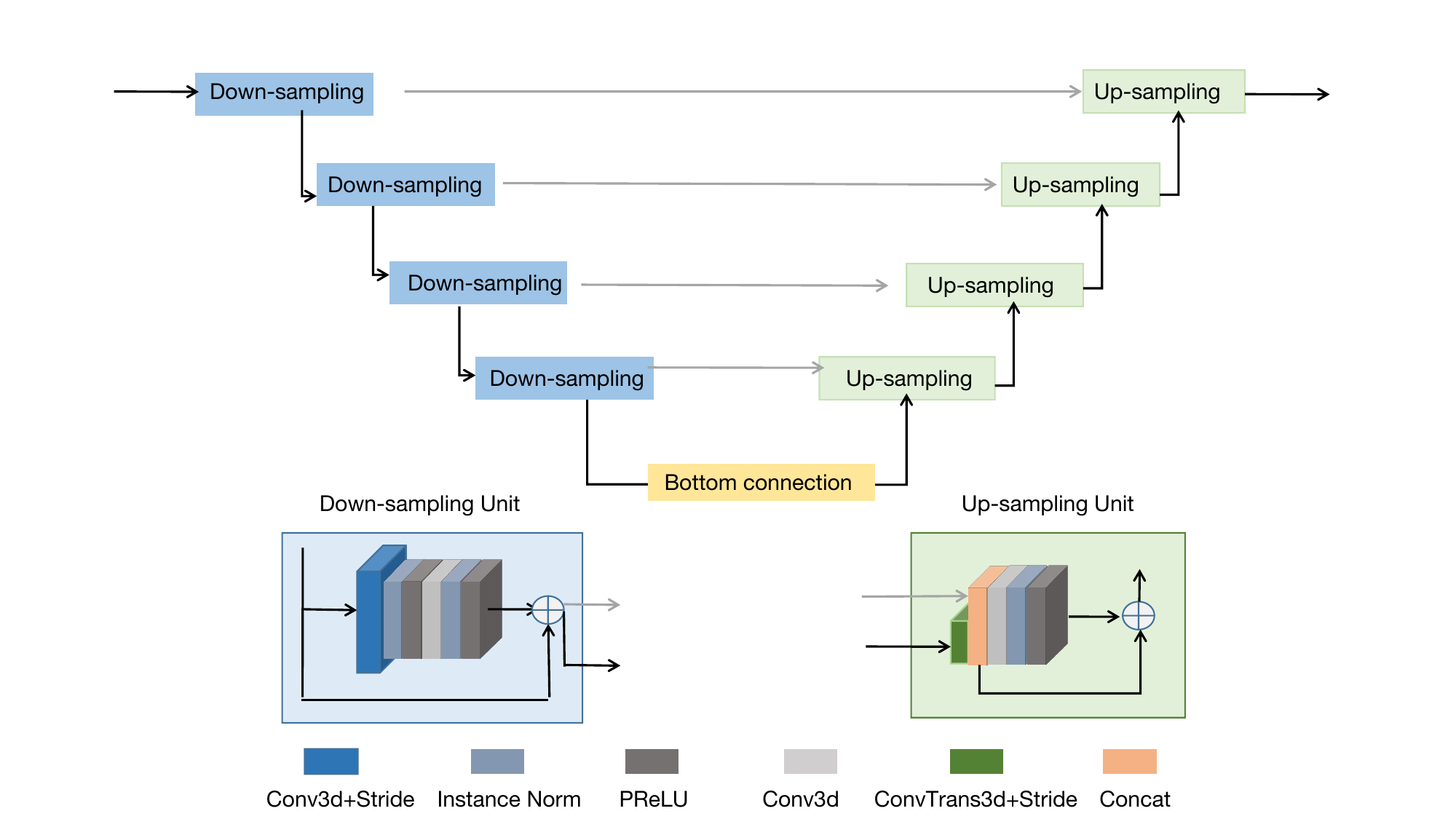}
\caption{Framework of Residual-UNet. Each Down-sampling layer (marked in blue) comprises convolution, normalization, dropout, and activation blocks. Each Up-sampling layer (marked in green) comprises transpose convolution, normalization, dropout, and activation blocks. The last layer (marked in yellow) is the bottom connection which does not down/up sample the data.}
\label{fig-unet}
\end{figure}
Residual-UNet \cite{kerfoot2018left} is a commonly used deep segmentation network with encoder and decoder layers defined using residual units. Data in the encoding path is down-sampled with convolution operation, and the decoding path is up-sampled using transpose convolution operation. Fig.~\ref{fig-unet} gives an example of Residual-UNet. Each Down-sampling layer (marked in blue) comprises convolution, normalization, dropout, and activation blocks, as well as a residual connection. Each Up-sampling layer (marked in green) comprises transpose convolution, normalization, dropout, activation blocks, and residual connection. The last layer (marked in yellow) is the bottom connection that does not down/up sample the data. The Residual-UNet will be used as the feature extraction module of the deep evidential segmentation model in Chapter \ref{Chapter5} and the multimodal medical image segmentation model in Chapter \ref{Chapter6}. 

\subsection{MFNet} 
\label{subsec:dmfnet} 
\begin{figure}
    \centering
    \includegraphics[width=\textwidth]{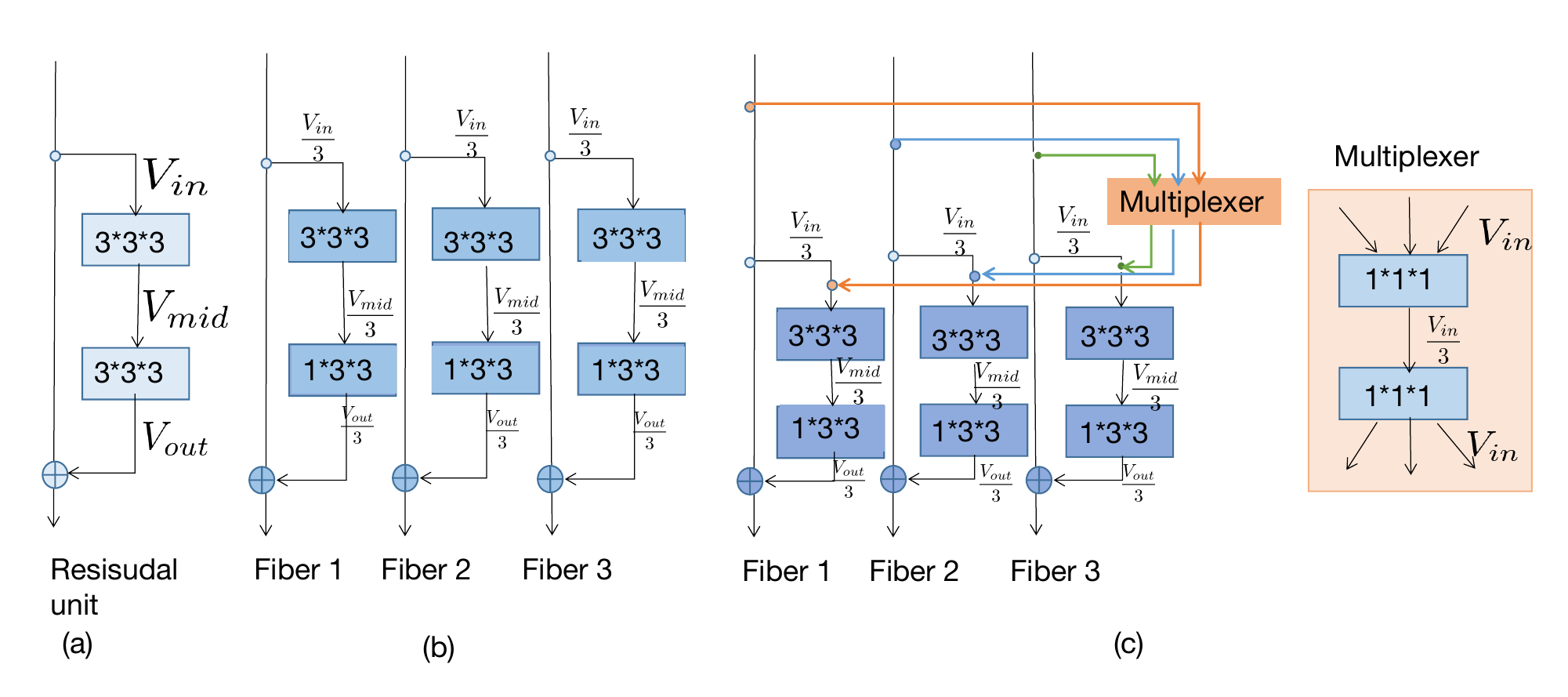}
    \caption{Illustration from residual unit to multi-fiber unit. (a) A single fiber with residual unit composed of two $3 \times 3 \times 3$ convolution layers \cite{he2016deep}. (b) Multi-fiber (MF) units \cite{chen2018multi}, 3 fibers for example. (c) Multi-fiber units with a multiplexer for transferring information across separated fibers \cite{chen2018multi}. Let $V_{in}$, $V_{mid}$, and $V_{out}$ denote the number of input channels, middle channels, and output channels, respectively. The total number of connections for the residual unit and multi-fiber units is $V_{in} \times V_{mid} + V_{mid} \times V_{out}$ and $(V_{in} \times V_{mid} + V_{mid} \times V_{out})/M$, respectively.}
    \label{fig:multi-fiber}
\end{figure}
In \cite{chen2018multi}, Chen et al. proposed a multi-fiber network (MFNet) whose framework is similar to UNet but can reduce by more than half the computation cost compared with UNet. It uses multi-fiber units instead of the residual units to construct the encoder-decoder module. As shown in Figure \ref{fig:multi-fiber} (a), the conventional residual unit uses two $3\times 3\times 3 $ convolutional layers to learn features, which is straightforward but computationally expensive. The multi-fiber unit is illustrated in Figure \ref{fig:multi-fiber} (b). It slices the residual unit into $M$ parallel and separated paths (called fibers here). The connection number is then reduced by the slicing parameter $M$. However, the slicing operation isolates each path from the others and blocks information flow across them, resulting in limited learning capacity for information representation. Thus, the authors propose a multiplexer layer (see Figure \ref{fig:multi-fiber} (c)) that acts as a router that operates across fibers. The multiplexer layer first gathers features from all fibers using a $1 \times 1 \times 1$ convolution layer and then redirects them to specific fibers using the following $1 \times 1 \times 1$ convolution layer. The MFNet unit will be used as the feature extraction module of the semi-supervised medical image segmentation model in Chapter \ref{Chapter4}. 

\subsection{No-New-UNet (nnUNet)} 
\label{subsec: nnunet}
\begin{figure}
    \centering
    \includegraphics[width=\textwidth]{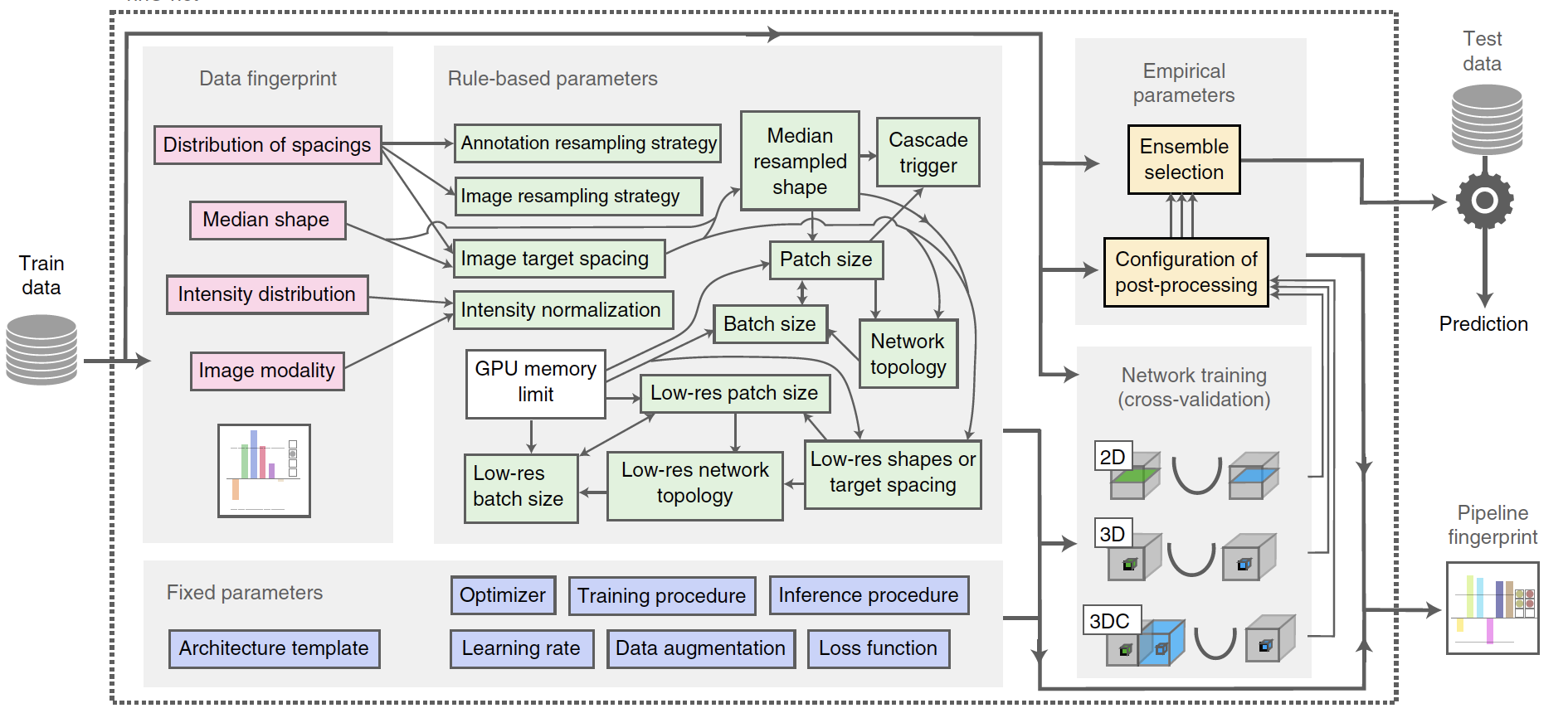}
    \caption{Proposed automated configuration by nnUNet \cite{isensee2021nnu}. Dataset properties are summarized in a  ``dataset fingerprint''. A set of heuristic rules operates on this fingerprint to infer the data-dependent hyperparameters of the pipeline. These are completed by blueprint parameters and the data-independent design choices to form ``pipeline fingerprints''. Three architectures are trained based on these pipeline fingerprints in a 5-fold cross-validation. Finally, nnU-Net automatically selects the optimal ensemble of these architectures and performs post-processing if required.}
    \label{fig:nnunet}
\end{figure}

nnUNet \cite{isensee2021nnu} is the first segmentation model designed as a segmentation pipeline for any given dataset and is a plug-and-play tool for state-of-the-art biomedical segmentation. Instead of designing a new encoder-decoder structure, the authors focus on studying a recipe that systematizes the configuration process on a task-agnostic level and drastically reduces the search space for empirical design choices by a set of fixed parameters, interdependent rules, and empirical decisions. Due to its modular structure, new architectures or algorithms can easily be integrated into nnUNet. Figure \ref{fig:nnunet} is the proposed automated configuration by nnUNet. The nnUNet will be used as another feature extraction module of the multimodal medical image segmentation model in Chapter \ref{Chapter6}.
\section{Conclusion }
\label{sec: conclusion}
\begin{figure}
\centering
\subfloat[\label{fig:demp_0}]{\includegraphics[width=0.5 \textwidth]{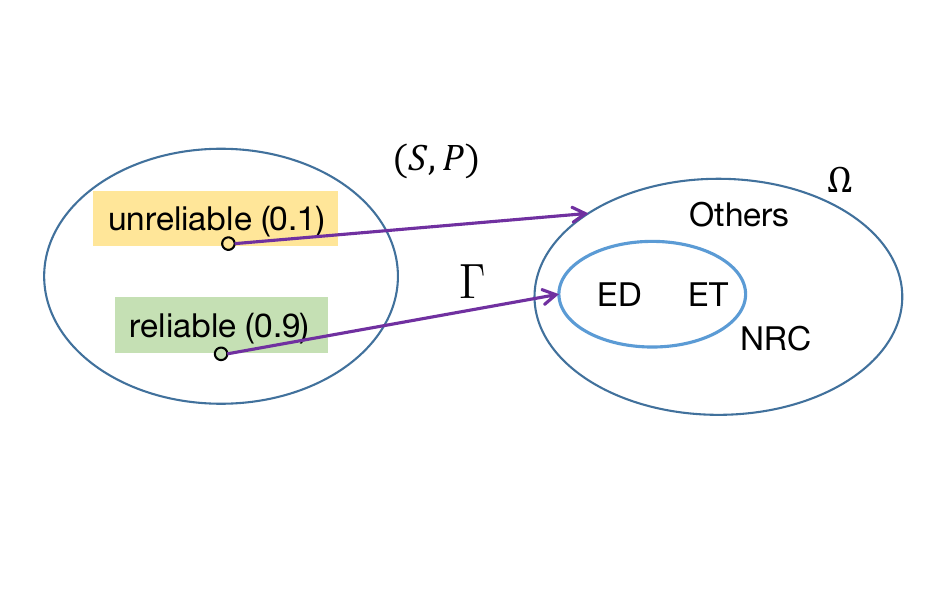}}
\subfloat[\label{fig:demp_1}]{\includegraphics[width=0.5\textwidth]{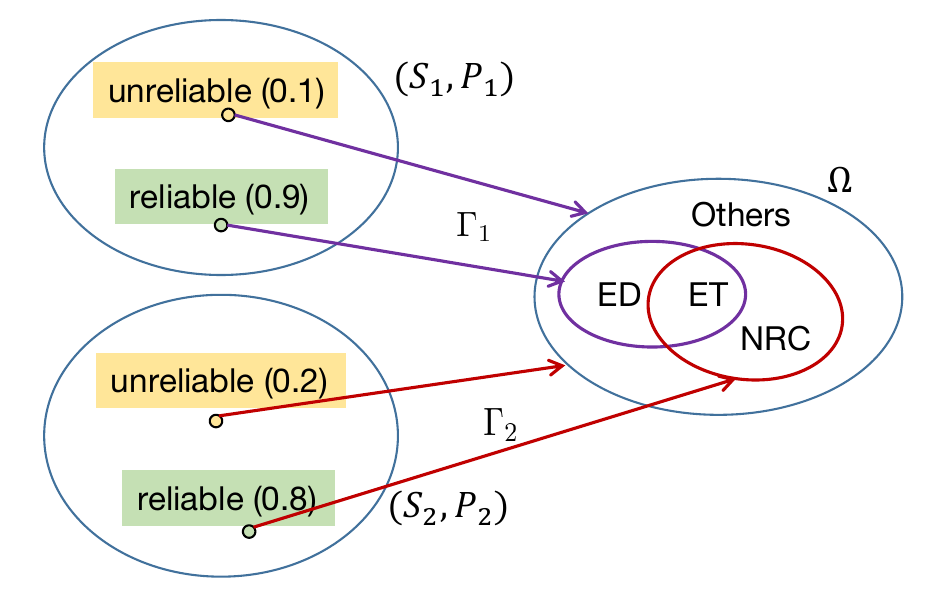}}\\
\subfloat[\label{fig:demp_2}]{\includegraphics[width=0.5\textwidth]{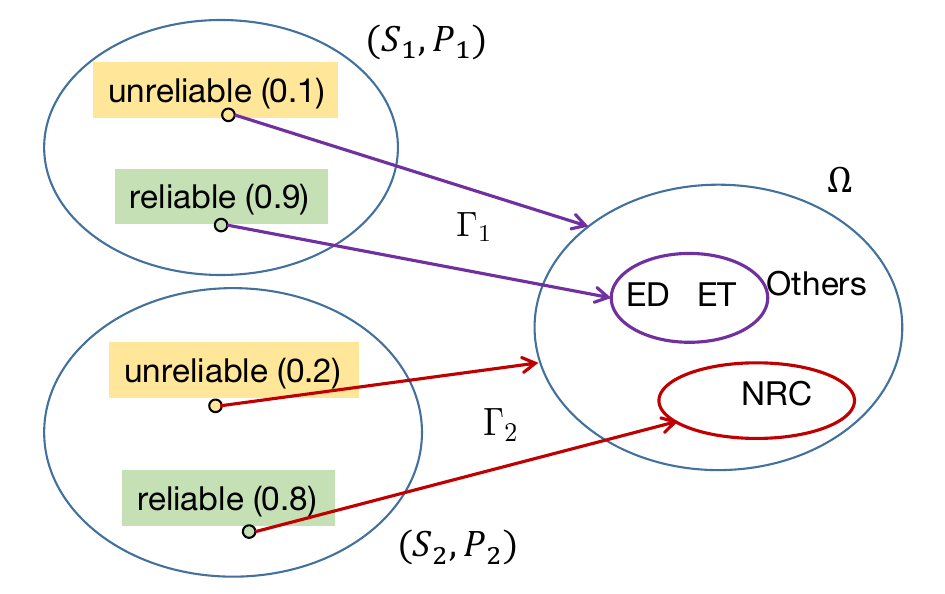}}
\caption{(a) Example of a segmentation task with uncertain information, (b) Example of a segmentation task with multiple sources of information, (c) Example of a segmentation task with conflicting sources of information.}
\label{fig:uncertainty_example}
\end{figure}

The research reviewed in this chapter focuses on improving the accuracy of segmentation performance under the assumption of adequate and perfect input information and accurate and appropriate prior knowledge. However, in reality, especially in the medical image segmentation domain, both the input information and prior knowledge are imperfect and contain a degree of uncertainty. Figure \ref{fig:demp_0} illustrates uncertain information taking a brain tumor segmentation task as an example. Let $X$ be the type of tumor of a voxel, and $\Omega=\{ED, ET, NRC, Others\}$, corresponding to the possibilities: edema, enhancing tumor, necrotic core, and others. Let us assume that a specialist provides the information $X \in \{ED, ET\}$, but there is a probability $p = 0.1$ that the information is unreliable. How to represent this situation by a probability function is a challenging problem. Another situation is when we have multiple information sources tainted with uncertainty, as illustrated in Figure \ref{fig:demp_1}; how can we model that kind of uncertainty and fuse the evidence? Furthermore, if the  information sources are in conflict and contain uncertainty as well, i.e., Figure~\ref{fig:demp_2}, it is difficult to represent and summarize that information by probabilistic models. Thanks to BFT, these challenges can be addressed by designing new frameworks for modeling, reasoning, and fusing imperfect (uncertain, imprecise) information. In the next chapter, we will give a brief introduction to BFT.

\stopcontents[chapters] 

\chapter{Fundamentals of belief function theory} 
\label{Chapter2}


\tocpartial

\begin{figure}
\centering
\includegraphics[width=\textwidth]{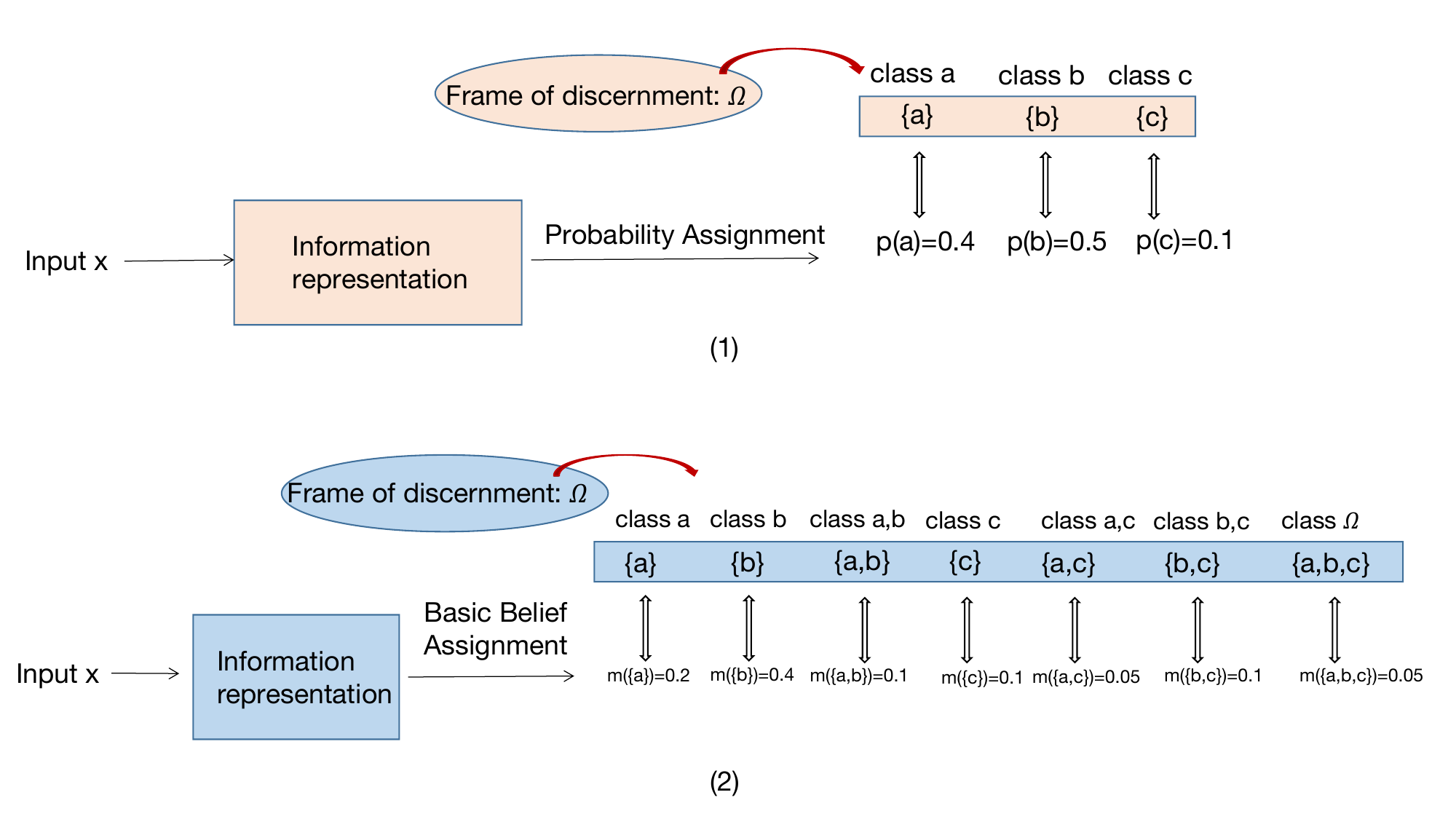}
\caption{An example of three class assignments: (1) Probabilistic model and (2) BFT model. In contrast with the probabilistic model, the BFT model can quantify uncertainty and assign it to the focal set $\{a,b\}, \{a,c\}, \{b,c\}$ and $\{a,b,c\}$ to represent its uncertainty or ignorance ($m$ here is the evidence (mass function) about a variable $\omega$ taking values in $\Omega$, which will be introduced in Section \ref{sec:mass}).}
\label{fig:probability_ds_three_class}
\end{figure}

BFT is a generalization of Bayesian theory, but it is more flexible than the Bayesian approach and suitable under weaker conditions~\cite{sun2018multi}, i.e.,~imperfect (uncertain, imprecise, partial) information. Figure~\ref{fig:probability_ds_three_class} shows the difference between a probabilistic model and a BFT model when applied to a three-class classification task ($\Omega=\{a, b, c\}$). For input $x$, the probabilistic model outputs the probability that $x$ belongs to classes $a$, $b$, and $c$ as 0.4, 0.5, and 0.1, respectively. In contrast, the BFT model can represent degrees of belief that $x$ belongs specifically to any subset of $\Omega$, e.g., $\{a, b\}$, $\{b, c\}$. Compared with the probabilistic model, the BFT model has more degrees of freedom to represent its uncertainty directly, as it shares a unit mass of belief among all subsets of $\Omega$.

In the past decades, BFT has generated considerable interest and has had great success in diverse fields, including uncertain reasoning~\cite{smets1990combination,yager1987dempster,dubois1988representation,denoeux2008conjunctive}, classification~\cite{denoeux1995k,denoeux2000neural} and clustering~\cite{denoeux2004evclus,masson2009recm}, etc. It was first originated by Dempster~\cite{dempster1967upper} in the context of statistical inference in 1968 and was later formalized by Shafer~\cite{shafer1976mathematical} as a theory of evidence in 1976. In 1986, Dubois and Prade proposed an approach to the computerized processing of uncertainty~\cite{dubois2012possibility}. In 1978, Yager proposed a new combination rule of the belief function framework~\cite{yager1987dempster}. In 1990, BFT was further popularized and developed by Smets~\cite{smets1990combination} as the 'Transferable Belief Model' with the pignistic transformation for decision making. Since then, there has been many developments. More detailed information about the development of BFT in 40 years can be found in~\cite{denzux201640}. 

In this chapter, we first introduce the basic notions of BFT, which includes evidence representation (i.e., mass functions, belief, plausibility and contour function, as well as simple mass function) in Section \ref{sec:mass}. Second, we introduce Dempster's rule in Section \ref{sec:fusion} to explain fusion operations of multiple sources of evidence. Third, we introduce the discounting operation for unreliable sources in Section \ref{sec:discounting}. Fourth, we introduce some commonly used decision-making methods in Section \ref{sec:dm}. Finally, in Section \ref{sec:bba}, we give a summary of basic belief assignment methods used to generate mass functions for medical image segmentation. 

\section{Representation of evidence}
\label{sec:mass}
Let $\Omega =\{\omega _{1},\omega _{2}, ..., \omega_{C}\} $ be a finite set of all possible hypotheses about some problem, called the frame of discernment. Evidence about a variable $\omega$ taking values in $\Omega$ can be represented by a mass function $m$, from the power set $2^{\Omega}$ to $[0, 1]$, such that
\begin{subequations}
\begin{align}
    \sum _{A\subseteq \Omega }m(A)=1,\\
    m(\emptyset)=0.
    \label{eq:evidence}
\end{align}    
\end{subequations}
Mapping $m$ can also be called \textbf{basic belief assignment} (BBA). Each subset $A \subseteq \Omega$ such that $m(A)>0$ is called a focal set of $m$. The mass $m(\Omega)$ represents the degree of ignorance about the problem. If all focal sets are singletons, then $m$ is said to be Bayesian and it is equivalent to a probability distribution. 
 
The information provided by a mass function $m$ can also be represented by a \textbf{belief function} $Bel$ or a \textbf{plausibility function} $Pl$ from $2^{\Omega }$ to $[0,1]$ defined, respectively, as
\begin{equation}
   Bel(A) = \sum _{ B\subseteq A}m(B)
   \label{eq:bel}
\end{equation}
and 
\begin{equation}
   Pl(A) = \sum _{B\cap A\neq \emptyset }m(B)=1-Bel(\bar{A}),
   \label{eq:plau}
\end{equation}
for all $A\subseteq \Omega$, where $\bar{A}$ denotes the complement of $A$. The quantity $Bel(A)$ can be interpreted as a degree of support to $A$, while $Pl(A)$ can be interpreted as a measure of lack of support given to the complement of $A$. The \textbf{contour function} $pl$ associated to $m$ is the function that maps each element $\omega$ of $\Omega$ to its plausibility:
\begin{equation}
    pl(\omega)=Pl(\{ \omega\}),  \quad \forall \omega \in \Omega.
\end{equation}

A mass function $m$ is said to be \emph{simple} if it can be obtained by discounting a logical mass function; it thus has the following form:
\begin{equation}
\label{eq:simple}
m(A)=s, \quad m(\Omega)=1-s,
\end{equation}
for some $A\subset \Omega$ such that $A\neq \emptyset$ and some $s\in [0,1]$, called the \emph{degree of support} in $A$. The quantity $w=-\ln(1-s)$ is called the \emph{weight of evidence} associated to $m$ \cite[page 77]{shafer1976mathematical}. In Chapter \ref{Chapter5}, a \textbf{simple mass function} with focal set $A$ and weight of evidence $w$ will be denoted as $A^w$.

\section{Dempster’s rule}
\label{sec:fusion}
In BFT, the belief about a certain question is elaborated by aggregating different belief functions over the same frame of discernment. Given two mass functions $m_{1}$ and $m_{2}$ derived from two independent items of evidence, the final belief that supports $A$ can be obtained by combining $m_{1}$ and $m_{2}$ with Dempster's rule~\cite{shafer1976mathematical} defined as
\begin{equation}
    (m_{1}\oplus m_{2})(A)=\frac{1}{1-\kappa }\sum _{B\cap D=A}m_{1}(B)m_{2}(D),
    \label{eq:demp1}
\end{equation}
for all $A\subseteq \Omega, A\neq \emptyset$, and $(m_{1}\oplus m_{2})(\emptyset)=0$. The coefficient $\kappa$ is the degree of conflict between $m_{1}$ and $m_{2}$, it is defined as
\begin{equation}
    \kappa=\sum _{B\cap D=\emptyset}m_{1}(B)m_{2}(D).
    \label{eq:demp2}
\end{equation}
Mass functions $m_1$ and $m_2$ can be combined if and only if $\kappa < 1$. The mass function $m_1 \oplus m_2$ is called the orthogonal sum of $m_1$ and $m_2$. Let $pl_1$, $pl_2$ and $pl_{12}$ denote the contour functions associated with, respectively, $m_1$, $m_2$ and $m_{1}\oplus m_{2}$. The following equation holds:
\begin{equation}
\label{eq:prodpl}
    pl_{12}=\frac{pl_1 pl_2}{1-\kappa}.
\end{equation}
Table \ref{tab:dempster's rule} shows an example of two source evidence fusion by using Dempster's rule for $A=\{a\}, \{b\}, \{a,b\}, \{c\}, \{a,c\}, \{b,c\}$, and $\{a,b,c\}$. 
\begin{table}
    \centering
    \caption{Example of Dempster's rule for evidence fusion.}
    \scalebox{0.75}{
    \begin{tabular}{l|lllllllll}
    \diagbox{$m_1$}{$m_2$} & $\{a\}$, 0.2 & $\{b\}$, 0.3 &$\{a, b\}$, 0.1 &$\{c\}$, 0.1 &  $\{a, c \}$, 0& $\{b, c \}$,0.2& $\{a, b, c\}$, 0.1\\
     \hline
     $\{a\}$, 0.3 & $\{a\}$, 0.06 & $\emptyset$, 0.09& $\{a\}$, 0.03 &$\emptyset$, 0.03 & $\{a\}$, 0 &$\emptyset$, 0.06& $\{a\}$, 0.03\\
    $\{b\}$, 0.3 & $\emptyset$, 0.06 & $\{b\}$, 0.09 &$\{b\}$, 0.03 &$\emptyset$, 0.03&  $\emptyset$ ,0 &$\{b\}$, 0.06&$\{b\}$, 0.03\\
    $\{a, b\}$, 0.1& $\{a\}$, 0.02& $\{b\}$, 0.03& $\{a,b\}$, 0.01 &$\emptyset$, 0.01 &$\{a\}$, 0& $\{b\}$, 0.02, & $\{a, b\}$, 0.01\\
    $\{c\}$, 0 & $\emptyset$, 0 &$\emptyset$, 0& $\emptyset$, 0& $\{c\}$, 0 & $\{c\}$, 0 &  $\{c\}$, 0& $\{c\}$, 0 \\
    $\{a, c\}$, 0.1 &  $\{a\}$, 0.02 &$\emptyset$, 0.03 & $\{a\}$, 0.01&  $\{c\}$, 0.01 & $\{a, c\}$, 0 & $\{c\}$, 0.02 & $\{a, c \}$, 0.01 \\
    $\{b, c\}$, 0.1& $\emptyset$, 0.02 & $\{b\}$, 0.03 &$\{b\}$, 0.01 & $\{c\}$, 0.01 & $\{ c\}$, 0 & $\{b, c\}$, 0.02 & $\{b, c\}$, 0.01\\
    $\{a, b, c\}$, 0.1& $\{a\}$, 0.02 & $\{b\}$, 0.03 &$\{a, b\}$, 0.01 & $\{c\}$, 0.01 &  $\{a, c \}$, 0& $\{b, c \}$, 0.02& $\{a, b, c\}$, 0.01\\
    \hline
    \end{tabular}
    }
    \label{tab:dempster's rule}
\end{table}
The degree of conflict is $\kappa=0.09+0.03+0.06+0.06+0.03+0.01+0.03+0.02=0.33$. The combined mass function is
\[(m_{1}\oplus m_{2})(\{a\})=(0.06+0.03*2+0.02*3+0.01)/(1-0.33)=19/67,\]
\[(m_{1}\oplus m_{2})(\{b\})=(0.09+0.03*5+0.06+0.02+0.01)/0.67=33/67,\]
\[(m_{1}\oplus m_{2})(\{a, b\})=(0.01*3)/0.67=3/67,\]
\[(m_{1}\oplus m_{2})(\{c\})=(0.01*3+0.02)/0.67=5/67,\]
\[(m_{1}\oplus m_{2})(\{a, c\})=0.01 /0.67=1/67,\]
\[(m_{1}\oplus m_{2})(\{b, c\})=(0.02*2+0.01) /0.67=5/67,\]
\[(m_{1}\oplus m_{2})(\{a, b, c\})=0.01 /0.67=1/67.\]

\section{Discounting}
\label{sec:discounting}

In \eqref{eq:demp2}, if $m_{1}$ and $m_{2}$ are logically contradictory, we cannot use Dempster's rule to combine them. Discounting strategies can be used to combine highly conflicting evidence~\cite{shafer1976mathematical,mercier2008refined,denoeux2019new}. Let $m$ be a mass function on $\Omega$ and $\beta$ a reliability coefficient in $[0,1]$. The \emph{discounting} operation ~\cite{shafer1976mathematical} with the discount rate $1-\beta$ transforms $m$ into a weaker, less informative mass function defined as follows:
\begin{equation}
^\beta m=\beta \, m +(1-\beta) \,m_?,
\label{eq:dis}
\end{equation}
where $m_?$ is the vacuous mass function defined by $m_?(\Omega)=1$, and coefficient $\beta$ is the degree of belief that the source mass function $m$ is reliable \cite{smets94a}. When $\beta=1$, we accept the mass function $m$ provided by the source and take it as a description of our knowledge; when $\beta=0$, we reject it, and we are left with the vacuous mass function $m_?$. In this paper, we focus on the situation when $\beta \in [0,1]$ and combine uncertain evidence with partial reliability using Dempster's rule. Mercier et al. extended the discounting operation with \emph{contextual discounting} in the BFT framework to refine the modeling of sensor reliability, allowing us to use more detailed information regarding the reliability of the source in different contexts \cite{mercier2008refined}, i.e., conditionally on different hypotheses regarding the variable of interest. More details about contextual discounting will be introduced in Chapter \ref{Chapter6}.

\section{Decision-making (DM)} 
\label{sec:dm}
After combining all the available evidence in the form of a mass function, it is necessary to make a decision. In this section, we introduce some classical BFT-based decision-making methods. 

\paragraph{Upper and lower expected utilities}
Let $u$ be a utility function. The lower and upper expectations of $u$ with respect to $m$ are defined, respectively, as the averages of the minima and the maxima of $u$ within each focal set of $m$:

\begin{subequations}
\begin{align}
\underline{E}_m(u)=\sum_{A\subseteq \Omega} m(A) \min_{\omega\in A} u(\omega),\\
\overline{E}_m(u)=\sum_{A\subseteq \Omega} m(A) \max_{\omega\in A} u(\omega).
\end{align}
\label{eq:utlity}
\end{subequations}
When $m$ is Bayesian, $\underline{E}_m(u)= \overline{E}_m(u) $. If $m$ is logical with focal set $A$, then $\underline{E}_m(u)$ and $\overline{E}_m(u)$ are, respectively, the minimum and maximum of $u$ in $A$. The lower or upper expectations can be chosen for the final decision according to the given task and decision-maker's attitude.

\paragraph{Pignistic criterion}

In 1990, Smets proposed a pignistic transformation~\cite{smets1990combination}
that distributes each mass of belief distributed equally among the elements of $\Omega$. The pignistic probability distribution is defined as
\begin{equation}
 BetP(\omega)=\sum _{\omega\in A}\frac{m(A)}{\mid A\mid }, \quad  \forall \omega\in \Omega,  
 \label{eq:pignistic}
\end{equation}
where $\left | A \right |$ denotes the cardinality of $A \subseteq \Omega$.

Besides the above methods, there are various decision-making methods proposed for BFT, such as Generalized OWA criterion \cite{yager1992decision},  Generalized minimax regret \cite{YAGER2004109}, \new{Generalized divergence \cite{xiao2022generalized}}, etc. More details about decision-making with BFT can be found in the review paper \cite{denoeux2019decision}. 

\section{Basic belief assignment (BBA) methods to generate mass functions}
\label{sec:bba}

\begin{figure*}
\includegraphics[width=\textwidth]{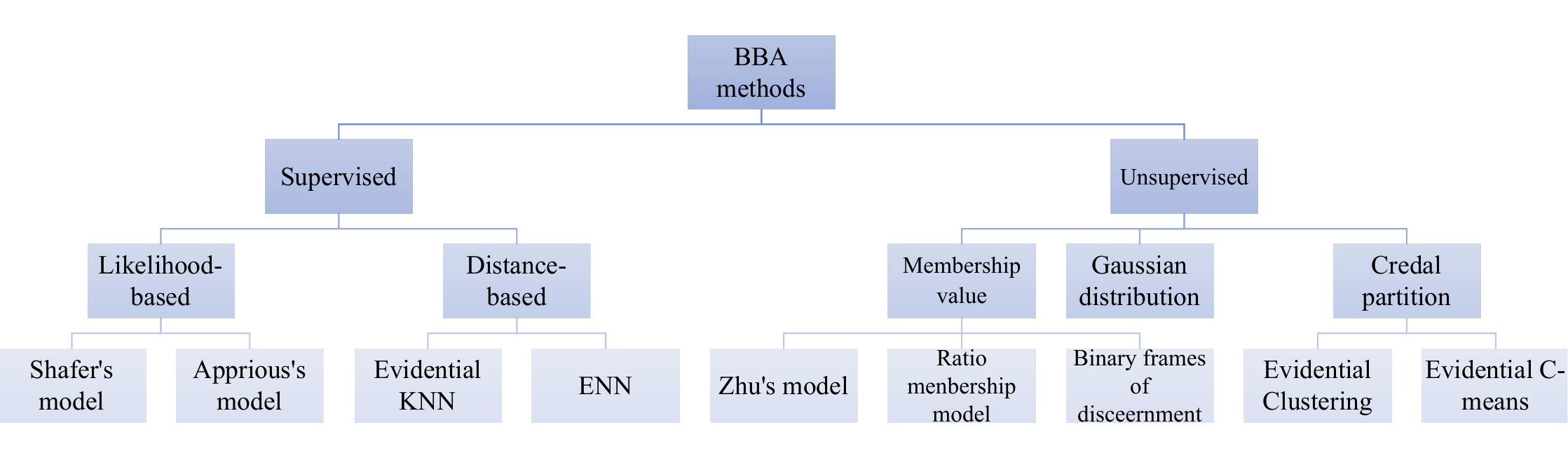}
\caption{Overview of Basic belief assignment (BBA) methods.\label{fig:bba}}
\end{figure*}

To model segmentation uncertainty, the first step is to generate mass functions. In this section, we introduce the main basic belief assignment methods applied to medical image segmentation. Figure~\ref{fig:bba} is an overview of basic belief assignment methods. In general, those methods can be separated into supervised and unsupervised methods according to whether annotations are used to optimize the parameters of basic belief assignment models or not. 

\subsection{Supervised basic belief assignment methods}
\label{subsec:supervised}

Supervised basic belief assignment methods can be classified into two categories. One is the likelihood-based methods, such as Shafer's model~\cite{shafer1976mathematical} and Appriou's model~\cite{appriou1999multisensor,appriou200501}. The other category is composed of distance-based methods, such as the evidential KNN rule~\cite{denoeux1995k}, the evidential neural network classifier~\cite{denoeux2000neural} and Radial basis function networks~\cite{denoeux2019decision}. It should be noted that the distance-based methods can easily be merged with popular deep segmentation models and have shown promising results~\cite{tong2021evidential, huang2021belief,huang2022lymphoma}.

\subsubsection{Likelihood-based methods}
\label{subsubsection:lik_methods}

\paragraph*{Shafer's model}
In~\cite{shafer1976mathematical}, Shafer proposed a likelihood-based evidential
model to calculate mass functions. Assuming that conditional density functions
$f(x\mid \omega_c)$ are known, then the conditional likelihood associated with the
pattern $X$ is defined by $\ell (\omega_c\mid x)=f(x\mid \omega_c)$. The mass functions are defined according to the knowledge of all hypotheses $\omega_1, \ldots, \omega_C $. Firstly, the plausibility of a simple hypothesis $\omega_c$ is proportional to its likelihood. The plausibility is, thus, given by
\begin{equation}
    Pl(\{\omega_c\})=\hslash \cdot \ell (\omega_c\mid x), \quad  \forall \omega_c \in \Omega,
    \label{eq:14}
\end{equation}
where $\hslash $ is a normalization factor with $\hslash=1/  \max_{\omega\in\Omega} \ell(\omega \vert x)$. The plausibility of a set $A$ is, thus, given by
\begin{equation}
  Pl(A)=\hslash\cdot \underset{\omega_c \in A}{\max} \ell (\omega_c\mid x).
\end{equation}

\paragraph*{Appriou's model}
Appriou~\cite{appriou1999multisensor,appriou200501} also proposed two likelihood-based models to calculate mass functions with the frame of discernment $\Omega= \{\omega_{c}, \neg{\omega_{c}} \}$. For the first model, the mass functions are defined by
\begin{subequations}
\begin{align}
m(\{\omega_{c}\})&=0,\\
m(\{\neg{\omega_{c}}\})&=\alpha_{c}  ({1-\hslash \cdot \ell (\omega_{c}\mid x) }) ,\\
m(\Omega)&=1-\alpha_{c}{ (1-\hslash \cdot \ell (\omega_{c}\mid x ))},
    \label{eq:15}
\end{align}
\end{subequations}
where $\alpha_{c}$ is a reliability factor depending on the hypothesis $\omega_{c}$ and on the source information. The second model is defined as
\begin{subequations}
\label{eq:appriou2}
\begin{align}
m(\{\omega_{c}\})&=\alpha_{c} \cdot \hslash \cdot \ell (\omega_{c}\mid x)/  {(1+\hslash \cdot \ell (\omega_{c}\mid x))}, \\
m(\{\neg{\omega_{c}}\})&=\alpha_{c}/ {(1+\hslash \cdot \ell (\omega_{c}\mid x))} ,\\
m(\Omega)&=1-\alpha_{c}.
    \label{eq:16}
\end{align}
\end{subequations}

\subsubsection{Distance-based methods}
\label{subsubsec: distance}
\paragraph*{Evidential KNN ({EKNN}) rule}
 
In~\cite{denoeux1995k}, Den{\oe}ux proposed a distance-based KNN classifier for classification tasks. Let $N_K(x)$ denote the set of the $K$ nearest neighbors of $x$ in learning set. Each $x_{i}\in N_K(x)$ is considered as a piece of evidence regarding the class label of $x$. The strength of evidence decreases with the distance between
$x$ and $x_{i}$. The evidence of $(x_{i},y_{i})$ support class $c$ is represented by
\begin{subequations}
\begin{align}
m_i(\{\omega_{c}\})&=\varphi _{c}(d_{i})y_{ic}, \quad  1 \le c \le C,\\
m_{i}(\Omega)&=1-\varphi_{c}(d_{i}),
\end{align}
\label{eq:18}
\end{subequations}
where $d_{i}$ is the distance between $x$ and $x_{i}$, which can be the Euclidean or other distance function; and $y_{ic}=1$ if $y_{i}=\omega_{c}$ and $y_{ic}=0$ otherwise. Function $\varphi _{c}$ is defined as
\begin{equation}
    \varphi _{c}(d)=\alpha \exp(-\gamma d^{2}),
    \label{eq:19}
\end{equation}
where $\alpha$ and $\gamma$ are two tuning parameters. The evidence of the $K$ nearest neighbors of $x$ is fused by Dempster's rule:
\begin{equation}
m=\bigoplus _{x_{i}\in N_K(x)}m_{i}.  
\label{eq:20}
\end{equation}

The final decision is made according to maximum plausibility. The detailed optimization of these parameters is described in~\cite{zouhal1998evidence}. Based on this first work, Den{\oe}ux et al. proposed the contextual discounting evidential KNN rule~\cite{denoeux2019new} with partially supervised learning to address the annotation limitation problem.

\paragraph*{Evidential neural network (ENN)}
The success of machine learning encouraged the exploration of applying belief function theory with learning methods. In~\cite{denoeux2000neural}, Den{\oe}ux proposed an ENN classifier in which mass functions are computed based on distances to prototypes. 

The ENN classifier is composed of an input layer of $H$ neurons, two hidden layers, and an output layer. The first input layer is composed of $I$ units, whose weights vectors are prototypes $p_1,\ldots, p_I$ in input space. The activation of unit $i$ in the prototype layer is
\begin{equation}
    s_i=\alpha _i \exp(-\gamma_i d_i^2),   
    \label{eq:s_i}
\end{equation}
where $d_i= \left |  x-p_i \right |  $ is the Euclidean distance between input vector $x$ and prototype $p_i$, $\gamma_i>0$ is a scale parameter,  and $\alpha_i \in [0,1]$ is an additional parameter. The second hidden layer computes mass functions $m_i$ representing the evidence of each prototype $p_i$, using the following equations: 
\begin{subequations}
\begin{align}
m_i(\{\omega _{c}\})&=u_{ic}s_i, \quad c=1, \ldots , C\\
m_{i}(\Omega)&=1-s_i, 
\end{align}
\label{eq:m_i}
\end{subequations}
where $u_{ic}$ is the membership degree of prototype $i$ to class $\omega_c$, and $\sum _{c=1}^C u_{ic}=1$. Finally, the third layer combines the $I$ mass functions $m_1,\ldots,m_I$ using Dempster's rule. The output mass function $m=\bigoplus_{i=1}^I m_i$ is a discounted Bayesian mass function that summarizes the evidence of the $I$ prototypes.

Let $\btheta$ denote the vector of all network parameters, composed of the $I$ prototypes $\bp_i$, their parameters $\gamma_i$ and $\alpha_i$, and their membership degrees $u_{ic}$, $c=1,\ldots,C$. In \cite{denoeux2000neural}, it was proposed to learn these parameters  by minimizing the regularized sum-of-squares loss function
\begin{equation}
 L_{SS}(\btheta)=\sum_{n=1}^{N}\sum_{c=1}^{C}(p_{nc}-y_{nc})^{2}+ \lambda  \sum_{i=1}^{I}\alpha_{i},  
 \label{eq:lossENN}
\end{equation}
where $p_{nc}$ is the pignistic probability of class $\omega_c$ for instance $n$, \new{$N$ is the number of training instances}, and $y_{nc}=1$ if the true class of instance $n$ is $\omega_c$, and  $y_{nc}=0$ otherwise.  The second term on the right-hand side of \eqref{eq:lossENN} is a regularization term, and $\lambda$ is a hyperparameter that can be tuned by cross-validation.


\begin{figure}
\centering
\includegraphics[width=0.8\textwidth]{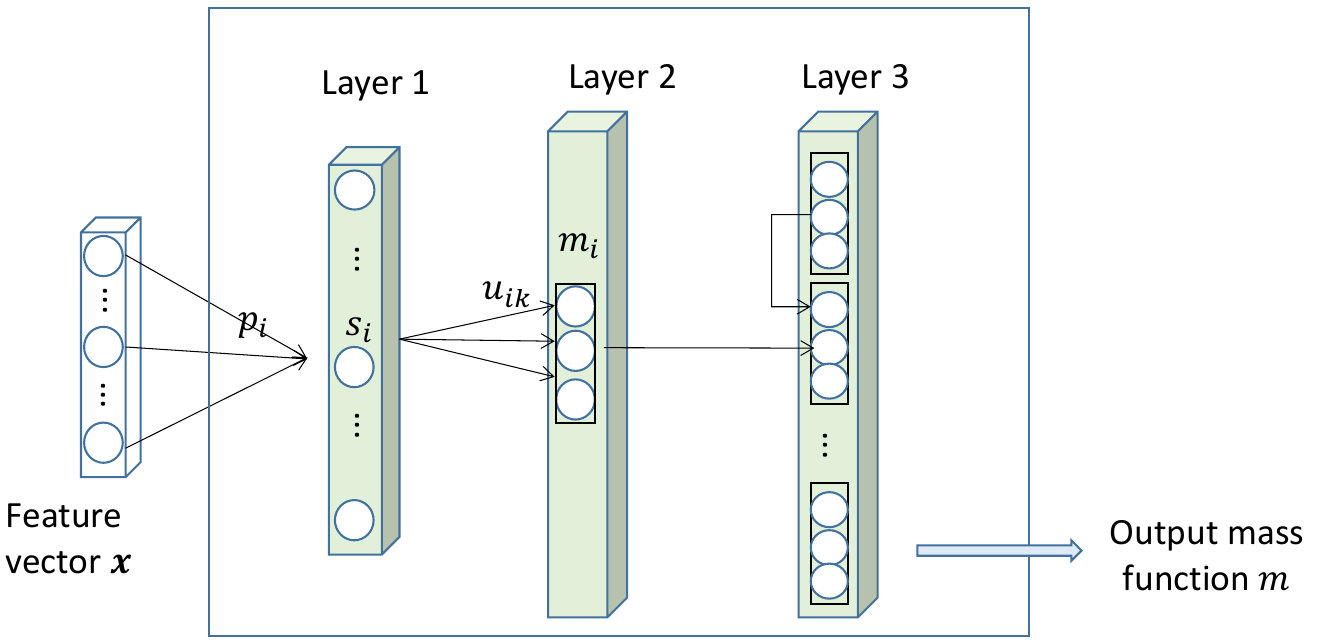}
\caption{Evidential neural network.}
\label{fig:ENN}
\end{figure}

\subsection{Unsupervised basic belief assignment methods}
\label{subsec:unsupervised}

The goal of unsupervised basic belief assignment methods is to generate mass functions without any label information. In earlier basic belief assignment studies, Fuzzy C-means (FCM)~\cite{dunn1973fuzzy} was the most popular algorithm used to generate membership values (MVs). Based on MVs, the authors can obtain mass functions according to some domain knowledge, e.g.,~threshold~\cite{zhu2002automatic}, or user-specific parameters~\cite{ghasemi2012brain}. The sigmoid and one-sided Gaussian function
can also be used to generate MVs~\cite{safranek1990evidence}. The notion of credal
partition~\cite{denoeux2004evclus}, an extension of fuzzy partition, enables us to generate mass functions directly~\cite{masson2008ecm}. Besides these two popular basic belief assignment methods, mass functions can also be generated from the Gaussian distribution of the input to the cluster center~\cite{chen2012manifold}.

\subsubsection{MV-based methods}
\label{subsub:MVs}
\paragraph*{FCM}

Considering that there are some BFT-based methods that use FCM to generate MVs,  we briefly summarize FCM here to offer a basic view for readers. With FCM, any $\mathbf{x}$ has a set of coefficients $w_{k}(x)$ representing the degree of membership in the $k$th cluster. The centroid of a cluster is the mean of all points, weighted by the $m$-th power of their membership degree,
\begin{equation}
c_{k}=\frac{{\sum _{x}{w_{k}(x)}^{m}x}}{ {\sum _{x}{w_{k}(x)}^{m}}},
\end{equation}
where $m$ is the hyper-parameter that controls how fuzzy the cluster will be. The higher it is, the fuzzier. Given a finite set of data, the FCM algorithm returns a list of cluster centers $P=\{c_{1}, \ldots, c_{C}\}$ and a partition matrix $W=(w_{ij}), i=1,\ldots, N, j=1, \ldots, C$,

\begin{equation}
    w_{{ij}}={\frac  {1}{\sum _{{k=1}}^{{C}}\left({\frac  {\left\|{\mathbf  {x}}_{i}-{\mathbf  {c}}_{j}\right\|}{\left\|{\mathbf  {x}}_{i}-{\mathbf  {c}}_{c}\right\|}}\right)^{{{\frac  {2}{m-1}}}}}},
\end{equation}
where $w_{ij}$, is the degree of membership of $\mathbf{x}_{i}$  to cluster $\mathbf{c}_{j}$. The objective function is defined as
\begin{equation}
    \underset{P }{arg \max} \sum_{i=1}^{N} \sum_{j=1}^{C} w_{ij}^m \left\|  \mathbf{x}_i-\mathbf{c}_j  \right\| . 
\end{equation}

\paragraph*{Zhu's model}

\begin{figure*}
\subfloat[]{\label{fig:overlaping}\includegraphics[width=0.5\textwidth]{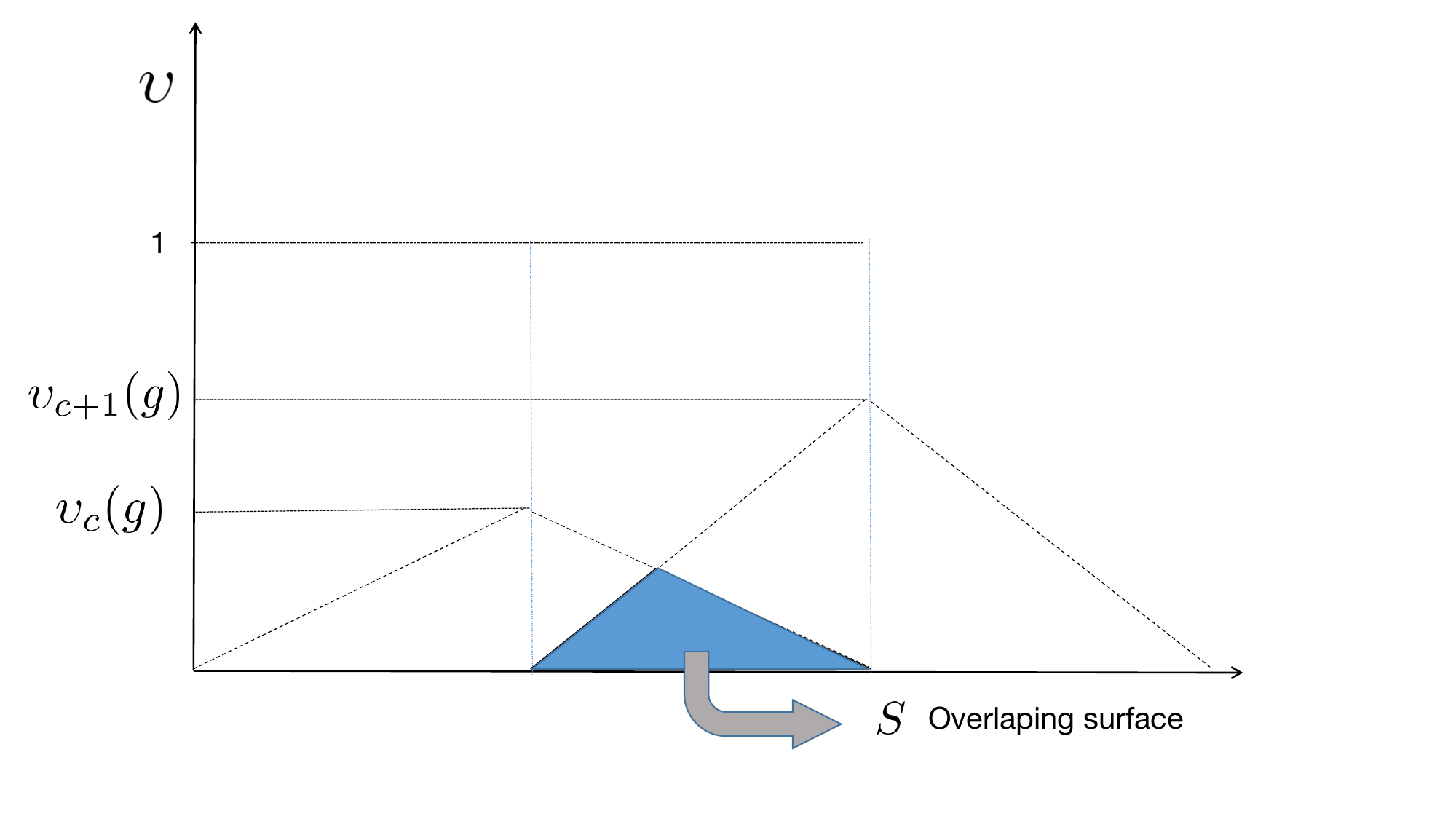}}
\subfloat[]{\label{fig:maximumambigulity}\includegraphics[width=0.5\textwidth]{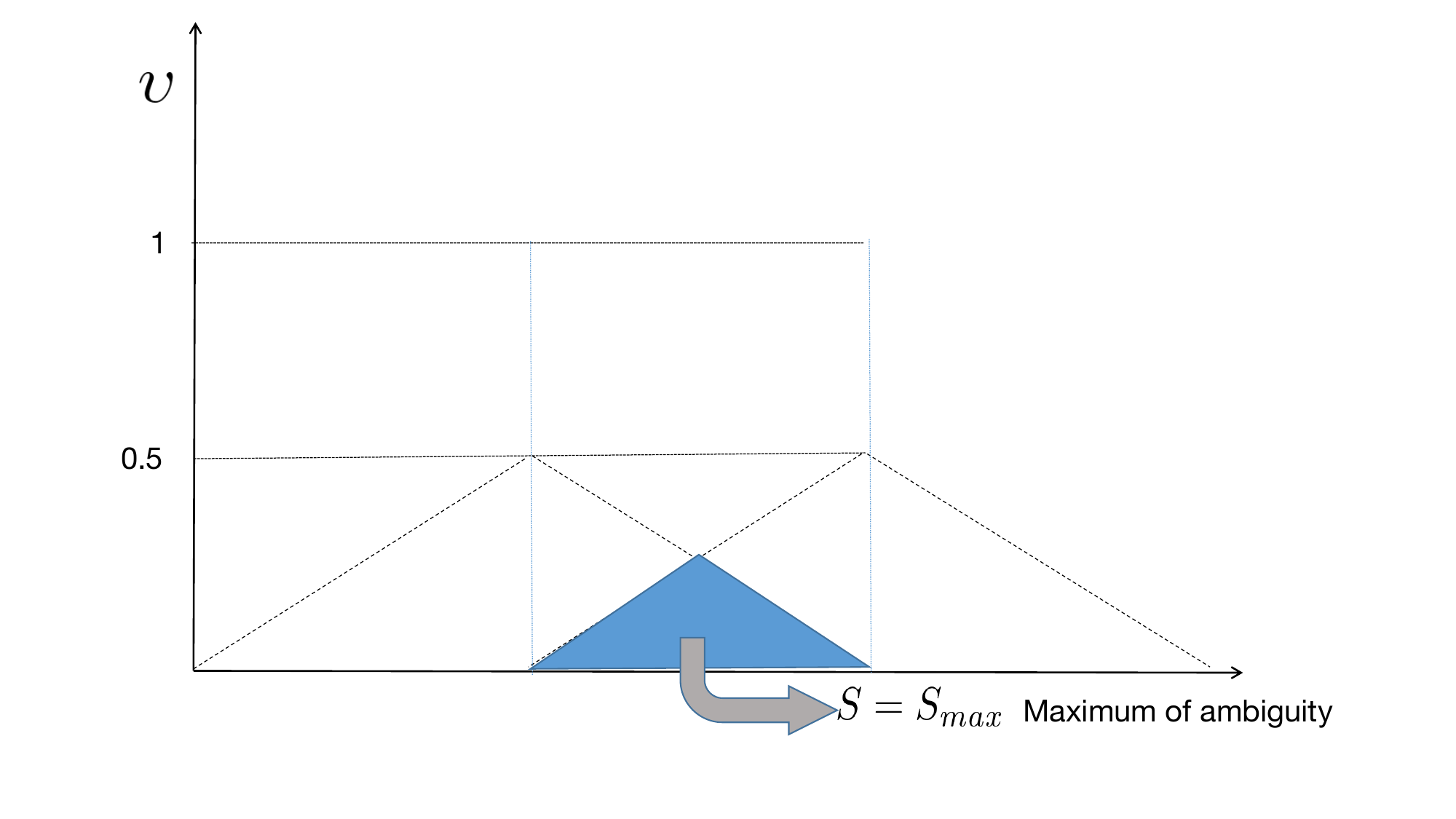}}
\caption{\label{fig:zhu}(a) Construction of triangular membership functions and (b) maximum ambiguity case: $\upsilon _c(g)=\upsilon _{c+1}(g)=0.5$. 
}
\end{figure*}

In~\cite{zhu2002automatic}, Zhu et al. proposed a method to determine mass
functions using FCM and neighborhood information. The mass assigned to a simple hypothesis
$\{ \omega_c \}$ is directly obtained from the filtered membership functions
$\upsilon _{c}(g)$ of the gray level $g(x,y)$ to cluster $c$ as
$m(\{\omega_{c}\}) = \upsilon_{c}(g)$. For a given gray level, the piece of evidence of
belonging to the cluster $c$ is, thus, directly given by its degree of
membership to the same cluster. If there is a high ambiguity in assigning a
gray level $g(x,y)$ to cluster $c$ or $c+1$, that is, $\left | \upsilon _c(g)-\upsilon _{c+1}(g) \right | < \varepsilon $, where $\varepsilon $ is a thresholding value, then a double hypothesis is formed. The value of the threshold $\varepsilon $ is chosen
depending on the application. The authors suggested fixing
$\varepsilon $ at 0.1. Once the double hypotheses are formed, their associated
mass is calculated according to the following formula:
\begin{equation}
   m(\{\omega_{c},\omega_{c+1}\})
   =\frac{S[\upsilon _{c}(g), \upsilon _{c+1}(g)]} { 2S_{max}}, 
   \label{eq:29}
\end{equation}
where $S$ represents the surface of a triangle and $S_{max}$ is the maximum of ambiguity. The surface of such a triangle depends both on the degrees of the membership functions of $g(x,y)$ to clusters $c$ and $c+1$ and on the conflicts between these MVs. Figure~\ref{fig:overlaping} shows how the triangle is constructed and how the mass of double hypotheses $\{\omega_c, \omega_{c+1} \}$ is derived from the surface of the triangle. The vertical axis of  Figure~\ref{fig:overlaping} represents the MVs. The surfaces of the two dotted triangles define two so-called triangular membership functions corresponding to classes $c$ and $c+1$. The two triangles are isosceles and have the same length for their bases. The heights of the triangles are equal to $m(\{\omega_c\})$ and $m(\{\omega_{c+1}\})$, respectively. The overlapping surface $S$ of the two triangles represents the MV to the double hypothesis $\{\omega_c, \omega_{c+1} \}$. Therefore, the mass value attributed to the double hypothesis $\{\omega_c, \omega_{c+1} \}$ can be directly calculated from the surface $S$.  Figure~\ref{fig:maximumambigulity} shows the condition of the maximum ambiguity case.

\paragraph*{Ratio MV transformation}

In~\cite{ghasemi2012brain}, Ghasemi et al. proposed a ratio membership value transformation method to calculate mass functions. The FCM algorithm was first used to generate MVs $f_{\omega_c}$ for each pixel. Then the MVs are used to build the mass functions. For this purpose, the three ratios of the available MVs are calculated, corresponding to three situations: ``no uncertainty'' (NU), ``semi-uncertainty'' (SU), and ``perfect-uncertainty'' (PU). First, PU is a critical situation in which the Ratio MVs are smaller than $\alpha$, then the mass function is calculated as $m(\{\omega_1\})=m (\{ \omega_2 \})=m (\Omega )=(f_{\omega_1}+f_{\omega_2})/3$. Second, two thresholds $\alpha$ and
$\beta$ with $\alpha=1.5$ and $\beta=3$ are selected to control the boundary between SU and PU, and between NU and SU, separately. For example, with 
\[
f_{\omega_1}=0.18,\quad
f_{\omega_2}=0.81,\quad
RMV=f_{\omega_1}/f_{\omega_2}= 4.5, \quad RMV>\beta,\]
the two MVs fall in the NU category. If
\[f_{\omega_1}=0.25,\quad f_{\omega_2}=0.65,\quad RMV=f_{\omega_1}/f_{\omega_2}=2.6,\quad \alpha<RMV<\beta,\]
the two MVs are in the SU category. The mass functions are calculated as
\begin{subequations}
\begin{align}
 m(\{\omega_1\})&=f_{\omega_1}-\frac{\lambda_{\omega_1,\omega_2}}{2},\\
 m(\{\omega_2\})&=f_{\omega_2}-\frac{\lambda_{\omega_1,\omega_2}}{2},\\
 m (\Omega)&=\lambda_{\omega_1,\omega_2},
\end{align} 
\label{eq:31}
\end{subequations}
where $\lambda$ is an uncertainty distance value  defined as $\lambda_{\omega_1, \omega_2}=\frac{\left | f_{\omega_1}-f_{\omega_2} \right |}{\beta-\alpha}$.

\subsubsection{Evidential C-means (ECM)}
\label{subsubsec:ecm}
In~\cite{denoeux2004evclus}, Den{\oe}ux et al. proposed an evidential clustering algorithm, called EVCLUS, based on the notion of credal partition,
which extends the existing concepts of hard, fuzzy (probabilistic), and
possibilistic partition by allocating to each object a ``mass of belief'', not only
to single clusters but also to any subsets of $\Omega= \{\omega_{1}, \ldots , \omega_{C}\}$. 

\paragraph*{Credal partition}
 Assuming there is a collection of five objects for two classes, mass functions for each source are given in Table \ref{tab:credal partition}. They represent different situations: the mass function of object 1 indicates strong evidence that the class of object 1 does not lie in $\Omega$; the class of object 2 is completely unknown, and the class of object 3 is known with certainty; the cases of objects 4 and 5 correspond to situations of partial knowledge ($m_{5}$ is Bayesian). The EVCLUS algorithm generates a credal partition for dissimilarity data by minimizing a cost function.

\begin{table}
    \centering
    \caption{Example of credal partition.}
    \begin{tabular}{|c|c|c|c|c|c|}
    \hline
    $A$& $m_{1}(A)$&$m_{2}(A)$&$m_{3}(A)$&$m_{4}(A) $&$m_{5}(A) $\\
    \hline
    $\{ \emptyset \}$&1&0&0&0&0 \\
    $\{a\}$&0&0&1&0.5&0.6\\
    $\{b\}$&0&0&0&0.3&0.4\\
    $\{a,b\}$&0&1&0&0.2&0\\
    \hline
    \end{tabular}
    \label{tab:credal partition}
\end{table}

\paragraph*{Evidential C-Means (ECM)}

The ECM algorithm~\cite{masson2008ecm} is another method for generating a credal partition from data. In ECM, a cluster is represented by a prototype $p_{c}$. For each non-empty set $A_{j}\subseteq\Omega$, a prototype $\bar{p_{j}}$ is defined as the center of mass of the prototypes $p_c$ such that $\omega_c\in A_j$. Then the non-empty focal set is defined as $F=\{A_{1}, \ldots , A_{f}\}\subseteq 2^{\Omega}\setminus\left\{\emptyset\right \} $. Deriving a credal partition from object data implies determining,
for each object $x_i$, the quantities $m_{ij}=m_i(A_j), A_i\ne \emptyset, A_j \subseteq \Omega$, in such a way that $m_{ij}$ is low (resp. high) when the distance between $x_i$ and the focal set $\bar{p_{j}}$ is high (resp. low). The distance between an object and
any nonempty subset of $\Omega$ is then defined by
\begin{equation}
    d_{ij}^2=\left\|  x_{i}-\bar{p_{j}}\right\| ^2.
    \label{eq:27}
\end{equation}

\subsubsection{Gaussian distribution (GD)-based model}
 \label{subsubsection: gd}
Besides the FCM-based and credal partition-based methods, the mass functions
can also be generated from GD~\cite{chen2012manifold}. The mass of simple hypotheses $\{\omega_c\}$ can be obtained from the assumption of GD according to the information $x_i$ of a pixel from an input image to cluster $c$ as follows:
\begin{equation}
m(\{\omega_c\})=\frac{1}{\sigma_c \sqrt{2\pi}} \exp{ \frac{-(x_i-\mu_{c})^2}{2 \sigma_{c}^{2}}},
\label{eq:32}
\end{equation}
where $\mu_c$ and $\sigma^2_c$ represent, respectively, the mean and the variance of the cluster $c$, which can be estimated by
\begin{equation}
    \mu_c=\frac{1}{n_c}\sum_{i=1}^{n_c}x_i,
    \label{eq:33}
\end{equation}
\begin{equation}
\sigma _c^2=\frac{1}{n_c}\sum_{i=1}^{n_c}(x_i-\mu_c)^2,
\label{eq:34}
\end{equation}
where $n_c$ is the number of pixels in the cluster $c$. The mass of multiple hypotheses  $\{ \omega_1, \omega_2, \ldots , \omega_T \}$ is determined as
\begin{equation}
    m(\{ \omega_1, \omega_2, \ldots , \omega_T\})=\frac{1}{\sigma_t\sqrt{2\pi}} \exp{\frac{-(x_i-\mu_{t})^2}{2\sigma_t^{2}}},
    \label{eq:35}
\end{equation}
where $\mu_t=\frac{1}{T} \sum_{i=1}^{T} \mu_i$, $\sigma_t=\max(\sigma_1, \sigma_2, \ldots ,\sigma_T), 2 \le T \le C$, $C$ is the number of clusters.

\subsubsection{Binary frames of Discernment (BFOD)}
 \label{subsubsection: bfod}
Under the assumption that the membership value is available, Safranek et al. introduced a BFOD-based method~\cite{safranek1990evidence} to transform membership values into mass functions. The BFOD is constructed as $\Omega = \{ \omega, \neg \omega \}$
with a function $cf(\nu)$, taking values in $\left [ 0,1 \right ] $ that assigns confidence factors. According to the authors, the sigmoid and one-sided Gaussian functions are the most appropriate functions for defining $cf(\nu)$. Once a confidence value is obtained, the transformation into mass functions can be accomplished by defining
appropriate transfer functions:
\begin{subequations}
\begin{align}
\label{eq:momega}
  m(\{\omega\})&=\frac{B}{1-A}cf(\nu)-\frac{AB}{1-A},\\
\label{eq:mnegomega}
 m(\{\neg \omega\})&=\frac{-B}{1-A}cf(\nu)+B,\\
 m(\Omega)&=1-m(\{\omega\})-m(\{\neg \omega\}),
\end{align}
\label{eq:28}
\end{subequations}
where $A$ and $B$ are user-specific parameters. In \eqref{eq:momega} and \eqref{eq:mnegomega}, the left-hand side stays clamped at zero when the right-hand side goes negative. Parameter $A$ is the confidence-factor axis intercept of the curve that depicts the dependence of $m(\{ \omega\})$ on confidence factors, and $B$ is the maximum support value assigned to $m(\{ \omega\})$ or $m(\{ \neg \omega\})$.

\section{Conclusion}
BFT is a powerful tool to represent imperfect information and combine different sources of information by Dempster's rule. In this chapter, we first recalled how imperfect information can be represented by mass functions and how multiple sources of evidence can be combined by Dempster's rule. We then introduced the discounting operation for unreliable evidence and decision-making based on mass functions. In addition, we gave a summary of existing basic belief assignment methods that have been used in medical image segmentation. In the next chapter, we will provide an overall review of BFT-based medical image segmentation methods.

\stopcontents[chapters] 

\chapter{Review of BFT-based medical image segmentation methods} 
\label{Chapter3}


\tocpartial

\begin{figure}
\subfloat[\label{fig:composition_modality}]{\includegraphics[width=0.5\textwidth]{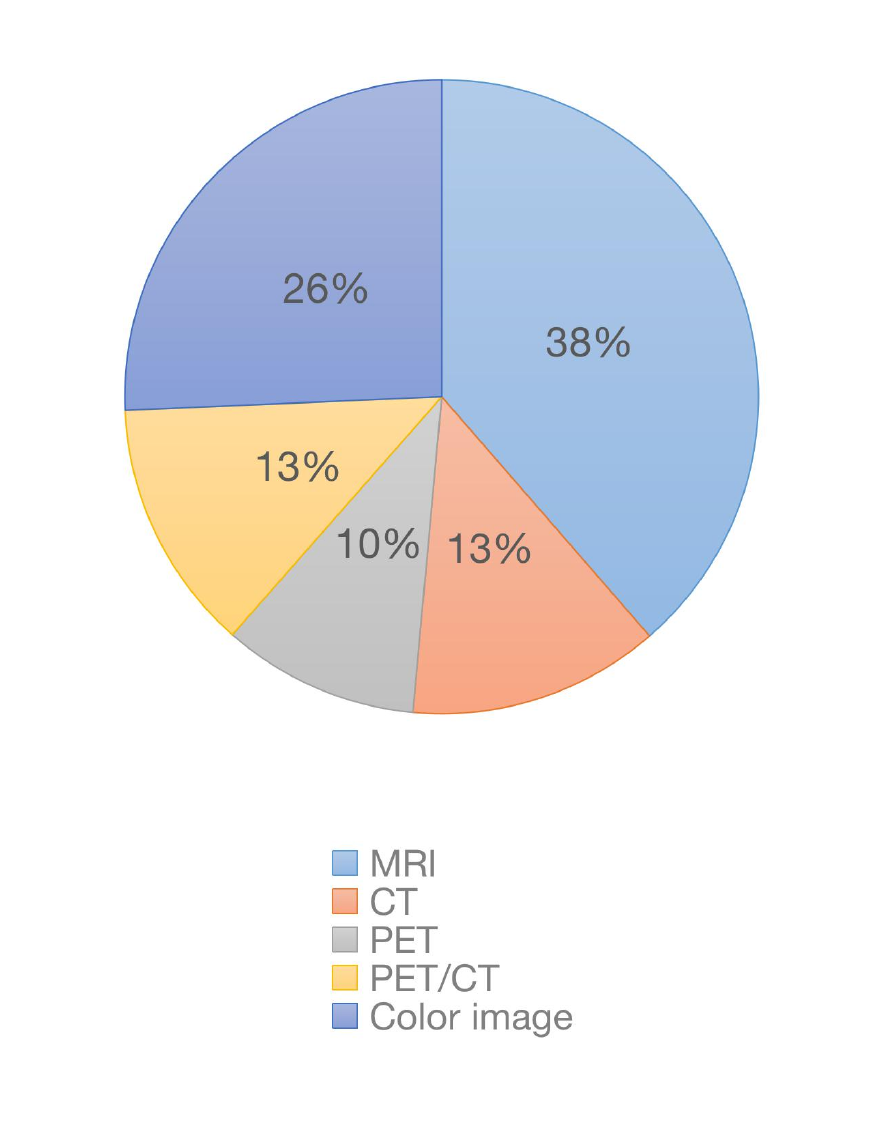}}
\subfloat[\label{fig:compsition_app}]{\includegraphics[width=0.5\textwidth]{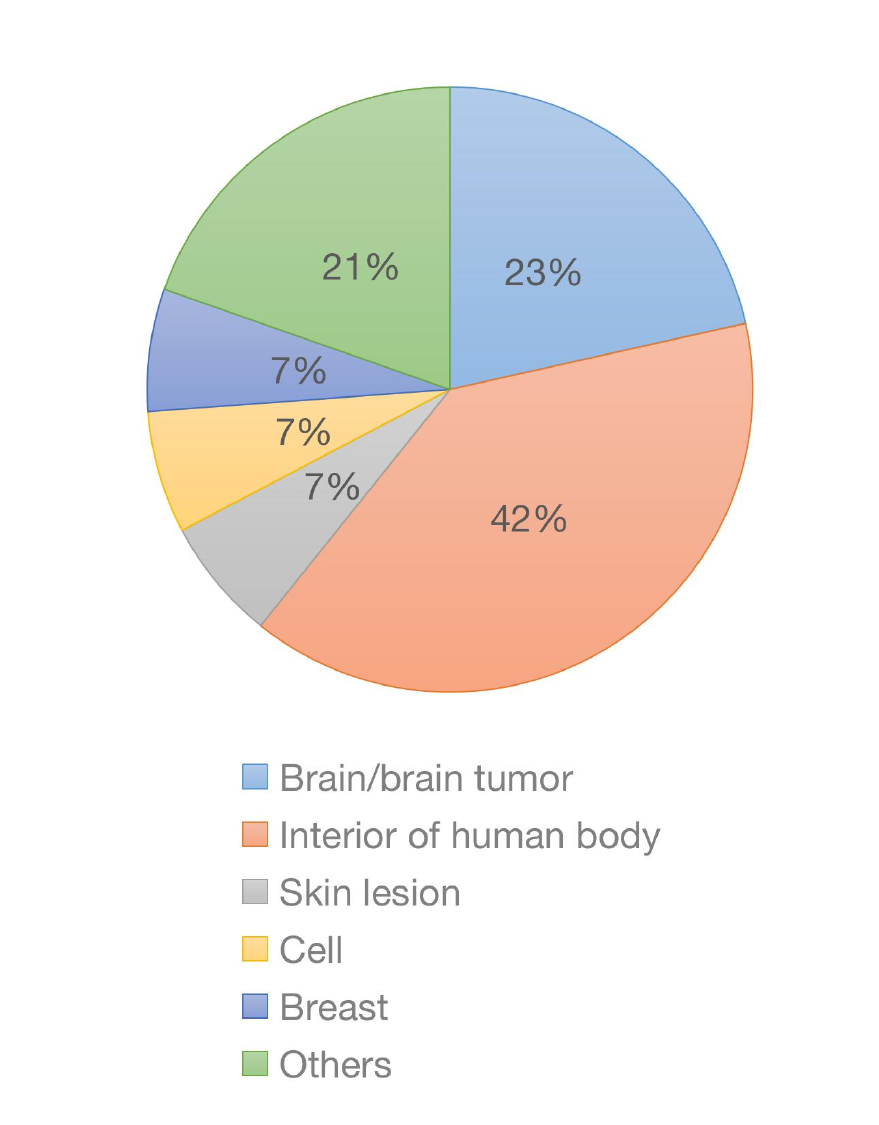}}
\caption{(a) The composition of input modality of the medical image, (b)The composition of specific medical application.}
\label{fig:compositation}
\end{figure}

To summarize the BFT-based medical image segmentation methods, we can either classify them by the input modality of the images or by the specific clinical application. Figure~\ref{fig:composition_modality} shows the proportion of types of medical images applied in the segmentation task and Figure~\ref{fig:compsition_app} displays the proportions of application in the medical image segmentation task. 
\begin{figure}
    \centering
    \includegraphics[width=\textwidth]{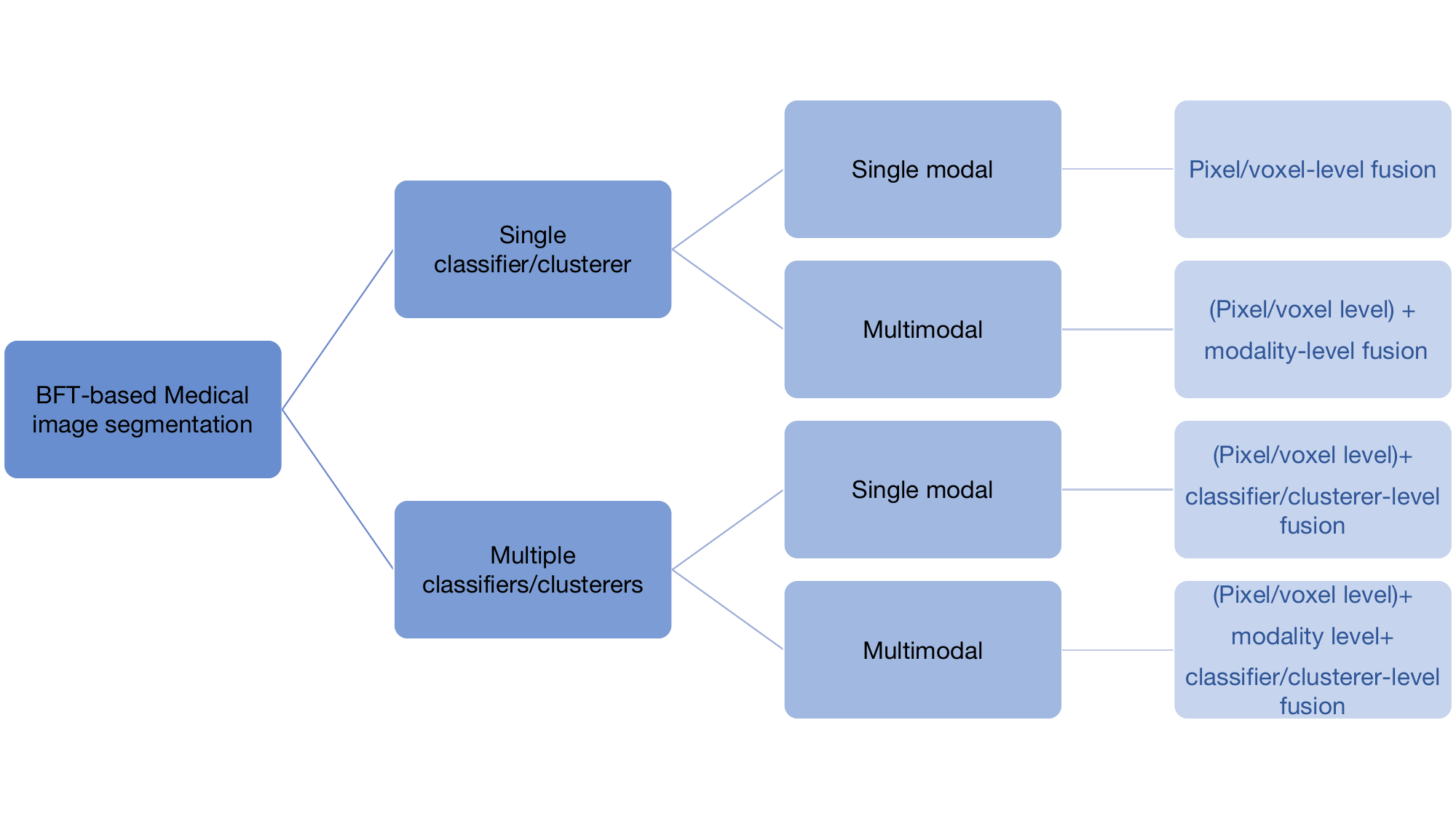}
    \caption{Overview of BFT-based medical image segmentation methods.}
    \label{fig:BFT-based-seg}
\end{figure}

\begin{figure}
\centering
\includegraphics[scale=0.4]{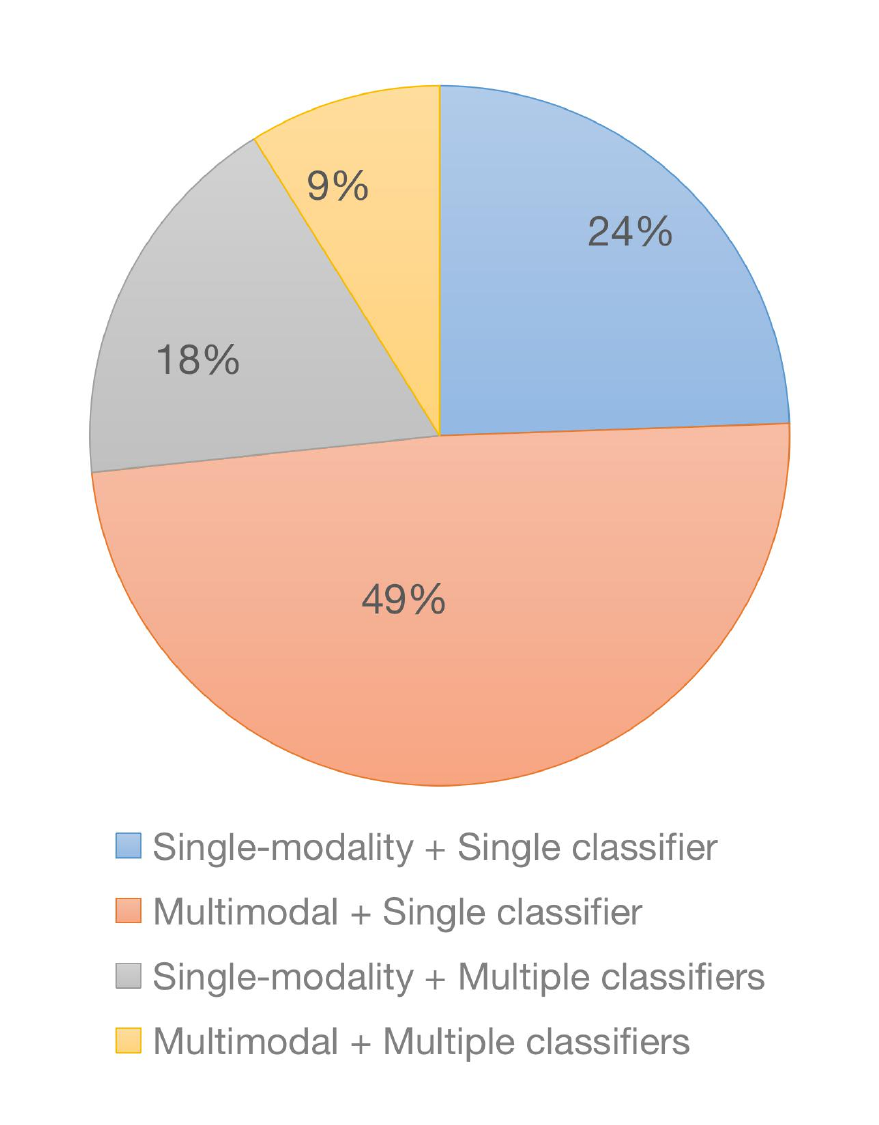}
\caption{The composition of the proportion of modalities and classifiers/clusterers.}
\label{fig:percentage_2}
\end{figure}

We have presented the common basic belief assignment methods to generate mass functions for medical images in Section \ref{sec:bba}. The most interesting way now to analyze those segmentation methods is to classify them according to the stage at which they fuse mass functions. Figure~\ref{fig:BFT-based-seg} gives an overview of the BFT-based medical image segmentation methods classified according to the fusion operation they perform. We first classify methods by the number of classifiers/clusterers, i.e., whether the fusion of mass functions is performed at the classifiers/clusterers level. Then we consider the number of input modalities, i.e.,~whether the fusion of mass functions is performed at the modality level. It should be noted that the medical image segmentation with single-modality input and single classifier/clusterer can also have a fusion operation, which is performed at the pixel/voxel level. Figure~\ref{fig:percentage_2} gives the proportions of the BFT-based segmentation methods if multiple classifiers/clusterers or multimodal medical images are used. 73\% of the methods use a single classifier or cluster. Among these methods, 24\% take single-modality medical images as input, and 49\% take multimodal medical images as input. The remaining 27\% of the methods use several classifiers or clusterers. Among those methods, 18\% take single-modality medical images as input, against 9\% for multimodal medical images. In Sections \ref{sec: single-clasisifer} and \ref{sec: multi-clasisifer} we give more details about the fusion operations and discuss their performances.  

\section{Segmentation with a single classifier or clusterer}
 \label{sec: single-clasisifer}
The medical image segmentation methods summarized in this section focus on using a single classifier or clusterer. To simplify the introduction, we classify them into single-modality and multimodal inputs according to whether the authors treat the input images as single or multiple source inputs. We will introduce them in Sections \ref{subsec: single} and \ref{subsec: multi}, respectively. 


\subsection{Single-modality evidence fusion}
\label{subsec: single}
\begin{figure}
\centering
\includegraphics[width=\textwidth]{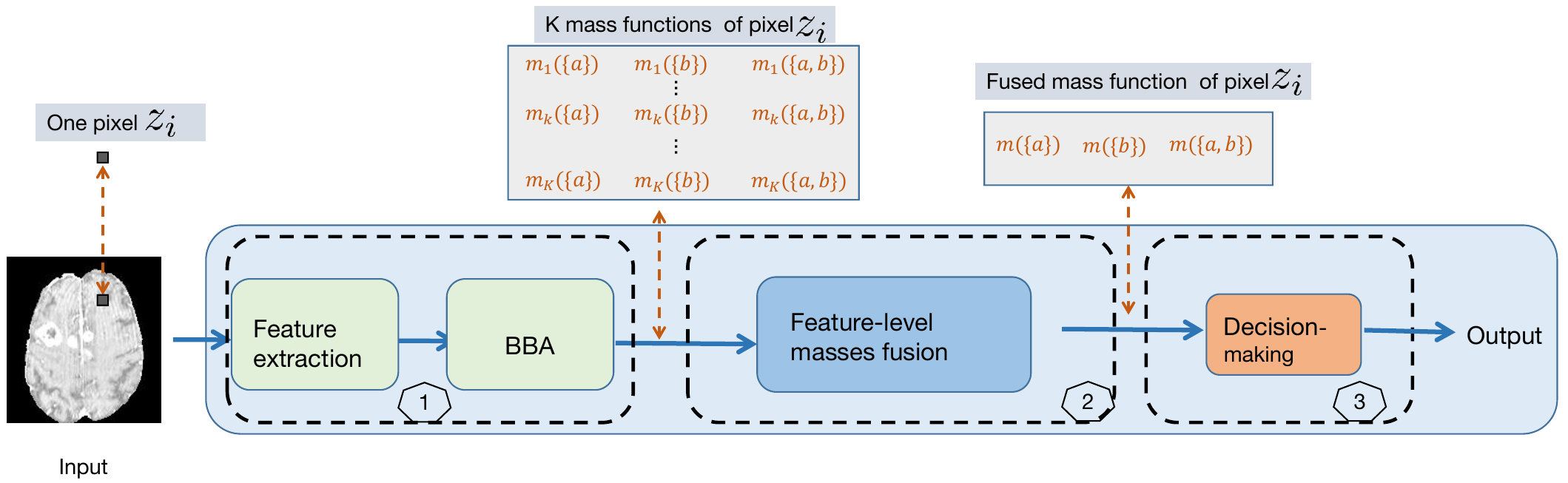}
\caption{Example framework of single-modality evidence fusion with single classifier. The segmentation process is composed of three steps: (1) image information is transferred into the feature extraction block, and some BBA methods are used to get feature-level mass functions; (2) Dempster's rule is used to fuse feature-level mass functions; (3) decision-making is made based on the fused mass functions to output a segmented mask. We take only one pixel $z_i$ as an example and show how BFT works on pixel-level evidence fusion under a binary segmentation task to simplify the process. For each pixel $z_i$, we can obtain $K$ mass functions from BBA. After feature-level evidence fusion, a fused mass function is assigned to pixel $z_i$ to represent the degree of belief this pixel belongs to classes a, b and ignorance.}  
\label{fig:fig6}
\end{figure}

Figure~\ref{fig:fig6} shows the framework of single-modality evidence fusion with a single classifier (we only take the classifier as an example in the rest of this chapter to simplify the explanation). The inputs of the framework are single-modality medical images. The segmentation process comprises three steps: mass calculation (including feature extraction and BBA), feature-level mass fusion, and decision-making. Since decision-making is not the focus of this chapter, we will not go into details about it; readers can refer to Ref.~\cite{denoeux2019decision} for a recent review of decision methods based on belief functions.

\begin{itemize}
    \item[(1)]First, the mass calculation step assigns each pixel/voxel {$K$} mass functions based on the evidence of $K$ different BBA methods,  $K$ input features, $K$ nearest neighbors, or $K$ prototype centers. 
    \item[(2)]Dempster's rule is then used for each pixel/voxel to fuse the feature-level mass functions. 
    \item[(3)]Finally, decisions are made to obtain final segmentation results.
\end{itemize}  
Here, a feature extraction method is used to generate MVs (corresponding to traditional medical image segmentation methods) or deep features (corresponding to deep learning-based medical image segmentation methods). The feature extraction method could be intensity-based methods such as threshold intensity, probabilistic-based methods such as SVM,  fuzzy set-based methods such as FCM, etc. There are various basic belief assignment methods, therefore we introduce them in specific tasks in the following. In general, the methods introduced in this section focus on feature-level evidence fusion. Table \ref{tab:fusion1} shows the segmentation methods with a single-modality input and classifier/cluster that focus on feature-level evidence fusion.

\begin{table}
  \centering
  \caption{Summary of BFT-based medical image segmentation methods with single-modality inputs and a single classifier/clusterer. }
  \scalebox{0.63}{
  \begin{tabular}{llllllll}
 \hline
 \multicolumn{1}{l}{Publications}&
Input image type& Application& BBA method \\

\hline
 \cite{SuhKnowledge}&MR images&cardiac segmentation& Shafer's model\\
\cite{gerig2000exploring}&MR images&brain tissue segmentation&BFOD\\
 \cite{1509843} &CT&lung and spinal canal segmentation &Shafer's model\\
 \cite{vannoorenberghe2006belief} &CT&thoracic segmentation& EKNN \\
 \cite{derraz2013globally} &optical imaging with color&cell segmentation &Appriou's model\\
 \cite{derraz2013image}&optical imaging with color&retinopathy segmentation &Appriou's model \\
 \cite{derraz2013globally}&optical imaging with color& retinopathy segmentation &Appriou's model\\
 \cite{lian2017spatial} &FDG-PET&lung tumor segmentation& ECM\\
 \cite{lian2017tumor} &FDG-PET&lung tumor segmentation &ECM\\
\cite{huang2021evidential}&PET-CT&lymphoma segmentation&ENN \\
\cite{huang2022lymphoma}&PET-CT&lymphoma segmentation&ENN, RBF\\
\hline
\end{tabular}
}
\label{tab:fusion1}
\end{table}

In~\cite{SuhKnowledge}, Suh et al. developed a knowledge-based endocardial and epicardial boundary detection and segmentation system with cardiac MR image sequences. Pixel and location information were mapped into mass functions by Shafer's model~\cite{shafer1976mathematical} (see Section \ref{subsubsection:lik_methods}). The  mass functions from the two sources  were fused by using Dempster's rule (see Section \ref{sec:fusion}). Experiments were applied to cardiac short-axis images and obtained an excellent success rate ($> 90\%$). However, this work only focused on cardiac boundary detection and did not discuss the details of the heart. In~\cite{1509843}, Vauclin et al. proposed a BFT-based lung and spinal canal segmentation model. The k-means clustering algorithm was first used to perform a pre-segmentation. Then a 3D filter exploits the results of the pre-segmentation to compute the MVs from spatial neighbors using Shafer's model. Segmentation results showed the credal partition permits the reduction of the connection risks between the trachea and the lung when they are very close and between the left and right lungs at the anterior or posterior junctions.

In~\cite{gerig2000exploring}, Gerig et al. presented a method for automatic segmentation and characterization of object changes in time series of 3D MR images. A set of MVs was derived from time series according to brightness changes. The BFOD transformation (see Section \ref{subsubsection: bfod}) was used to map the obtained features into mass functions. Then the set of evidence was combined by Dempster's rule (See Section \ref{sec:fusion}). A comparison with results from alternative segmentation methods revealed an excellent sensitivity and specificity performance in the brain lesion region. The author also pointed out that better performance could be obtained with multimodal and multiple time-series evidence fusion. Simulation  results showed that about 80\% of the implanted voxels could be detected for most generated lesions.

In~\cite{vannoorenberghe2006belief}, Vannoorenberghe et al. presented a BFT-based thoracic image segmentation method. First, a K-means algorithm performed coarse segmentation on the original CT images. Second, the EKNN rule (see Section \ref{subsubsec: distance}) was applied by considering spatial information and calculating feature-level mass functions. The authors claimed that using this segmentation scheme leads to a complementary approach combining region segmentation and edge detection. Experimental results showed promising results on 2D and 3D CT images for lung segmentation.

In~\cite{derraz2013globally}, Derraz et al. proposed an active contour
(AC)-based~\cite{chan2000active} global segmentation method for vector-valued image that incorporated both probability and mass functions. All features issued from the vector-valued image were integrated with inside/outside descriptors to drive the segmentation results by maximizing the Maximum-Likelihood distance between foreground and background. Appriou's second model \eqref{eq:appriou2} was used to calculate the imprecision caused by low contrast and noise between inside and outside descriptors issued from the multiple channels. Then the fast algorithm based on Split Bregman~\cite{goldstein2009split} was used for final segmentation by forming a fast and accurate minimization algorithm for the Total Variation (TV) problem. Experiments were conducted on color biomedical images (eosinophil, lymphocyte, eosinophil, monocyte, and neutrophil cell~\cite{mohamed2012enhanced}) and achieve around 6\% improvements by using F-score on five image groups. In the same year, Derraz et al. proposed a new segmentation method~\cite{derraz2013image} based on Active Contours (AC) for the vector-valued image that incorporates Bhattacharyya's distance~\cite{michailovich2007image}. The only difference is that the authors calculate the probability functions by Bhattacharyya distance instead of the Maximum-Likelihood distance in this paper. The performance of the proposed algorithm was demonstrated on the retinopathy dataset~\cite{quellec2008optimal,niemeijer2009retinopathy}, showing an increase of 3\% in F-score compared with the best-performed methods.

In~\cite{lian2017spatial}, Lian et al. introduced a tumor delineation method in fluorodeoxyglucose positron emission tomography (FDG-PET) images by using spatial ECM~\cite{lelandais2014fusion} with adaptive distance metric, a variant of the ECM algorithm recalled in Section \ref{subsubsec:ecm}. The authors proposed the adaptive distance metric to extract the most valuable features, and spatial ECM was used to calculate mass functions. Compared with ECM and spatial ECM, the proposed method showed a 14\% and 10\% increase in Dice score when evaluated on the FDG-PET images of non-small cell lung cancer (NSCLC) patients, which constitutes a very good performance.

In~\cite{huang2021evidential}, we proposed a 3D PET/CT lymphoma segmentation framework with BFT and deep learning. In this paper, the PET and CT images were concatenated as a signal modal input method and transferred into UNet to get high-level semantic features. Then the ENN classifier (see Section \ref{subsubsec: distance}) was used to map high-level semantic features into mass functions by fusing the contribution of $K$ prototypes. Moreover, the segmentation uncertainty was considered in this paper with an uncertainty loss during training. Based on the first work, we verified the similarity of radial basis function (RBF, see Section \ref{subsubsec: rbf} ) network and ENN in uncertainty quantification and merged them with the deep neural network (UNet, see Section \ref{subsec:Unet}) for lymphoma segmentation~\cite{huang2022lymphoma}. The segmentation performance confirmed that RBF is an alternative approach of ENN to act as an evidential classifier and showed that the proposal outperforms the baseline methods and the state-of-the-art both in accuracy and reliability. We will introduce more detail about this work in Chapter \ref{Chapter5}.

Before 2020, the BFT-based medical image segmentation methods with single-modality medical used low-level image features, e.g.,~grayscale and shape features, to generate mass functions, which limits the segmentation accuracy. However, none of them discussed the reliability of the segmentation results. We are the first to merge BFT with a deep segmentation model (UNet) and optimize the mass functions with some learning algorithms~\cite{huang2021evidential}. Based on this, we further discuss the relationship between segmentation accuracy and reliability in~\cite{huang2022lymphoma}, which takes a new direction to study reliable medical image segmentation methods and bridge the gap between experimental results and clinical application.

\subsection{Multimodal evidence fusion}
\label{subsec: multi}
Single-modality medical images often do not contain enough information to present the information about the disease and are often tainted with uncertainty. In addition to feature-level evidence fusion, the fusion of multimodal evidence is also important to achieve accurate and reliable medical image segmentation performance. Approaches for modality-level evidence fusion can be summarized into three main categories according to the way they calculate the evidence: probabilistic-based fusion, fuzzy set-based fusion, and BFT-based fusion. The development of convolution neural networks (CNNs) further contributes to the probabilistic-based fusion methods~\cite{zhou2019review}, which can be summarized into image-level fusion (e.g.,~data concatenation~\cite{peiris2021volumetric}), feature-level fusion (e.g.,~attention mechanism concatenation~\cite{zhou2020fusion,zhou2022tri}), and decision-level fusion (e.g.,~model ensembles~\cite{kamnitsas2017ensembles}). However, none of those methods considers the conflict between sources of evidence, i.e., the modality-level uncertainty is not well studied, which limits the reliability and explainability of the performance.

\begin{figure}
\centering
\includegraphics[width=\textwidth]{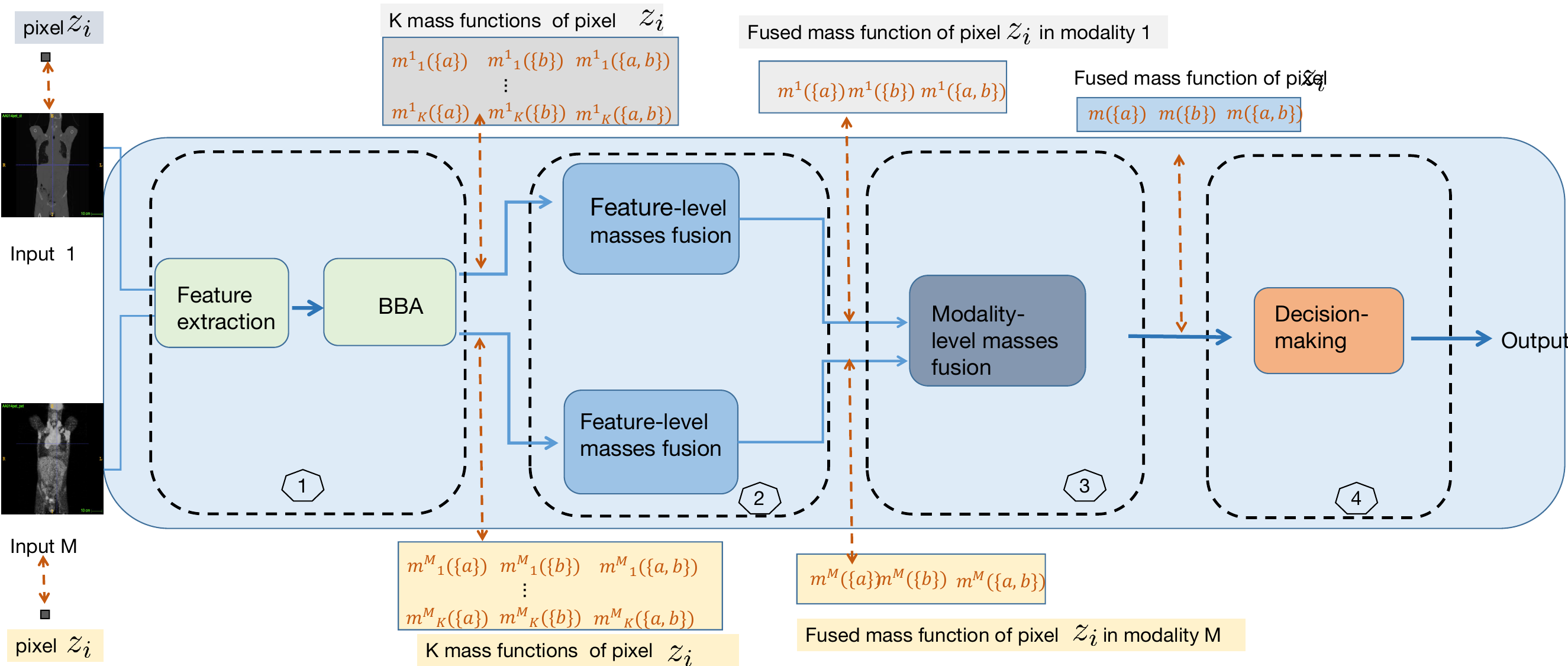}
\caption{Example framework of multimodal evidence fusion with a single classifier. The segmentation process is composed of four steps: (1) for each modal input, image features are fed into the classifier, and some BBA methods are used to get feature-level mass functions; (2) inside each modal, the calculated feature-level mass functions are fused by using Dempster's rule to generate modality-level mass functions; (3) between the modalities, the calculated modality-level mass functions are fused by using Dempster's rule again; (4) decision-making is made based on the fused mass functions to output a segmented mask.}  
\label{fig:fig7}
\end{figure}

This section focuses on the BFT-based segmentation methods with modality-level evidence fusion. Figure~\ref{fig:fig7} shows an example framework of multimodal evidence fusion with a single classifier. We separate the segmentation process into four steps: mass calculation, feature-level evidence fusion (optional), modality-level evidence fusion, and decision-making. Compared with single-modality evidence fusion reviewed in Section \ref{subsec: single}, multimodal evidence fusion focuses here not only on feature-level but also on modality-level evidence fusion. It should be noted that feature-level evidence fusion is not necessary in this case. Multimodal evidence fusion is the most popular application for BFT in the medical image segmentation domain. Therefore we classify those methods according to their input modal for better analysis.  Table \ref{tab:fusion2} summarizes the segmentation methods with multimodal inputs and a single classifier/cluster with the main focus on modality-level evidence fusion.

\begin{table}
  \centering
  \caption{Summary of BFT-based medical image segmentation methods with multimodal input and single classifier/clusterer. }
  \scalebox{0.55}{
  \begin{tabular}{llllllll}
 \hline
 \multicolumn{1}{l}{Publications}&Input image type& Application & BBA method \\

\hline
\cite{vannoorenberghe1999dempster}&optical imaging with color&skin cancer segmentation& GD-based model\\
 \cite{bloch1996some}&multimodal MR images&brain tissue segmentation& prior knowledge \\
 \cite{taleb2002information}&multimodal MR images&vertebrae segmenta& threshold+contour distance \\
 \cite{chaabane2009relevance}&optical imaging with color&cell lesion segmentation & Appriou's model \\
 \cite{zhu2002automatic} &multimodal MR images& brain tissue segmentation &Zhu's model \\
 \cite{bloch2008defining}&multimodal MR images &brain tissue segmentation& prior knowledge \\
 \cite{chaabane2011new}  &optical imaging with color&cell lesion segmentation&Aprrious's method\\
 \cite{ghasemi2012brain}&multimodal MR images& brain tissue segmentation  &Ratio MV\\
 \cite{harrabi2012color} & optical imaging with color& breast cancer segmentation &GD-based model\\
 \cite{lelandais2012segmentation} &PET &biological target tumor segmentation &ECM\\
 \cite{Rui2013Lesion} &multimodal MR images &cerebral infraction segmentation &Zhu's model \\
 \cite{ghasemi2013novel} &multimodal MR images& brain tumor segmentation & Ratio MV\\
 \cite{lelandais2014fusion}& multi-tracer PET&biological target tumor segmentation& ECM\\
 \cite{makni2014introducing}&multi-parametric MR image & prostate zonal anatomy &ECM \\
 \cite{derraz2015joint}  &PET/CT &lung tumor segmentation  & Appriou's model\\
 \cite{trabelsi2015skin} &optical imaging with color &skin lesion segmentation&None \\
 \cite{xiao2017vascular} &multimodal MR images& vascular segmentation &GD-based model\\
\cite{lian2017accurate} &FDG-PET/CT& lung tumor segmentation&ECM \\
 \cite{lian2018unsupervised} &FDG-PET/CT &lung cancer segmentation  &ECM\\
 \cite{tavakoli2018brain} &multimodal MR images &brain tissue segmentation &Ratio MV\\
\cite{lima2019modified} &multimodal MR images& brain tissue segmentation&Ratio MV \\
\cite{lian2018joint} &PET/CT& lung tumor segmentation &ECM\\
\hline
\end{tabular}
}
\label{tab:fusion2}
\end{table}

\paragraph*{Fusion of multimodal MR images}
 
In~\cite{bloch1996some}, Bloch first proposed a BFT-based dual-echo MR pathological brains tissue segmentation model with uncertainty and imprecision quantification. The author assigned mass functions based on a reasoning approach that uses gray-level histograms provided by each image to choose focal elements. After defining mass functions, Dempster's rule combined the mass from dual-echo MR images for each pixel. Based on the first work with BFT, in~\cite{bloch2008defining}, Bloch proposed to use fuzzy mathematical morphology~\cite{bloch1995fuzzy}, i.e.,~erosion and dilation, to generate mass functions by introducing imprecision in the probability functions and estimating compound hypotheses. Then Dempster's rule (see Section \ref{sec:fusion}) is used to fuse mass functions from multimodal images. It should be noted that in this paper, the strong assumption is made that it is possible to represent imprecision by a fuzzy set, also called the structuring element. Application examples on dual-echo MR image fusion showed that the fuzzy mathematical morphology operations could represent either spatial domain imprecision or feature space imprecision (i.e., gray levels features). The visualized brain tissue segmentation results show the robustness of the proposed method.

As we mentioned in Section \ref{sec:bba}, Zhu et al. proposed modeling mass functions in BFT using FCM and spatial neighborhood information for image segmentation \cite{zhu2002automatic}. The visualized segmentation results on MR brain images showed that the fusion-based segmented regions are relatively homogeneous, enabling accurate measurement of brain tissue volumes compared with single-modality input MR image input.

In~\cite{ghasemi2012brain}, Ghasemi et al. presented a brain tissue segmentation approach based on FCM and BFT. The authors used the FCM to model two different input features: pixel intensity and spatial information, as membership values (MVs). Then for each pixel, the Ratio MV transformation (see Section \ref{subsub:MVs}) was used to map MVs into mass functions. Last, the authors used Dempster's rule (see Section \ref{sec:fusion}) to fuse intensity-based and spatial-based mass functions to get final segmentation results. Compared with FCM, the authors reported an increase in Dice score and accuracy. As an extension of~\cite{ghasemi2012brain}, Ghasemi et al. proposed an unsupervised brain tumor segmentation method that modeled pixel intensity and spatial information into mass functions with Ratio MV transformation and fused the two mass functions with Dempster's rule in~\cite{ghasemi2013novel}. 

In~\cite{Rui2013Lesion}, Wang et al. proposed a lesion segmentation method for infarction and cytotoxic brain edema. The authors used a method similar to Zhu's model (see Section \ref{subsub:MVs}) to define simple and double hypotheses. FCM was used first to construct the mass functions of simple hypotheses $\{a\}$ and $\{ b\}$. Then masses were assigned to double hypotheses as $m(\{a, b\})=\frac{1}{4} \times \frac{min(m(\{a\}),m(\{b\}))}{m(\{a\})+m(\{b\})}$. Finally, the authors used Dempster's rule for modality-level evidence fusion. The results showed that infarction and cytotoxic brain edema around the infarction could be well segmented by final segmentation.

In~\cite{makni2014introducing}, Makni et al. introduced a  multi-parametric MR image segmentation model by using spatial neighborhood in ECM for prostate zonal anatomy. The authors extended the ECM (see Section \ref{subsubsec:ecm}) with neighboring voxels information to map multi-parametric MR images into mass functions. Then they used prior knowledge related to defects in the acquisition system to reduce uncertainty and imprecision. Finally, the authors used Dempster's rule to fuse the mass functions from the multi-parametric MR images. The method achieved good performance on prostate multi-parametric MR image segmentation.

In~\cite{tavakoli2018brain}, Tavakoli et al. proposed a segmentation method based on the evidential fusion of different modalities (T1, T2, and Proton density (PD)) for brain tissue. The authors used FCM to get MVs and used the Ratio MV transformation (see Section \ref{subsub:MVs}) to transform the clustering MVs into mass functions. The authors first formed the belief structure for each modal image and used Dempster's rule to fuse the three modalities' mass functions of T1, T2, and PD. Compared with FCM, this method achieved a 5\% improvement in the Dice score. Based on Tavakoli's method~\cite{tavakoli2018brain}, Lima et al. proposed a modified brain tissue segmentation method in~\cite{lima2019modified}. The authors tested their method with four modality inputs: T1, T2, PD, and Flair. The reported results outperformed both the baseline method FCM and Tavakoli's method with four-modality evidence fusion. 

In~\cite{xiao2017vascular}, Xiao et al. proposed an automatic vascular segmentation algorithm, which combines the grayscale and shape features of blood vessels and extracts 3D vascular structures from the head phase-contrast MR angiography dataset. First, grayscale and shape features are mapped into mass functions by using the GD-based method (see Section \ref{subsubsection: gd}). Second, a new reconstructed vascular image was established according to the fusion of vascular grayscale and shape features by Dempster's rule. Third, a segmentation ratio coefficient was proposed to control the segmentation result according to the noise distribution characteristic. Experiment results showed that vascular structures could be detected when both grayscale and shape features are robust. Compared with traditional grayscale feature-based or shape feature-based methods, the proposal showed better performance in segmentation accuracy with the decreased over-segmentation and under-segmentation ratios by fusing two sources of information.

Since Bloch's early work fully demonstrated the advantages of BFT in modeling uncertain and imprecision, introducing partial or global ignorance, and fusing conflicting evidence in a multimodal MR images segmentation task, researchers in this domain have gone further to explore the advantages of BFT in multi MR image fusion. Among these research works, FCM is the most popular clustering algorithm to map input information into MVs. Ration MV transformation or Zhu's model (see Section \ref{subsub:MVs}) is usually used to generate mass functions from MVs. Apart from these two-step methods, the GD-based model (see Section \ref{subsubsection: gd}) can also be used to generate mass functions directly. 

\paragraph*{Fusion of RGB channels}

In~\cite{vannoorenberghe1999dempster}, Vannoorenberghe et al. pointed out that taking the R, G, and B channels as three independent information sources can be limited and nonoptimal for medical image tasks and proposed a color image segmentation method based on BFT. They calculated the degree of evidence by mapping R, G, and B channel intensity into mass functions using the Gaussian distribution information (similar to GD-based model) with an additional discounting operation). Then three pieces of evidence were fused with Dempster's rule (see Section \ref{sec:fusion}). The proposed segmentation method was applied to biomedical images to detect skin cancer (melanoma). Experiments showed a significant part of the lesion is correctly extracted from the safe skin. The segmentation performance is limited by feature representation, e.g., some regions correspond to pixels that cannot be classified as either the safe skin or the lesion because only the pixel-level feature is insufficient for hard-example segmentation.

In~\cite{chaabane2009relevance} Chaabane et al. proposed a color medical image segmentation method based on fusion operation. Compared to~\cite{vannoorenberghe1999dempster}, the authors first modeled  probabilities for each region by a Gaussian distribution, then used Appriou's second model \eqref{eq:appriou2} to map probability into mass functions. Dempster's rule then combined the evidence from the three color channels. Compared with single-channel segmentation results, the fused results achieved a 10\% increase in segmentation sensitivity. 

Different from the methods described in~\cite{vannoorenberghe1999dempster} and \cite{chaabane2009relevance} that decompose color images into R, G, B channels, Harrabi et al.~\cite{harrabi2012color} presented a color image segmentation method that represents the color image with 6 color spaces (RGB, HSV, YIQ, XYZ, LAB, LUV). The segmentation operation is based on multi-level thresholding and evidence fusion techniques. First, the authors identified the most significant peaks of the histogram by multi-level thresholding with the two-stage Otsu optimization approach. Second, the GD-based model (see Section \ref{subsubsection: gd}) was used to calculate the mass functions for each color space. Then the authors used Dempster's rule to combine six sources of information. Compared with single color spaces, such as RGB and HSV, the combined result taking into account six color spaces, has a
significant increase in segmentation sensitivity, for example, an increase of 4\% and
7\% as compared to RGB and HSV, respectively.

In~\cite{trabelsi2015skin}, Trabelsi et al. applied BFT in optical imaging with color to improve skin lesion segmentation performance. The authors decomposed color images into R, G, and B channels and applied the FCM method on each channel to get probability functions for pixel $x$ in each color space. In this paper, the authors take the probability functions as mass functions and calculate the orthogonal sum of the probability functions from the three-channel images. Even though experiments showed about 10\% improvements in segmentation accuracy compared with single-channel results, this work does not harness the full power of BFT as it only considers Bayesian mass functions.

In general, the BFT-based RGB medical image segmentation approaches are used to generate mass functions from possibility distributions, e.g.,~Gaussian distribution and Possibility C-means distribution, and fuse them by Dempster's rule. Though the authors claimed they could get better performance compared with single color input, the segmentation performance is limited by features because gray-scale and intensity features are not robust and efficient in representing image information. Further research could take both feature extraction and uncertainty quantification into consideration, e.g., a deep feature extraction model with an evidential classifier, to improve the performance.

\paragraph*{Fusion of PET/CT}

In~\cite{lelandais2012segmentation}, Lelandais et al. proposed a BFT-based multi-trace PET images segmentation method to segment biological target volumes. First, the authors used a modified FCM algorithm with the discounting algorithm to determine mass functions. The modification integrated a disjunctive combination of neighboring voxels inside the iterative process. Second, the operation of reduction of imperfect data was conducted by fusing neighboring voxels using Dempster's rule (see Section \ref{sec:fusion}). Based on this first work, Lelandais et al. proposed an ECM-based fusion model~\cite{lelandais2014fusion} for biological target volume segmentation with multi-tracer PET images. The segmentation method introduced in this paper is similar to the one introduced in~\cite{makni2014introducing} with a different application.

In~\cite{derraz2015joint}, Derraz et al. proposed a multimodal tumor segmentation framework for PET and CT inputs. Different from Lelandais et al.'s work that uses ECM or optimized ECM to generate mass functions directly, the authors construct mass functions in two steps. They first proposed a NonLocal Active Contours (NLAC) based variational segmentation framework to get probability results. Then, similar to the authors' previous
work~\cite{derraz2013globally,derraz2013image}, they used Appriou's second model~\eqref{eq:appriou2} to map MVs into mass functions. Then Dempster's rule was used to fuse the mass functions from PET and CT modalities. The framework was evaluated on a lung tumor segmentation task. Compared with the state-of-the-art methods, this framework yielded the highest Dice score for tumor segmentation. 

Based on Lelandais et al.'s work, Lian et al. proposed a tumor segmentation method~\cite{lian2017accurate} using Spatial ECM~\cite{lelandais2014fusion} with Adaptive Distance Metric~\cite{lian2017spatial} in FDG-PET images with the guidance of CT images. Based on the first work, Lian et al. proposed a co-segmentation method of lung tumor
segmentation~\cite{lian2018joint,lian2018unsupervised}. They took PET and CT as independent inputs and used ECM (see Section \ref{subsubsec:ecm}) to generate mass functions. At the same time, an adaptive distance metric was used to quantify clustering distortions and spatial relationships during the evidential clustering procedure. Then Dempster's rule was used to fuse mass functions from PET and CT modalities. The quantitative and qualitative evaluation results showed superior performance compared with single-modality segmentation results with an increase of 1\% and 58\% in PET and CT in Dice scores, respectively.

ECM is the most common BBA method to generate mass functions for PET/CT medical image segmentation approaches. Similar to the BFT-based RGB medical image segmentation methods, the segmentation performance here is limited by feature extraction methods. Further research could build on recent advancements in deep feature representation and combine ECM with deep neural networks to learn mass feature representation. 

\section{Segmentation with several classifiers or clusterers}
\label{sec: multi-clasisifer}

It is common practice that two or more physicians cooperate  for disease diagnosis, which can minimize the impact of physicians' misjudgments. Similarly, combining the results from multiple decision mechanisms as well as addressing the conflicts is critical to achieving a more reliable diagnosis. This section introduces the BFT-based medical image segmentation methods with several classifiers or clusterers. We follow the same approach as in Section \ref{sec: single-clasisifer} and separate the methods into single-modality and multimodal evidence fusion reviewed, respectively, in Sections \ref{subsec: single-modality} and \ref{subsec: multi-modality}.
 
\subsection{Single-modality evidence fusion}
\label{subsec: single-modality}
\begin{figure}
\centering
\includegraphics[width=\textwidth]{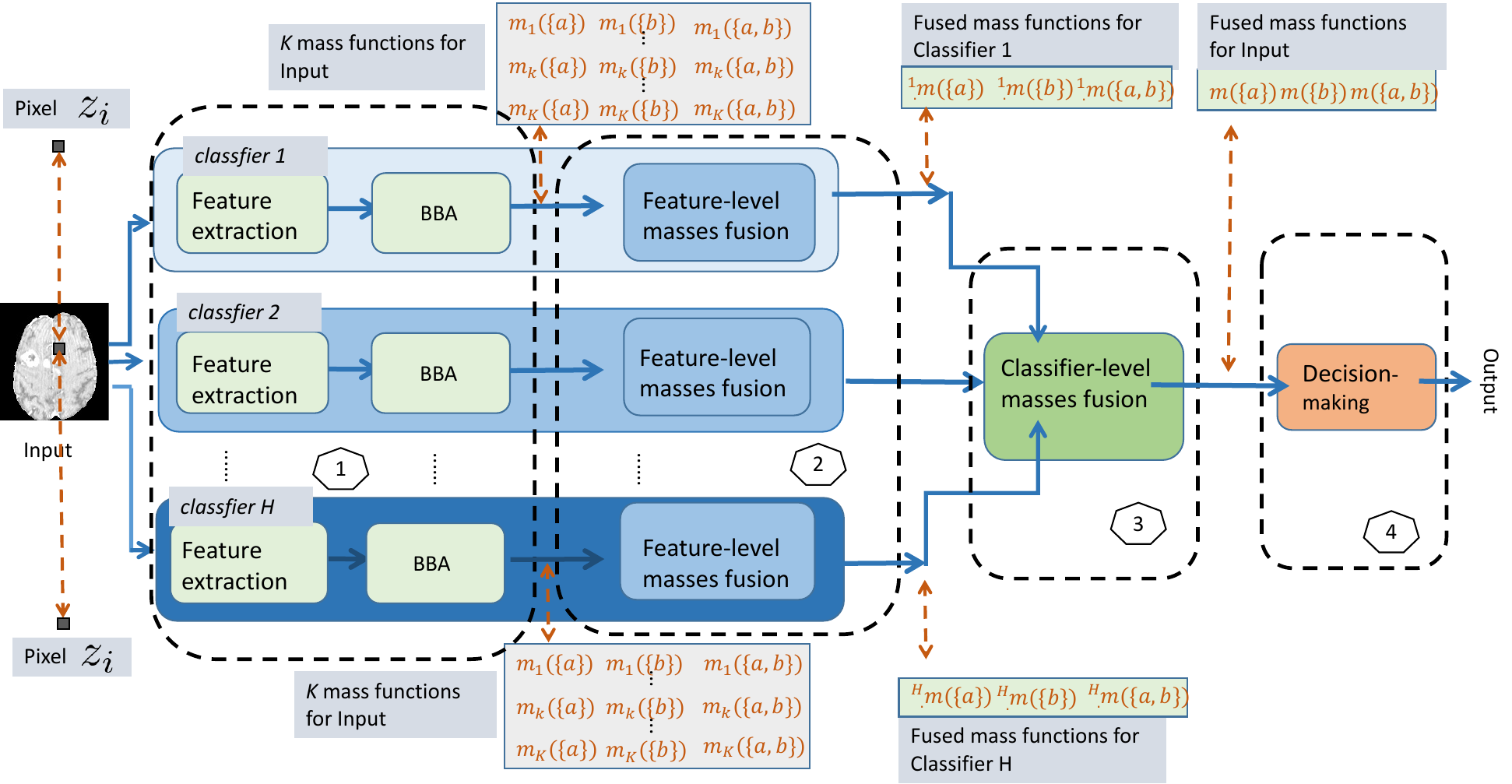}
\caption{Example framework of single-modality evidence fusion with several classifiers. The segmentation process is composed of four steps: (1) for each pixel, image features are transferred into the different classifiers, and some BBA methods are used to get feature-level mass functions; (2) inside each classifier, the calculated feature-level mass functions are fused by using Dempster's rule to generate classifier-level mass functions; (3) between the classifiers, the calculated classifier-level mass functions are fused by using Dempster's rule again; (4) decision-making is made based on the final fused mass functions and output a segmented mask.}
\label{fig:fig8}
\end{figure}

\begin{table}
  \centering
  \caption{Summary of BFT-based medical image segmentation methods with multimodal inputs and several classifiers/clusterers. }
  \scalebox{0.65}{
  \begin{tabular}{llllllll}
 \hline
 \multicolumn{1}{l}{Publications} &Input image type& Application&BBA method \\
\hline
 \cite{capelle2002segmentation}&MR images &brain tumor segmentation&EKNN\\
 \multirow{2}*{\cite{capelle2004evidential}}&\multirow{2}*{MR images}& \multirow{2}*{brain tumor segmentation}&EKNN+Shafer's\\
 &&&model+Appriou's model\\
 \cite{barhoumi2007collaborative} &optical imaging with color &skin lesion malignancy tracking& None \\
 \cite{guan2011study}&MR images& brain tissue segmentation&Zhu's model\\
 \cite{ketout2012improved} &optical imaging with color& endocardial contour detection& threshold \\
 \multirow{2}*{\cite{wen2018improved}} &\multirow{2}*{MR images} &\multirow{2}*{brain tissue segmentation} &Zhu's model+\\
 &&&GD-based model\\
 \cite{george2020breast} &optical imaging with color &breast cancer segmentation&Discounting \\
 \cite{huang2021belief} &PET-CT &lymphoma segmentation &ENN\\

\hline
\end{tabular}
}
\label{tab:fusion3}
\end{table}
Compared with the methods presented in Section \ref{subsec: single}, the methods introduced in this section focus on feature-level and classifier-level evidence fusion, which aims to minimize the impact of misjudgments caused by a single model's inner shortcomings. Figure \ref{fig:fig8} shows an example of a medical image segmentation framework with single-modality input and several classifiers. We separate the segmentation process into four steps: mass calculation, feature-level evidence fusion (optional), classifier-level evidence fusion, and decision-making. Similar to Figure \ref{fig:fig6}, we assume for each pixel, we can obtain $K$ mass functions by one classifier. After feature-level evidence fusion, for each pixel $z_i$, we can get $H$ mass functions corresponding to $H$ classifiers. Then we fuse the $H$ mass functions and assign a fused mass function to the pixel $z_i$, representing the degree of belief this pixel belongs to classes a, b and ignorance. Table \ref{tab:fusion3} summarizes the segmentation methods with single-modality inputs and several single classifiers/clusterers with the main focus on classifier/clusterer level evidence fusion.

In~\cite{capelle2002segmentation}, Capelle et al. proposed a segmentation scheme for MR images based on BFT. The authors used the Evidential K-NN rule recalled in Section \ref{subsubsec: distance} to map image features into mass functions. Then, they applied the evidential fusion process to classify the voxels. Based on this first work, Capelle et al. later proposed an evidential segmentation scheme of multimodal MR images for brain tumor detection in~\cite{capelle2004evidential}. This work focused on analyzing different evidential modeling techniques and on the influence of the introduction of spatial information to find the most effective brain tumor segmentation method. Three different BFT-based models: the distance-based BFT model (EKNN, see Section \ref{subsubsec: distance}), the likelihood function-based BFT method (Shafer's model, see Section \ref{subsubsection:lik_methods}), and Appriou's first model~\eqref{eq:15} were used to model information; Dempster's rule (see Section \ref{sec:fusion}) was then used to combine the three mass functions. This study concluded that neighborhood information increases the evidence of class membership for each voxel, thus making the final decision more reliable. Experimental results showed better segmentation performance compared with the state-of-the-art methods when the paper was published.


In~\cite{taleb2002information}, Taleb-Ahmed proposed a segmentation method for MR sequences of vertebrae in the form of images of their multiple slices with BFT. The authors used three different classifiers to calculate three kinds of mass functions. Firstly, the authors used gray-level intensity and standard deviation information to calculate two pixel-level mass functions with two fixed thresholds. Then the distance between two matching contours of consecutive slices ($P$ and $Q$) was used to calculate contour-level mass functions as follows:
\begin{subequations}
\begin{equation}
m(\{S_{PQ}\})=\begin{cases}
 1-e^{-\eta \left |   d(P_{i},Q_{i})-\beta \right |   } & \text{if } d(P_{i},Q_{i}) \in [\rho,\beta],\\ 
 0 & \text{otherwise}
 \end{cases}
 \end{equation}
 \begin{equation}
 m(\overline{\{S_{PQ}\}})= 
 \begin{cases}
 1-e^{-\eta \left |  d(P_{i},Q_{i})-\beta \right |   }& \text{if } d(P_{i},Q_{i}) \in (\beta,+\infty),\\ 
 0 & \text{otherwise}
 \end{cases}
 \end{equation}
  \begin{equation}
 m(\Omega)=e^{-\eta \left |   d(P_{i},Q_{i})-\beta \right |  },
\end{equation} 
 \label{eq:36}
\end{subequations}
where $P_{i}$ and $Q_{i}$ are two matching points of the slices $P$ and $Q$, $d(P_{i},Q_{i})$ is the corresponding distance; $\Omega=\{S_{PQ},\overline{S_{PQ}}\}$, $S_{PQ}$ means that  points $P_{i}$ and $Q_{i}$  both belong to cortical osseous. Parameter $\beta$ represents the tolerance that the expert associates with the value $d(P_{i}, Q_{i})$, $\rho$ is the inter-slice distance and $\eta$ makes it possible to tolerate a greater inaccuracy in the geometrical resemblance of two consecutive contours. Dempster's rule was then used to combine the three mass functions for final segmentation.

In~\cite{barhoumi2007collaborative}, Barhoumi et al. introduced a new collaborative computer-aided diagnosis system for skin malignancy tracking. First, two diagnosis opinions were produced by perceptron neural network classification and content-based image retrieval (CBIR) schemes. Then Dempster's rule was used to fuse the two opinions to achieve a final malignancy segmentation. Simulation results showed that this BFT-based combination could generate accurate diagnosis results. In this work, the frame of discernment is composed of two elements, and the mass functions are Bayesian.

In~\cite{guan2011study}, Guan et al. proposed a human brain tissue segmentation method with BFT. The authors first used Markov random forest (MRF)~\cite{li2009markov} for spatial information segmentation and then a two-dimensional histogram (TDH) method of fuzzy clustering~\cite{duan2008multi} to get a vague segmentation. Then a redundancy image was generated, representing the difference between the MRF and TDH methods, and Zhu's model (see Section \ref{subsub:MVs}) was used to calculate mass functions. Finally, Dempster's rule fused the two segmentation results and the generated redundancy image to handle controversial pixels. The visual segmentation results showed that this method has higher segmentation accuracy compared with the state-of-the-art.

As discussed in Section \ref{sec:fusion}, Dempster's rule becomes numerically unstable when combining highly conflicting mass functions. In this case, the fused results can be unreliable, as a small change in mass functions can result in sharp changes in the fusion results. Researchers in the medical domain have also recognized this limitation. In~\cite{ketout2012improved}, Ketout et al. proposed a modified mass computation method to address this limitation and applied their proposal to endocardial contour detection. First, the outputs of each active contour model (ACM)~\cite{kass1988snakes} were represented as mass functions. Second, a threshold was proposed to check if the evidence $m$ conflicts with others. If there was conflict, a modifying operation was used on the conflicting evidence. Finally, the results of edge set-based segmentation~\cite{li2005level} and region set-based segmentation~\cite{chan2001active} were fused by using the ``improved BFT''~\cite{xin2005improved} to get a more accurate contour of the left ventricle. Experimental results showed that the fused contour is closer to the ground truth than the contour from the edge or region. False detection of the two contours was suppressed in the resulting image by rejecting the conflicting events by the fusion algorithm. Meanwhile, the proposed method could detect the edges of the endocardial borders even in low-contrast regions.

In~\cite{wen2018improved}, Wen et al. proposed an improved MR image segmentation method based on FCM and BFT. First, the authors fused two modality images $A$ and $B$ with  function $F(x,y) = w_1  g_{A}(x, y)+w_2  g_{B}(x, y)$, where $x$ and $y$ are image pixels and $w_1$ and $w_2$ are weighs used to adjust the influence of different images on the final fusion result and verifying $w_1 + w_2=1$. Second, the authors calculated the MV by FCM and calculated the mass functions of simple and double hypotheses by Zhu's model (see Section \ref{subsub:MVs}) without the neighboring pixel information. Third, the authors generated another kind of mass functions by weighting those of its neighboring pixels with the GD-based model (see Section \ref{subsubsection: gd}) and used Zhu's model again to construct simple and double hypotheses mass functions. Finally, the authors used Dempster's rule to complete the fusion of the two mass functions to achieve the final segmentation. Compared with the FCM-based method, the proposed method can better decrease the conflict in multiple sources to achieve easy convergence and significant improvement by using Dempster's rule for classifier-level evidence fusion.

Besides, with the development of CNNs, the research community used Dempster's rule for the fusion of multiple CNN classifiers. In~\cite{george2020breast}, George et al. proposed a breast cancer detection system using transfer learning and BFT. This first work first applied BFT in multiple evidence fusion with deep learning. High-level features were extracted using a convolutional neural network such as ResNet-18, ResNet-50, ResNet-101, GoogleNet, and AlexNet. A patch-based feature extraction strategy was used to avoid wrong segmentation of the boundaries and provide features with good discriminative power for classification. The extracted features were classified into benign and malignant classes using a support vector machine (SVM). A discounting operation was applied to transfer probability-based segmentation maps into mass functions. The discounted outputs from the different CNN-SVM frameworks were then combined using Dempster's rule. This work takes advantage of deep learning and BFT and has achieved good performance. Compared with majority voting-based fusion methods, BFT-based fusion showed superior segmentation accuracy. Compared with a single classifier, such as ResNet-101, the fused framework achieved an increase of 1\%, 0.5\%, 3\%, and 2\% for, respectively, sensitivity, specificity, and AUC. Also, the authors compared their results with the state-of-the-art method and achieved comparable segmentation accuracy.

Apart from using BFT to combine the discounted probabilities from the CNN classifiers~\cite{george2020breast}, another solution is to construct a deep evidential segmentation framework directly. In~\cite{huang2021belief}, We proposed a BFT-based semi-supervised learning framework (EUNet) for brain tumor segmentation. This work applied BFT in a deep neural network to quantify segmentation uncertainty directly. During training, two kinds of evidence were obtained: the segmentation probability functions and mass functions generated by UNet and EUNet, respectively. Dempster's rule was then used to fuse the two pieces of evidence. This is the first work that embeds BFT into CNN and achieves an end-to-end deep evidential segmentation model. We will introduce more detail about EUNet in Chapter \ref{Chapter4}.

The approaches introduced in this section use several classifiers or clusterers to generate different mass functions  and fuse them by Dempster's rule. Among those approaches, George et al.~\cite{george2020breast} first applied Dempster's rule to combine the discounted probabilities from different deep segmentation models. In \cite{huang2021belief}, we merged ENN with UNet to construct an end-to-end segmentation model and fuse two kinds of evidence by Dempster's rule. Compared to George et al.'s approach~\cite{george2020breast}, our approach can generate mass functions directly from a deep segmentation model, which is more promising.
\subsection{Multimodal evidence fusion}
\label{subsec: multi-modality}
\begin{figure}
\centering
\includegraphics[width=\textwidth]{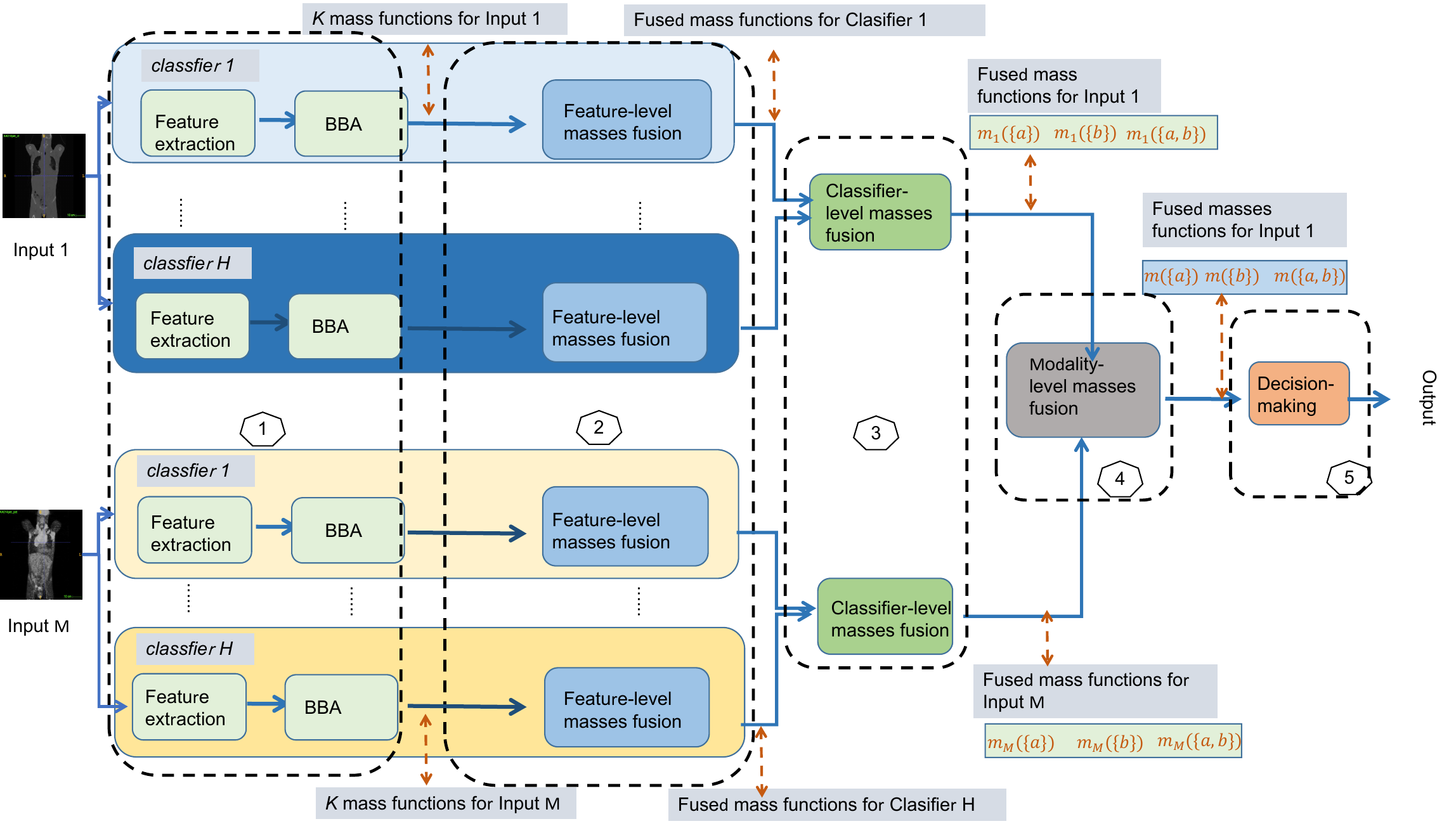}
\caption{Example framework of multimodal evidence fusion with several classifiers. The segmentation process is composed of five steps: (1) for each modality input, image features are fed into different classifiers, and some BBA methods are used to get feature-level mass functions; (2) inside each classifier, the feature-level mass functions are fused by Dempster's rule to get classifier-level mass functions; (3) inside each classifier, the calculated classifier-level mass functions are fused by Dempster's rule to get modality-level mass function; (4) inside each modal, the calculated modality-level mass functions are fused by Dempster's rule; (5) decision-making is made based on the final fused mass functions and outputs a segmented mask.}
\label{fig:fig9}
\end{figure} 

\begin{table}
  \centering
  \caption{Summary of BFT-based medical image segmentation methods with multimodal inputs and several classifiers/clusterers. }
  \scalebox{0.73}{
  \begin{tabular}{llllllll}
 \hline
 \multicolumn{1}{l}{Publications} &Input image type& Application& BBA method &\\

\hline
 \cite{gautier2000belief} &multimodal MR images & lumbar sprain segmentation & Prior knowledge \\
 \cite{lajili2018two}&CT& breast segmentation&threshold  \\

 \cite{huang2021deep}&PET/CT &lymphoma segmentation &None\\
 \cite{huang2022contextual}&multimodal MR images  &brain tumor segmentation& ENN\\

\hline
\end{tabular}
}
\label{tab:fusion4}
\end{table}

Figure \ref{fig:fig9} shows an example of a medical image segmentation framework with several classifiers and multimodal inputs, which is  more complex than the frameworks  introduced in Sections \ref{subsec: single}, \ref{subsec: multi} and \ref{subsec: single-modality}. The segmentation process comprises five steps: mass calculation, feature-level evidence fusion (optional), classifier-level evidence fusion, modality-level evidence fusion, and decision-making. Pixels at the same position from different modalities are fed into different classifiers and different BBA methods to get pixel-level mass functions. Dempster's rule is used first for feature-level evidence fusion, then to fuse classifier-level evidence, and last to fuse modality-level evidence. Here we show the segmentation example of the same located pixels $z_i^1$ and $z_i^M$ that are obtained from modal $1$ and $M$. The same pixel from different modalities is transferred separately into $H$ classifiers, and some BBAs are used to get pixel-level mass functions. Similar to Figure~\ref{fig:fig6}, we assume we can obtain $K$ mass functions with one classifier for each pixel. After the fusion of feature-level, classifier-level, and modality-level evidence fusion, a final mass function is assigned to the pixel $z_i$ to represent the degree of belief this pixel belongs to class a, b and ignorance. Table \ref{tab:fusion4} summarizes the segmentation methods focusing on classifier/clusterer fusion and modality-level evidence fusion.

In~\cite{gautier2000belief}, Gautier et al. proposed a method for helping physicians monitor  spinal column diseases with multimodal MR images. At first, an initial segmentation was applied with active contour~\cite{lai1994deformable}. Then several mass functions were obtained from expert opinions on constructing the frame of discernment. Thus, the mass functions were human-defined. Finally, Dempster's rule (see Section \ref{sec:fusion}) was then used to fuse the mass functions from different experts. The method yielded the most reliable segmentation maps when the paper was published.

Based on their previous work on multimodal medical image fusion with a single cluster~\cite{chaabane2009relevance}, Chaabane et al. presented another BFT-based segmentation method with several clusterers~\cite{chaabane2011new}. First, possibilistic C-means clustering~\cite{bezdek2013pattern} was used on R, G, and B channels to get three MVs. Then the MVs were mapped into mass functions with focal sets of cardinality 1 and 2 using Zhu's model (see Section \ref{subsub:MVs}). Dempster's rule was used first to fuse three mass functions from three corresponding color spaces. Based on the initial segmentation results, another mass function was calculated for each pixel and its neighboring pixels for each color space. Finally, the authors used Dempster's rule again to fuse the two mass functions from two corresponding clusterers. Experimental segmentation performance with cell images showed the effectiveness of the proposed method. Compared with the
FCM-based segmentation method, the proposal increased by 15\% the segmentation sensitivity.

In~\cite{lajili2018two}, Lajili et al. proposed a two-step evidential fusion approach for breast region segmentation. The first evidential segmentation results were obtained by a gray-scale-based K-means clustering method, resulting in $k$ classes. A sigmoid function was then used to define a mass function on the frame $\Omega=\{\text{Breast},\text{Background}\}$ at each pixel $z$ depending on its class.  
For local-homogeneity-based segmentation, the authors modeled the uncertainty with a threshold value $\alpha$, by defining $m(\{\text{Breast}\})=1-\alpha$, $m(\{\text{Background}\})=0$, $m(\Omega)=\alpha$, where $\alpha$ represents the belief mass corresponding to the uncertainty on the membership of the pixel $z$. A final fusion strategy with Dempster's rule was applied to combine evidence from the two mass functions. Experiments were conducted on two breast datasets~\cite{suckling1994mammographic,bowyer1996digital}. The proposed segmentation approach yielded more accurate results than the best-performed method. It extracted the breast region with correctness equal to 0.97, which was 9\% higher than the best-performing method. 

In~\cite{huang2021deep}, we proposed to consider PET and CT as two modalities and used to UNet model to segment lymphoma separately. Then the two segmentation masks were fused by Dempster's rule. Although this is the first work that applied BFT in multimodal evidence fusion with several deep segmentation models, a limitation of this work is that only Bayesian mass functions are used for evidence fusion. To improve this first work, we proposed a multimodal evidence fusion model with contextual discounting for brain tumor segmentation~\cite{huang2022contextual}. In this work, using four modules for feature extraction and evidential segmentation, the framework assigns each voxel a mass function. In addition, a contextual discounting layer is designed to take into account the reliability of each source when classifying different classes. Finally, Dempster's rule is used to combine the discounted evidence to obtain a final segmentation. This method can be used together with any state-of-the-art segmentation module to improve the final performance. We will introduce more detail about the multimodal evidence fusion model in Chapter \ref{Chapter6}.

Few studies have considered multimodal medical images as independent inputs and used independent classifiers to generate mass functions. The performance of this kind of approach is limited by the representation of image features and the ability to quantify the model uncertainty. We first merged ENN with UNet for the fusion of multimodal MR images with contextual discounting~\cite{huang2022contextual}. This approach enables the model to generate a learned reliability metric from input modalities during different segmentation tasks, which can potentially make the results more explainable.

\section{Conclusion}

The choice of using single-modality or multimodal depends on the dataset. Generally, the more source data we have, the more reliable segmentation results we will get. The choice of a single or several classifiers/clusterers depends on the limitation of the computation source and the requirement of computation efficiency. Prior to 2020, BFT-based medical image segmentation methods had limited segmentation accuracy due to the use of low-level image features, such as grayscale and shape features, to generate mass functions. Moreover, none of them considered the segmentation reliability. Since the application of deep learning in medical image segmentation has been very successful, the use of BFT in deep neural networks is a promising research direction, in particular, to quantify the uncertainty and reliability of the segmentation results, as well as fuse evidence from multimodal medical images. In the next part, we will introduce our contributions related to the construction of the deep segmentation model with BFT and deep learning algorithms (Chapter \ref{Chapter4}), the uncertainty quantification with deep segmentation models (Chapter \ref{Chapter5}), and multimodal evidence fusion (Chapter \ref{Chapter6}).
\stopcontents[chapters] 

\part{Deep Evidential Segmentation}
\chapter{Semi-supervised medical image segmentation } 
\label{Chapter4}


\tocpartial
\section{Introduction}


The performance of deep learning-based methods depends mainly on large-scale learning examples and their annotations. However, obtaining precisely annotated data is very challenging in the medical domain. Recently, some semi-supervised deep learning methods have been proposed to train the models with fewer labels. Techniques for semi-supervised medical image segmentation can be divided into three classes: graph-constrained methods \cite{xu2017deep}, self-learning methods \cite{li2019transformation,min2019two}, and generative adversarial learning methods \cite{mondal2018few,sun2019parasitic}. Though experimental results are promising, only a few authors focus on studying the uncertainty caused by the low quality of the images and the lack of annotations. In this work, we follow the main idea of self-training and use the image information to construct a semi-supervised brain tumor segmentation framework. We first propose to use two parallel segmentation modules: a probabilistic segmentation module and an evidential segmentation module, to obtain two segmentation results. We then use an evidential fusion module to combine evidence in order to decrease uncertainty. Results from a  series of experiments on the BraTS2019 brain tumor dataset showed that our framework achieves encouraging results when only some training data are labeled.

We organize this chapter as follows:  Section \ref{sec: model4} introduces the semi-supervised medical image segmentation framework. Section \ref{sec: exp4} reports numerical experiments by comparing the segmentation performance with the state-of-the-art methods under supervised and semi-supervised learning. Finally, we conclude this work in Section \ref{sec:conclu4}.

\section{Proposed framework}
\label{sec: model4}
\begin{figure}
    \centering
\includegraphics[width=0.9\textwidth]{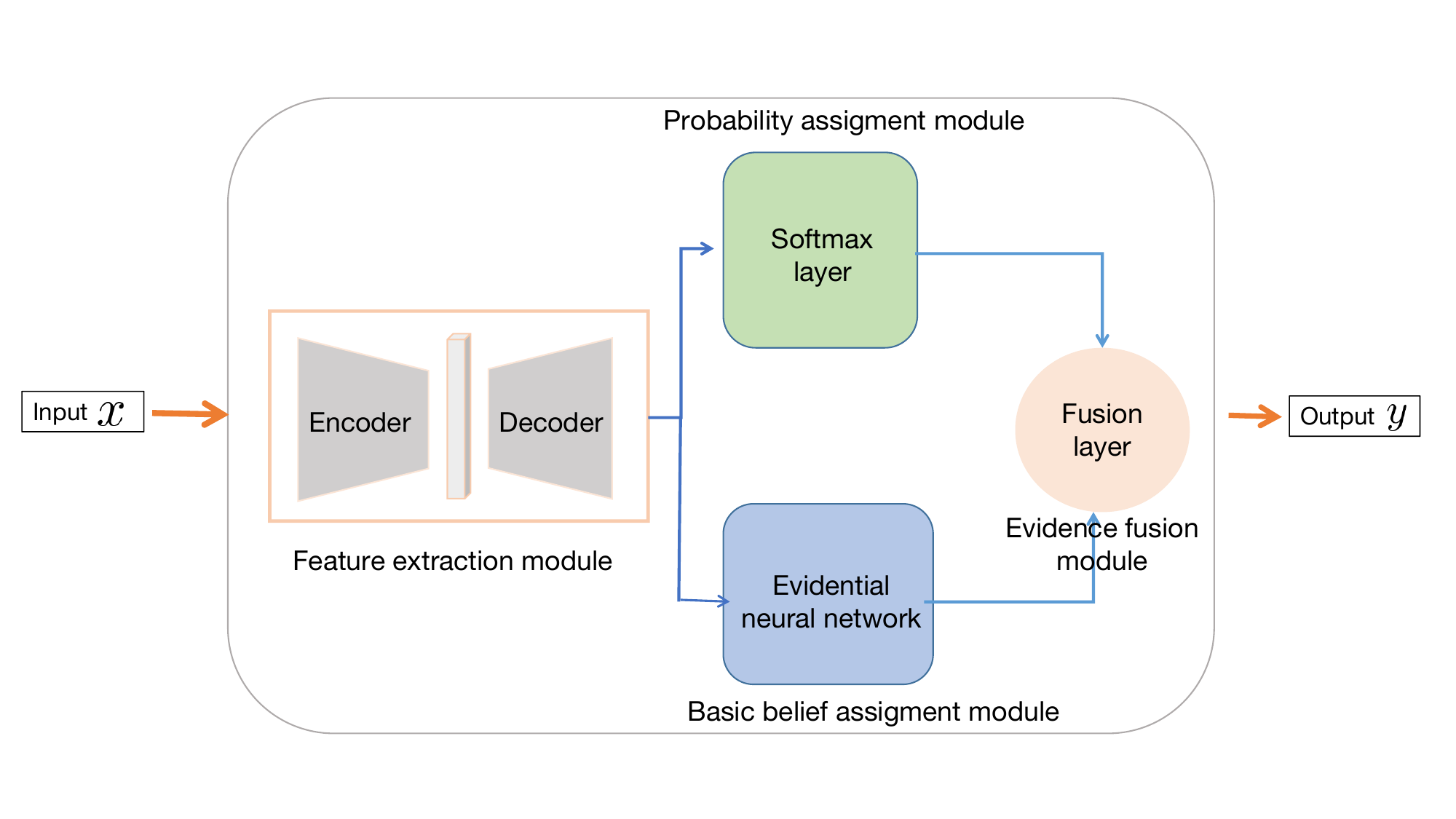}
\caption{Overall flowchart of our proposal. It is composed of four modules for feature extraction, probability assignment, basic belief assignment, and evidence fusion.}
    \label{fig:overall}
\end{figure}

In this section, we describe the proposed framework, named SEFNet, which is composed of a feature extraction module similar to UNet, a probability assignment module to obtain a probability segmentation, a basic belief assignment module to obtain an evidential segmentation, and an evidence fusion module to combine both probability and evidential segmentations. Figure \ref{fig:overall} shows the overall flowchart of our proposal. \new{The first step is to extract features from input images.} Second, two modules are used in parallel to map features into probabilities or mass functions. \new{The probability assignment module uses the softmax transformation to map the extracted features into probabilities. The mass function assignment module uses ENN (see Section \ref{subsubsec: distance}) to map the extracted features into mass functions. Third, the two sources of evidence, probabilities and mass functions, are combined by Dempster’s rule in the evidence fusion module. The semi-supervised learning algorithm will be introduced in Section \ref{subsec:semi-learning}. }

\subsection{Evidential segmentation with multiple evidence fusion}
\begin{figure}
\centering
\includegraphics[width=\textwidth]{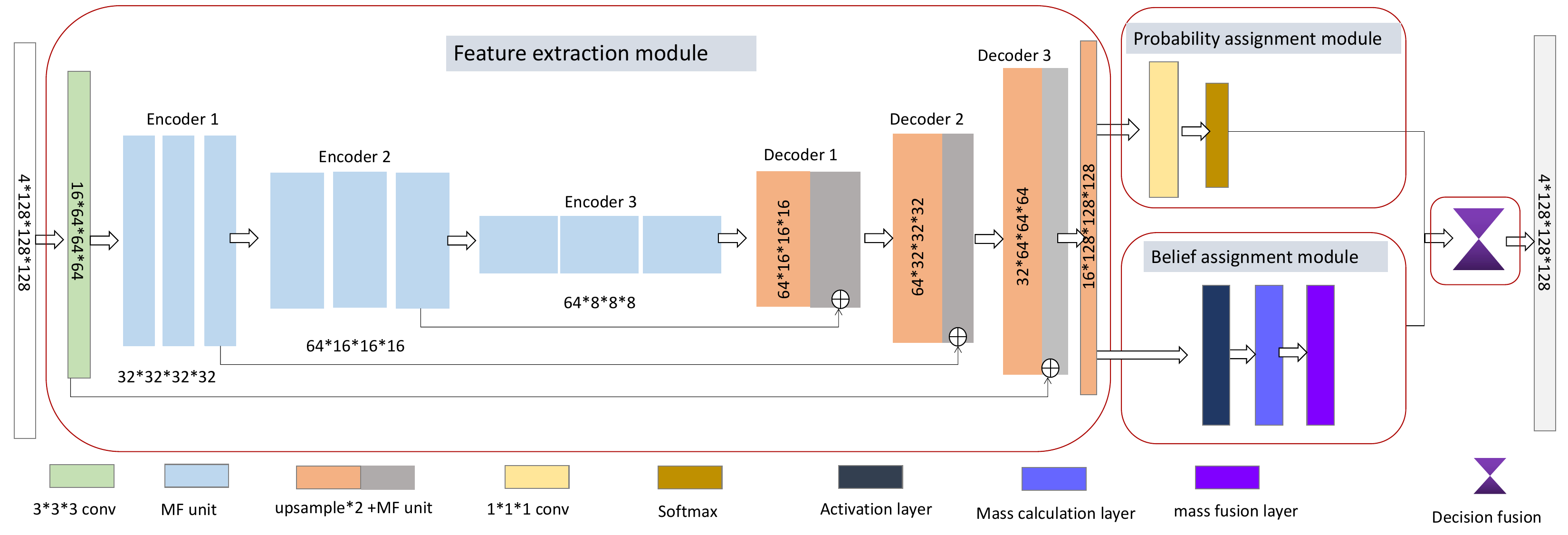}
\caption{Proposed evidential medical image segmentation framework.}
\label{fig:brats}
\end{figure}

Figure  \ref{fig:brats} shows the detailed evidential segmentation framework. The input  is composed of four MRI modalities. (Here, we show an example of input data of size $4 \times 128 \times 128 \times 128$.) 
\label{subsec:esf}

\paragraph{Feature extraction} As shown in Figure \ref{fig:brats}, we apply multi-fiber units (see Figure \ref{fig:multi-fiber} from Section \ref{subsec:dmfnet}) in the encoding stage to achieve multi-scale representation. In the decoding stage, the high-resolution features from the encoding stage are concatenated with the upsampled features, which makes the whole feature extraction module similar to UNet \cite{ronnebergerconvolutional}.

\paragraph{Probability and belief assignment modules} The \emph{probability assignment} module comprises a $1 \times 1 \times 1 $ projection layer followed by a softmax layer, which maps the feature vectors into  probabilities directly. The output of the probability assignment module is denoted as $p_{CNN}$, and we assume that each voxel belongs to one of four classes denoted as $\{0,1,2,4\}$.  The  \emph{belief assignment} module is based on the ENN model recalled in Section \ref{subsubsec: distance}; it comprises three layers: an prototype activation layer that computes distance-based activations using \eqref{eq:s_i}, a mass calculation layer that computes mass functions using \eqref{eq:m_i}, and a combination layer that combined mass functions derived from prototypes using Dempster's rule.


\paragraph{Evidence fusion} The objective of this module is to make a final segmentation decision. The decision based on several information sources can be expected to be more accurate and reliable than using a single source of information. In our case, if only part of the training data is labeled, the uncertainty is higher than it is in the fully supervised case. To increase the segmentation performance, we propose an additional \emph{evidence fusion} layer to combine evidence from the probability and belief assignment modules. Here, the  voxel-wise output probability distributions $p_{CNN}$ from the probability assignment module can be seen as a Bayesian mass functions, which can be combined with the  voxel-wise output mass functions $m_{ENN}$ from the belief assignment module using Dempster’s rule. The combined mass functions are Bayesian, and are given by
\begin{equation}
    (p_{CNN}\oplus m_{ENN})(\{\omega_{c}\})=\frac{p_{CNN}(\omega_c)pl_{ENN}(\omega_c)}{\sum_{l=1}^C p_{CNN}(\omega_l)pl_{ENN}(\omega_l)} , \quad c=1,\ldots,C,
    \label{eq:11}
\end{equation}
where $pl_{ENN}(\omega_c)=m_{ENN}(\{\omega_c\})+m_{ENN}(\Omega)$ is the plausibility of class $\omega_c$ derived from mass function $m_{ENN}$.

\subsection{Semi-supervised learning}
\label{subsec:semi-learning}

We propose a semi-supervised learning algorithm to optimize the framework when only part of the training data is labeled, with the aim of obtaining an accuracy as close as possible to that of a fully supervised learning method. The general idea is that similar images are expected to produce similar classification  or segmentation results even if some transformations have been performed because the relevant characteristics are preserved despite the transformation. During each learning epoch, a transformed copy $x_t$ of  each input image $x$ is computed using one of several transformations, namely, random intensity change, Gaussian blur and exponential noise. This transformation operation will be described in Section \ref{subsubsec: pre-processing}. Two loss functions are proposed for training data with and without labels. 
 
\paragraph{Training with labels} We train the network with the labeled data using the \new{following \textsf{loss1} function, class-independent Dice loss,} which measures the overlap region between the output $S$ and the ground truth $G$:
\new{
\begin{equation}
    \textsf{loss1}=\sum_{c=1}^{C} \left ( 1-2\frac{  { \sum_{n=1}^{N_{1}}}  S_{cn} G_{cn} }{ {
{ \sum_{n=1}^{N_{1}}} S_{cn}+ G_{cn} } }\right ),
    \label{eq:8}
\end{equation}
}
where $G_{cn}=1$ if voxel $n$ belongs to class $c$, and $G_{cn}=0$ otherwise, and $S_{cn}$ represents the corresponding predicted mask; $C$ is the number of classes and $N_1$ is the number of voxels with labels.

\paragraph{Training without labels} For the data without labels, we use the mean squared error loss, denoted here as \textsf{loss2}, to optimize the feature representation by minimizing the difference between the original output $S$ and the transformed output $S_{t}$:
\begin{equation}
\textsf{loss2}=\frac{1}{2N_{2}C}{ \sum_{c=1}^{C} \sum_{n=1}^{N}}\left \| S_{cn}- S^{t}_{cn} \right \|^{2},
\label{eq:10}
\end{equation}
where $S^{t}_{cn}$ is the segmented label at point $n$ corresponding to the generated input $x_t$, $N_2$ is the number of voxels without labels.

\section{Experiments and results}
\label{sec: exp4}
In this section, we present numerical experiments to verify the effectiveness of the proposed model. \new{In Section \ref{subsec: experimental settings}, we introduce the dataset, the preprocessing of the data, the parameter setting, and the evaluation protocols. The sensitivity analysis of hyper-parameters and the comparative analysis of segmentation performance are then introduced in Sections \ref{subsec: sensitivity analysis} and \ref{subsec: comparative analysis}, respectively.}

\subsection{Experiment settings}
\label{subsec: experimental settings}
\subsubsection{Dataset}
\label{subsubsec: data}
\begin{figure}
\centering
\includegraphics[width=\textwidth]{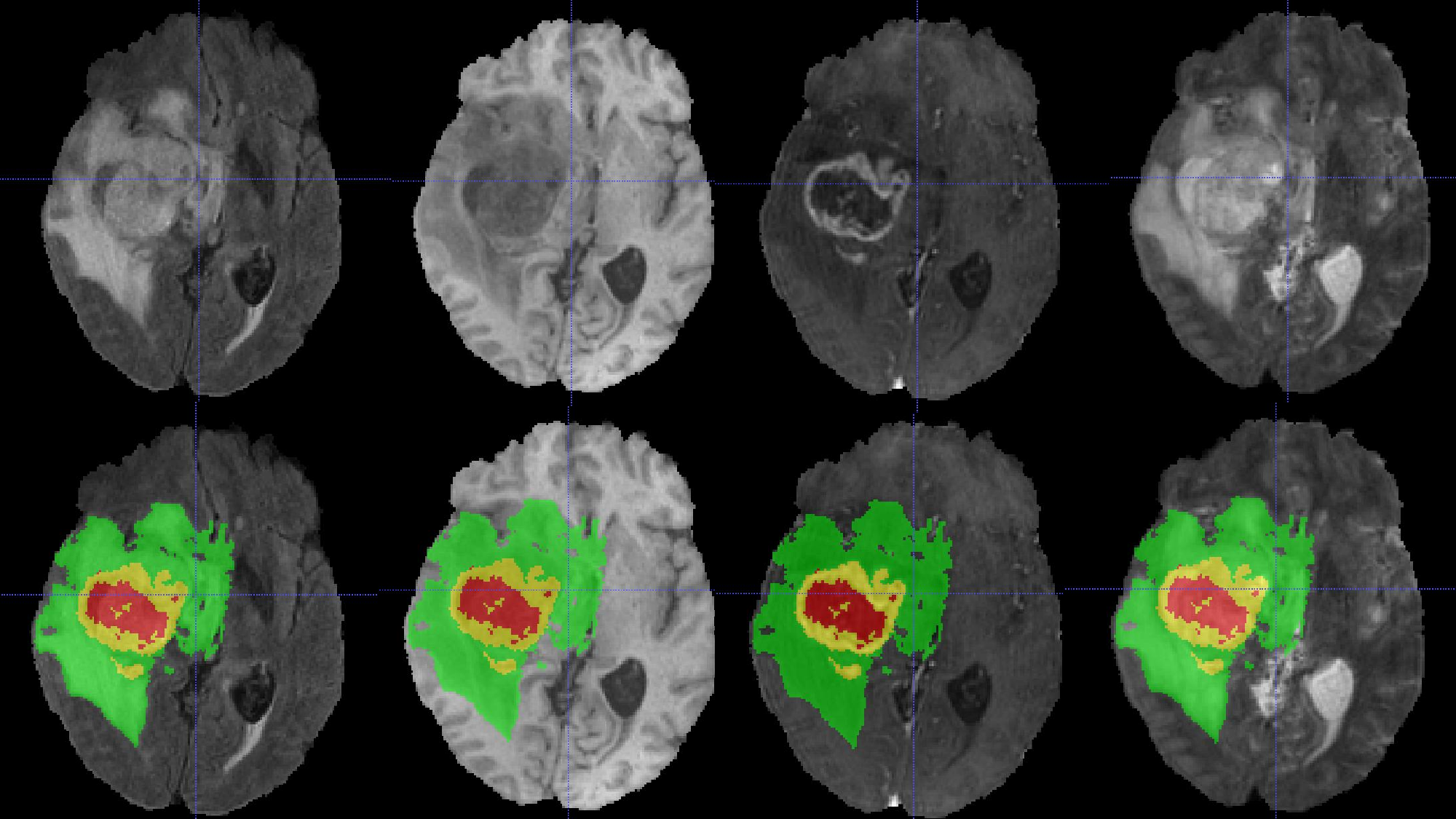}
\caption{Example of one patient from BraTs2019 dataset. The first and second rows show MR images and MR images with labeled tumor masks, respectively. From left to right: FLAIR, T1, T1Gd, T2. Labels 1, 2, and 4 are marked in red, green, and yellow, respectively.}
\label{fig:example}
\end{figure}

The experiment data is provided by the Brain Tumor Segmentation (BraTS) 2019 challenge \cite{menze2014multimodal,bakas2017advancing,bakas2018identifying}. The dataset consists of 335 cases of patients for training, 125 cases for validation and 166 cases for test. Since the test set is not available now, in this chapter, we use the official validation set to test our model, i.e., take it as a test set. For each patient, we have four kinds of MR sequences: T1, T1-weighted (T1Gd), T2, and FLAIR. Each of them has a volume of $155\times240\times240$. For data from the training set, each case was annotated into three heterogeneous histological sub-regions: peritumoral edema (ED, label 2), necrotic core and non-enhancing tumor (NRC/NET, label 1), and enhancing tumor(ET, label 4). The background is marked as label 0. Figure~\ref{fig:example} shows an example of the four modalities and the corresponding tumor region. The evaluation was based on the segmentation accuracy of three overlap regions: enhancing tumor (ET, label 4), tumor core (TC, the composition of label 1 and 4), and whole tumor (WT, the composition of label 1, 2, and 4). \new{For data from the validation set, only four modalities of MR sequence information are available.}

\new{We used five-fold cross-validation to train our SEFNet. During training, we randomly divided the BraTS 2019 training set into 5 equal-sized datasets. The training process was then repeated five times, with each of the 5 datasets used exactly once as the validation data. The segmentation performance with cross-validation is reported by the average of five models. For a fair comparison with the state-of-the-art, we fine-tuned the best-performing model with the full training set and tested the performance on the validation set. The segmentation performances were assessed by the online evaluation server CBICA’s Image Processing Portal \footnote{https://ipp.cbica.upenn.edu/}.} 

\subsubsection{Pre-processing}
\label{subsubsec: pre-processing}
Before feeding the data into the framework, several pre-processing methods were used to process the input data. We first applied intensity normalization to each MRI modality from each patient independently by subtracting the mean and dividing by the standard deviation of the brain region only. Moreover, to prevent overfitting, we used four types of data augmentation. First, we applied a random intensity shift between $\left[-0.1, 0.1\right]$ and random intensity scaling between $\left[0.9, 1.1 \right]$ to MRI data. Second, we randomly cropped the MRI data from $155\times240\times240$ to $128\times128\times 128$. Third, we used random rotation with a rotation angle of 10. Finally, we used random mirror flipping for MRI data along each 3D axis with a flip probability of 50\%. The data augmentation operation is applied during each training epoch. Since the data augmentation operation is randomly chosen, the input $x$ in each training epoch varies.   

\begin{figure}
    \centering
    \includegraphics[width=\textwidth]{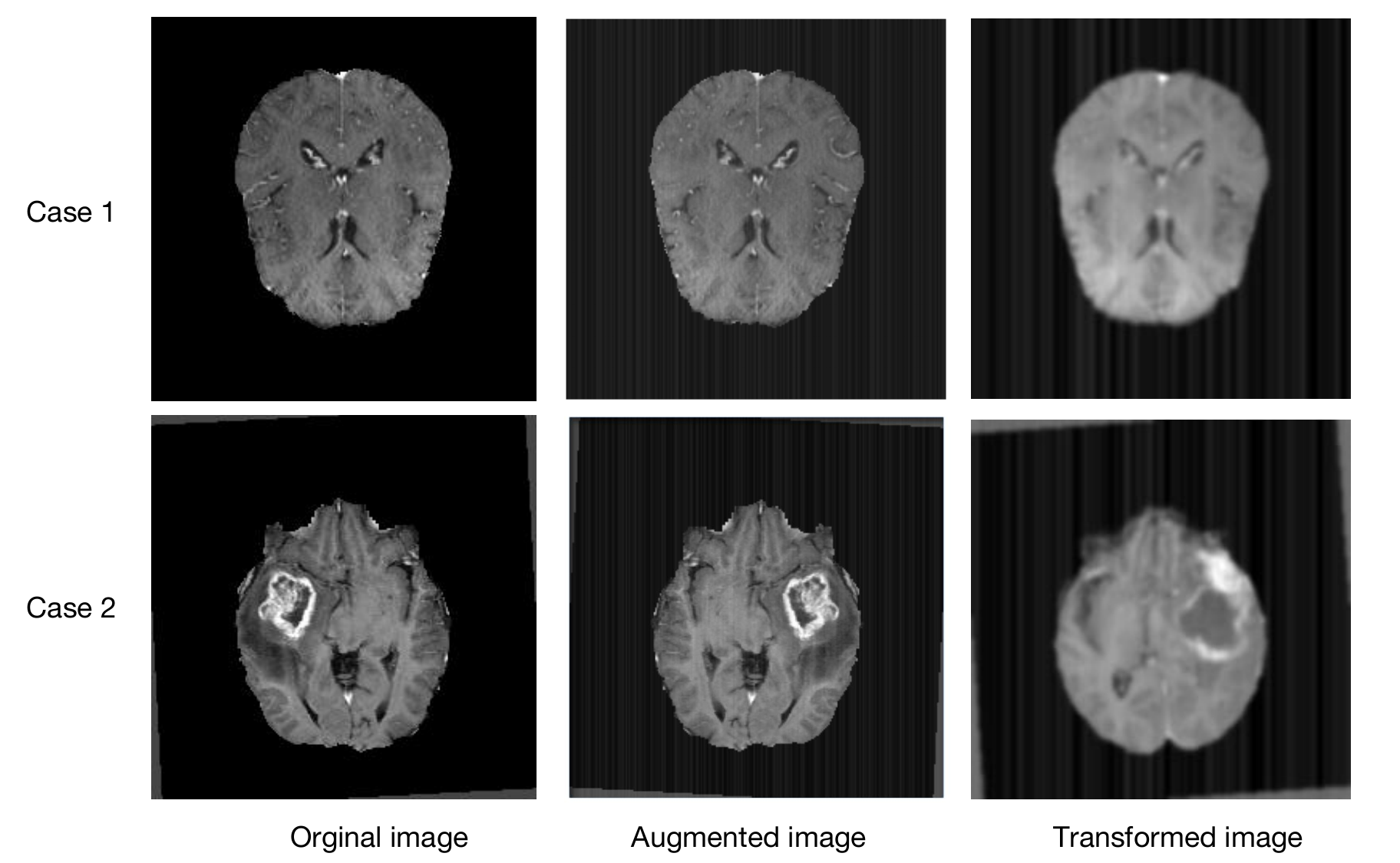}
    \caption{Example of pre-processed images. From left to right: the original image, the augmented image and the transformed image, respectively.}
    \label{fig: preprocessing}
\end{figure}

For semi-supervised training, we used transformations for each preprocessed input $x$. We first applied random intensity change on the input with the shift between $\left[-0.2, 0.2\right]$ and scaling between $\left[0.9, 1.1\right]$. Then we added Gaussian Blur with a standard deviation of 3 to the image. Finally, we added exponential noise with an exponent of 3 to the input. After transformation, for each input $x$, we will generate transformed data $x_t$. Figure \ref{fig: preprocessing} shows two examples of input images before and after prepossessing. (To better show the difference between images, here we only show the aligned images without random cropping). Compared with the original image, the augmented image is randomly flipped with a small intensity change and image rotation. Compared with augmented input $x$, the transformed input $x_t$ is more blurred and noisy.

\subsubsection{Parameter settings}
\label{subsubsec: setting}
\new{For the feature extraction module, the spatial dimension and input channel were set, respectively, as 3 and 4. The channels \new{(number of filters)} of the input layer were set as 16. The channels \new{(number of filters)} of MF units were set as $32, 64, 64$ for three corresponding encoders and each MF unit has $16$ parallel fibers (see Section \ref{subsec:dmfnet}).}

For the mass assignment module, the prototypes \textsf{$p$} \eqref{eq:s_i} were initialized randomly with Xavier uniform \cite{hanin2018start} distributions to simplify the model, as well as the parameter of the membership degrees \textsf{$u$} \eqref{eq:m_i}. The impact of prototype numbers will be discussed in Section \ref{subsubsec: prorotype}. Then we initialized parameters \textsf{$\alpha$} \eqref{eq:s_i} and \textsf{$\gamma$} \eqref{eq:s_i} with constants 0.5 and 0.01. 

\new{The maximum training epoch was set to 300. The training process was stopped when the performance did not increase in 20 epochs. The Adam optimization algorithm with batch size 8 was used to train the model. The initial learning rate was set to 0.001} at the beginning and decayed with an adjusted learning rate 
\begin{equation}
    lr=lr_0 \times \left ( 1-\frac{e}{N_e} \right ) ^{0.9},
    \label{eq:12}
\end{equation}
where $e$ is an epoch counter, and $N_e$ is a total number of epochs. \new{The model with the best performance with cross-validation was saved as the final model for testing.} All experiments were implemented in Python with the PyTorch framework and were trained and tested on a desktop with a 2.20GHz Intel(R) Xeon(R) CPU E5-2698 v4 and a Tesla V100-SXM2 graphics card with 32 GB GPU memory. 

\subsubsection{Evaluation criteria}
\label{subsubsec: evaluation}

We used two evaluation criteria: the Dice score (see \eqref{eq: dice_score}) and the Hausdorff Distance (HD) (see \eqref{eq:hds}), to measure the segmentation performance. For each patient, we separately computed these two indices for the three classes and then averaged indices over the patients. Results are reported in Sections \ref{subsec: sensitivity analysis} and \ref{subsec: comparative analysis}.

\subsection{Sensitivity analysis}
\label{subsec: sensitivity analysis}
\subsubsection{Evaluating the impact of number of prototypes}
\label{subsubsec: prorotype}
\begin{figure}
\centering
\subfloat[\label{fig:Dice}]{\includegraphics[width=0.5\textwidth]{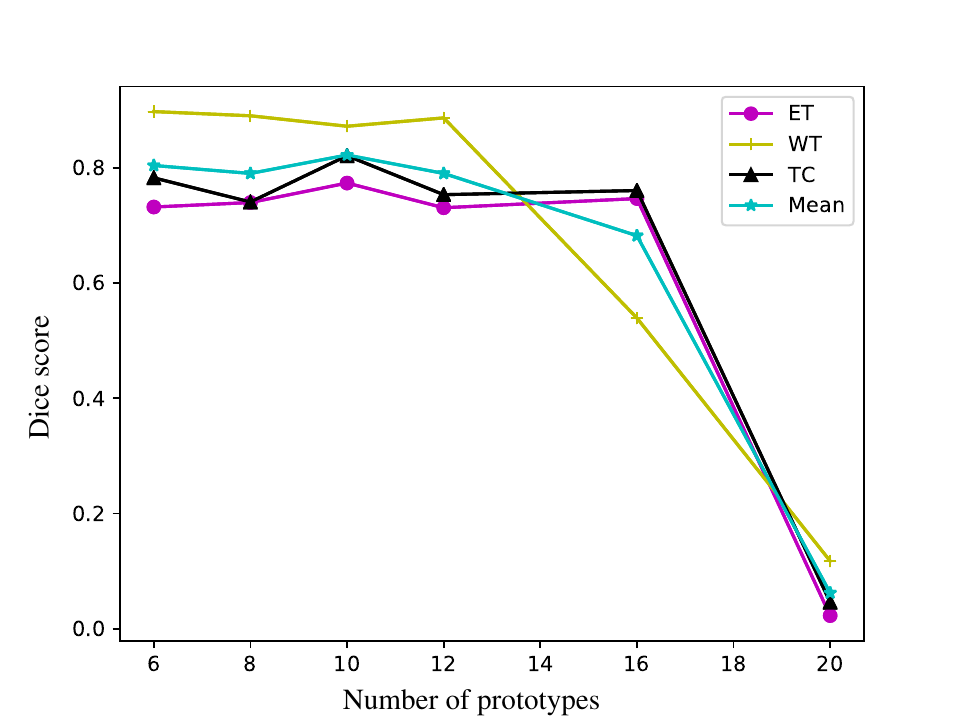}}
\subfloat[\label{fig:Hausdorff}]{\includegraphics[width=0.5\textwidth]{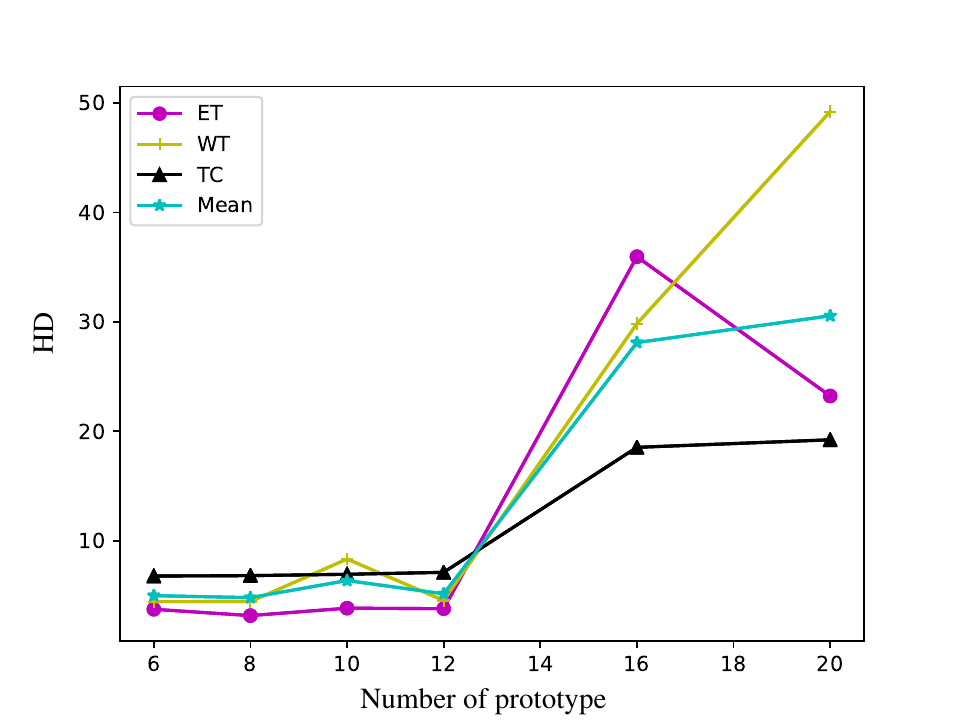}}\\
\caption{Values of the (a) Dice score (the higher the better), (b) Hausdorff distance (HD) (the lower the better) of ET, TC, WT with the different number of prototypes.}
\label{fig:prototype}
\end{figure}

The number of prototypes is an important hyper-parameter that may impact segmentation performance. We trained the model with 6, 8, 10, 12, 16, and 20 prototypes. Figure \ref{fig:prototype} shows the corresponding results. We can see from Figure \ref{fig:prototype} when the number of prototypes is between 6 and 12, the performance is stable. SEFNet achieves the best performance in terms of Dice score and Hausdorff Distance with 10 prototypes. With more than 12 prototypes, the performance decreases. Thus, in the following experiments, the number of prototypes in SEFNet is fixed as 10.

\subsubsection{Evaluating the impact of semi-supervised learning}
\label{subsubsec: semi}

\begin{figure}
\centering
\subfloat[\label{fig:train_loss}]{\includegraphics[width=\textwidth]{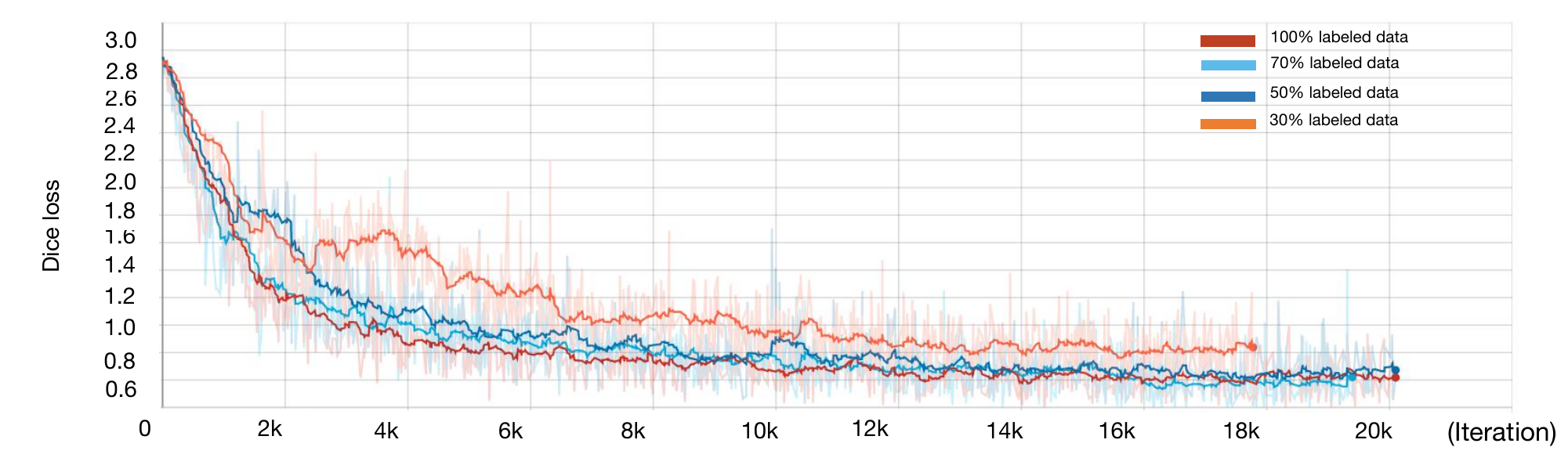}}\\
\subfloat[\label{fig:train_dice}]{\includegraphics[width=\textwidth]{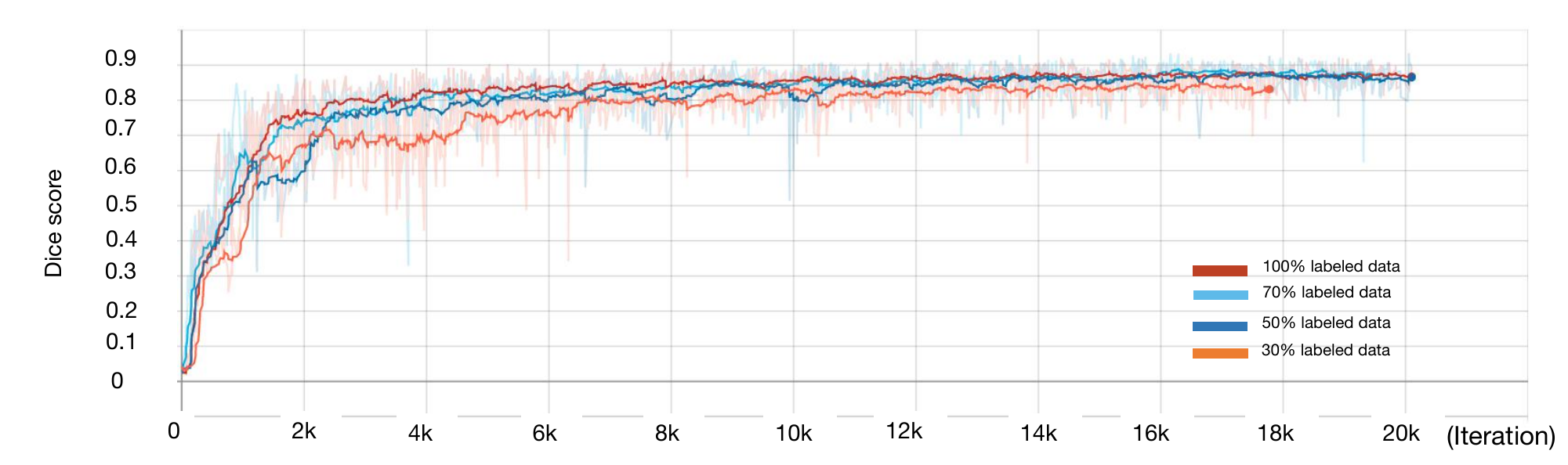}}\\
\subfloat[\label{fig:val_loss}]{\includegraphics[width=\textwidth]{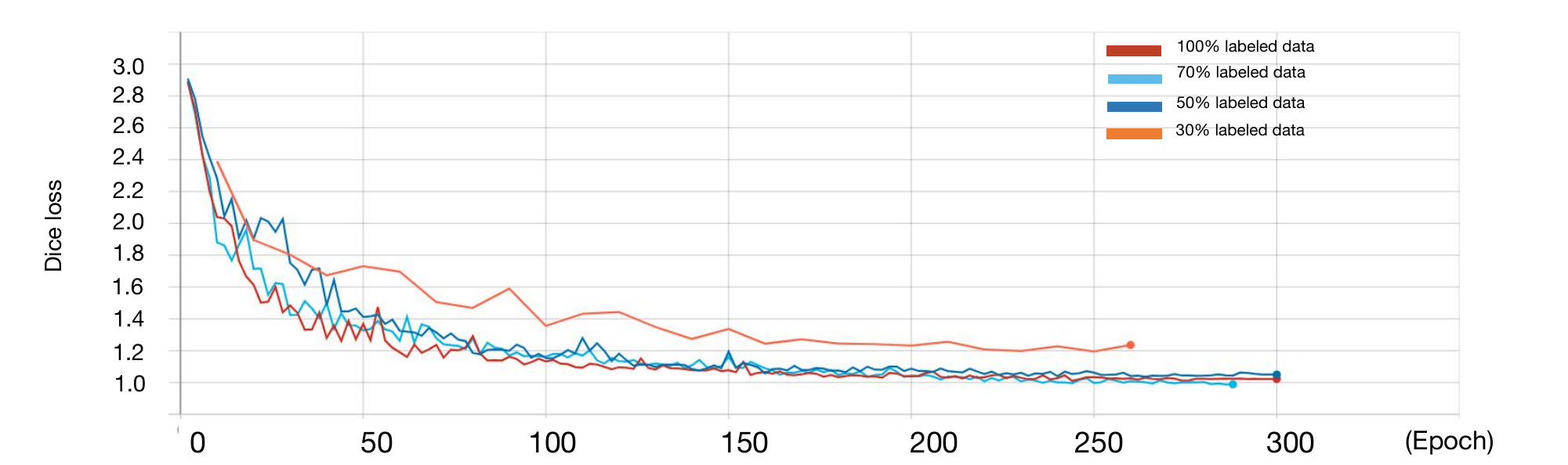}}\\
\subfloat[\label{fig:val_dice}]{\includegraphics[width=\textwidth]{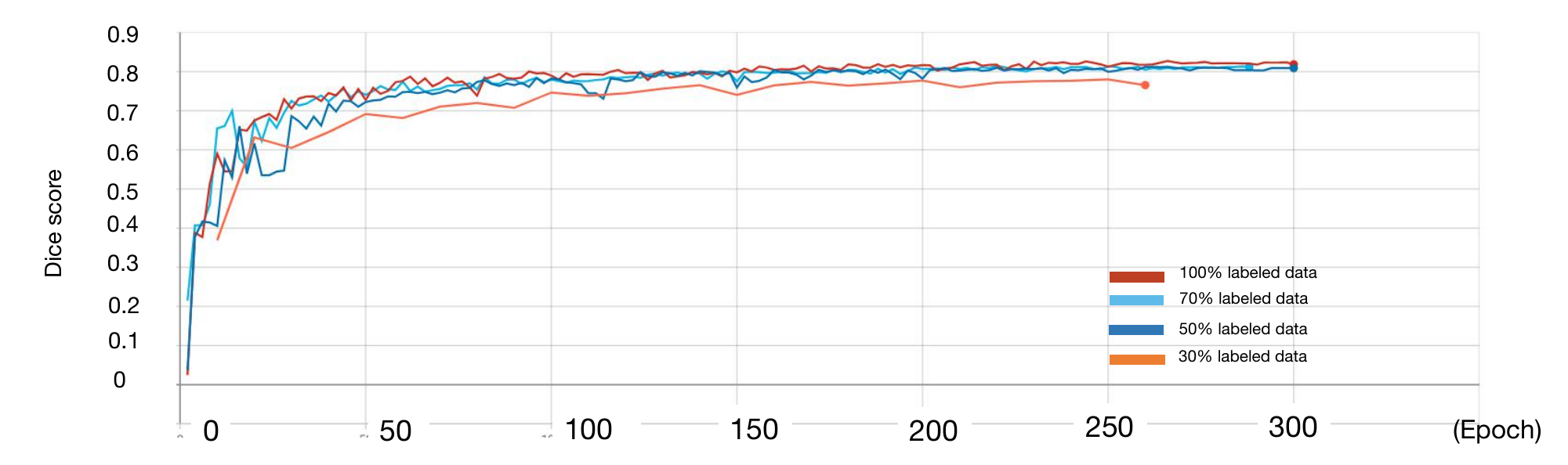}}\\
\caption{\new{Plots of Dice loss and Dice score during training. (a) Mean Dice loss during training, (b) Mean Dice score during training, (c) Mean Dice loss with cross-validation, (d) Mean Dice score with cross-validation.}}
\label{fig: training-process}
\end{figure}

To evaluate the impact of the proportions of the labeled training data, we trained our model with 100\%, 70\%, 50\%, and 30\% labeled data and reported the cross-validation performance in Table \ref{tab:2}. It shows the corresponding segmentation performance of the Dice score and the Hausdorff distance (HD), as well as the corresponding standard deviations (SD) among the evaluated cases. Compared with 100\% training data labeled, the Dice score decreased by 2.1\%, 5.6\%, and 7\%, respectively, when only 70\%, 50\%, and 30\% training data are labeled. With different proportions of the labeled training data, the model shows comparable performance on Hausdorff Distance. Compared with WT (ET+NRC/NET+ED) segmentation, the standard deviations (SD) are quite large for ET and TC (ET+NRC/NET) segmentation. Two main reasons can explain the large SD: the segmentation between ET and NRC/NET is challenging; some cases do not contain ET or NRC/NET tumors, while the model segment DE tumors into ET or NRC/NET class. Figure \ref{fig: training-process} shows plots of the Dice loss and Dice score under different proportions of training labels during training. With the decrease in training labels, the model only slightly decreases segmentation performance. Table \ref{tab:2_2} shows the segmentation performance on the online validation dataset with the five corresponding models obtained by cross-validation. When only 70\%, 50\%, and 30\% training data are labeled, the Dice score decreased by 2.9\%, 4.2\%, and 7.5\%, respectively, compared with 100\% training data labeled. The above results demonstrate the effectiveness of semi-supervised learning.
\begin{table}
  \centering
  \caption{Performance comparison on the BraTS2019 training set with cross-validation.}
  \scalebox{0.75}{
  \begin{tabular}{llllllllll}
  \hline
  \multicolumn{1}{l}{Label}  &\multicolumn{3}{c}{Dice Score(\ttpm SD)}&&
  \multicolumn{3}{c}{HD(\ttpm SD)} \\
  \cline{2-4} 
  \cline{6-8} 
proportion& ET&WT&TC && ET&WT&TC\\
 \hline
100 \% &0.783\ttpm 0.227&0.906\ttpm 0.056&0.805\ttpm 0.220
&&3.481\ttpm 4.516&4.618\ttpm 3.829&6.978\ttpm 7.595\\
70 \% &0.762\ttpm 0.215&0.905\ttpm 0.056&0.814\ttpm0.201
&&5.151\ttpm 7.254&6.219\ttpm 11.037&6.507\ttpm 7.732\\
50 \%&0.727\ttpm 0.261&0.904\ttpm 0.066&0.806\ttpm 0.215
	&&5.115\ttpm 6.149&4.858\ttpm 5.399&6.672\ttpm 7.529\\
30 \%&0.713\ttpm 0.247&0.897\ttpm 0.070&0.799\ttpm 0.215
&&5.904\ttpm 7.214&5.554\ttpm 6.867&6.459\ttpm 6.489\\
\hline
\end{tabular}}
\label{tab:2}
\end{table}

\begin{table}
  \centering
  \caption{\new{Performance comparison on the BraTS2019 online validation set.}}
  \scalebox{0.75}{
  \begin{tabular}{llllllllll}
  \hline
  \multicolumn{1}{l}{Label}  &\multicolumn{3}{c}{Dice Score (\ttpm SD)}&&
  \multicolumn{3}{c}{HD(\ttpm SD)} \\
  \cline{2-4} 
  \cline{6-8} 
    proportion& ET&WT&TC && ET&WT&TC \\
 \hline
100\% & 0.763\ttpm0.254 &0.883\ttpm0.095  &0.808\ttpm0.201  &&  4.592\ttpm8.401	&5.983\ttpm6.202 &7.671\ttpm10.608\\
70\% &0.734\ttpm0.286&0.884\ttpm0.110&0.767\ttpm0.242
	&&5.021\ttpm8.929&5.718\ttpm7.929&8.427+11.799\\
50\% &0.721\ttpm0.294&0.877\ttpm0.129&0.777\ttpm0.227
	&&7.204\ttpm17.002&7.942\ttpm15.235&11.830\ttpm19.809\\
30\%&0.688\ttpm0.289&0.887\ttpm0.113&0.744\ttpm0.267&&8.609\ttpm17.533	&7.448\ttpm12.585	&12.236\ttpm19.690\\
\hline
\end{tabular}
}
\label{tab:2_2}
\end{table}

\begin{figure}
    \centering
    \includegraphics[width=\textwidth]{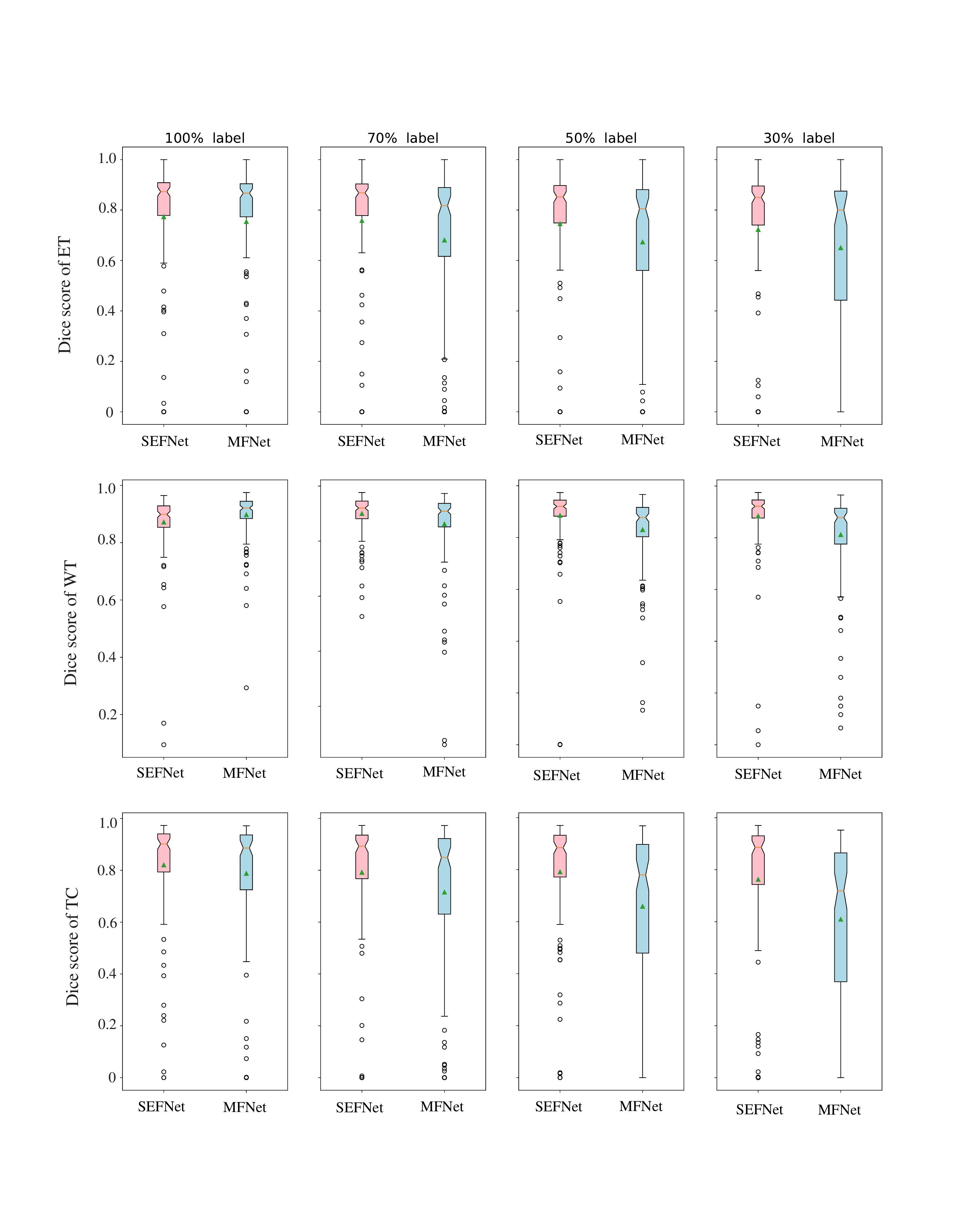}
    \caption{Dice score of ET, TC, and WT with different percentages of labeled training data. From left to right: the results of training with 100\%, 70\%, 50\%, and 30\% labels, respectively. The first, second, and third rows show the Dice score of ET, WT, and TC, respectively. The pink and light-blue boxplot represents the results of our proposal (SEFNet) and the baseline model (MFNet), respectively. The orange line and green triangle represent the median and mean value of the Dice score, respectively.}
    \label{fig:boxall}
\end{figure}

Furthermore, we compared the distribution of the Dice score among 125 validation cases in our model SEFNet and the baseline model MFNet, under the different proportions of labeled training data in Figure \ref{fig:boxall}. To simplify the comparison, we only showed the mean value of the Dice score here. The boxplot displays the data based on a five-number summary: the minimum (the lowest data point excluding any outliers), the maximum (the largest data point excluding any outliers), the median, and the 25\% and 75\% percentile. As shown in Figure \ref{fig:boxall}, SEFNet yields better performance than MFNet with different proportions of labeled training data. When 100\% training data are labeled, SEFNet and MFNet can achieve high segmentation accuracy. With the decreasing proportion of labeled training data, the SEFNet shows increasing advantages. Compared with MFNet, when only 30\% training data are labeled, SEFNet yields around 4\%, 5\%, and 8\% increase, respectively, in ET, WT, and TC. 

\subsection{Comparative analysis}
\label{subsec: comparative analysis}
\subsubsection{Comparison with full-supervised methods}
\label{subsubsec: full-supervised}
We first compare our results with the state-of-the-art under full-supervised learning on the BraTs2019 validation set. The comparison is presented in Table \ref{tab:state-of-the-art}. We highlight the best-performed results in bold characters and underline the second-best results. SEFNet achieves a Dice score of 0.793, 0.868, 0.861, and 0.841, respectively, for ET, WT, TC, and the mean over the three regions. Compared with MFNet, it has an increase of 4\%, 6.2\%, and 4.2\% of Dice score in ET, TC and the mean, respectively. Also, SEFNet surpasses most of the reported methods, i.e., UNet, attention UNet, and MCNet. The performance of SEFNet is not as good as the top one solution of the BraTs2019 challenge segmentation task, which uses a Two-stage cascaded U-Net \cite{jiang2019two} (double model), with marginal performance gaps of 0.9\%, 4.1\%, and 0.3\% for ET, WT, and TC in Dice score, respectively. The two-stage cascaded U-Net framework uses one UNet for coarse segmentation and then another UNet for accurate segmentation, which could increase the segmentation accuracy to a certain degree. However, the computation cost is high and the reported memory requirement is over 12 GB during the experiment with a batch size of 1.
\begin{table}
  \centering
  \caption{Performance comparison on the BraTs2019 validation set (in case of full-supervised learning). The best results are in bold, and the second best results are underlined.}
  \scalebox{0.65}{
  \begin{tabular}{llllllllll}
  \hline
  \multicolumn{1}{l}{Methods}  &\multicolumn{4}{c}{Dice score}& &
  \multicolumn{4}{c}{HD} \\
   \cline{2-5}
   \cline{7-10}
   & ET&WT&TC&Mean &&ET&WT&TC&Mean \\
\hline
SEFNet (ours) & \underline{0.793} & 0.868 &  \underline{0.861} &\underline{0.841}&& 5.616& 8.329 & 6.618 &6.854 \\
MFNet \cite{chen20193d} &0.753& 0.880& 0.765&0.799&& \underline{4.872}& 8.022& 9.706&7.562\\
DMFNet \cite{chen20193d} &0.756&0.890&0.799&0.815& & 5.069 & 6.531&7.454& 6.351\\
3D-UNet \cite{wang20193d}&0.737&0.894&0.807&0.812 &&5.994&\underline{5.567}&7.357&\underline{6.342}\\
Dense-UNeT\cite{agravat2019brain}&0.600& 0.700& 0.630&0.643 &&11.690& 14.330& 17.100&14.373\\
AttentionUNet \cite{islam2019brain} &0.704&\underline{0.898}& 0.792& 0.798& &7.050& 6.290& 8.760& 7.370\\
MCNet \cite{li2019multi} & 0.771& 0.886&0.813 & 0.823 &&6.232 &7.409 & \underline{6.033} &6.558\\
Two-stage cascaded U-Net \cite{jiang2019two}&\bfseries 0.802 \mdseries & \bfseries 0.909 \mdseries & \bfseries 0.864 \mdseries & \bfseries  0.858 \mdseries &&  \bfseries 3.145 \mdseries &  \bfseries 4.263 \mdseries &  \bfseries 5.439 \mdseries & \bfseries 4.282 \mdseries\\
\hline
\end{tabular}}
\label{tab:state-of-the-art}
\end{table}

\begin{figure}
    \centering
     \includegraphics[width=\textwidth]{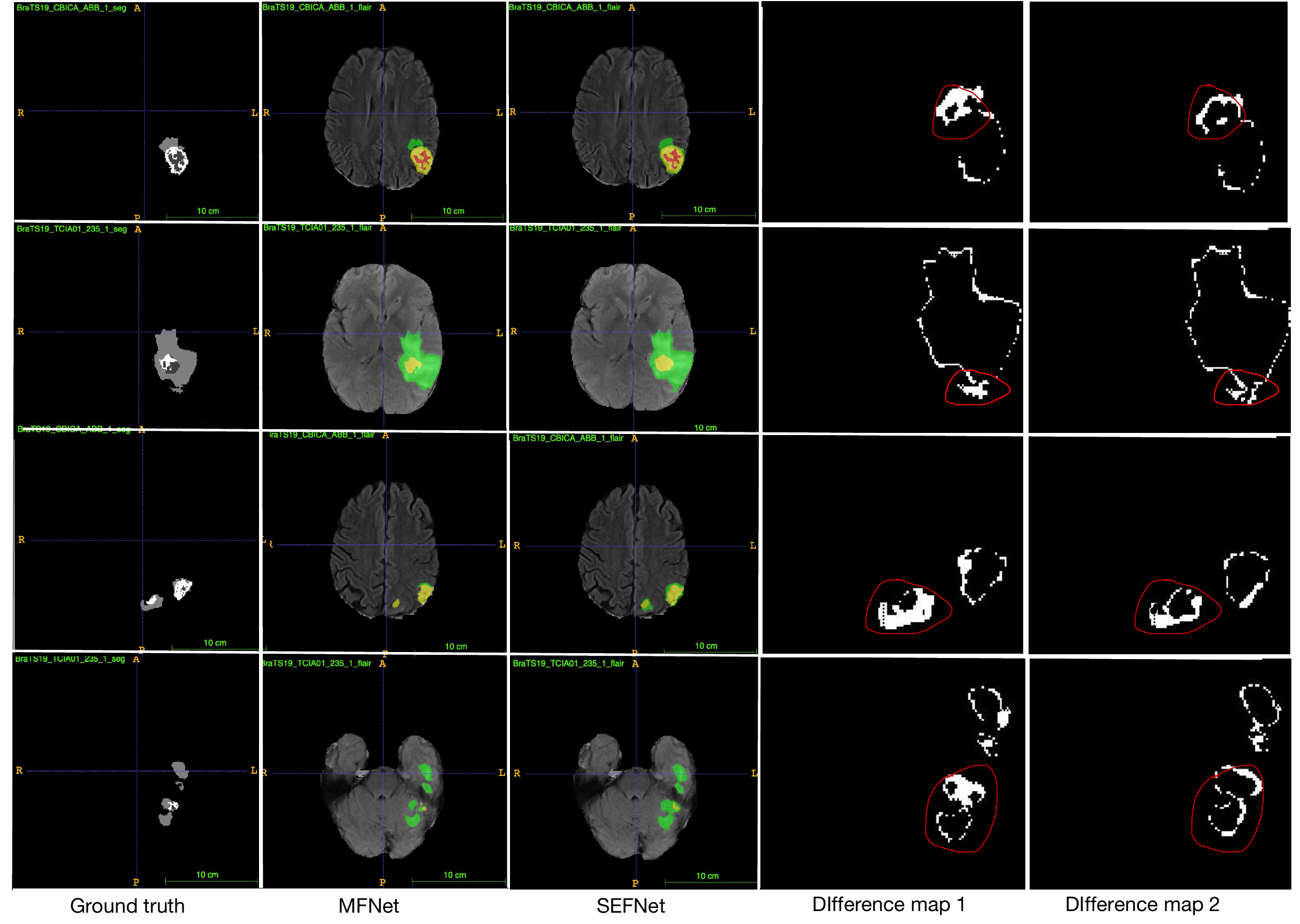}
    \caption{Example segmentation results on the BRATS2019 training dataset. The figures in rows show the results of different axial slices from left to right: the ground truth (GT), the segmentation results of MFNet and SEFNet, and the \textbf{difference map} between GT and the corresponding segmentation map. The points in white indicate where the segmented result is wrong. Labels 1, 2, and 4 are marked in red, green, and yellow, respectively. Moreover, we highlight the regions with fewer misclassification voxels by red circles in \textbf{difference map}.}
    \label{fig:imshow}
\end{figure}

Figure \ref{fig:imshow} presents a visual comparison of the brain tumor segmentation results obtained from different slices. From left to right, we can see the ground truth (GT), the segmentation result of the baseline method (MFNet) and our proposal (SEFNet), and the \textbf{difference map} between MFNet and GT on the one hand and between SEFNet and GT on the other hand. The white points in the \textbf{difference map} indicate the position the voxels are wrong-segmented.  We highlight the regions with fewer misclassification voxels by red circles in Figure \ref{fig:imshow}, where there are fewer white points. Compared with MFNet, SEFNet can generate more precise segmentation results.

\subsubsection{Comparison with semi-supervised methods}
\label{subsubsec: semi-supervised}
We also compared the segmentation performance with the state-of-the-art under semi-supervised learning. Comparing our result with those of other semi-supervised methods is difficult because BraTs datasets from different years (2015, 2017, 2018, or 2019) were tested with those methods, leading to the various training and validation set components. Also, the different propositions of training labels make the comparison difficult. In this work, we simplify the comparison by only comparing the results when 50\% of the training data are labeled. As we can see from Table \ref{tab:state-of-the-art-semi}, SEFNet achieves the best performance on validation data with the reported 0.792 mean Dice score on BraTS2019 dataset. Compared with the performance of MASSL and SAM-GAN on the BraTs2018 dataset, SEFNet yields an increase of 10\% and 4.1\% in the Dice score of WT and Mean, respectively. Compared with the best performance of PGAN on the BraTs2017 dataset, SEFNet has an increases of 16.6\%, 12.8\%, and 8.9\% in the Dice score of WT, TC, and Mean, respectively. Compared with the performance of TSMAN on the BraTs2015 dataset, SEFNet has an increase of 9\% in the mean Dice score. The above comparison demonstrates the effectiveness of SEFNet. 
\begin{table}
\centering
\caption{Performance comparison on the BraTs2019 validation set (in case of semi-supervised learning). Symbol $*$ indicates the results are not available from the published paper.}
\scalebox{0.65}{
\begin{tabular}{lllllllll}
\hline
\multicolumn{1}{c}{Dataset}&
\multicolumn{1}{c}{Method}&
\multicolumn{1}{c}{Training}&
\multicolumn{1}{c}{Test}&
\multicolumn{1}{c}{Proportion of}&
\multicolumn{4}{c}{Dice score}
\\
\cline{6-9}
&&Number&Number& labeled data&ET&WT&TC&Mean\\
\hline
\multirow{1}*{BraTs2019} & SEFNet (ours) &285&125& \multirow{1}*{50\%} & 0.721 & \bfseries 0.877  \mdseries & \bfseries 0.777 \mdseries  & \bfseries 0.792 \mdseries \\ 
\hline

\multirow{2}*{BraTs2018} &MASSL \cite{chen2019multi} &200&50& \multirow{2}*{50\%}  & *&0.770&*&*\\
& SAM-GAN\cite{xi2019semi} &285&66&& *&*&*&0.751\\
\hline
\multirow{2}*{BraTs2017} & Transfer-UNet\cite{zeng20173d}&  285&46& \multirow{2}*{50\%} &0.734 &0.690& 0.631&0.685 \\
& PGAN \cite{sun2019parasitic}&285&46&  &\bfseries 0.751 \mdseries & 0.711 &0.649&0.703\\
\hline
\multirow{3}*{BraTs2015} & Transfer-UNet \cite{zeng20173d} &140&80& \multirow{3}*{50\%}  & 0.633 & 0.616 & 0.642&0.630 \\
&PGAN \cite{sun2019parasitic}&140&80 && 0.668 & 0.652 &0.667 &0.662 \\
& TSMAN \cite{min2019two} &244&30&& * & * &* &0.707\\
\hline
\end{tabular}}
\label{tab:state-of-the-art-semi}
\end{table}

\section{Conclusion}
\label{sec:conclu4}
In this chapter, we have presented a semi-supervised multiple evidence fusion framework (SEFNet) for medical image segmentation. With the SEFNet framework, we compute two pieces of segmentation evidence: probability functions generated by a softmax layer, and mass functions generated by an ENN module. Dempster's rule is then used to fuse the two pieces of evidence and to decrease segmentation uncertainty. For images with labels, we use the supervised class-independent Dice loss to guide the training process. For images without labels, we use information constraints through image transformation operations to provide training guidance. Quantitative and qualitative results on the BraTs2019 dataset show that using Dempster's rule with semi-supervised learning makes it possible to efficiently deal with segmentation uncertainty, resulting in comparable performance under semi-supervised conditions.

There are some limitations to this work. First, the evidence from the basic belief assignment module needs to be well studied; in particular, the study of masses in the frame of discernment could be interesting and meaningful. Second, one of the potential problems in this work may be the independence of sources of information. The two pieces of evidence obtained from the probability assignment module and the basic belief assignment module are not independent because they share the same features. In the next chapter, we will first study the segmentation uncertainty and then further evaluate the framework by applying it to other medical image segmentation problems. Lastly, we will investigate the reliability of segmentation results with expected calibration error (ECE). 
 
 \stopcontents[chapters] 

\chapter{Uncertainty quantification in medical image segmentation } 
\label{Chapter5}


\tocpartial

\section{Introduction}

Despite the excellent performance of deep learning methods, the issue of quantifying prediction uncertainty remains \cite{hullermeier2021aleatoric}. This uncertainty can be classified into three types: distribution, model, and data uncertainty. Distribution uncertainty is caused by training-test distribution mismatch (dataset shift) \cite{quinonero2008dataset}. Model uncertainty arises from limited training set size and model misspecification  \cite{mehta2019propagating,maddox2019simple,yu2019uncertainty}. Finally, sources of data uncertainty include class overlap, label noise, and homo or hetero-scedastic noise. Because of the limitations of medical imaging and labeling technology, as well as the need to use a large nonlinear parametric segmentation model, medical image segmentation results are particularly tainted with uncertainty, which limits the segmentation reliability. 

\begin{figure}
\includegraphics[width=\textwidth]{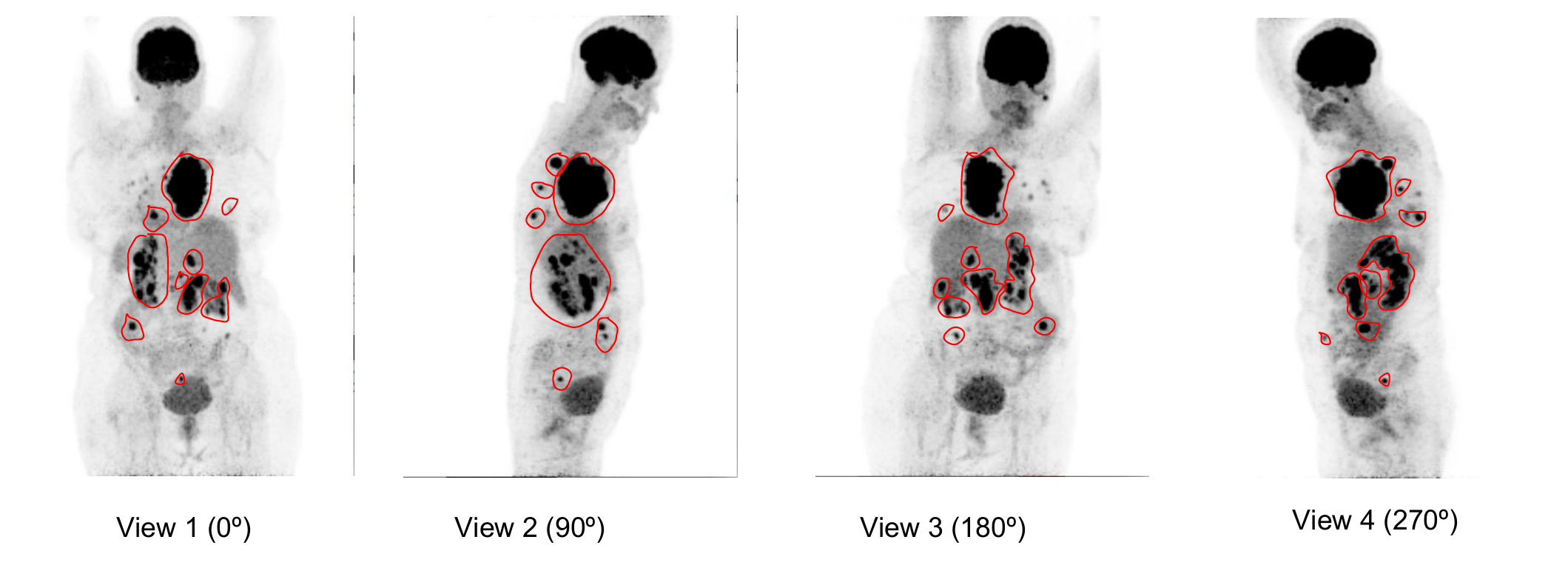}
\caption{Example of a patient with lymphomas in 3D PET-CT views. The lymphomas are the dark areas circled in red.}
\label{fig1}
\end{figure}

In clinical lymphoma diagnosis and radiotherapy planning, PET-CT scanning is an effective imaging tool to locate and segment lymphomas. The standardized uptake value (SUV), defined as the measured activity normalized for body weight and injected dose to remove variability in image intensity between patients, is widely used to locate and segment lymphomas thanks to its high sensitivity and specificity to the metabolic activity of tumors \cite{jhanwar2006role}. However, PET images have a low resolution and suffer from the partial volume effect blurring the contours of objects \cite{zaidi2010pet}. For that reason, CT images are usually used jointly with PET images because of their anatomical feature-representation capability and high resolution. Figure \ref{fig1} shows 3D PET-CT views of a lymphoma patient in which PET and CT are merged for showing. The lymphomas are in black, as well as the brain and the bladder. As we can see from this figure, lymphomas vary in intensity distribution, shape, type, and number. The attributes of lymphomas make the study of automatic lymphoma segmentation challenging and unreliable.
 
In this chapter, we introduce a different uncertainty quantification approach in medical image segmentation. The main idea is to hybridize a deep medical image segmentation model with evidential classifiers (one of the classifiers, ENN, was introduced in Section \ref{subsubsec: distance}, and the other is the Radial basis function (RBF) network-based evidential classifier). Two evidential layers are first compared based on different ways of using distances to prototypes for computing mass functions, to show the effectiveness of uncertainty quantification. Then the two evidential classifiers are plugged with deep segmentation models to construct deep evidential segmentation models to quantify segmentation uncertainty. The whole model is trained end-to-end by minimizing the Dice loss function. The proposed combination of deep feature extraction and evidential segmentation is shown to outperform the baseline model as well as three other state-of-the-art models on a dataset of 173 patients with lymphomas.

We organize this chapter as follows: Section \ref{sec:model5} first compares the performance of uncertainty quantification of two evidential classifiers and then introduces two deep evidential segmentation models with the application on lymphoma segmentation. Section \ref{sec: exper5} reports the numerical experiments, which demonstrate the advantages of uncertainty quantification of the proposed segmentation models. Finally, we conclude this work in Section \ref{sec:conc5}.

\section{Proposed method}
\label{sec:model5}

\subsection{Uncertainty quantification with evidential classifiers}
\label{sec:evclass}

In this section, we review two methods for designing classifiers that output mass functions to quantify uncertainty, referred to as \emph{evidential classifiers}. The Evidential neural network (ENN) classifier introduced in \cite{denoeux2000neural} is first recalled in Section \ref{subsubsec: distance}. A new model based on the interpretation of a radial basis function (RBF) network as combining of simple mass functions by Dempster's rule, inspired by \cite{denoeux19d}, is then described in Section \ref{subsubsec: rbf}. The two models are compared experimentally in Section \ref{subsubsec:compar}.

\subsubsection{Evidential neural network (ENN)}
\label{subsubsec: enn}

In \cite{denoeux2000neural}, Den{\oe}ux proposed the ENN classifier, in which mass functions are computed based on distances to prototypes. The basic idea is to consider each prototype as a piece of evidence, which is discounted based on its distance to the input vector. The evidence from different prototypes is then pooled by Dempster's rule.

The ENN classifier is composed of an input layer of $H$ neurons (where $H$ is the dimension of input space), two hidden layers, and an output layer. The first input layer is composed of $I$ units, whose weights vectors are prototypes \new{$\bp_1,\ldots, \bp_I$} in input space. The second hidden layer computes mass functions $m_i$ representing the evidence of each prototype $\bp_i$. The third layer combines the $I$ mass functions $m_1,\ldots,m_I$ using Dempster's rule. A short description of the ENN model is given in Section \ref{subsubsec: distance} section. 

The idea of applying the above model to features extracted by a convolutional neural network (CNN) was first proposed by Tong et al. in \cite{tong21b}. In this approach, the ENN module becomes an ``evidential layer'', which is plugged into the output of a CNN instead of the usual softmax layer. The feature extraction and evidential modules are trained simultaneously. A similar approach was applied in  \cite{tong2021evidential} to semantic segmentation.  In the next section, we present an alternative approach based on a radial basis function (RBF) network and weights of evidence.

\subsubsection{Radial basis function (RBF) network} 
\label{subsubsec: rbf}

As shown in \cite{denoeux19d}, the calculations performed in the softmax layer of a feedforward neural network can be interpreted in terms of a combination of evidence by Dempster's rule. The output class probabilities can be seen as normalized plausibilities according to an underlying belief function. Applying these ideas to an RBF network, it is possible to derive an alternative evidential classifier with  properties similar to those of the ENN model recalled in Section \ref{subsubsec: distance}. Consider an RBF network with $I$ prototype (hidden) units. The activation of hidden unit $i$ is
\begin{equation}
    s_i=\exp(-\gamma_i d_i^2),   
    \label{eq:activRBF}
\end{equation}
where, as before, $d_i=\left \| \bx-\bp_i \right \| $ is the Euclidean distance between input vector $\bx$ and prototype $\bp_i$, and $\gamma_i>0$ is a scale parameter. For the application considered in this chapter, we only need to consider  the case of binary classification with $C=2$ and $\Omega=\{\omega_1,\omega_2\}$. (The case where $C>2$ is also analyzed in \cite{denoeux19d}). Let $v_{i}$ be the weight of the connection between hidden unit $i$ and the output unit, and let $w_i=s_i v_i$ be the product of the output of unit $i$ and weight $v_i$. The quantities $w_i$ can be interpreted as weights of evidence for class $\omega_1$ or $\omega_2$, depending on the sign of $v_i$:
\begin{itemize}
    \item If $v_{i}\ge 0$, $w_i$  a weight of evidence for class $\omega_1$;
    \item If $v_{i}<0$, $-w_i$ is a weight of evidence for class $\omega_2$.
\end{itemize}
To each prototype, $i$ can, thus, be associated with the following simple mass function:
\[
m_{i}=\{\omega_1\}^{w_{i}^+} \oplus \somega{2}^{w_{i}^-},
\]
where \new{$w_{i}^+=\max(0,w_i)$ and $w_{i}^-=-\min(0,w_i)$} denote, respectively, the positive and negative parts of $w_i$. Combining the evidence of all prototypes in favor of $\omega_1$ or $\omega_2$ by Dempster's rule, we get the mass function
\begin{equation}
\label{eq:mRBF1}
m=\bigoplus_{i=1}^I m_{i}=\{\omega_1\}^{w^+} \oplus \somega{2}^{w^-},
\end{equation}
with $w^+=\sum_{i=1}^I w_{i}^+$ and $w^-=\sum_{i=1}^I w_{i}^-$. 
In \cite{denoeux19d}, the normalized plausibility of $\omega_1$ corresponding to mass function $m$ was shown to have the following expression:
\begin{equation}
\label{eq:p}
    p(\omega_1)=\frac{Pl(\somega{1})}{Pl(\somega{1})+Pl(\somega{2})}=\frac{1}{1+\exp(-\sum_{i=1}^I v_i s_i)},
\end{equation}
i.e., it is the output of a unit with a logistic activation function. When training an RBF network with a logistic output unit, we thus actually combine evidence from each of the prototypes, but the combined mass function remains latent. In \cite{denoeux19d}, mass function $m$ defined by \eqref{eq:mRBF1} was shown to have the following expression:
\begin{subequations}
\label{eq:m12}
\begin{align}
m(\somega{1}) &= \frac{[1-\exp(-w^+)]\exp(-w^-)}{1-\kappa} \\
m(\somega{2}) &= \frac{[1-\exp(-w^-)]\exp(-w^+)}{1-\kappa}\\
\label{eq:m12Omega}
m(\Omega) &= \frac{\exp(-w^+-w^-)}{1-\kappa}=\frac{\exp(-\sum_{i=1}^I|w_i|)}{1-\kappa},
\end{align}
where
\begin{equation}
\label{eq:conf12}
\kappa=[1-\exp(-w^+)] [1-\exp(-w^-)]
\end{equation}
\end{subequations}
is the degree of conflict between mass functions $\{\omega_1\}^{w^+}$ and $\{\omega_2\}^{w^-}$.

In this approach, we thus need to train a standard RBF network with $I$ prototype units and one output unit with a logistic activation function by minimizing a loss function. Here we define it as the regularized cross-entropy loss
\begin{equation}
L_{CE}(\btheta)=-\sum_{n=1}^N \left(y_n\log p_n  + (1-y_n)\log(1-p_n)\right) + \lambda \sum_{i=1}^I w_i^2,
\label{eq:lossRBF}
\end{equation}
where $p_n$ is the normalized plausibility of class $\omega_1$ computed from \eqref{eq:p} for instance $n$, \new{$y_n$ is class label of instance $n$ ($y_n=1$ if the true class of instance $n$ is $\omega_1$, and $y_n=0$ otherwise),} and $\lambda$ is a hyperparameter. We note that increasing $\lambda$ has the effect of decreasing the weights of evidence from prototypes and, thus, obtaining less informative mass functions. 

\subsubsection{Comparison between ENN and RBF network}
\label{subsubsec:compar}

\begin{figure}
\begin{center}
\includegraphics[width=0.5\textwidth]{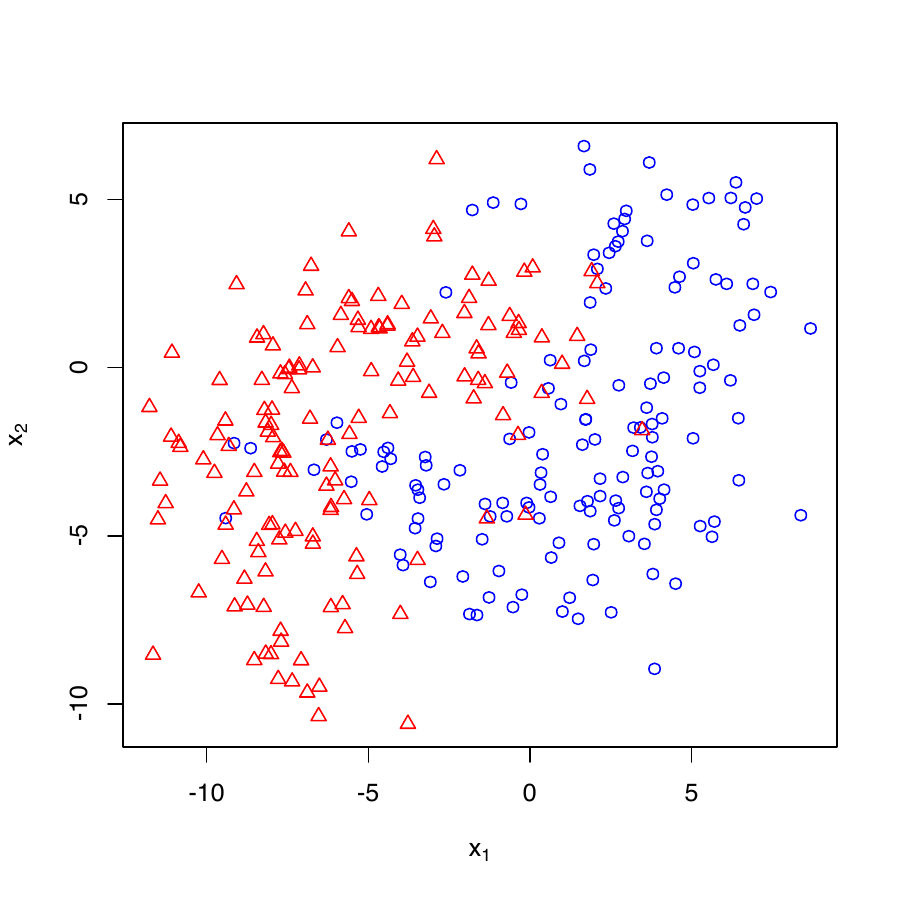}
\end{center}
\caption{Simulated data. \label{fig:bananas}}
\end{figure}

To compare the uncertainty quantification performance of the two classifiers (ENN and RBF network), we consider the two-class dataset shown in Figure \ref{fig:bananas} and show the test error rate and the mean uncertainty of the two classifiers. Furthermore, we also plot the contours of the mass assigned to given classes and the uncertainty. The two classes are randomly distributed around half circles with Gaussian noise and are separated by a nonlinear boundary. A learning set of size $N=300$ and a test set of size 1000 were generated from the same distribution. 

\begin{figure}
\subfloat[\label{fig:error}]{\includegraphics[width=0.5\textwidth]{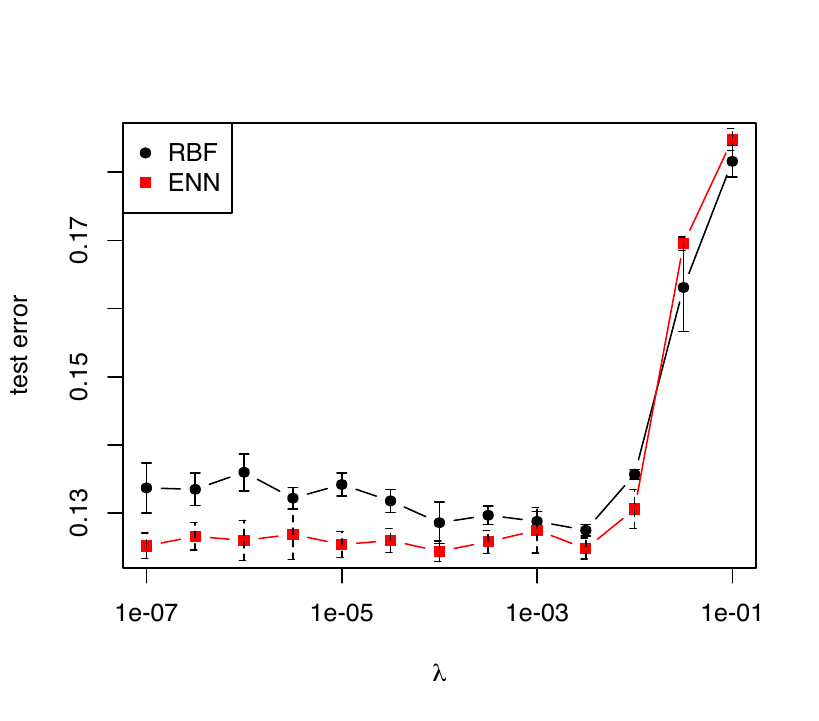}}
\subfloat[\label{fig:uncert}]{\includegraphics[width=0.5\textwidth]{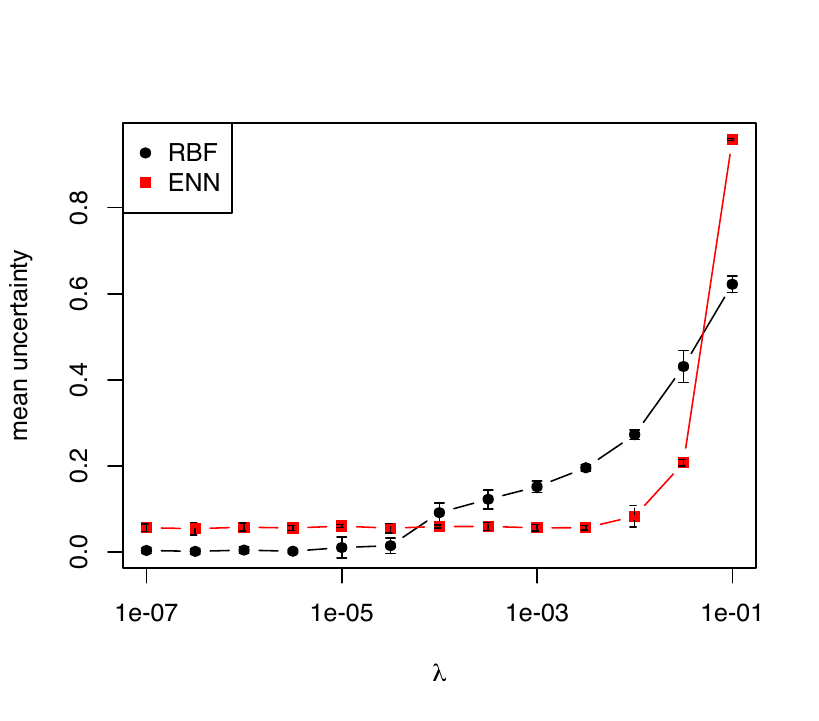}}
\caption{Test error rates (a) and mean uncertainty (b) for the ENN and RBF models, as functions of regularization parameter $\lambda$. \label{fig:compar_err}}
\end{figure}

An ENN and an RBF network were initialized with $I=6$ prototypes generated by the $k$-means algorithm and were trained on the learning data. Figures \ref{fig:error} and \ref{fig:uncert} show, respectively, the test error rate and the mean uncertainty (defined as the average mass assigned to the frame $\Omega$), as functions of hyperparameter $\lambda$ in \eqref{eq:lossENN} and \eqref{eq:lossRBF}, for 10 different runs of both algorithms with different initializations. As expected, uncertainty increases with $\lambda$ for both models, but the ENN model appears to be less sensitive to $\lambda$ as compared to the RBF model. Both models achieve similar minimum error rates for $\lambda$ around $10^{-3}$, and have similar mean uncertainties for $\lambda=10^{-4}$. 

\begin{figure}
\centering
\subfloat[\label{fig:contourRBFomega1}]{\includegraphics[width=0.35\textwidth]{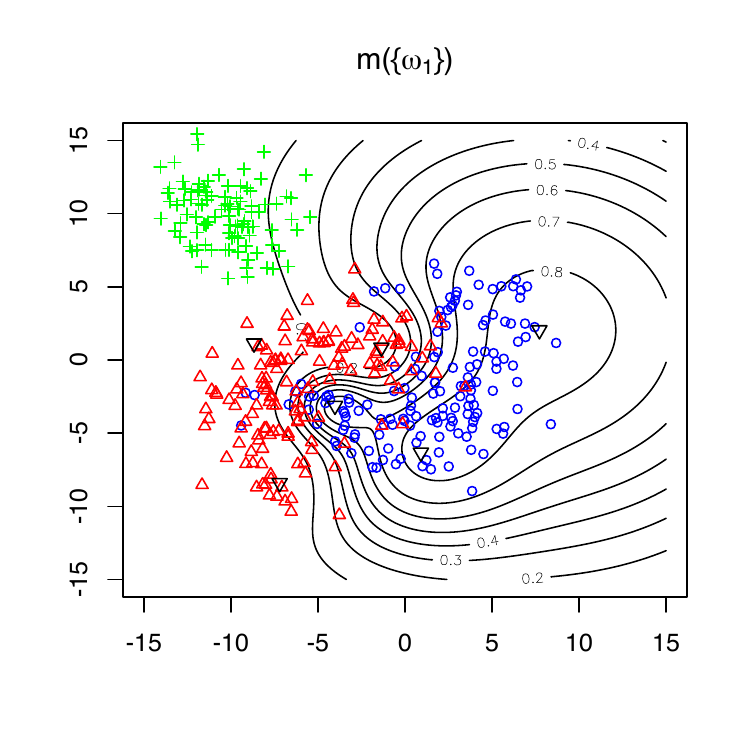}}
\subfloat[\label{fig:contourENNomega1}]{\includegraphics[width=0.35\textwidth]{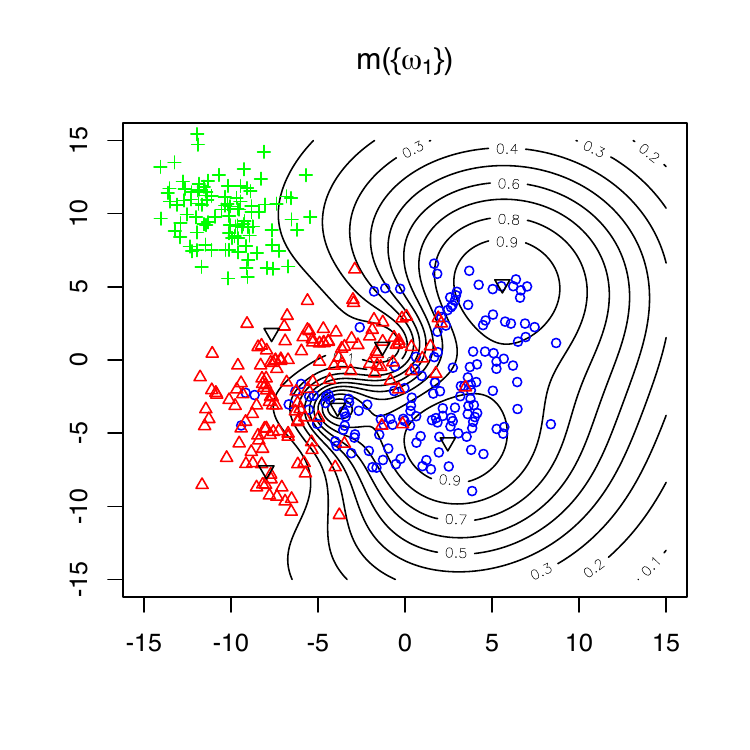}}\\
\subfloat[\label{fig:contourRBFomega2}]{\includegraphics[width=0.35\textwidth]{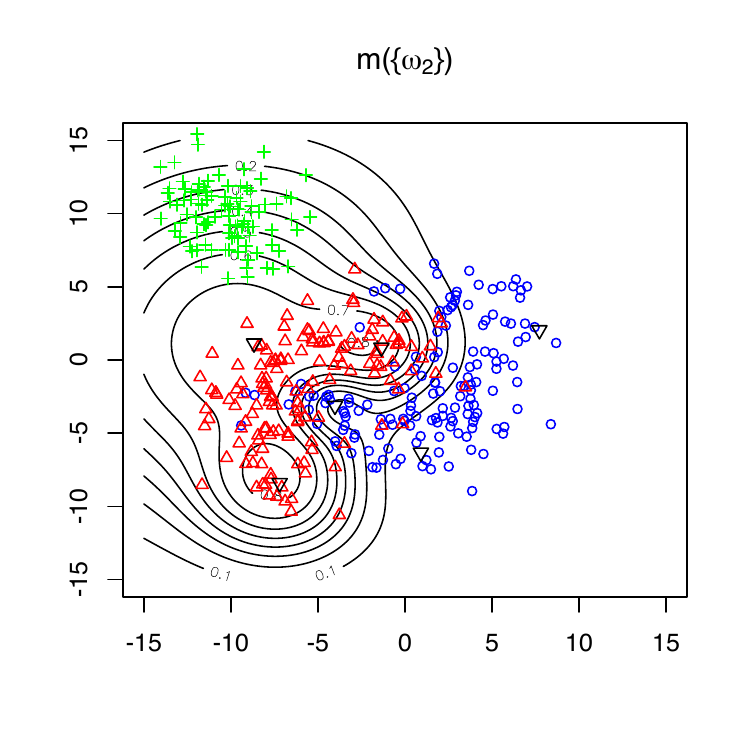}}
\subfloat[\label{fig:contourENNomega2}]{\includegraphics[width=0.35\textwidth]{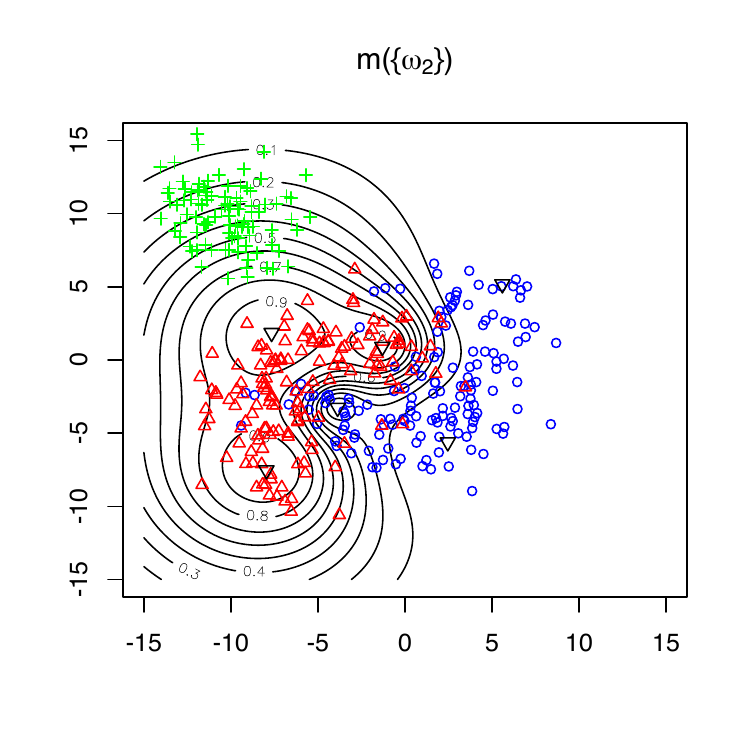}}\\
\subfloat[\label{fig:contourRBFOmega}]{\includegraphics[width=0.35\textwidth]{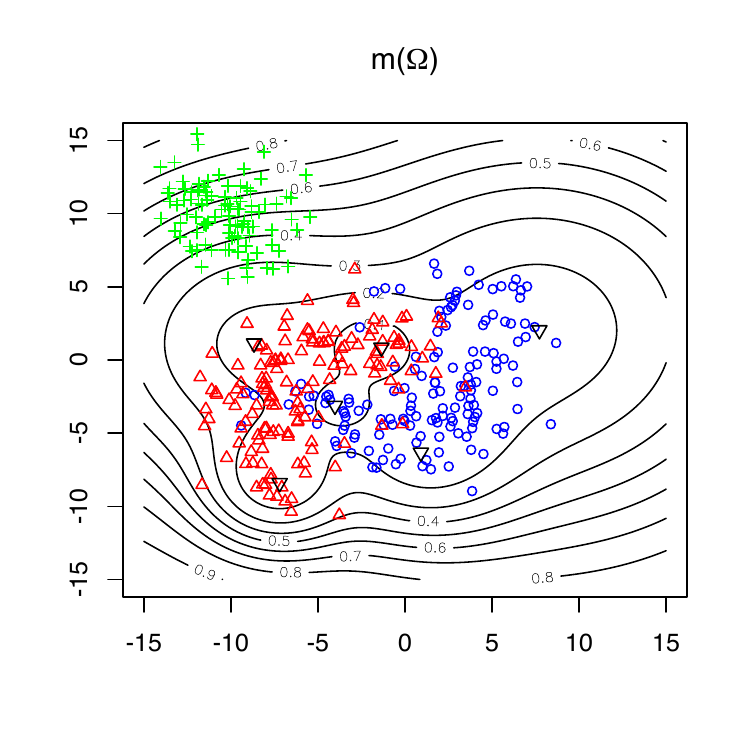}}
\subfloat[\label{fig:contourENNOmega}]{\includegraphics[width=0.35\textwidth]{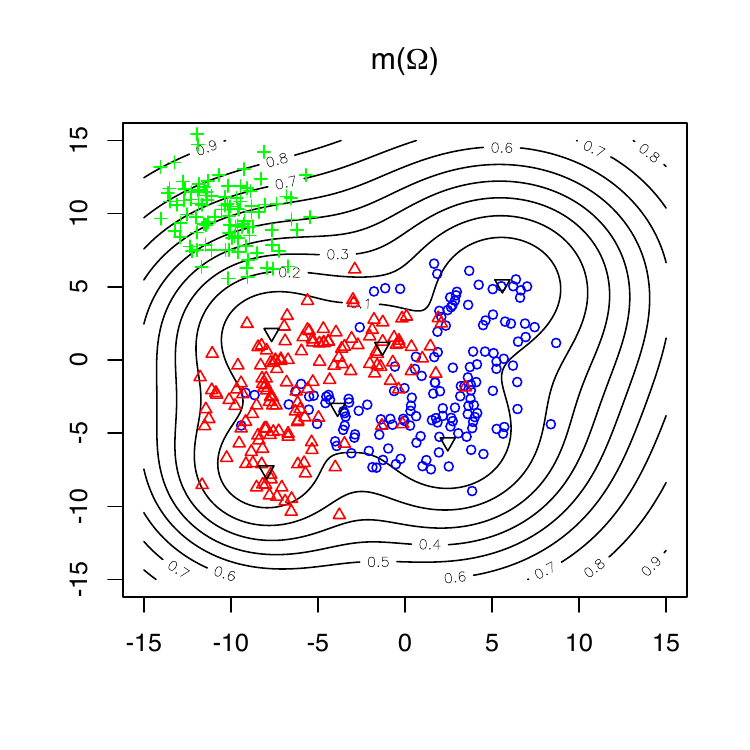}}
\caption{Contours of the mass assigned to $\{\omega_1\}$, $\{\omega_2\}$ and $\Omega$ by the RBF  (left column) and ENN (right column) models. The training data are displayed in blue and red, and the third class (absent from the training data) is shown in green. The training was done with $\lambda=0.001$ for the two models. 
\label{fig:contours}}
\end{figure}

\begin{figure}
\centering
\subfloat[\label{fig:omega1}]{\includegraphics[width=0.4\textwidth]{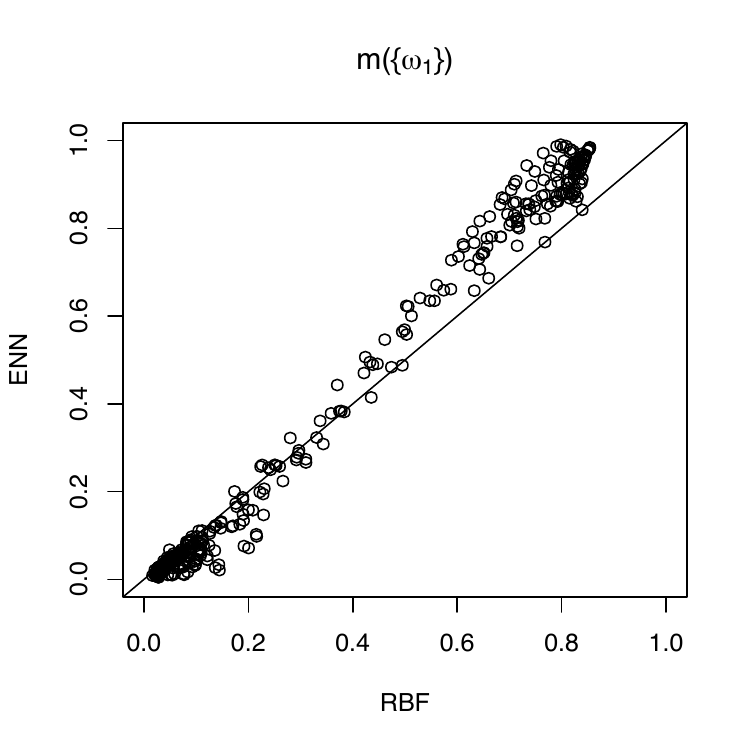}}
\subfloat[\label{fig:omega2}]{\includegraphics[width=0.4\textwidth]{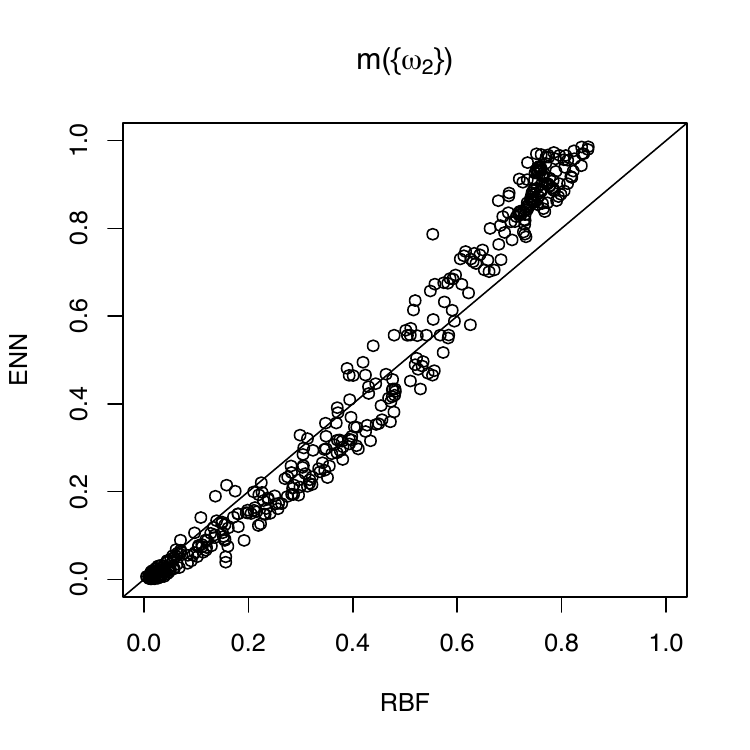}}\\
\subfloat[\label{fig:Omega}]{\includegraphics[width=0.4\textwidth]{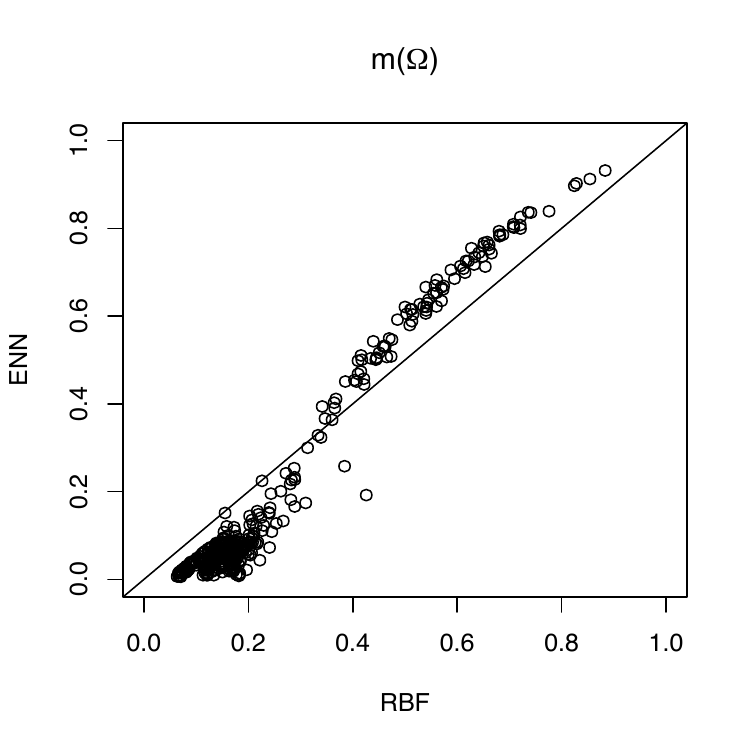}}
\caption{Masses computed by the RBF network (horizontal axis) versus the ENN model (vertical axis) for the extended dataset. \label{fig:compar_mass}}
\end{figure}

\new{As shown in \cite{denoeux2000neural}, the robustness of the ENN model arises from the fact that, when the input $\bx$ is far from all prototypes, the output mass function $m$ is close to the vacuous mass function. This property, in particular, makes the network capable of detecting observations generated from a distribution that is not represented in the learning set. From \eqref{eq:m12Omega}, we can expect the RBF network model to have a similar property: if $\bx$ is far from all prototypes, all weights of evidence $w_i$ will be small and the mass $m(\Omega)$ will be close to unity. To compare the mass functions computed by the two models, not only in regions of high density where training data are present, but also in regions of low density, we introduced  a third class in the test set, as shown in Figure \ref{fig:contours}}. Figure \ref{fig:compar_mass} shows scatter plots of masses on each of the focal sets computed for the two models trained with $\lambda=10^{-3}$ and applied to an extended dataset \new{composed of the learning data and the third class}. We can see that the mass functions are quite similar. Contour plots  shown in Figure \ref{fig:contours} confirm this similarity.

\subsection{Uncertainty quantification in evidential segmentation}
\label{subsec: evidential segmentation}
\begin{figure}
\includegraphics[width=\textwidth]{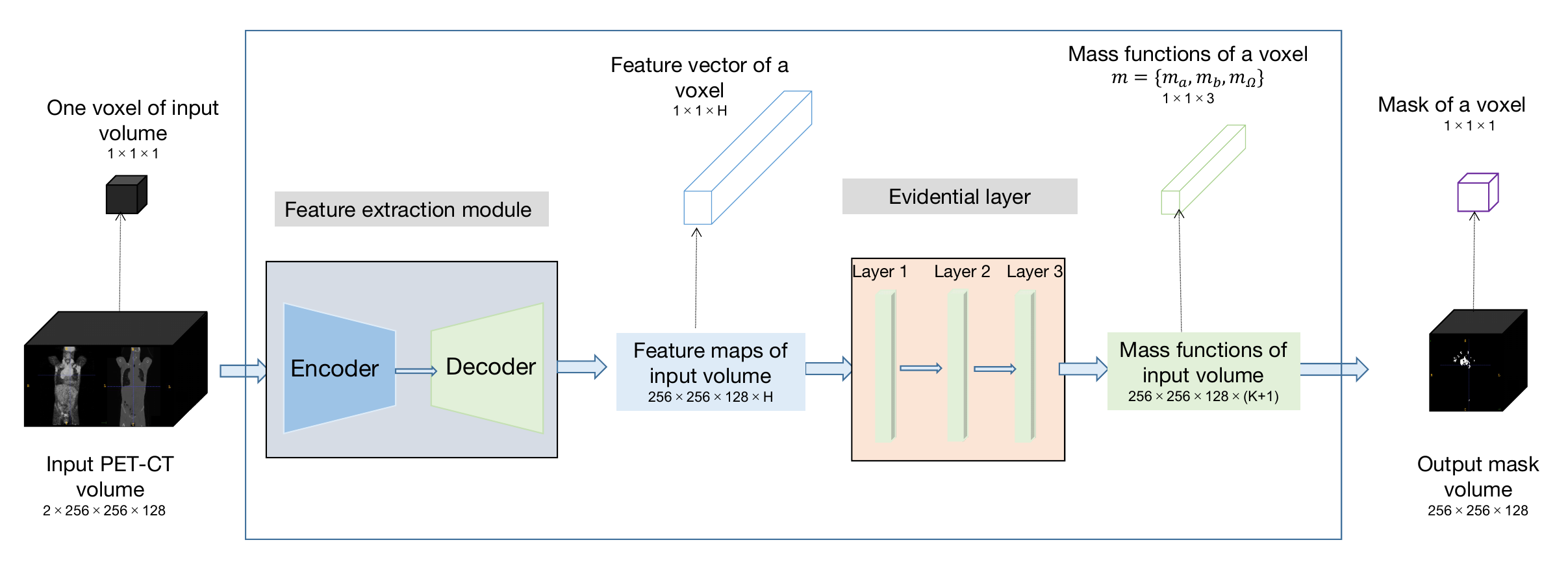}
\caption{Global lymphoma segmentation model.}
\label{fig:arch}
\end{figure}

We propose an evidential segmentation model to delineate lymphomas, shown in Figure \ref{fig:arch}. It is composed of an encoder-decoder feature extraction module, and an evidential layer based on one of the two models, ENN and RBF network. The input is the concatenated PET-CT image volume provided as  a tensor of size $2 \times256 \times256 \times 128$, where 2 corresponds to the number of modality channels, and $256 \times256 \times 128$ is the size of each input volume. The PET-CT image volumes are first fed into the feature extraction module, which outputs high-level features in the form of a  tensor of size $256 \times256 \times 128 \times H$, where $H$ is the number of features computed at each voxel. This tensor is then fed into the evidential layer, which outputs mass functions representing evidence about the class or the ignorance (uncertainty) of each voxel, resulting in a tensor of size $256 \times256 \times 128 \times (C+1)$, where $C+1$ is the number of masses (one for each class and one for the frame of discernment $\Omega$). The whole network is trained end-to-end by minimizing a regularized Dice loss. The different components of this model are described in greater detail below.

\paragraph{Feature extraction module} The feature extraction module is based on a UNet with residual encoder and decoder layers (see Section \ref{subsec:Unet}), as shown in Figure~\ref{fig-unet}. Each down-sampling layer (marked in blue) is composed of convolution, normalization, dropout and activation blocks. Each up-sampling layer (marked in green) is composed of transpose convolution, normalization, dropout and activation blocks. The last layer (marked in yellow) is the bottom connection which does not down or up-sample the data. In the experiments reported in Section \ref{sec: exper5}, the channels (number of filters) were set as $(8, 16, 32, 64, 128)$ with kernel size equal to 5 and convolutional strides equal to $(2, 2, 2, 2)$. The spatial dimension, input channel and output channel of the module were set, respectively, as 3, 2, and the number $H$ of extracted features. (Experiments with several values of $H$ are reported in Section \ref{subsec:sensitivity}). The dropout rate was set as 0 and no padding operation was applied. Instance normalization \cite{ulyanov2017improved} was used to perform intensity normalization across the width, height and depth of a single feature map of a single example. The Parametric Rectified Linear Unit (PReLU) function \cite{he2015delving}, which generalizes the traditional rectified unit with a slope for negative values, was used as the activation function. For each input voxel, the feature extraction module outputs a $1 \times H$ feature vector, which is fed into the evidential layer.

\paragraph{Evidential layer}
A probabilistic network with a softmax output layer may assign voxels a high probability of belonging to one class while the segmentation uncertainty is actually high because, e.g., the feature vector describing that voxel is far away from feature vectors presented during training. Here, we propose to plug in one of the evidential classifiers described in Sections \ref{subsubsec: enn} and \ref{subsubsec: rbf} at the output of the feature extraction module. The ENN or RBF classifier then takes as inputs the high-level feature vectors computed by the UNet and computes, for each voxel $n$, a mass function $m_n$ on the frame $\Omega=\{\omega_1,\omega_2\}$, where $\omega_1$ and $\omega_2$ denote, respectively, the background and the lymphoma class, and $\Omega$ denotes the uncertainty (see Section \ref{subsec:calibration} for more details about uncertainty quantification). We will use the names ``ENN-UNet''  and ``RBF-UNet'' to designate the two variants of the model. 

\paragraph{Loss function}
The whole network is trained end-to-end by minimizing a regularized Dice loss (see Section \ref{subsec: evaluation cri}). We use the Dice loss instead of the original cross-entropy loss in UNet because the quality of the segmentation is finally assessed by the Dice coefficient. The Dice loss is defined as
\begin{equation}
    \textsf{loss}_{D}=1-\frac{2 \sum_{n=1}^{N} S_n G_n}{ \sum_{n=1}^{N} S_n+ \sum_{n=1}^{N} G_n},
    \label{eq:Dice_loss}
\end{equation}
where $N$ is the number of voxels in the image volume, $S_n$ is the output pignistic probability of the tumor class (i.e., $m_n(\{\omega_2\})+ m_n(\Omega)/2$) for voxel $n$, and $G_n$ is ground truth for voxel $n$, defined as $G_n=1$ if voxel $n$ corresponds to a tumor, and $G_n=0$ otherwise. The regularized loss function is 
    \begin{equation}
        \textsf{loss}=\textsf{loss}_{D} +\lambda R,  
    \label{eq:22}
    \end{equation}
where $\lambda$ is the regularization coefficient and $R$ is a regularizer defined either as $R=\sum_i \alpha_i$ \eqref{eq:lossENN} if the ENN classifier is used in the ES module or as $R=\sum_i v_i^2$ \eqref{eq:lossRBF} if the RBF classifier is used. The regularization term allows us to decrease the influence of unimportant prototypes and avoid overfitting. 


\section{Experiments and results}
\label{sec: exper5}

The model introduced in Section \ref{subsec: evidential segmentation} was applied to a set of PET-CT data recorded on patients with lymphomas\footnote{The code is available at \url{https://github.com/iWeisskohl.}}. The experimental settings are first described in Section \ref{subsec:settings}. A sensitivity analysis with respect to the main hyperparameters is first reported in Section \ref{subsec:sensitivity}. We then compare the segmentation accuracy and reliability of our models with those of state-of-the-art methods in Sections \ref{subsec:state-of-art} and \ref{subsec:calibration}, respectively.

\subsection{Experiment settings}
\label{subsec:settings}
\begin{figure}
\includegraphics[width=\textwidth]{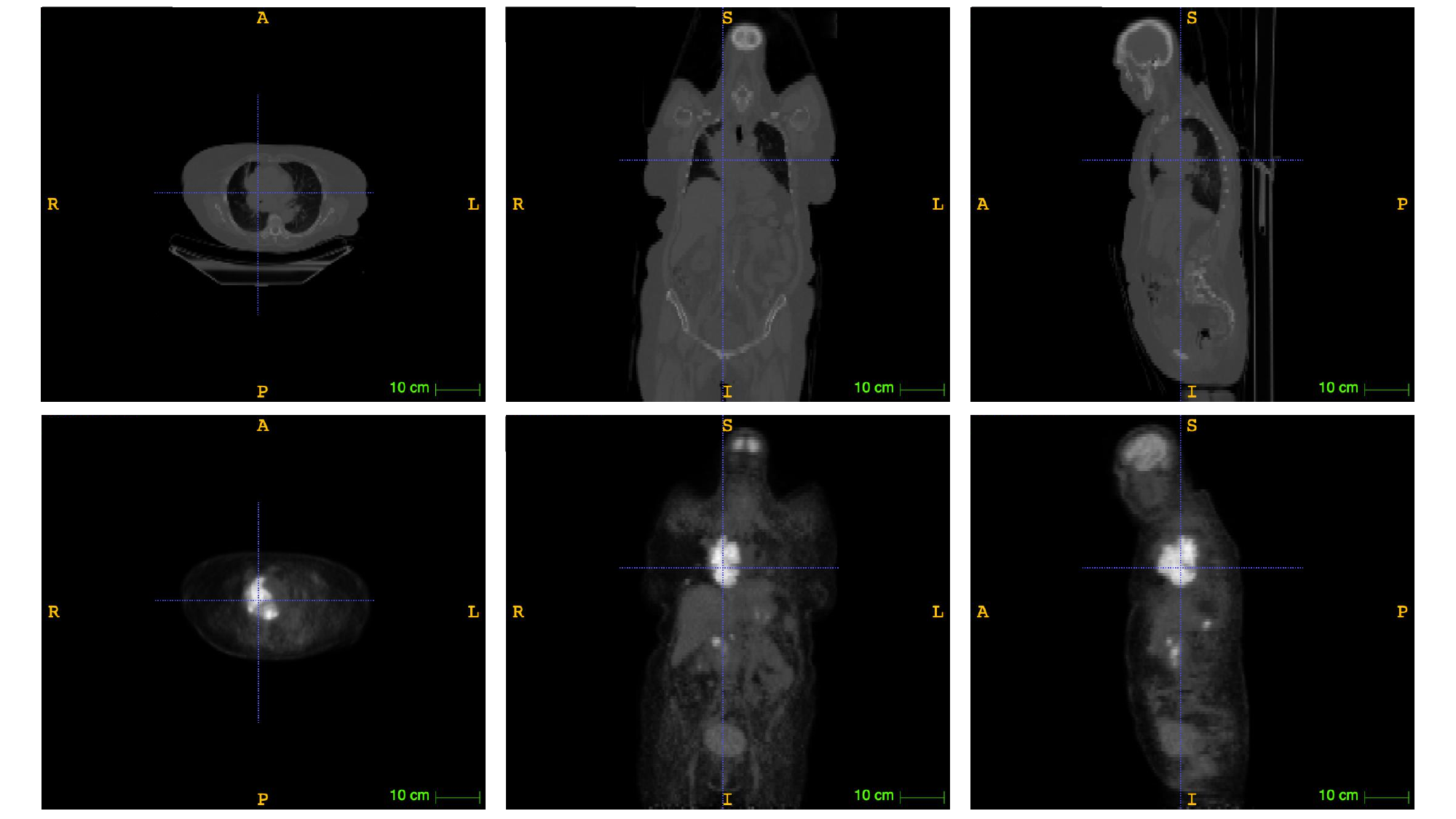}
\caption{Example of a patient with lymphomas. The first and second rows show, respectively, CT and PET slices for one patient in axial, sagittal and coronal views.}
\label{fig2}
\end{figure}

\paragraph{Dataset}
The dataset considered in this chapter contains 3D images from 173 patients who were diagnosed with large B-cell lymphomas and underwent PET-CT examination. (The study was approved as a retrospective study by the Henri Becquerel Center Institutional Review Board, Rouen, France). The lymphomas in mask images were delineated manually by experts and considered as ground truth. Figure \ref{fig2} shows examples of PET and CT image slices for one patient with lymphomas. As can be seen, lymphomas in PET images  usually correspond to the brightest pixels, but organs such as the brain and bladder are also located in bright pixel areas, which may result in segmentation errors. Moreover, lymphoma boundaries are blurred, which makes it hard to delineate lymphomas precisely. 
All PET/CT data were stored in the  DICOM (Digital Imaging and Communication in Medicine) format. The size and spatial resolution of PET and CT images and the corresponding mask images vary due to the use of different imaging machines and operations. For CT images, the size varies from $267\times 512\times512$ to $478\times 512\times512$. For PET images, the size varies from $276\times 144\times144$ to $407\times 256\times256$. 

\paragraph{Pre-processing}
Several pre-processing methods were used to process the PET/CT data. At first, the data in DICOM format were transferred into the NIFTI (Neuroimaging Informatics Technology Initiative) format for further processing. Second, the PET, CT and mask images were normalized: (1) for PET images, we applied a random intensity shift and scale of each channel with the shift value of 0 and scale value of 0.1; (2) for CT images, the shift and scale values were set to 1000 and 1/2000; (3) for mask images, the intensity value was normalized into the $[0,1]$ interval by replacing the outside value by $1$. Third, PET and CT images were resized to $256\times 256\times 128$ by linear interpolation, and mask images were resized to $256\times 256\times 128$ by nearest neighbor interpolation. Lastly, the registration of CT and PET images was performed by B-spline interpolation. All the prepossessing methods can be found in the SimpleITK \cite{lowekamp2013}\cite{yaniv2018} toolkit. During training, PET and CT images were concatenated as a two-channel input. We randomly selected 80\% of the data for training, 10\% for validation, and 10\% for testing. This partition was fixed and used in all the experiments reported below.  

\paragraph{Parameter initialization}
For the evidential layer module, we considered two variants based on the ENN classifier recalled in Section \ref{subsubsec: enn} on the one hand, and an RBF network as described in Section \ref{subsubsec: rbf} on the other hand. Both approaches are based on prototypes in the space of features extracted by the UNet module. Two ways can be used to initialize the prototypes: $k$-means initialization and random initialization. When using ENN or RBF classifiers as stand-alone classifiers, prototypes are usually initialized by a clustering algorithm such as the $k$-means. Here, this approach is not so easy, because the whole network is trained in an end-to-end way, and the features are constructed during the training process. However, $k$-means initialization can still be performed by a four-step process:
\begin{enumerate}
    \item A standard UNet architecture (with a softmax output layer) is trained end-to-end;
    \item The $k$-means algorithm is run in the space of features extracted by the trained UNet;
    \item The evidential layer is trained alone, starting from the initial prototypes computed by the $k$-means;
    \item The whole model (feature extraction module and evidential layer) is fine-tuned by end-to-end learning with a small learning step.
\end{enumerate}
As an alternative method, we also considered training the feature extraction module and the evidential layer simultaneously, in which case the prototypes were initialized randomly from a normal distribution with zero mean and identity covariance matrix. For the ENN module, the initial values of parameters $\alpha_i$ and $\gamma_i$ were set, respectively, at 0.5 and 0.01, and membership degrees $u_{ik}$ were initialized randomly by drawing uniform random numbers and normalizing. For the RBF module, the initial value of the scale parameter \textsf{$\gamma_i$} of RBF was set to 0.01, and the weight $v_i$ was drawn randomly from a standard normal distribution.

\paragraph{Learning algorithm} Each model was trained on the learning set with 100 epochs using the Adam optimization algorithm. The initial learning rate was set to $10^{-3}$. An adjusted learning rate schedule was applied by reducing the learning rate when the training loss did not decrease in 10 epochs. The model with the best performance on the validation set was saved as the final model for testing.  
All methods were implemented in Python with the PyTorch-based medical image framework MONAI, and were trained and tested on a desktop with a 2.20GHz Intel(R) Xeon(R) CPU E5-2698 v4 and a Tesla V100-SXM2 graphics card with 32 GB GPU memory.

\paragraph{Evaluation criteria}
Dice score, Sensitivity and Precision were used to assess the quality of medical image segmentation algorithms. These criteria are defined as follows in \eqref{eq: dice_score}, \eqref{eq: sensi} and \eqref{eq: precision}, respectively. In addition to the quality of the segmentation, we also wish to evaluate the reliability of output probabilities or belief functions. For that purpose, we used an additional evaluation criterion, the Expected Calibration Error (ECE) \cite{guo2017calibration}. The output pignistic probabilities from the evidential layer are first discretized into $R$ equally spaced bins $B_r$, $r=1,\ldots,R$ (we used $R=10$). The accuracy of bin $B_r$ is defined as
 \begin{equation}
     \acc(B_r)=\frac{1}{\mid B_r \mid}\sum_{i\in B_r}^{} \boldsymbol{1} (P_i=G_i),
 \end{equation}
 where $P_i$ and $G_i$ are, respectively, the predicted and true class labels for sample $i$. The average confidence of bin $B_r$ is defined as
 \begin{equation}
\conf(B_r)=\frac{1}{\mid B_r \mid}\sum_{i\in B_r}^{}S_i,
 \end{equation}
where $S_i$ is the confidence for sample $i$. The ECE is the weighted average of the difference in accuracy and confidence of the bins:
\begin{equation}
\label{eq:ECE}
\textsf{ECE}= \sum_{r=1}^{R} \frac{\mid B_r \mid }{N}\mid \acc(B_r)-\conf(B_r)\mid,
\end{equation}
where $N$ is the total number of elements in all bins, and $\mid B_r\mid$ is the number of elements in bin $B_r$. A model is perfectly calibrated when $\acc(B_r)=\conf(B_r)$ for all $r\in \{1,\ldots,R\}$. Through the bin-size weighting in the ECE metric, the highly confident and accurate background voxels significantly affect the results. Because  our dataset has an imbalanced foreground and background proportions, we  only considered voxels belonging to the tumor to calculate the ECE, similar to \cite{jungo2020analyzing}\cite{rousseau2021post}. For each patient in the test set, we defined a bounding box covering the lymphoma region and calculated the ECE in this bounding box. We are interested in the patient-level ECE and thus reported the mean patient ECE instead of the voxel-level ECE (i.e., considering all voxels in the test set to calculate the ECE).


\subsection{Sensitivity analysis}
\label{subsec:sensitivity}

We analyzed the sensitivity of the results to the main design hyperparameters, which are: the number $H$ of extracted features, the number $I$ of prototypes and the regulation coefficient $\lambda$ \eqref{eq:22}. The influence of the initialization method was also studied. In all the experiments reported in this section as well as in Section \ref{subsec:state-of-art}, learning in each of the configurations was repeated five times with different random initial conditions.

\paragraph{Influence of the number of features}
Table \ref{tab:input_dim} shows the means and standard deviations (over five runs) of the three performance indices for ENN-UNet and RBF-UNet with different numbers of features ($H\in\{2,5,8\}$). Here the features come from the feature extraction module. The number of prototypes and the regularization coefficient were set, respectively, to $I=10$ and $\lambda=0$. The prototypes were initialized randomly. ENN-UNet achieves the highest Dice score and sensitivity with $H=2$ features, but the highest precision with $H=8$. However, the differences are small and concern only the third decimal point. Similarly, RBF-UNet had the best values of the Dice score and precision for $H=5$ features, but again the differences are small. Overall, it seems that only two features are sufficient to discriminate between tumor and background voxels.

\begin{table}
\caption{Means and standard deviations (over five runs) of the performance measures for different input dimensions $H$, with $I=10$ randomly initialized prototypes and  $\lambda=0$. The best values are shown in bold.}
\centering
\label{tab:input_dim}
\begin{tabular}{cccccccc}
\hline
 Model &$H$ &\multicolumn{2}{c}{Dice score} &\multicolumn{2}{c}{Sensitivity}&  \multicolumn{2}{c}{Precision} \\
\cline{3-8} 
&&Mean&SD&Mean&SD&Mean&SD\\
\hline
\multirow{3}*{ENN-UNet} & 2  & 0.833&0.009 & 0.819 &0.019& 0.872&0.018
\\
&5& 0.831 &0.012 &0.817&0.016 &0.870&0.011\\
&8  &0.829& 0.006
&0.816& 0.010 & 0.877& 0.019\\
\hline
\multirow{3}*{RBF-UNet}&2& 0.824&0.009&0.832&0.008&  0.845 &0.016\\
&5&  0.825&0.006&0.817&0.016&0.862&0.010\\
&8& 0.821&0.011&0.813&0.010&  0.862&0.022\\
\hline
\end{tabular}
\end{table}

\paragraph{Influence of the regularization coefficient}
In the previous experiment, the networks were trained without regularization. Tables \ref{tab:lambda1} and \ref{tab:lambda2} show the performances of ENN-UNet and RBF-UNet for different values of $\lambda$, with $I=10$ randomly initialized prototypes and, respectively, $H=2$ and $H=8$ inputs. With both settings, ENN-UNet does not benefit from regularization (the best results are obtained with $\lambda=0$). In contrast, RBF-UNet is more sensitive to regularization and achieves the highest Dice score  with  $\lambda=0.01$. This finding confirms the remark already made in Section \ref{subsubsec:compar}, where it was observed that an ENN classifier seems to be less sensitive to regularization than an RBF classifier (see Figure \ref{fig:error}).

\begin{table}
\caption{Means and standard deviations (over five runs) of the performance measures for  different values of the regularization coefficient $\lambda$, with $I=10$ randomly initialized prototypes and $H=2$ features. The best values are shown in bold.}
\centering
\label{tab:lambda1}
\begin{tabular}{ccccccccc}
\hline
 Model &$\lambda$ &\multicolumn{2}{c}{Dice score} &\multicolumn{2}{c}{Sensitivity}&  \multicolumn{2}{c}{Precision} \\
\cline{3-8} 
 & &Mean&SD&Mean&SD&Mean&SD\\
\hline
\multirow{3}*{ENN-UNet} & 0  &0.833 &0.009&0.819 & 0.019&0.872&0.018\\
&1e-4& 0.822&	0.007&0.818&0.026&	0.839&0.035\\
&1e-2  &0.823&0.004&0.817&0.023	&0.856&0.023\\
\hline
\multirow{3}*{RBF-UNet}&0&0.824&0.009& 0.832&0.008&0.845&0.016\\
&1e-4& 0.825&	0.011&0.811&0.022&	0.869&0.020\\
&1e-2& 0.829&0.010&0.818	&0.022 &0.867&0.016\\
\hline
\end{tabular}
\end{table}

\begin{table}
\caption{Means and standard deviations (over five runs) of the performance measures for  different values of the regularization coefficient $\lambda$, with $I=10$ randomly initialized prototypes and  $H=8$ features. The best values are shown in bold.}
\centering
\label{tab:lambda2}
\begin{tabular}{ccccccccc}
\hline
 Model &$\lambda$ &\multicolumn{2}{c}{Dice score} &\multicolumn{2}{c}{Sensitivity}&  \multicolumn{2}{c}{Precision} \\
\cline{3-8} 
 & &Mean&SD&Mean&SD&Mean&SD\\
\hline
\multirow{3}*{ENN-UNet} & 0  &0.829 &0.006&0.811 & 0.010&0.877&0.019\\
&1e-4& 0.827&	0.008&0.809&0.019&0.873&0.024\\
&1e-2  &0.822&0.009&	0.807&0.021	&0.867&0.011\\
\hline
\multirow{3}*{RBF-UNet}&0&0.821&0.010&0.813&0.010&0.862&0.022\\
&1e-4& 0.827&	0.004&0.830& 0.005&	0.852& 0.012\\
&1e-2& 0.832&0.006&0.825	&0.022&  0.867&0.020\\
\hline
\end{tabular}
\end{table}

\paragraph{Influence of the number of prototypes}
The number $I$ of prototypes is another hyperparameter that may impact segmentation performance. Table \ref{tab:protos} shows the performances of ENN-UNet and RBF-UNet with 10 and 20 randomly initialized prototypes, the other hyperparameters being fixed at  $H=2$ and $\lambda=0$. Increasing the number of prototypes beyond 10 does not seem to improve the performance of ENN-UNet, while it does slightly improve the performance of RBF-UNet in terms of Dice score and precision, at the expense of an increased computing time.

\begin{table}
\caption{Means and standard deviations (over five runs) of the performance measures for  different numbers $I$ of randomly initialized prototypes, with $H=2$ features and  $\lambda=0$. The best values are shown in bold.}
\centering
\label{tab:protos}
\begin{tabular}{cccccccccc}
\hline
 Model &$I$ &\multicolumn{2}{c}{Dice score} &\multicolumn{2}{c}{Sensitivity}&  \multicolumn{2}{c}{Precision} \\
\cline{3-8} 
 &&Mean&SD&Mean&SD&Mean&SD\\
\hline
\multirow{2}*{ENN-UNet} & $10$  & 0.833 &0.009&0.819 & 0.019&0.872&0.018\\
&$20$& 0.823&0.007&0.804&0.006&0.864&0.012\\
\hline
\multirow{2}*{RBF-UNet}&$10$&0.824&0.009& 0.832&0.008&0.845&0.016\\
&$20$& 0.830&0.007& 0.810&0.012&0.867&0.010\\
\hline
\end{tabular}
\end{table}

\paragraph{Influence of the prototype initialization method}

Finally, we compared the two initialization methods mentioned in Section \ref{subsec:settings}. For $k$-means initialization, in the first step, a UNet model was trained with the following settings: kernel size=$5$, channels =$(8, 16, 32, 64, 128)$ and strides=$(2,2,2,2)$. The spatial dimension, input, and output channel were set, respectively to, 3, 2, and 2. This  pre-trained UNet was used to extract $H=2$ features, and 10 prototypes were obtained by running the $k$-means algorithm in the space of  extracted features. These prototypes were fed into ENN or RBF layers, which were trained separately, with fixed features. For this step, the  learning rate was set to $10^{-2}$. Finally, the whole model was fine-tuned end-to-end, with  a smaller learning rate equal to $10^{-4}$. 
Table \ref{tab:initial} shows the performances of ENN-UNet and RBF-UNet with random and $k$-means initialization. Both ENN-UNet and RBF-UNet achieve a higher Dice score when using the $k$-means initialization method, and the variability of the results is also reduced with this method.

\begin{table}
\caption{Means and standard deviations (over five runs) of the performance measures for different initialization methods, with $I=10$ prototypes, $H=2$ features and $\lambda=0$. The best values are shown in bold.}
\centering
\label{tab:initial}
\begin{tabular}{cccccccccc}
\hline
 Model & Initialization&\multicolumn{2}{c}{Dice score} &\multicolumn{2}{c}{Sensitivity}&  \multicolumn{2}{c}{Precision} \\
\cline{3-8} 
 & &Mean&SD&Mean&SD&Mean&SD\\
\hline

\multirow{2}*{ENN-UNet} &Random &0.833 &0.009&0.819 & 0.019&0.872&0.018\\
&  $k$-means&0.846&0.002&0.830&0.004&0.879&0.008
\\

\hline
\multirow{2}*{RBF-UNet}&Random&0.824&0.009& 0.832&0.008&0.845&0.016\\
&  $k$-means&0.839&0.003&0.824&0.001&0.879&0.008\\

\hline
\end{tabular}
\end{table}

\begin{figure}
\centering
\subfloat[\label{fig:contourESomega1}]{\includegraphics[width=0.5\textwidth]{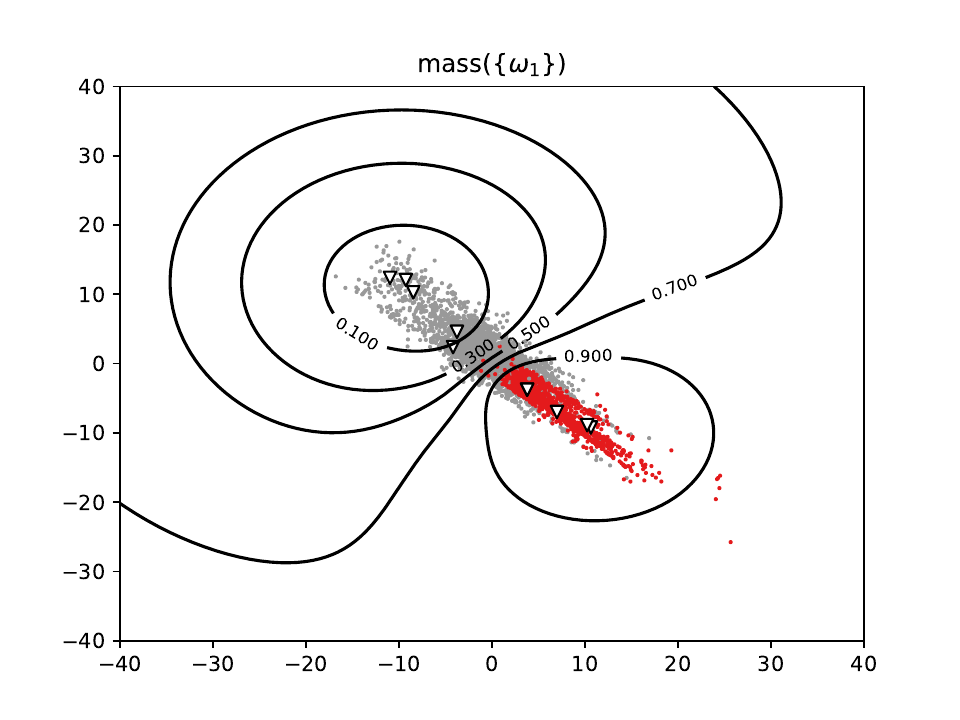}}
\subfloat[\label{fig:contourESomega2}]{\includegraphics[width=0.5\textwidth]{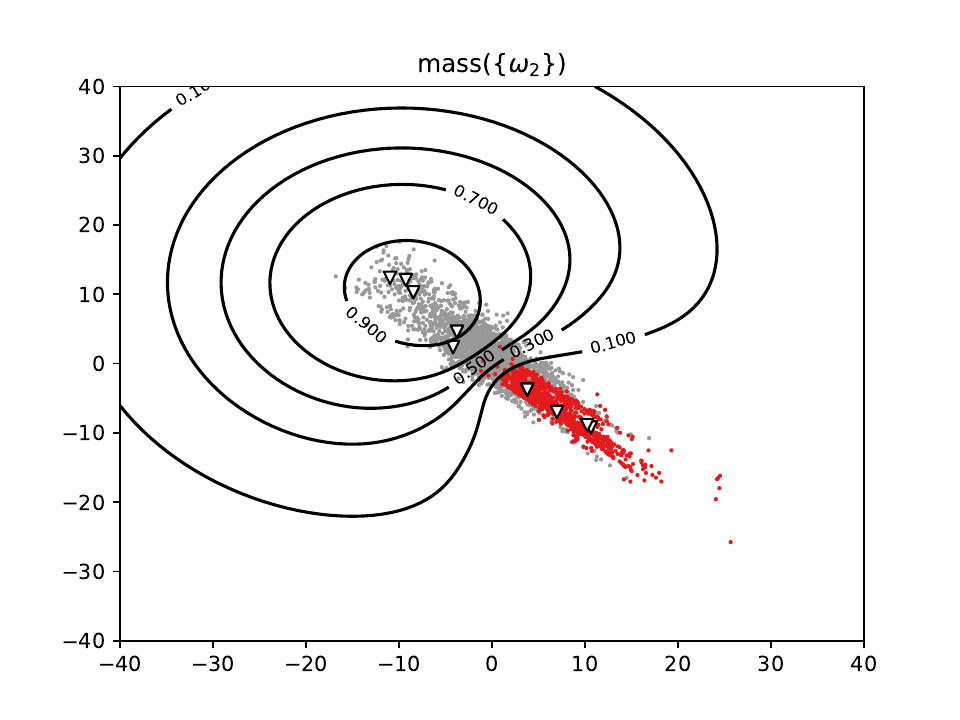}}\\
\subfloat[\label{fig:contourESOmega}]{\includegraphics[width=0.5\textwidth]{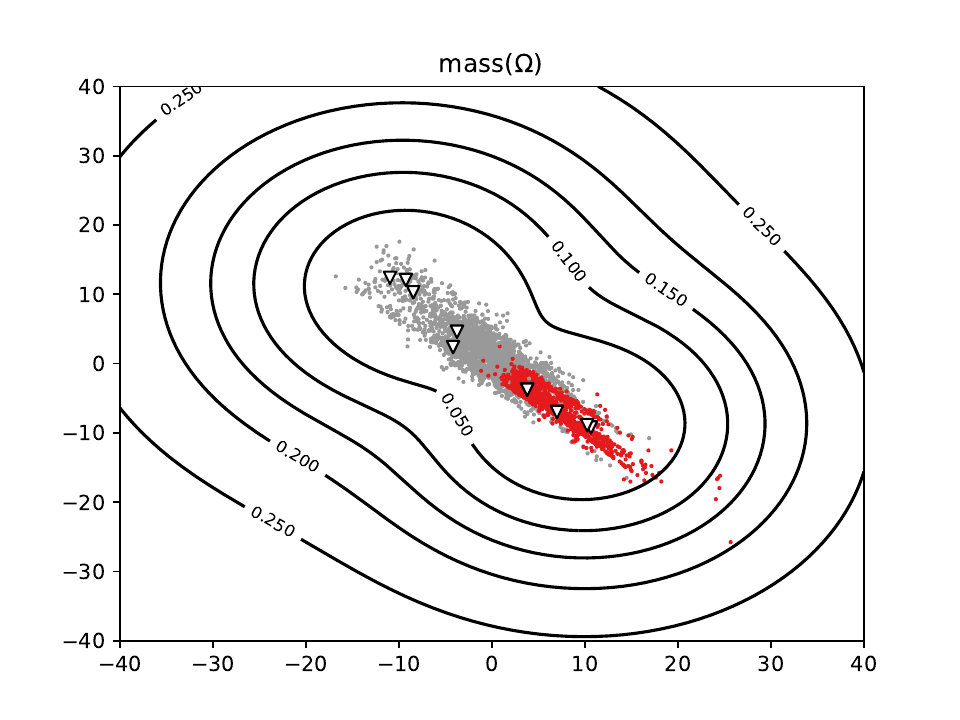}}
\caption{Contours in feature space of the masses assigned to the background (a), the tumor class (b), and the frame of discernment (c) by the ENN-UNet model initialized by $k$-means. Training was done with $\lambda=10^{-2}$, $H=2$ and $I=10$. Sampled feature vectors from the tumor and background classes are marked in gray and red, respectively. 
\label{fig:ES-kmeans}}
\end{figure}

\begin{figure}
\centering
\subfloat[\label{fig:contourRBFUomega1}]{\includegraphics[width=0.5\textwidth]{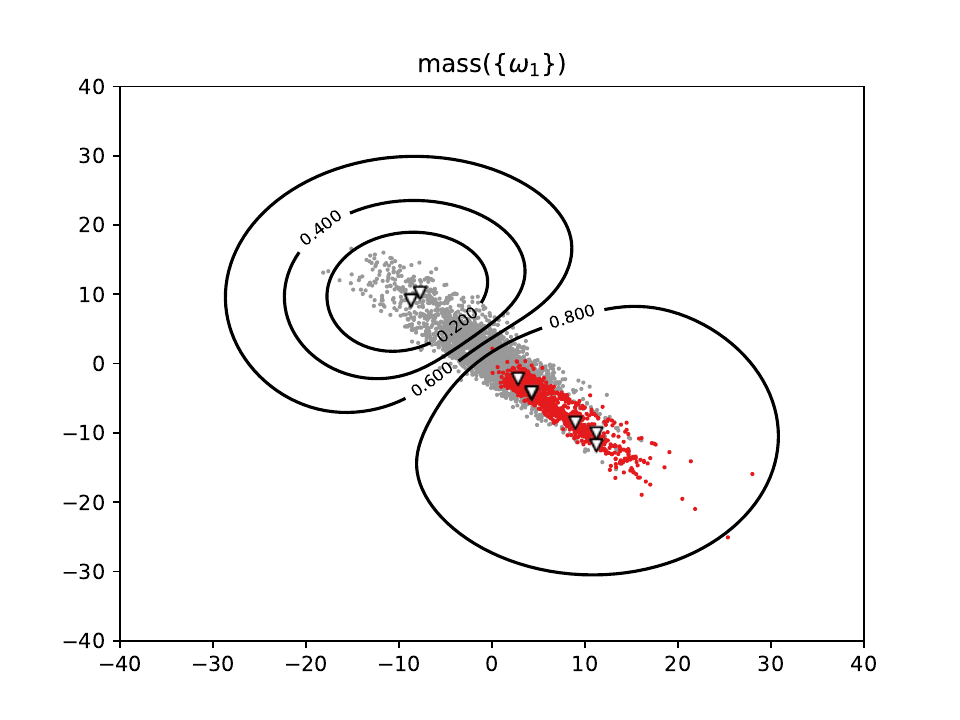}}
\subfloat[\label{fig:contourRBFUomega2}]{\includegraphics[width=0.5\textwidth]{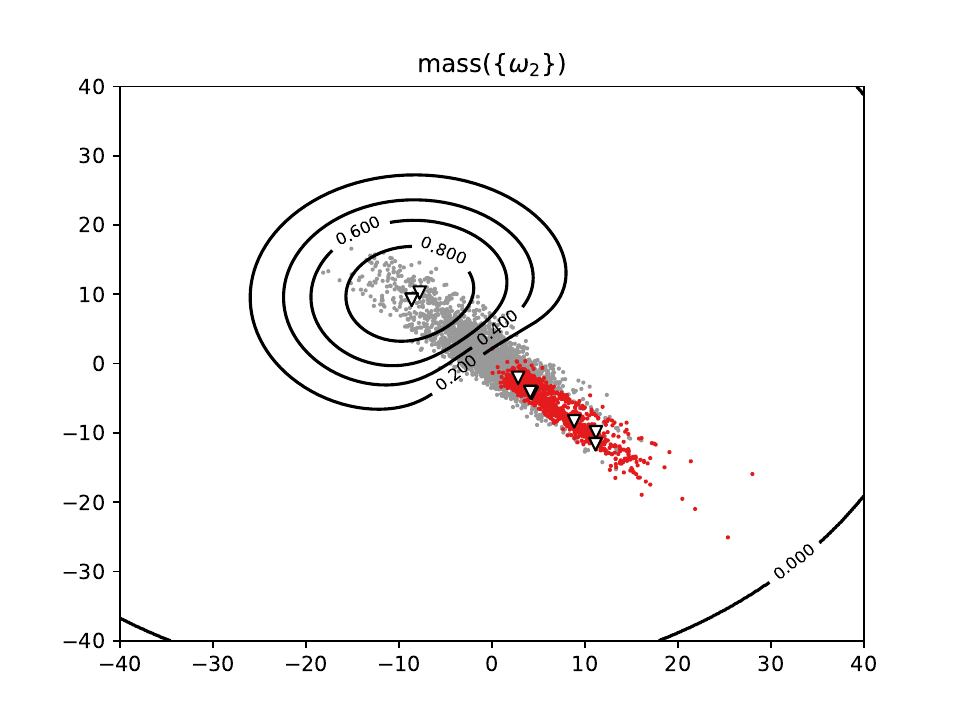}}\\
\subfloat[\label{fig:contourRBFUOmega}]{\includegraphics[width=0.5\textwidth]{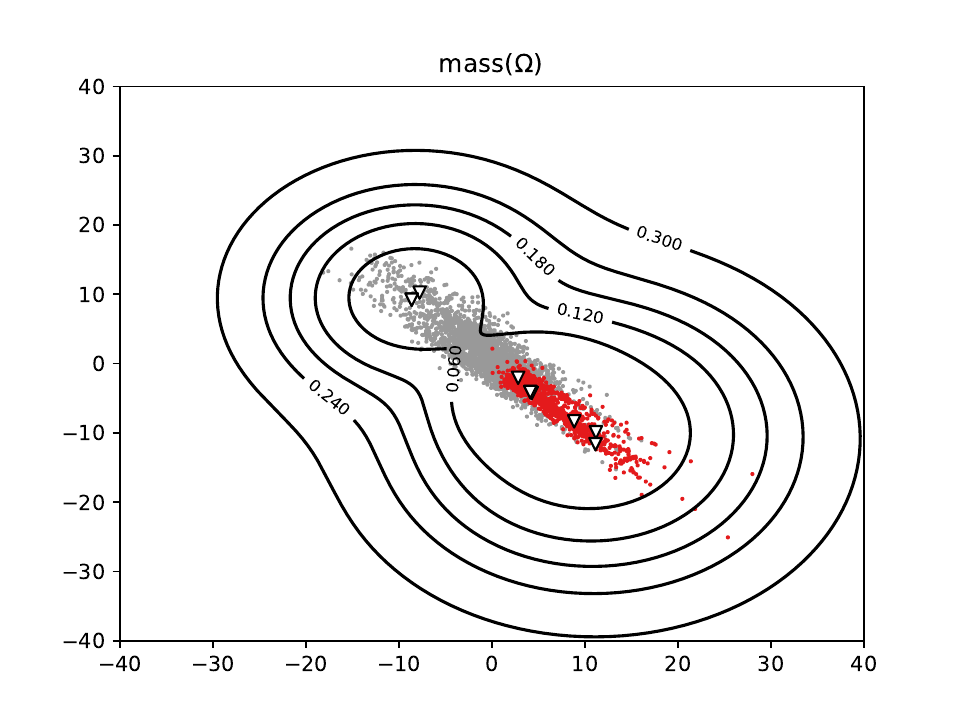}}
\caption{Contours in feature space of the masses assigned to the background (a), the tumor class (b) and the frame of discernment (c) by the RBF-UNet model initialized by $k$-means. Training was done with $\lambda=10^{-2}$, $H=2$ and $I=10$. Sampled feature vectors from the tumor and background classes are marked in gray and red, respectively. 
\label{fig:RBF-kmeans}}
\end{figure}

\begin{figure}
\centering
\subfloat[\label{fig:contourESRomega1}]{\includegraphics[width=0.5\textwidth]{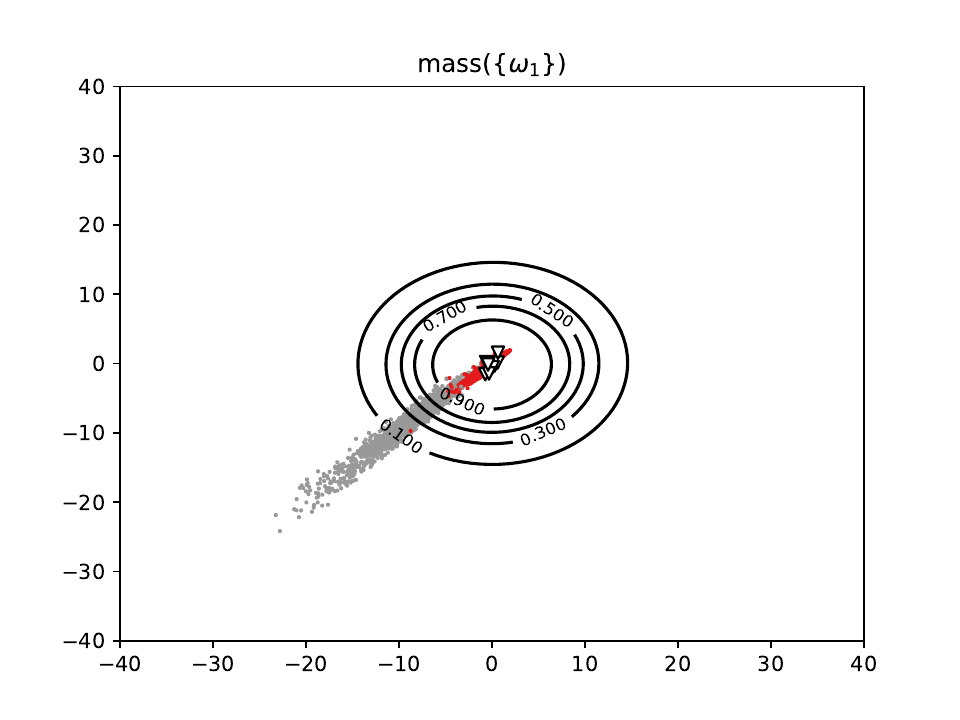}}
\subfloat[\label{fig:contourESRomega2}]{\includegraphics[width=0.5\textwidth]{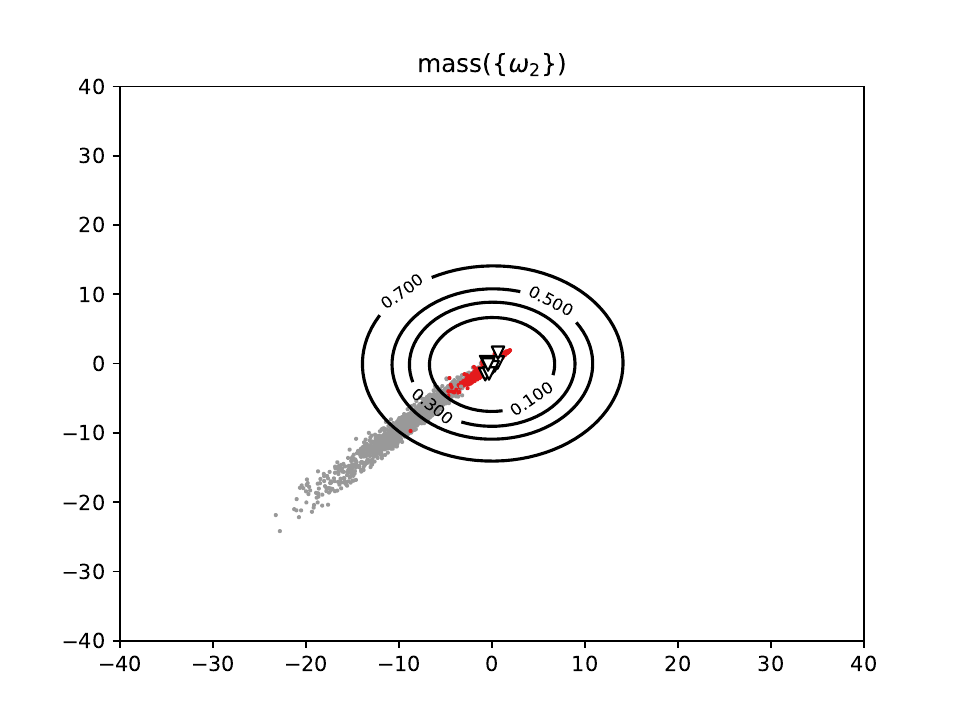}}\\
\subfloat[\label{fig:contourESROmega}]{\includegraphics[width=0.5\textwidth]{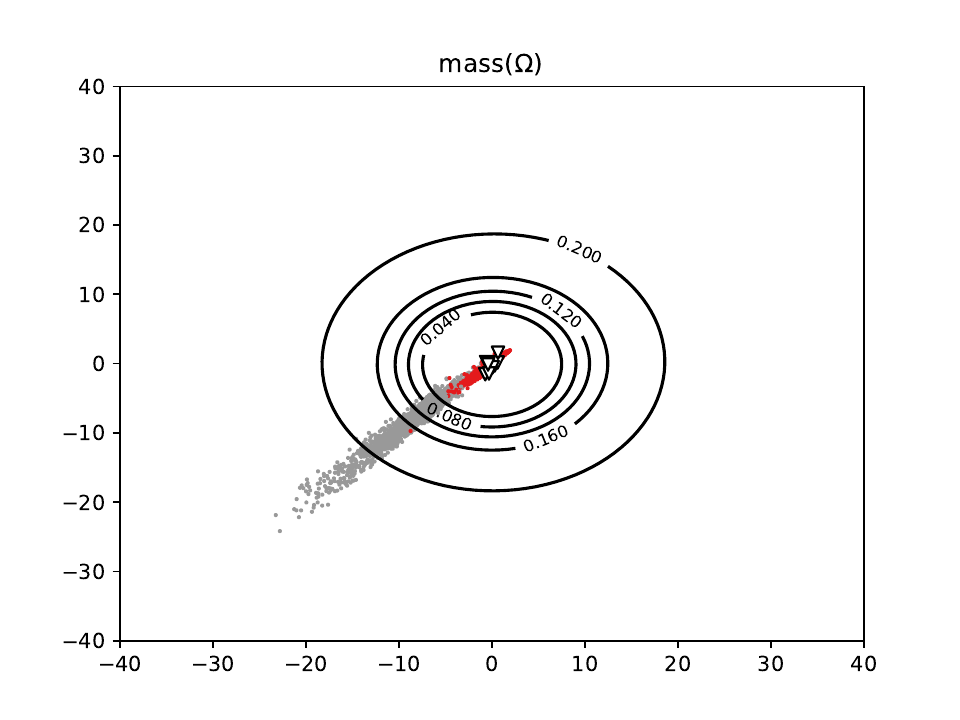}}
\caption{Contours in feature space of the masses assigned to the background (a), the tumor class (b), and the frame of discernment (c) by the ENN-UNet model initialized randomly. Training was done with $\lambda=10^{-2}$, $H=2$ and $I=10$. Sampled feature vectors from the tumor and background classes are marked in gray and red, respectively. 
\label{fig:ES-random}}
\end{figure}

Not only does the $k$-means initialization method slightly improve the performances of ENN-UNet and RBF-UNet quantitatively, but it also tends to position the prototypes in regions of high data density. As a result, a high  output mass $m(\Omega)$ signals that the input data is atypical. In that sense, the output mass function is more interpretable. This point is illustrated by Figures \ref{fig:ES-kmeans} and \ref{fig:RBF-kmeans}, which show the contours, in the two-dimensional feature space, of the masses assigned to the background, the tumor class and the frame of discernment when using $k$-means initialization (with $\lambda=10^{-2}$ and $I=10$) with, respectively, ENN-UNet and RBF-UNet. For both models, the prototypes are well distributed over the two classes, and the mass on $\Omega$ decreases with the distance to the data, as expected. In contrast, when using random initialization (as shown in Figure \ref{fig:ES-random} for the ENN-UNet model -- results are similar to the RBF-UNet model), the prototypes  are located in the background region, and the mass $m(\Omega)$ does not have a clear meaning (although the decision boundary still ensures good discrimination between the two classes).

From this sensitivity analysis, we can conclude that the performances of both ENN-UNet and RBF-UNet are quite robust to the values of the hyperparameters and that the two models achieve comparable performances. The $k$-means initialization method seems to yield better results, both quantitatively and qualitatively. The next section is devoted to a comparison with alternative models.


\subsection{Comparative analysis: segmentation accuracy} 
\label{subsec:state-of-art}

In this section, we compare the performances of the ENN-UNet and RBF-UNet models with those of the baseline model, UNet \cite{ronnebergerconvolutional}, as well as three state-of-the-art models: VNet \cite{milletari2016v}, SegResNet \cite{myronenko20183d} and nnUNet \cite{isensee2018nnu}. VNet is a variant of UNet that introduces short residual connections at each stage. Compared with UNet, SegResNet contains an additional variational autoencoder branch. nnUNet ( see Section \ref{subsec: nnunet}) is the first segmentation model designed as a segmentation pipeline for any given dataset. For all compared methods, the same learning set and pre-processing steps were used. All the compared methods were trained with the Dice loss function \eqref{eq:Dice_loss}. Details about the optimization algorithm were given in Section \ref{subsec:settings}. All methods were implemented based on the MONAI framework\footnote{More details about how to use those models can be found from MONAI core tutorials \url{https://monai.io/started.html##monaicore}.} and can be called directly. For UNet, the kernel size was set as 5 and the channels were set to $(8, 16, 32, 64, 128)$ with strides=$(2,2,2,2)$. 
For nnUNet, the kernel size was set as $(3, (1,1, 3), 3, 3)$ and the upsample kernel size was set as $(2,2,1)$ with strides $((1,1,1), 2, 2, 1)$. For SegResNet \cite{myronenko20183d} and VNet \cite{milletari2016v}, we used the pre-defined model without changing any parameter. The spatial dimension, input channel and output channel were set, respectively, 3, 2, and 2 for the four compared models. 
As for other hyperparameters not mentioned here, we used the pre-defined value given in MONAI. As shown by the sensitivity analysis performed in Section \ref{subsec:sensitivity}, the best results for ENN-UNet and RBF-UNet are achieved with  $\lambda=0$, $I=10$, $H=2$ and $k$-means initialization.

\begin{figure}
\centering
\subfloat[\label{fig:dotplot_Dice}]{\includegraphics[width=0.4\textwidth]{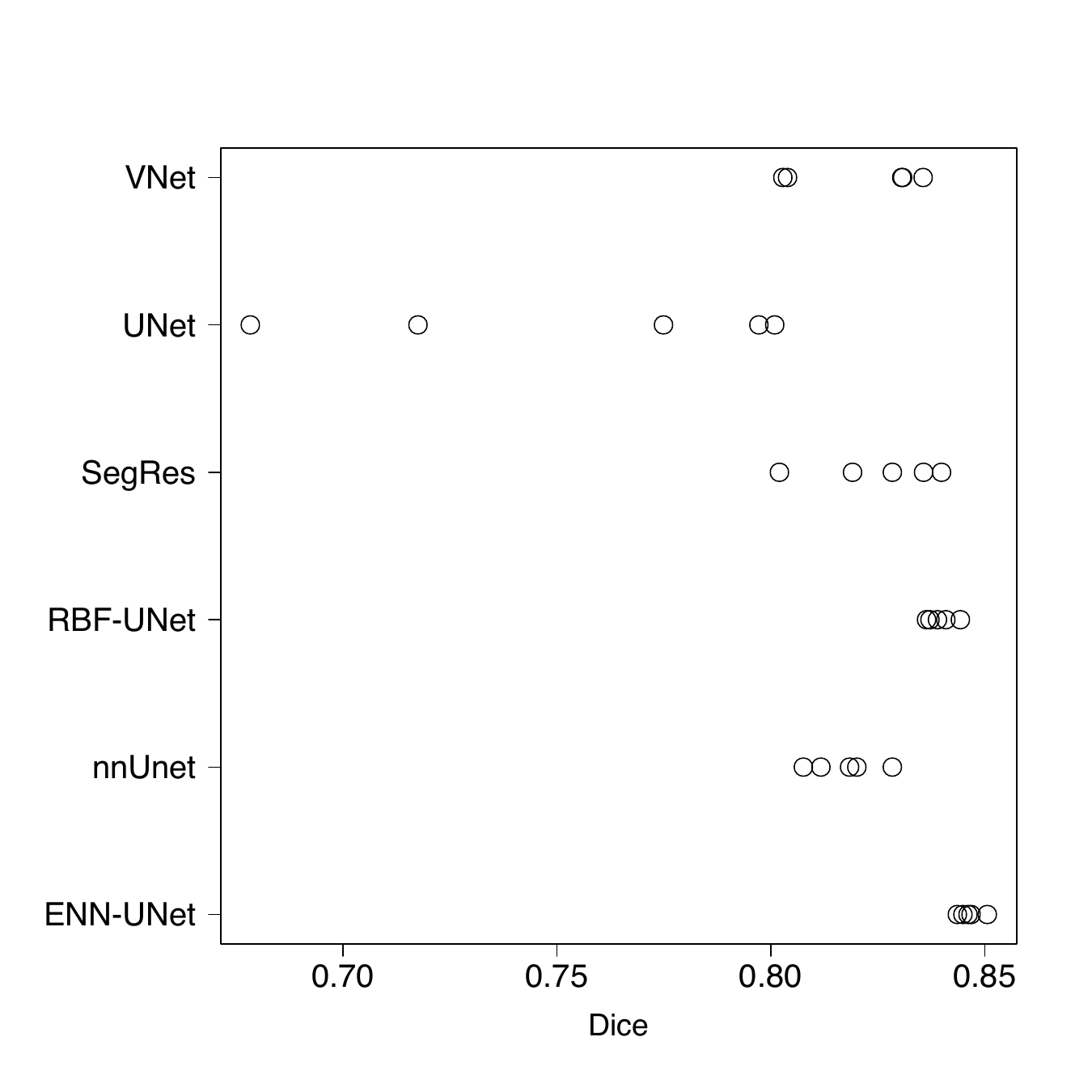}}
\subfloat[\label{fig:dotplot_Sensitivity}]{\includegraphics[width=0.4\textwidth]{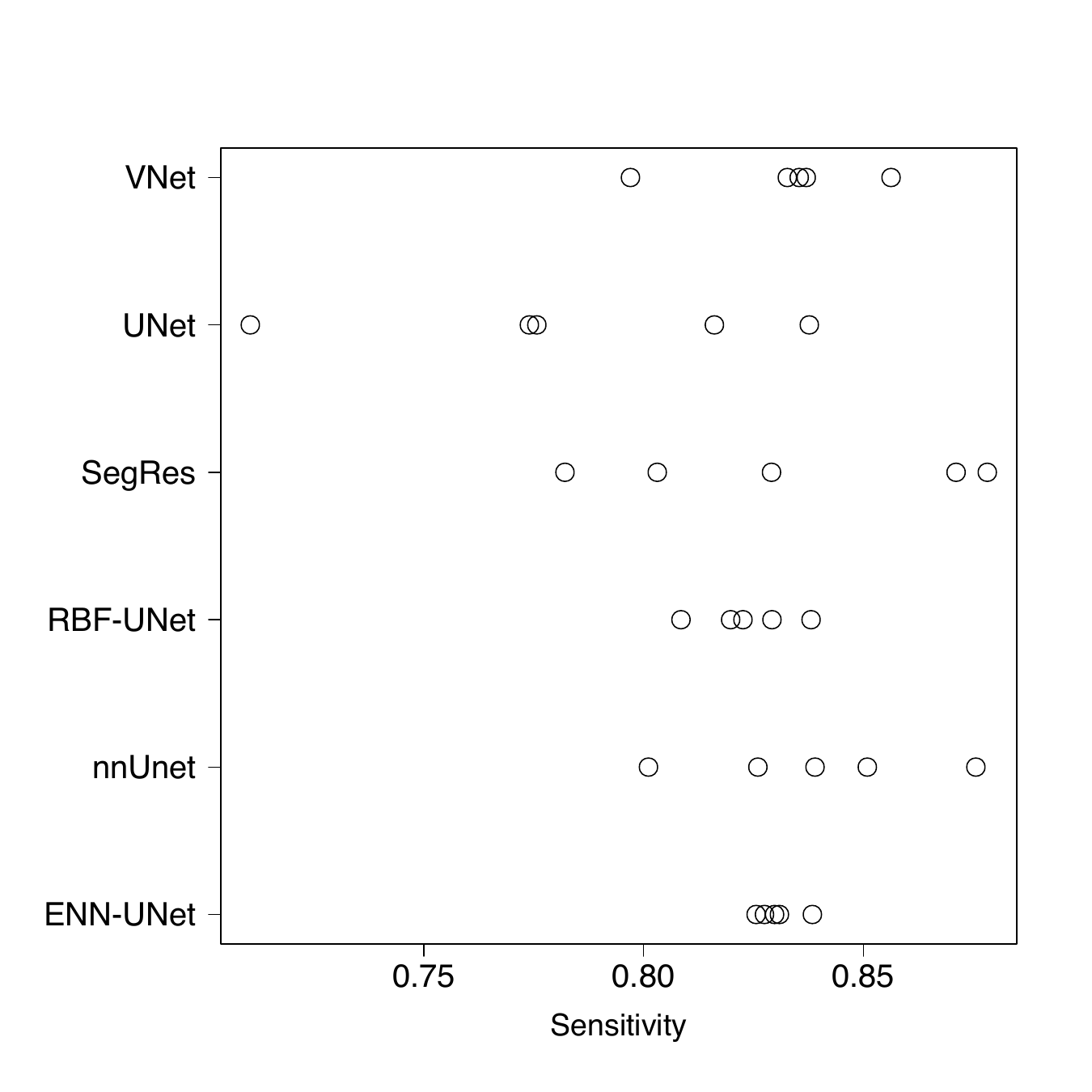}}\\
\subfloat[\label{fig:dotplot_Precision}]{\includegraphics[width=0.4\textwidth]{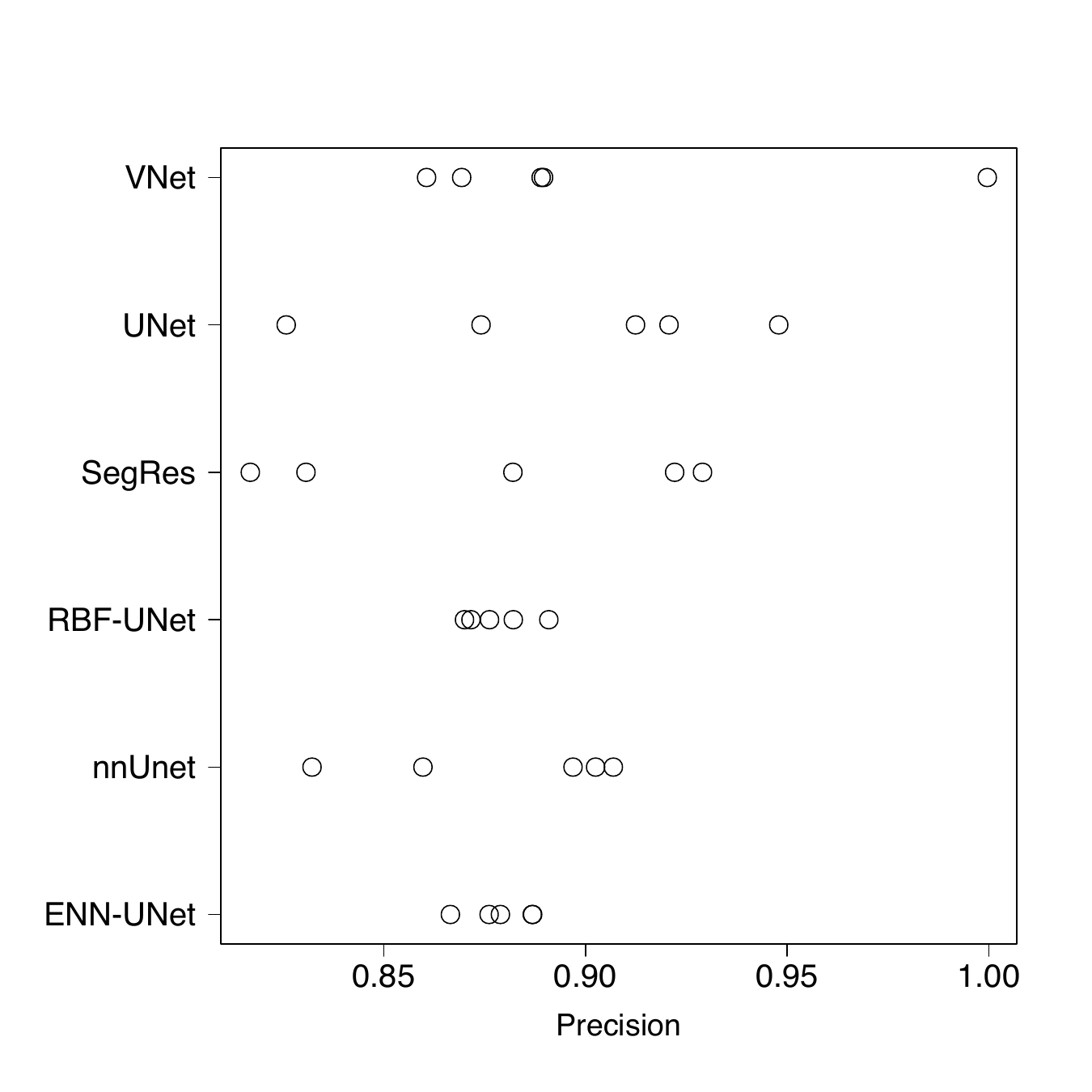}}
\caption{Values of the Dice score (a), sensitivity (b) and precision (c) for five runs of the six methods. \label{fig:dotplots}}
\end{figure}

 The means and standard deviations of the Dice score, sensitivity, and precision over five runs with random initialization for the six methods are shown in Table~\ref{tab:compar}, and the raw values are plotted in Figure \ref{fig:dotplots}. We can see that ENN-UNet and RBF-UNet achieve, respectively, the highest and the second-highest mean Dice scores. A Kruskal-Wallis test performed on the whole data concludes with a significant difference between the distributions of the Dice score for the six methods (p-value = 0.0001743), while the differences are not significant for sensitivity (p-value = 0.2644) and precision (p-value = 0.9496). Table \ref{tab:conover} shows the results of the Conover-Iman test of multiple comparisons \cite{conover79}\cite{dinno17} with Benjamini-Yekutieli adjustment \cite{BY01}. We can see that the differences between the Dice scores obtained by ENN-UNet and RBF-UNet on the one hand, and the four other methods on the other hand are highly significant (p-values $< 10^{-2}$),  while the difference between ENN-UNet and RBF-UNet is only weakly significant (p-value = 0.0857).

\begin{table}
\caption{Means and standard deviations (over five runs) of the performance measures for ENN-UNet, RBF-UNet and four reference methods. The best result is shown in bold, and the second best is underlined.}
\centering
\label{tab:compar}
\scalebox{0.9}{
\begin{tabular}{ccccccc}
\hline
 Model  &\multicolumn{2}{c}{Dice score} &\multicolumn{2}{c}{Sensitivity}&  \multicolumn{2}{c}{Precision}\\
 \cline{2-7} 
 &Mean&SD&Mean&SD&Mean&SD\\
\hline
UNet \cite{kerfoot2018left} & 0.753&0.054& 0.782 &0.048&\underline{0.896}&0.047\\
nnUNet \cite{isensee2018nnu} & 0.817 &0.008& \textbf{0.838}&0.028  &0.879&0.032 \\
VNet \cite{milletari2016v}&0.820&0.016&0.831&0.021 & \textbf{0.901}&0.056 \\
SegResNet \cite{myronenko20183d}& 0.825&0.015& \underline{0.832}&0.042&0.876&0.051 \\
\hline
ENN-UNet&\textbf{0.846}&0.002&0.830&0.004&0.879&0.008\\
RBF-UNet&\underline{0.839}&0.003&0.824&0.001&0.879&0.008\\

\hline
\end{tabular}
}
\end{table}

\begin{table}
    \centering
       \caption{Conover-Iman test of multiple comparisons between the Dice scores obtained by the six models: t-test statistics and p-values. P-values less than 0.01 are printed in bold. \label{tab:conover}}
    
    \begin{tabular}{lccccc}
    \hline
        &    ENN-UNet  &   nnUnet &  RBF-UNet  &   SegResNet  &     UNet\\
        \hline
  nnUnet &   6.759 & & & & \\
         &    \textbf{0.0000} & & & & \\
RBF-UNet &   2.156 & -4.602 & & &  \\
         &     0.0857 &   \textbf{0.0004}& & &  \\
  SegResNet &   5.349 & -1.410 &  3.193 & &   \\
         &    \textbf{0.0001}  &   0.3282 &   \textbf{0.0088}& &   \\
    UNet &   10.283 &  3.524  & 8.127 &  4.934 &    \\
         &    \textbf{0.0000}  &  \textbf{0.0043} &   \textbf{0.0000}  &  \textbf{0.0002}&    \\
    VNet &   6.054&  -0.705 &  3.898  & 0.705 & -4.229\\
         &    \textbf{0.0000}  &   0.8091  &  \textbf{0.0019}   &  0.8669  &  \textbf{0.0009}\\
         \hline
    \end{tabular}
\end{table}

\begin{figure}
\centering
\includegraphics[scale=0.45]{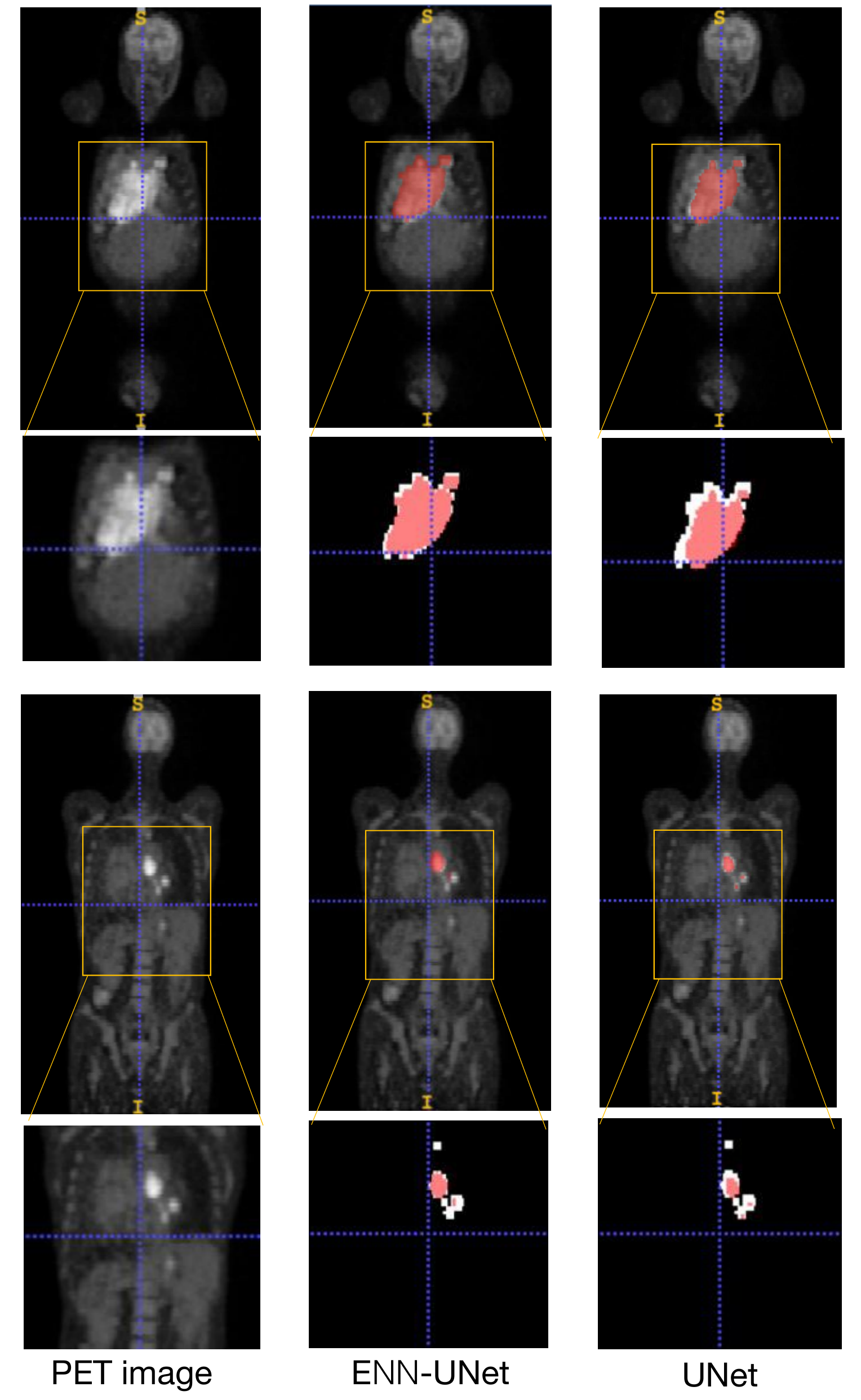}
\caption{Two examples of segmentation results by ENN-UNet and UNet. The first and the second row are, respectively, 
 representative of large and isolated small lymphomas. The three columns correspond, from left to right, to the PET images and the  segmentation results obtained by ENN-UNet and UNet. The white and red regions represent, respectively, the ground truth and the segmentation result. \label{fig:seg_result}}
\end{figure}

Figure \ref{fig:seg_result} shows two examples of segmentation results obtained by ENN-UNet and UNet, corresponding to large and isolated lymphomas. We can see, in these two examples, that UNet is more conservative (it correctly detects only a subset of the tumor voxels), which may explain why it has relatively high precision. However, the tumor regions predicted by ENN-UNet better overlap the ground-truth tumor region, which is also reflected by the higher Dice score. 

\subsection{Comparative analysis: segmentation uncertainty} 
\label{subsec:calibration}

Besides segmentation accuracy, another important issue concerns the quality of uncertainty quantification. Monte-Carlo dropout (MCD) \cite{gal2016dropout} is a state-of-the-art technique for improving the uncertainty quantification capabilities of deep networks. In this section, we compare the  ECE \eqref{eq:ECE} achieved by  UNet (the baseline),  SegResNet (the best alternative method found in Section \ref{subsec:state-of-art}), and our proposals: ENN-UNet, and RBF-UNet, with and without MCD. For the four methods, the dropout rate was set to 0.5 and the sample number was set to 20; we averaged the 20 output probabilities (the pignistic probabilities for the two evidential models) at each voxel as the final output of the model. 

The results are reported in Table \ref{tab:ECE}. We can see that MCD enhances the segmentation performance (measured by the Dice index) of UNet et SegResNet, and improves the calibration of all methods, except SegResNet. Overall, the smallest average ECE is achieved by  RBF-UNet and ENN-UNet with MCD, but the standard deviations are quite large. A Kruskal-Wallis test concludes with a significant difference between the distributions of ECE for the eight methods (p-value = 0.01). The p-values of the Conover-Iman test of multiple comparisons with Benjamini-Yekutieli adjustment reported in Table \ref{tab:conover-calib} show significant differences between the ECE of RBF-UNet with MCD on the one hand, and those of RBF-UNet without MCD, SegResNet with MCD, and  UNet without MCD on the other hand. We also tested the pairwise differences between the ECE values obtained by RBF-UNet and ENN-UNet with MCD on the one hand, and UNet with and without MCD as well as SegResNet with and without MCD, on the other hand, using the Wilcoxon rank sum test. The corresponding p-values are shown in Table \ref{tab:wilcox}. We find significant differences between the ECE  RBF-UNet with MCD and those of the other methods, but only a weakly significant difference between ENN-UNet with MCD and UNet without MCD. In summary, there is some  evidence that MCD improves calibration, even for evidential models, and that the  best calibration is achieved by the  RBF-UNet model, but this evidence is not fully conclusive due to the limited size of the dataset; our findings  will  have to be confirmed by further experiments with larger datasets.

\begin{table}
\centering
\caption{Means and standard deviations (over five runs) of the Dice score and ECE for UNet, SegResNet, ENN-UNet, and RBF-UNet, with and without MCD. The best results are shown in bold, and the second best is underlined.}
\scalebox{0.9}{
    \begin{tabular}{cccccc}
    \hline
     Model  &\multicolumn{2}{c}{Dice score} &\multicolumn{2}{c}{ECE(\%)}\\
 \cline{2-3} 
 \cline{4-5} 
 &Mean&SD&Mean&SD\\
  \cline{1-5} 
    UNet       &0.754&0.054&  2.22&0.205\\
    SegResNet &0.825&0.015&  1.97&0.488\\
    ENN-UNet   & \textbf{0.846} & 0.002 &  1.99 &0.110\\
    RBF-UNet   &0.839&0.003&  2.12&0.028\\
    \hline
    UNet with MCD &0.828&0.005&1.93&0.337\\
    SegResNet with MCD &\underline{0.844}&0.009&2.53&0.973\\
    ENN-UNet with MCD& 0.841&0.003&\underline{1.53}&0.075 \\
    RBF-UNet with MCD &0.840&0.003&\textbf{1.52}&0.041\\
    \hline
    \end{tabular}}
    
    \label{tab:ECE}
\end{table}

\begin{table}
    \centering
       \caption{Conover-Iman test of multiple comparisons between the ECE  obtained by UNet, SegResNet, ENN and RBF, with and without MCD: t-test statistics and p-values. P-values less than 0.01 are printed in bold.} \label{tab:conover-calib}
    \scalebox{0.87}{
    \begin{tabular}{lccccccc}
    \hline
&       ENN  &   ENN-MC   &     RBF  &   RBF-MC &  SegRes  & SegRes-MC& UNet\\
\hline
  ENN-MC &   0.926 &&&&&&\\
         &     1.0000 &&&&&&\\
     RBF &  -1.191 & -2.118 &&&&&\\
         &     0.7403 &    0.2892 &&&&&\\
  RBF-MC &   2.812 &  1.886 &  4.004 &&&&\\
         &     0.1145 &    0.3419 &   \textbf{0.0095} &&&&\\
SegRes &   0.695 & -0.232 &  1.886 & -2.117 &&&\\
         &     1.0000 &    1.0000 &    0.3761 &    0.3305 &&&\\
SegRes-MC &  -0.860 & -1.787 &  0.331 & -3.673  &-1.555 &&\\
         &     1.0000 &    0.3530 &    1.0000 &   \textbf{0.0159} &    0.4756 &&\\
    UNet &  -1.357 & -2.283 & -0.165 & -4.169 & -2.051 & -0.496 & \\
         &     0.6337 &    0.2677 &    1.0000 &   \textbf{0.0119} &    0.2962 &    1.0000 & \\
 UNet-MC &   0.430 & -0.496 &  1.621 & -2.382 & -0.265  & 1.290  &  1.787\\
         &     1.0000 &    1.0000  &   0.4507  &   0.2564  &   1.0000  &   0.6667  &  0.3824\\
 \hline
    \end{tabular}}
\end{table}   

\begin{table}
    \centering
       \caption{P-values for the Wilcoxon rank sum test applied to the comparison of ECE obtained by ENN-UNet and RBF UNet with MCD on the one hand and the four other methods on the other hand (UNet and SegResNet with and without MCD).} \label{tab:wilcox}
    
    \begin{tabular}{lcccc}
    \hline
 & UNet &   UNet-MC &  SegRes  & SegRes-MC\\
    \hline
ENN-MC& 0.095 &0.67 &0.69& 0.31\\
RBF-MC& 0.0079 &0.012 &0.055 &0.0079\\
\hline
\end{tabular}
\end{table}

\section{Conclusion}
\label{sec:conc5}

An evidential model for segmenting lymphomas from 3D PET-CT images with uncertainty quantification has been proposed in this chapter. Our model is based on the concatenation of a UNet, which extracts high-level features from the input images, and an evidential segmentation module, which computes output mass functions for each voxel. Two versions of this evidential module, both involving prototypes, have been studied: one is based on the ENN classifier initially proposed as a stand-alone classifier in \cite{denoeux2000neural}, while the other one relies on an RBF classifier and the addition of the weight of evidence. The whole model is trained end-to-end by minimizing the Dice loss. Our model has been shown to outperform the baseline UNet model and other state-of-the-art segmentation methods on a dataset of 173 patients with lymphomas. Preliminary results also suggest the outputs of the evidential models (in particular, the one with an RBF layer) are better calibrated and that calibration error can be further decreased by Monte Carlo dropout. These results, however, will have to be confirmed by further experiments with larger datasets. One of the potential problems that may arise is related to the dimensionality of the feature space. In the application considered in this work, good results were obtained with only two extracted features. If some other learning tasks require a much higher feature dimension, we may need a much higher number of prototypes, and learning may be slow. This issue could be addressed by adapting the loss function as proposed, e.g., in \cite{Hryniowski20}.

Multimodal medical images indicate different information about the presence of disease, with varying reliability of each modality image. In this work, we use two modalities: PET providing information on the lymphoma and CT providing anatomical information, to better guide radiotherapy planning. We fuse images from PET and CT modalities by concatenating them into a single input and quantifying the segmentation uncertainty. This work can be extended in many research directions. For example, we can further evaluate the approach by applying it to other medical image segmentation problems and study the quantification and fusion of modality-level uncertainty. In the next chapter, we will introduce our work on quantifying modality-level uncertainty and combining unreliable sources of information with an application to a large-scale brain tumor dataset.

 \stopcontents[chapters] 
\chapter{Multimodal medical image segmentation with contextual discounting } 
\label{Chapter6}


\tocpartial

\section{Introduction}
Single-modality medical images often do not contain enough information to reach an accurate and reliable diagnosis. This is why physicians generally use multiple sources of information, such as multi-MR images for brain tumor segmentation, or PET-CT images for lymphoma segmentation. The effective fusion of multimodal information is of great importance in the medical domain for better disease diagnosis and radiotherapy. Using convolutional neural networks (CNNs), researchers have mainly adopted probabilistic approaches to information fusion, which can be classified into three strategies: image-level fusion, such as input data concatenation \cite{peiris2021volumetric}; feature-level fusion, such as attention mechanism concatenation \cite{zhou2020fusion}; and decision-level fusion such as weighted averaging \cite{kamnitsas2017ensembles}. However, probabilistic fusion is unable to effectively manage conflict that occurs when different labels are assigned to the same voxel based on different modalities. Also, it is important, when combining multimodal information, to take into account the reliability of the different sources of information.

In this chapter, we introduce a multimodal evidence fusion framework based on contextual discounting \cite{pichon2016proposition, mercier2012belief} and deep learning. In the BFT framework, the reliability of a source of information can be taken into account using the discounting operation \cite{shafer1976mathematical}, which transforms each piece of evidence provided by a source into a weaker, less informative one. To our knowledge, this work is the first attempt to apply evidence theory with contextual discounting to the fusion of deep neural networks. The idea of considering multimodal images as independent inputs and quantifying their reliability is simple and reasonable. However, modeling the reliability of sources is important and challenging. Our method computes mass functions to assign degrees of belief to each class and assign an ignorance degree to the whole set of classes. It thus has one more degree of freedom than a probabilistic model, which allows us to model source uncertainty directly. The contributions of this work are the following: (1) Four BFT-based evidential segmentation modules are used to compute the belief of each voxel belonging to the tumor for four modality MR images; (2) An evidence-discounting mechanism is applied to each of the single-modality MR images to take into account its reliability; (3) A multimodal evidence fusion strategy is then applied to combine the discounted evidence with BFT and achieve more reliable results. End-to-end learning is performed by minimizing a new loss function based on the discounting mechanism, allowing us to increase the segmentation performance and  reliability. The framework is evaluated on the BraTs 2021 database of 1251 patients with brain tumors. Quantitative and qualitative results show that our method outperforms the state-of-the-art and implements an effective new idea for merging multiple information within deep neural networks.

We organize this chapter as follows: Section \ref{sec: model6} introduces the multimodal medical image segmentation model. Section \ref{sec:exp6} reports the numerical experiments by evaluating the performance in segmentation accuracy and reliability. Finally, we conclude this work in Section \ref{sec:conclu6}.

\section{Proposed approach}
\label{sec: model6}

\begin{figure}
\includegraphics[width=\textwidth]{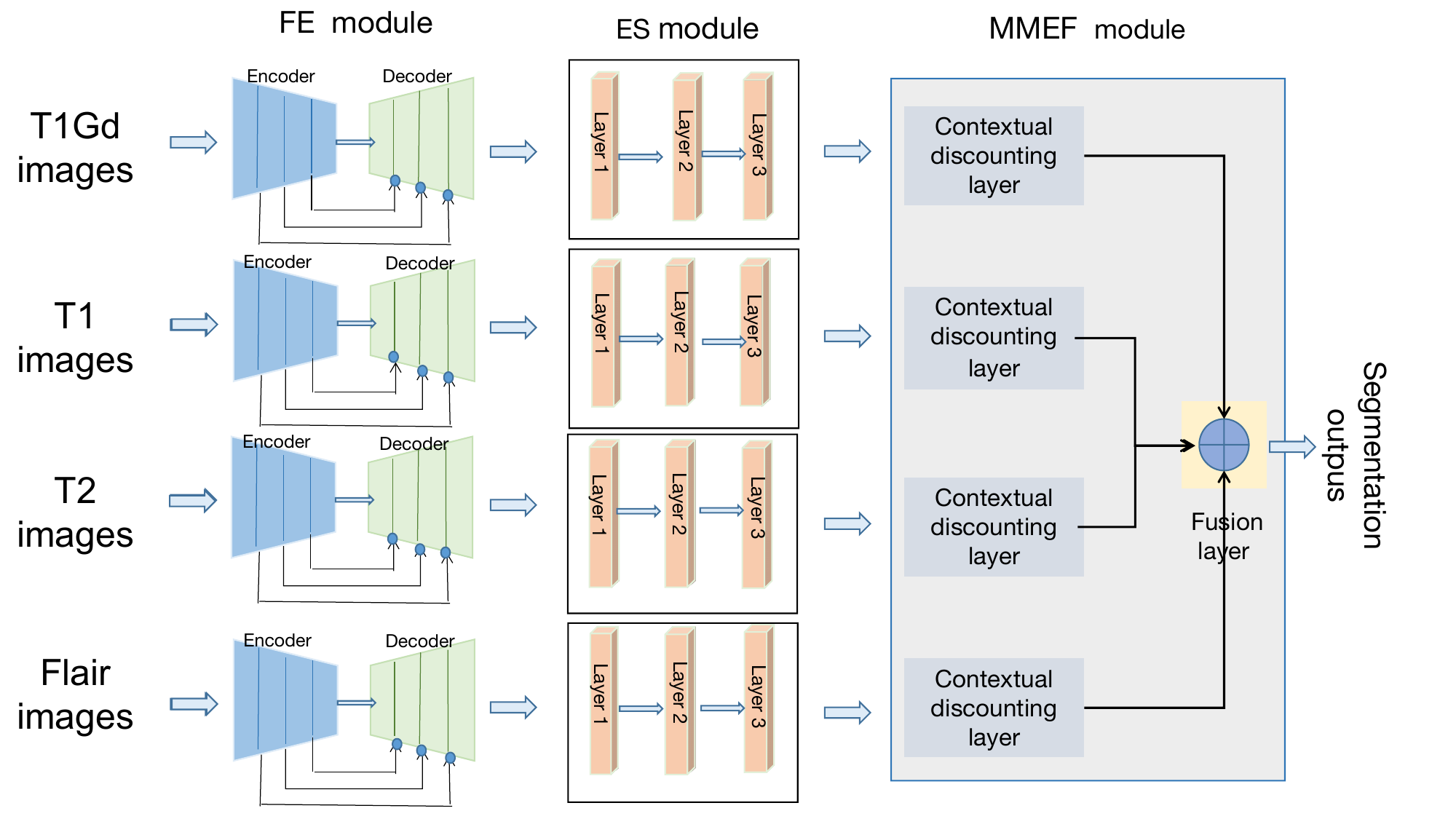}
\caption{Multimodal evidence fusion framework. It is composed of four encoder-decoder feature extraction (FE) modules corresponding to T1Gd, T1, T2 and FLAIR modality inputs; four evidential segmentation (ES) modules corresponding to each of the inputs; and a multimodal evidence fusion (MMEF) module.}
\label{fig:architecture}
\end{figure}

Our proposed evidence fusion framework is shown in Figure~\ref{fig:architecture}. It is composed of (1) four encoder-decoder feature extraction (FE) modules, (2) four evidential segmentation (ES) modules, and (3) a multimodal evidence fusion (MMEF) module. Since this chapter focuses on uncertainty representation and multimodal evidence fusion, the FE module here can be any state-of-the-art medical image segmentation models, such as Residual-UNet (see Section \ref{subsec:Unet}), nnUNet (see Section \ref{subsec: nnunet}). Details about the evidential segmentation and multimodal evidence fusion modules will be given in Sections \ref{subsec:ES} and \ref{subsec:MMEF}, respectively. The loss function used for optimizing the framework will be described in Section \ref{subsec:loss}.

\subsection{Evidential segmentation (ES) module}
\label{subsec:ES}
Based on ENN introduced in \ref{subsubsec: distance}, we propose a BFT-based evidential segmentation module to quantify the uncertainty about the class of each voxel by a mass function. The basic idea of the evidential segmentation module is to assign a mass to each of the $C$ classes and to the whole set of classes $\Omega$, based on the distance between the input $x$ of each voxel and $I$ prototypes. The input $x$ is the high-level semantic feature vector generated by the feature extraction module. For each prototype $p_i$, the evidential segmentation module will first calculate mass functions $m_i$ representing the corresponding evidence using the following equations:
\begin{subequations}
\begin{align}
m_i(\{\omega _{c}\})&=u_{ic} \alpha _i \exp(-\gamma_i \| x-p_i\|^2), \quad c=1,..., C\\
m_{i}(\Omega)&=1-\alpha _i \exp(-\gamma_i \| x-p_i\|^2), 
\label{eq:enn2}
\end{align}
\end{subequations}
where $\gamma_i>0$ and $\alpha_i \in [0,1]$ are two parameters, $u_{ic}$ is the membership degree of prototype $i$ to class $\omega_c$, and $\sum _{c=1}^C u_{ic}=1$. Then the $I$ pieces of evidence are combined by Dempster's rule (see Section \ref{sec:fusion}). As shown in Fig \ref{fig:architecture}, for each input modality, we have an independent evidential segmentation module plugged into the feature extraction module's output. For each voxel, the evidential segmentation module generates a mass function $m$ to each of the $C$ classes and a mass function $m(\Omega)$ to the whole set of classes $\Omega$. Here $m(\Omega)$ is regarded as the degree of uncertainty when assigning a voxel to the given classes.

\subsection{Multimodal evidence fusion (MMEF) module}
\label{subsec:MMEF}
In this work, we address the problem of modeling the reliability of single-modality medical images by the discounting operation under the BFT framework. The multiple discounted evidence is then combined by Dempster's rule (see Section \ref{sec:fusion}). In this section, we explain the use of contextual discounting in multimodal MR image fusion.

\paragraph{Discounting source evidence with reliability coefficient}
We follow Mercier's idea \cite{mercier2008refined} to extend the \emph{discounting} operation into \emph{contextual discounting}. With \emph{contextual discounting}, we can represent richer meta-knowledge regarding the reliability of the single modality image in different contexts, i.e., different types of MR images are conditionally reliable regarding different brain tumor segmentation tasks. We assume that we have evidence regarding the reliability of a source modality $S$, conditionally on each hypothesis $\omega_c \in \Omega$, i.e., in a context where the quantity $x$ of interest is known to be equal to $\omega_c$. 
When it is known that the actual value of $x$ is $\omega_c$, $1-\beta_c $ is the plausibility that the source is not reliable in the same context, and thus our knowledge about $x$ taking value in $\{\omega_c\}$ is vacuous, i.e., $m_{?}$. For each source of information, we thus have a vector $\bbeta=[\beta_1,...,\beta_C]$ representing its reliability in different contexts. The contour function of the discounted mass function can be shown \cite{mercier2008refined} to be
\begin{equation}
     ^{\bbeta} pl(\{\omega_c \})=1-\beta_c+\beta_c \, pl(\{\omega_c\}), c=1,...,C,
     \label{eq:c_discounting}
\end{equation}
where $^{\bbeta} pl(\{\omega_c\})$ is the discounted plausibility of class $m(\{\omega_c\})$, $pl(\{\omega_c\}) =m(\{\omega_c\})+m(\Omega)$. By representing the evidence with the contour function, we can decrease the calculation complexity.

\paragraph{Fusion with discounted source evidence}
\label{subsubsec:fusion}
The four independent discounted evidence corresponding to four input modalities must be combined to generate final evidence. Equation \eqref{eq:prodpl} allows us to compute the contour function in time proportional to the size of $\Omega$, without having to compute the combined mass $m_{1}\oplus m_{2}$ with \eqref{eq:demp1}. In our case, $^{\bbeta^1} pl_{T1}$, $^{\bbeta^2} pl_{T2}$, $^{\bbeta^3} pl_{T1Gd}$ and $^{\bbeta^4} pl_{FLAIR}$ represent the contour functions provided, respectively, by modality T1, T2, T1Gd and FLAIR, with discount rate vectors $1-\bbeta^1$, $1-\bbeta^2$, $1-\bbeta^3$ and $1-\bbeta^4$. As a consequence of \eqref{eq:prodpl}, the normalized contour function of multiple sources of information is proportional to the product of the contour function of each source information and can be used to simplify the processes of the orthogonal sum of $m_1$ and $m_2$ using Dempster’s rule. The combined contour function $^{\bbeta} pl$ of each singleton is, thus,
\begin{equation}
      ^{\bbeta} pl (\{ \omega_c \})= \frac {\prod_{t=1}^{T} {^{\bbeta^t_c}} pl_{t}(\{ \omega_c \})} {\sum_{c=1}^{C} \prod_{t=1}^{T} {^{\bbeta^t_c}} pl_{t}(\{ \omega_c \})}, c=1, ... , C, 
      \label{eq:dis_pl}
\end{equation}
where $T$ is the number of information sources, and $C$ is the number of classes on $\Omega$. 
\subsection{Learning with contextual discounted Dice loss}
\label{subsec:loss}
To optimize the whole framework, we maximize the overlap region between the output $S$ and the ground truth $G$ by minimizing the following loss function \textsf{$loss_D$}
, given as follows:
\begin{equation}
    loss_{D}=1-\frac{2 \sum_{n=1}^{N}\sum_{c=1}^{C} {S_{cn} \times G_{cn}}} {\sum_{n=1}^{N}\sum_{c=1}^{C} {S_{cn} + G_{cn}}},
\label{eq:loss}
\end{equation}
where $C$ and $N$ are the number of classes and voxels; $G_{cn}=1$ if voxel $n$ belongs to class $c$, and $G_{cn}=0$ otherwise, and $S_{cn}$ represents the corresponding predicted mask based on the discounted source evidence after fusion. 

\section{Experiments and results}
\label{sec:exp6}
\subsection{Experiment settings}

\begin{figure}
\centering
\includegraphics[width=\textwidth]{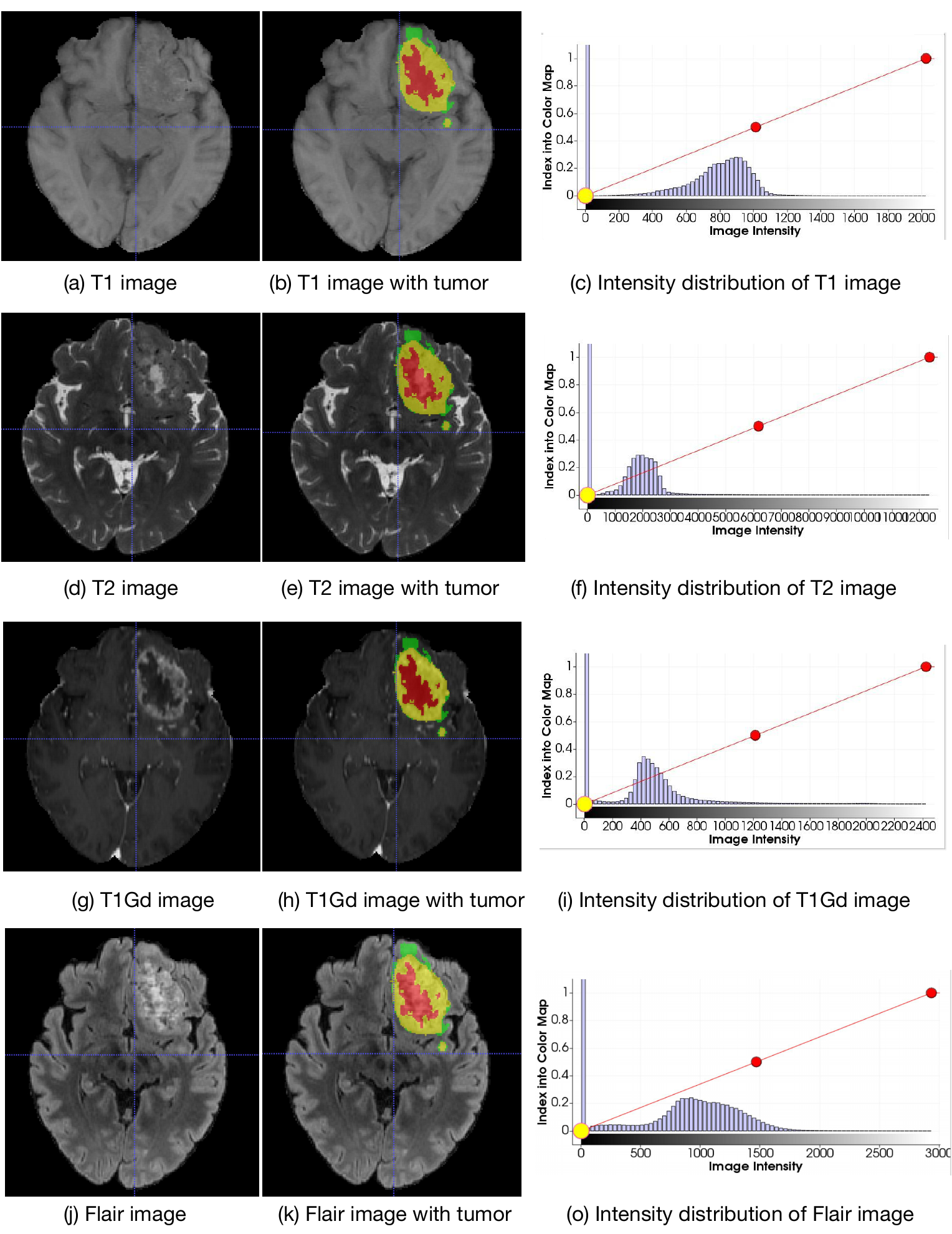}
\caption{Examples of a patient with the brain tumor in four modalities: T1, post-contrast T1-weighted (T1Gd), T2, FLAIR. The three tumor classes: enhancing core (yellow), necrotic/cystic core (red), and edema/invasion (green). The first, second and third columns show, respectively, the original image, the image with the tumor mask in three classes, and the intensity histogram of the image.}
\label{fig:image_tumor_intensity}
\end{figure}

\paragraph{Dataset}
\label{subsec:data}
We used a multimodal brain tumor dataset from the BraTS 2021 challenge \cite{baid2021rsna} to evaluate our framework. The original BraTS2021 dataset comprises training, validation and test sets with the corresponding number of cases as 1251, 219 and 570, respectively. Compared to the BraTS2019 dataset we used in Chapter \ref{Chapter4}, BraTS2021 is a more recent version with more brain tumor cases available. For BraTS2021, the validation and test set and the corresponding ground truth are not available; thus, in this chapter, we train and evaluate our framework with the training set. Following \cite{peiris2021volumetric}, we randomly divided the 1251 scans into 834, 208, and 209 for new training, validation, and test set, respectively. 
Each case has four modalities: T1, T1Gd, T2, and FLAIR with $240\times 240 \times 155$ voxels. Figure \ref{fig:image_tumor_intensity} shows four modality MR image slices from one patient. The appearance of a brain tumor varies in different modalities, and the tumor boundaries are blurred, making it hard to delineate different tumors precisely. Annotations of 1251 scans comprise the GD-enhancing tumor (ET-label 4), the peritumoral edema (ED-label 2), and the necrotic and non-enhancing tumor core (NRC/NET -label1). Similar to BraTS2019, the task of BraTS2021 is to segment three overlap regions: enhancing tumor (ET, label 4), tumor core (TC, the composition of label 1 and 4), and whole tumor (WT, the composition of label 1, 2, and 4).

\paragraph{Pre-processing}
Similar to the pre-processing methods in \cite{peiris2021volumetric}, we performed a min-max scaling operation followed by clipping intensity values to standardize all volumes and crop/padding the volumes to a fixed size of $128 \times 128 \times 128$ by removing the unnecessary background. No data augmentation technique was applied and no additional data was used in this study.

\paragraph{Parameter initialization and learning}  
The initial values of parameters (see \eqref{eq:enn2})  $\alpha_i$ and $\gamma_i$ were set, respectively, to 0.5 and 0.01, and membership degrees $u_{ic}$ were initialized randomly by drawing uniform random numbers and normalizing. The prototypes were initialized by the $k$-means clustering algorithm. For the multimodal evidence fusion module, the initial values of parameter $\beta_c$ were set to 0.5. Each model was trained on the learning set with 100 epochs using the Adam optimization algorithm. The initial learning rate was set to $10^{-3}$. The model with the best performance on the validation set was saved as the final model for testing\footnote{The code is available at \url{https://github.com/iWeisskohl}.}.

The multimodal evidence fusion framework was implemented in Python with the PyTorch-based medical image framework MONAI, and was trained and tested on a desktop with a 2.20GHz Intel(R) Xeon(R) CPU E5-2698 v4 and a Tesla V100-SXM2 graphics card with 32 GB GPU memory.

\subsection{Segmentation accuracy}
\label{subsubsec:acc}
We used the Dice Score (see \eqref{eq: dice_score}) and the Hausdorff Distance (HD) (see \eqref{eq:hds}) as our evaluation metrics to compare the segmentation accuracy with the baseline methods and state-of-the-art methods. For each patient, we separately computed these two indices for the three classes and then averaged indices over the patients, following a similar evaluation strategy as in \cite{peiris2021volumetric}. We use Residual-UNet (see Section \ref{subsec:Unet}) and nnUNet (see Section \ref{subsec: nnunet}) as feature extraction modules to construct two models, named MMEF-UNet and MMEF-nnUNet, respectively. We compared our models with two recent transformer-based models (nnFormer \cite{zhou2021nnformer}, VT-UNet \cite{peiris2021volumetric}), two classical CNN-based methods (UNet \cite{cciccek20163d}, V-Net \cite{milletari2016v}), and the two baseline models. The quantitative results are reported in Table~\ref{sota}. Our models outperform the two classical CNN-based models and two recent transformer-based methods in terms of the Dice score; the best result is obtained by MMEF-nnUNet according to this criterion. In contrast, MMEF-UNet achieves the lowest HD.
\begin{table}
  \centering
  \caption{Segmentation results on the BraTS 2021 test set. The best result is shown in bold, and the second best is underlined.}
  \label{sota}
  \scalebox{0.78}{
  \begin{tabular}{lllllllll}
  \hline
  \multicolumn{1}{l}{Methods}  &\multicolumn{4}{c}{Dice score} &\multicolumn{4}{c}{HD}\\
  \cline{2-5}
  \cline{6-9}
 & ET&TC&WT&Mean & ET&TC&WT&Mean\\
\hline
UNet \cite{cciccek20163d}&83.39& 86.28 &89.59& 86.42 &11.49& 6.18& 6.15 &7.94\\
VNet \cite{milletari2016v} & 81.04& 84.71 &90.32& 85.36& 17.20 &7.48&  7.53 &10.73\\
nnFormer \cite{zhou2021nnformer}&82.83 &86.48 &90.37& 86.56&11.66 & 7.89 & 8.00 &9.18\\
VT-UNet \cite{peiris2021volumetric}&85.59 &87.41 &91.20& 88.07& \textbf{10.03} &6.29&  6.23 & \underline{7.52}\\

Residual-UNet \cite{kerfoot2018left} &	85.07 &87.61 &89.78&87.48 & 11.76 & \underline{6.14}& 6.31 & 8.07 	\\
nnUNet \cite{isensee2018nnu} &\underline{87.12}& \textbf{90.31}& \underline{91.47}& \underline{89.68} & 12.46 & 11.04& 5.97&9.82\\
MMEF-UNet (Ours) & 86.96 & 87.46 & 90.68 &88.36& \underline{10.20} & \textbf{6.07} & \underline{5.29} & \textbf{7.18}\\
MMEF-nnUNet (Ours) &\textbf{87.26} &\underline{90.05} &\textbf{92.83}& \textbf{90.05} & 10.09& 9.68& \textbf{5.10} &8.29 \\
\hline
\end{tabular}}
\end{table}

\begin{figure}
\centering
\includegraphics[width=\textwidth]{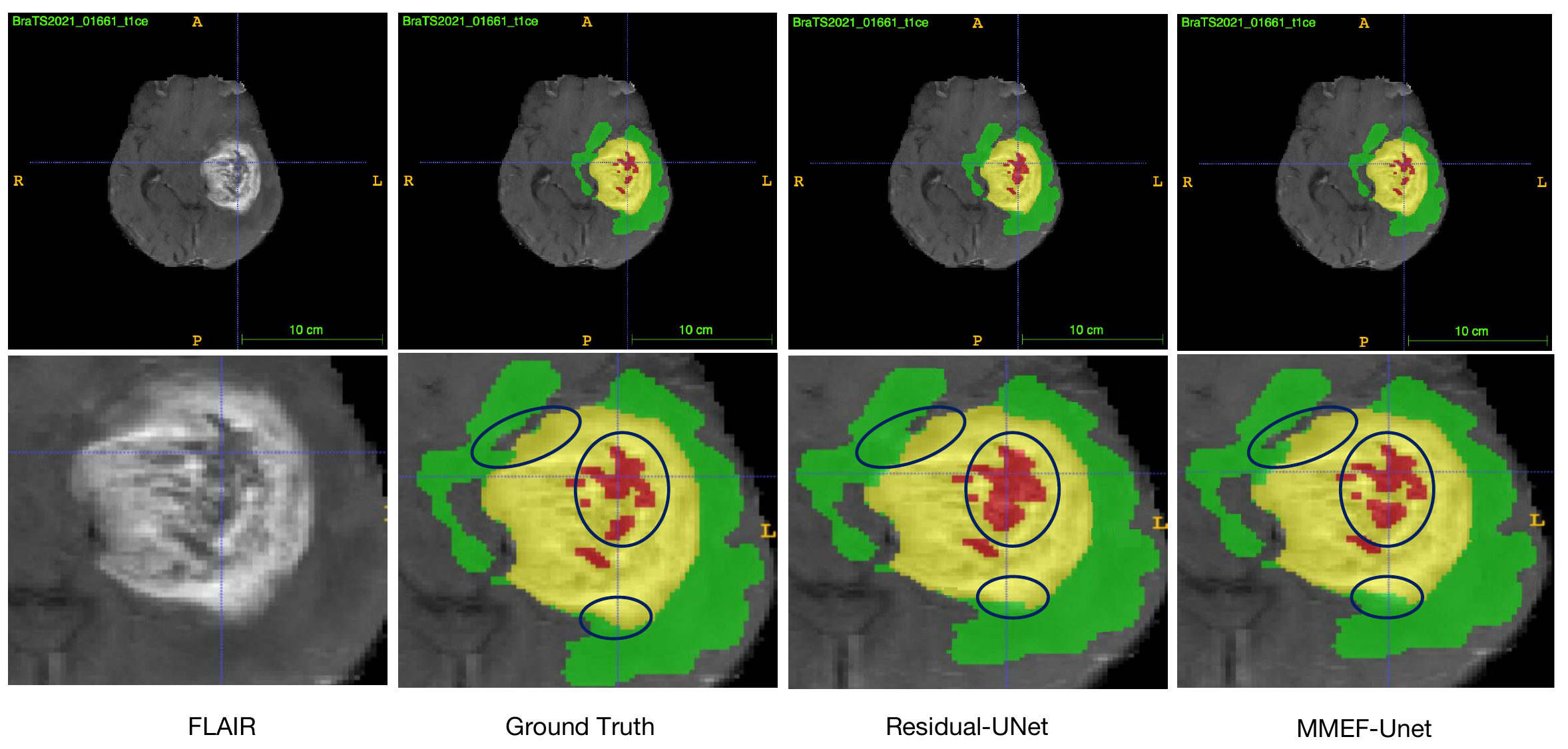}
\caption{Visualization of the results on BraTs2021 data. The first and the second row are, respectively, the whole brain region and the tumor subregions (the main differences are marked in blue circles). The three columns correspond, from left to right, to the FLAIR image, the ground truth, and the segmentation results obtained by Residual-UNet and MMEF-UNet. The green, yellow and red represent the ET, ED, and NRC/NET, respectively.}
\label{fig:visu}
\end{figure}

\paragraph{Visualized segmentation results}
Figure ~\ref{fig:visu} shows the segmentation results of Residual-UNet with the inputs of four concatenated modalities and MMEF-UNet with the inputs of four separate modalities. Our model locates and segments brain tumor subregions precisely, especially at the tumor boundary where uncertainty is high.

\subsection{Segmentation reliability}
\label{subsubsec:reliab}

A very important question related to the design of a multimodal medical image segmentation model is the reliability of its fusion results. Although all the recent publications report improved segmentation performance as the result of merging multimodal medical images into deep neural networks, the reliability of the results obtained with such fusion processes has seldom been investigated. There are two approaches to measuring the segmentation reliability. One is to test the reliability of the model by calculating the expected calibration error (ECE, see \eqref{eq:ECE}). The other is to calculate the reliability of source information \cite{kobayashi1999sensor}. In this chapter, the reliability of source information is represented by the reliability coefficients $\bbeta$ we mentioned in \eqref{eq:c_discounting}. 

\paragraph{Model's reliability}
To test the reliability of our models, we computed the ECE for the two baseline models and ours. We obtain ECE values of 2.35 \% and 2.04\%, respectively, for Residual-UNet and MMEF-UNet, against ECE values of 4.46\% and 4.05\%, respectively, for nnUNet and MMEF-nnUNet. Since the probabilities computed by our models are better calibrated, the segmentation results of our models are more reliable.
\begin{table}[ht]
    \centering
        \caption{Dice score and ECE on the BraTs2021 test set.}
    \begin{tabular}{c|c|c}
    \hline
    Methods & Dice score &  ECE (\%) \\
        \hline
         Residual-UNet& 86.42&2.35\\
         nnUNet& 89.68&4.46\\
         MMEF-UNet& 88.36 &2.04\\
         MMEF-nnUNet& 90.05 &4.05\\
\hline
    \end{tabular}
    \label{tab:my_label}
\end{table}

\paragraph{Interpretation of reliability coefficients} 

Table~\ref{beta} shows the learned reliability coefficients $\bbeta$ (see \eqref{eq:c_discounting}) for the four modalities with three different classes. The evidence from the T1Gd modality is reliable for ET, ED, and NRC classes with the highest reliability value (close to 1). In contrast, the evidence from the FLAIR modality is only reliable for the ED class, with a reliability value coefficient to 0.86. The reliability coefficient of the T2 modality is only around 0.4 for the three classes, which means the T2 modality has a limited contribution to the final decision. The evidence from the T1 modality is less reliable for the three classes compared to the evidence of the other three modalities. These reliability results are consistent with domain knowledge about these modalities reported in \cite{baid2021rsna}, i.e., ET is described by areas with both visually avid, as well as faint, enhancement on T1Gd, the appearance of NRC/NET is hypointense on T1Gd. That is why its reliability coefficient is high for these areas. ED is defined by the abnormal hyperintense signal envelope on the FLAIR volumes. This interpretation of the reliability coefficients offers an effective way to explain the segmentation results to physicians and patients.

\begin{table}
  \centering
  \caption{Estimated reliability coefficient $\beta$ (after training) for classes ET, ED and NRC/NET and the four modalities. Higher values correspond to greater contribution to the segmentation.}
  \label{beta}
  \begin{tabular}{lcccc}
  \hline
  \multicolumn{1}{l}{$\beta$}& &\multicolumn{1}{l}{ET}  & \multicolumn{1}{l}{ED}&
  \multicolumn{1}{l}{NRC/NET} \\
  \hline
  T1Gd&&0.9996 &	0.9726  &0.9998\\
  T1&&0.4900&	0.0401&	0.2655\\
  T2&&0.4814&0.3881&	0.4563\\
  FLAIR&&0.0748&	0.86207	&0.07512\\
\hline
\end{tabular}
\end{table}

\subsection{Ablation analysis}
\label{subsubsec:abla}
We also investigated the contribution of each module component to the performance of the framework. Table~\ref{ablition:es} highlights the importance of introducing the evidential segmentation and multimodal evidence fusion modules. Residual-UNet is the baseline model that uses the softmax transformation to map feature vectors into probabilities. Compared to Residual-UNet, Residual-UNet-ES uses the ES module instead of softmax that map feature vectors into mass functions. MMEF-UNet, our final proposal, fuses the four single-modality outputs from Residual-UNet-ES with the MMEF module.

\paragraph{Illustrative results with single-modality inputs (Residual-UNet vs. Residual-UNet-ES)}
Table \ref{ablition:es} shows the performance of segmentation results with and without the evidential segmentation module when only single-modality MR images are used. Compared to the baseline method Residual-UNet, our proposal, which plugs the evidential segmentation module after Residual-UNet, improves the segmentation performance based on single inputs. For example, compared with Residual-UNet, Residual-UNet-ES has an increase in 1.38\% and 0.62\% of Dice score on T1Gd and FLAIR modality, respectively. Furthermore, the introduction of evidential segmentation decreases the ECE on four modalities, which means the segmentation results with the evidential segmentation module are more reliable.

\paragraph{Illustrative results with different fusion methods (concatenating fusion vs. multimodal evidence fusion)} We also compared the performance of fusion methods with image concatenation and multimodal evidence fusion. As we can see from Table \ref{ablition:es}, compared to any single-modality input, the fusion of multimodal medical images increases the Dice score by a significant amount. Furthermore, the use of the multimodal evidence fusion module improves the performance in terms of segmentation accuracy (Dice score) and reliability (ECE) in the subregions of ET, ED and NRC/NET compared to the input-level fusion that concatenates T1, T1Gd, T2, and FLAIR images as one single input. The above results demonstrate the effectiveness of our evidence fusion models.

\begin{table}
  \caption{Segmentation results on BraTS 2021 validation set($\uparrow $ means higher is better).}
  \centering
  \label{ablition:es}
  \scalebox{0.85}{
  \begin{tabular}{llllllcll}
  \hline
  \multicolumn{1}{l}{Methods} &\multicolumn{1}{l}{Input} &\multicolumn{3}{c}{Dice score in single class}&\multicolumn{1}{c}{Mean Dice}  &\multicolumn{1}{c}{ECE(\%)} 
\\
  \cline{3-5} 
  & Modality&ET&ED&NRC/NET &&\\
\hline
 Residual-UNet&T1Gd&79.55 & 61.16& 71.26& 71.30 & 3.67\\
 Residual-UNet-ES&T1Gd& 80.50 $\uparrow $ & 62.95$\uparrow $ & 73.39$\uparrow $ & 72.68$\uparrow $ &3.24$\uparrow$\\
\hline
 Residual-UNet&T1& 44.32$\uparrow $ & 50.58$\uparrow $ &44.85$\uparrow $& 48.97$\uparrow $&4.27\\
 Residual-UNet-ES&T1& 44.41& 56.08 &44.06 & 48.55&3.33$\uparrow$\\
\hline
Residual-UNet&T2& 45.56&65.38&46.06&53.29&3.69
\\
Residual-UNet-ES&T2& 46.37$\uparrow $ & 66.36$\uparrow $ &46.98$\uparrow $ &54.06$\uparrow $ &3.18 $\uparrow$\\
\hline
Residual-UNet&FLAIR& 43.83 &70.85&39.99&52.54&3.45
\\
Residual-UNet-ES&FLAIR&45.45$\uparrow $
&71.58$\uparrow $&40.12$\uparrow $&53.16 $\uparrow $ &3.43$\uparrow$\\
\hline 
\multirow{2}*{Residual-UNet}  &T1Gd,T1&	86.06 &81.68 &77.07 & 81.60&2.35 \\
&T2, FLAIR&\\
\multirow{2}*{MMEF-UNet}&T1Gd,T1&86.46$\uparrow $&83.79$\uparrow $&77.50$\uparrow $&82.57$\uparrow $ &2.04 $\uparrow$\\
&T2, FLAIR&\\
\hline
\end{tabular}}
\end{table}

\section{Conclusion}

\label{sec:conclu6}

Based on BFT, a multimodal evidence fusion framework considering segmentation uncertainty and source reliability has been proposed for multi-MRI brain tumor segmentation. The originality of our method is that the evidential segmentation module performs both tumor segmentation and uncertainty quantification, and the multimodal evidence fusion module carries out multimodal evidence fusion with contextual discounting and Dempster's rule.  

This work is the first to implement contextual discounting for multimodal information fusion with BFT and deep neural networks. The contextual discounting operation allows us to take into account the uncertainty of the different sources of information directly, and it reveals the reliability of different modalities in different contexts as well. Our method can be considered as decision-level fusion and can be used together with any state-of-the-art feature extraction module to achieve better performance.

Some limitations of this work remain in the computation cost. We treat single modality images as independent inputs using independent feature extraction and evidential segmentation modules, which introduces additional computation costs compared to image concatenation methods (e.g., the FLOPs and parameter numbers are equal to 280.07G and 76.85M for UNet-MMEF, against 73.32G, and 19.21M for Residual-UNet). In future research, the refinement of the framework to improve segmentation performance and reduce its complexity will be considered.

 \stopcontents[chapters]


\chapter{Conclusion and perspective}
\label{Chapter7}

\section{Conclusion}

The issue of uncertainty and reliability of machine learning methods has recently come to the forefront. It has incited researchers in the medical image segmentation domain to study both accurate and reliable segmentation methods~\cite{kwon2020uncertainty,mehrtash2020confidence,ghoshal2021estimating,zhou2022trustworthy}. In this thesis, we studied on medical image segmentation with a specific focus on segmentation uncertainty and reliability under the framework of BFT and deep neural networks. In particular, we first briefly introduced medical image segmentation in Chapter \ref{Chapter1} by describing the context of medical image segmentation, the deep learning medical segmentation methods and three baseline deep segmentation models. In Chapter \ref{Chapter2}, we then summarized the main concepts of BFT, i.e., representation of evidence, Dempster's combination rule, discounting, decision-making and the methods to generate mass functions. In Chapter \ref{Chapter3}, we introduced the BFT-based medical image segmentation methods in detail by grouping them according to the number of input modalities and classifiers used to generate mass functions, and we showed how unreliable or conflicting sources of information can be combined to reach reliable fusion results. 

Our three main technique contributions have been presented in Chapters \ref{Chapter4}, \ref{Chapter5} and \ref{Chapter6}. In Chapter \ref{Chapter4}, we have described a semi-supervised medical image segmentation model. The main idea is to introduce constraints for unlabelled data by generating pseudo labels and decrease the uncertainty caused by the lack of labels by fusing evidence from the probability distribution and evidence distribution. In Chapter \ref{Chapter5}, we introduced two evidential classifiers for uncertainty quantification and showed how they can be used with deep segmentation models to improve segmentation accuracy and reliability. In Chapter \ref{Chapter6}, we presented a multimodal medical image segmentation model that learns the reliability of each modality image when segmenting different types of tumors. We showed that using contextual discounting when fusing evidence from different modalities allows us to reach more reliable and explainable results. Our experimental results show that in many applications, BFT-based methods have the potential to outperform probability-based methods by modeling information more effectively and combining multiple piece of evidence at different stages.

\section{Future work}
Despite the advantages of BFT for medical image segmentation, existing methods still have limitations that need to be addressed. Some directions for further research are discussed below.

First, acquiring large amounts of labeled training data is particularly challenging for medical image segmentation tasks and has become the bottleneck of learning-based segmentation performance. The successful application of unsupervised basic belief assignment methods to medical image segmentation points to a new direction to address the problem of lack of annotated training data. In this thesis, we have applied semi-supervised learning with BFT to improve the segmentation accuracy when only partial training data are labeled (see Chapter \ref{Chapter4}). To our best knowledge, there is no published paper dealing with the combination of unsupervised basic belief assignment methods with deep learning. The neural-network-based evidential clustering method described in~\cite{denoeux21b} and the EKNN rule with partially supervised learning introduced in~\cite{denoeux2019new} are two steps in these directions. These works provide insights into using unsupervised or semi-supervised learning to quantify segmentation uncertainty with unannotated or partially annotated data sets. Future research could study the combination of unsupervised basic belief assignment methods with deep learning to overcome the annotation limitation.

Second, most of the existing BFT-based medical image segmentation methods still use low-level features and do not fully exploit the advantages of deep learning. Combining BFT with deep segmentation networks should allow us to develop accurate and reliable segmentation models, particularly in medical image segmentation tasks for which medical knowledge is available and can be modeled by mass functions. In this thesis, we have revealed the similarity of ENN and RBF network when acting as an evidential classifier and integrated both classifiers within a deep segmentation network to improve the segmentation accuracy and reliability (see Chapter \ref{Chapter5}). We believe that more promising results will be obtained by blending BFT with the existing powerful deep medical image segmentation models, e.g., transformer \cite{shamshad2022transformers} and diffusion models \cite{kazerouni2022diffusion}.

Third, even though the BFT-based fusion methods have considered the conflict of multiple sources of evidence, a problem remains that multimodal medical images may have different degrees of reliability when segmenting different type of tumors or organs. Multimodal medical image fusion may fail if some sources of information are unreliable or partially reliable for performing various segmentation tasks. In this thesis, we have proposed a method to learn reliability coefficients conditionally on different segmentation tasks in a deep neural network (see Chapter \ref{Chapter6}). Another interesting research direction, accordingly, is to estimate task-specific reliability and enhance the explainability of deep evidential neural networks using contextual discounting for multimodal or cross-modal medical image segmentation tasks.

We hope our work will increase awareness of the challenges of existing BFT-based medical image segmentation methods, call for future contributions to bridge the gap between experimental performance and clinical application, and develop accurate, reliable, and explainable deep segmentation models.

 \stopcontents[chapters] 
\printbibliography[heading=bibintoc]

\end{document}